\documentclass[a4paper,11pt]{book}

\usepackage[T1]{fontenc}
\usepackage[utf8]{inputenc}

\usepackage[french,german,english]{babel}

\usepackage{lmodern} 

\usepackage{fourier} 
\usepackage{setspace} 

\usepackage{graphicx,xcolor}
\graphicspath{{images/}}

\usepackage{booktabs}
\usepackage{lipsum}
\usepackage{microtype}
\usepackage{url}

\usepackage{fancyhdr}

\fancypagestyle{plain}{
	\fancyhf{}

	\fancyfoot[C,C]{\thepage}}
\fancypagestyle{addpagenumbersforpdfimports}{
	\fancyhead{}
	
	\fancyfoot{}
	\fancyfoot[RO,LE]{\thepage}
}

\usepackage{listings}
\usepackage{hyperref}
\hypersetup{pdfborder={0 0 0},
	colorlinks=true,
	linkcolor=black,
	citecolor=black,
	urlcolor=black}
\ifpdf
\usepackage[final]{pdfpages}
\else
\usepackage{calc}
\usepackage{breakurl}
\usepackage[nlwarning=false]{hypdvips}
\usepackage{backref}
\renewcommand*{\backref}[1]{}
\fi
\usepackage{bookmark}

\makeatletter
\renewcommand\@pnumwidth{20pt}
\makeatother

\makeatletter
\def\cleardoublepage{\clearpage\if@twoside \ifodd\c@page\else
    \hbox{}
    \thispagestyle{empty}
    \newpage
    \if@twocolumn\hbox{}\newpage\fi\fi\fi}
\makeatother \clearpage{\pagestyle{plain}\cleardoublepage}

\usepackage{color}
\usepackage{tikz}
\usepackage[explicit]{titlesec}
\newcommand*\chapterlabel{}
\titleformat{\chapter}[display]  
	{\normalfont\bfseries\Huge} 
	{\gdef\chapterlabel{\thechapter\ }}     
 	{0pt} 
 	  {\begin{tikzpicture}[remember picture,overlay]
    \node[yshift=-8cm] at (current page.north west)
      {\begin{tikzpicture}[remember picture, overlay]
        \draw[fill=black] (0,0) rectangle(35.5mm,15mm);
        \node[anchor=north east,yshift=-7.2cm,xshift=34mm,minimum height=30mm,inner sep=0mm] at (current page.north west)
        {\parbox[top][30mm][t]{15mm}{\raggedleft \rule{0cm}{0.6cm}\color{white}\chapterlabel}};  
        \node[anchor=north west,yshift=-7.2cm,xshift=37mm,text width=\textwidth,minimum height=30mm,inner sep=0mm] at (current page.north west)
              {\parbox[top][30mm][t]{\textwidth}{\rule{0cm}{0.6cm}\color{black}#1}};
       \end{tikzpicture}
      };
   \end{tikzpicture}
   \gdef\chapterlabel{}
  } 
\titlespacing*{name=\chapter,numberless}{-3.7cm}{83.2pt-\parskip}{-3.2pt+\parskip}
\titlespacing*{\chapter}{-3.7cm}{50pt-\parskip-\parskip}{30pt+\parskip+\parskip}
\titlespacing*{\section}{0pt}{13.2pt}{1em-\parskip}  
\titlespacing*{\subsection}{0pt}{13.2pt}{1em-\parskip}
\titlespacing*{\subsubsection}{0pt}{13.2pt}{1em-\parskip}
\titlespacing*{\paragraph}{0pt}{13.2pt}{1em-\parskip}

\newcounter{myparts}
\newcommand*\partlabel{}
\titleformat{\part}[display]  
	{\normalfont\bfseries\Huge} 
	{\gdef\partlabel{\thepart\ }}     
 	{0pt} 
 	  {\ifpdf\setlength{\unitlength}{20mm}\else\setlength{\unitlength}{0mm}\fi
	  \addtocounter{myparts}{1}
	  \begin{tikzpicture}[remember picture,overlay]
    \node[anchor=north west,xshift=-65mm,yshift=-6.9cm-\value{myparts}*20mm] at (current page.north east) 
      {\begin{tikzpicture}[remember picture, overlay]
        \draw[fill=black] (0,0) rectangle(62mm,20mm);   
        \node[anchor=north west,yshift=-6.1cm-\value{myparts}*\unitlength,xshift=-60.5mm,minimum height=30mm,inner sep=0mm] at (current page.north east)
        {\parbox[top][30mm][t]{55mm}{\raggedright \color{white}Part \partlabel \rule{0cm}{0.6cm}}};  
        \node[anchor=north east,yshift=-6.1cm-\value{myparts}*\unitlength,xshift=-63.5mm,text width=\textwidth,minimum height=30mm,inner sep=0mm] at (current page.north east)
              {\parbox[top][30mm][t]{\textwidth}{\raggedleft \rule{0cm}{0.6cm}\color{black}#1}};
       \end{tikzpicture}
      };
   \end{tikzpicture}
   \gdef\partlabel{}
  } 
\titlespacing*{\part}{11.06cm}{26.4pt-\parskip-\parskip}{0pt}

\usepackage{amsmath}
\usepackage{amsfonts}
\usepackage{amssymb}
\usepackage{mathtools}
\makeatletter
\def\resetMathstrut@{%
  \setbox\z@\hbox{%
    \mathchardef\@tempa\mathcode`\(\relax
      \def\@tempb##1"##2##3{\the\textfont"##3\char"}%
      \expandafter\@tempb\meaning\@tempa \relax
  }%
  \ht\Mathstrutbox@1.2\ht\z@ \dp\Mathstrutbox@1.2\dp\z@
}
\makeatother
\usepackage{hyperref}
\usepackage{cleveref}
\usepackage{amsmath}

\makeatletter
\def\bstctlcite{\@ifnextchar[{\@bstctlcite}{\@bstctlcite[@auxout]}}
\def\@bstctlcite[#1]#2{\@bsphack
  \@for\@citeb:=#2\do{%
    \edef\@citeb{\expandafter\@firstofone\@citeb}%
    \if@filesw\immediate\write\csname #1\endcsname{\string\citation{\@citeb}}\fi}%
  \@esphack}
\makeatother

\tikzset{
 int/.style={
  draw=black,
  minimum size=8mm,
  alias=thisone,
  append after command={(thisone.south west) -- (thisone.north east)}
 }
}

\tikzstyle{sum} = [circle,draw,minimum size=2mm,inner sep=0pt]

\definecolor{boschred}{RGB}{215,0,18}
\definecolor{fuchsia}{RGB}{168,1,99}
\definecolor{violet}{RGB}{63,19,108}
\definecolor{darkblue}{RGB}{8,66,126}
\definecolor{lightblue}{RGB}{14,120,197}
\definecolor{blueAkzent3}{RGB}{184,217,251}
\definecolor{blueAkzent4}{RGB}{199,229,251}
\definecolor{lightblue}{RGB}{14,120,197}
\definecolor{turquoise}{RGB}{19,153,160}
\definecolor{lightgreen}{RGB}{103,180,25}
\definecolor{darkgreen}{RGB}{10,81,57}
\definecolor{darkgray}{RGB}{66,76,88}
\definecolor{lightgray}{RGB}{178,179,181}

\definecolor{matlab1}{rgb}{0,    	  0.4470,    0.7410}
\definecolor{matlab2}{rgb}{0.8500,    0.3250,    0.0980}
\definecolor{matlab3}{rgb}{0.9290,    0.6940,    0.1250}
\definecolor{matlab4}{rgb}{0.4940,    0.1840,    0.5560}
\definecolor{matlab5}{rgb}{0.4660,    0.6740,    0.1880}
\definecolor{matlab6}{rgb}{0.3010,    0.7450,    0.9330}
\definecolor{matlab7}{rgb}{0.6350,    0.0780,    0.1840}

\pgfdeclarelayer{background}
\pgfsetlayers{background,main}

\newcommand{\Ts}{T_\text{s}}
\newcommand{\npdf}[3]{\mathcal{N}\! \left(#1 \, \left| \, #2, #3 \right. \right)}
\newcommand{\condProb}[2]{p\! \left(#1  \, \left| \, #2\right. \right)}
\newcommand{\bE}[2]{\mathbb{E}_{#1}\left[#2\right]}
\newcommand{\Qfun}[2]{\mathcal{Q}\! \left(#1 , #2 \right)}
\newcommand{\gfun}[1]{\gamma\! \left(#1 \right)}
\newcommand{\zk}[1]{z_{#1}(k)}
\newcommand{\zkm}[1]{z_{#1}(k-1)}

\newcommand{\xg}{x_{\text{g}}}
\newcommand{\sg}{s_{\text{g}}}
\newcommand{\xh}{x_{\text{h}}}
\newcommand{\yh}{y_{\text{h}}}

\newcommand{\xe}{x_{\text{e}}}

\newcommand{\xhi}{x_{\text{h},i}}

\newcommand{\bR}{\mathbb{R}}

\newcommand{\rsc}{r_\text{sclera}}
\newcommand{\rrt}{r_\text{retina}}

\newcommand{\dt}{\; d \text{t}}
\newcommand{\diris}{d_\text{iris}}
\newcommand{\niris}{n_\text{iris}}
\newcommand{\nirisx}{n_{\text{iris},x}}
\newcommand{\nirisy}{n_{\text{iris},y}}
\newcommand{\nirisz}{n_{\text{iris},z}}

\usepackage{enumitem}
\renewcommand{\theenumi}{\Alph{enumi}}
\usepackage{algorithmic}
\usepackage{algorithm}
\usepackage{array}
\usepackage[caption=false,font=normalsize,labelfont=sf,textfont=sf]{subfig}

\usepackage{fontawesome5}
\usepackage{placeins}
\usepackage{siunitx}
\usepackage{multirow}
\newcommand*\circled[1]{\tikz[baseline=(char.base)]{
		\node[shape=circle,draw,inner sep=1pt] (char) {#1};}}
\usepackage{acronym}
\usepackage{todonotes}
\let\oldquote\quote
\let\endoldquote\endquote
\renewenvironment{quote}[2][]
{\if\relax\detokenize{#1}\relax
	\def\quoteauthor{#2}%
	\else
	\def\quoteauthor{#2~---~#1}%
	\fi
	\oldquote}
{\par\nobreak\smallskip\hfill(\quoteauthor)%
	\endoldquote\addvspace{\bigskipamount}}
	
\newcommand\blfootnote[1]{%
\begingroup
\renewcommand\thefootnote{}\footnote{#1}%
\addtocounter{footnote}{-1}%
\endgroup
}

\begin{document}
\bstctlcite{IEEEexample:BSTcontrol}
\setlength{\parindent}{0pt}
\setlength{\parskip}{0pt} 
\frontmatter
\begin{titlepage}


\vspace{2cm}
{\centering \huge Towards Energy Efficient Mobile Eye Tracking For AR Glasses Through Optical Sensor Technology}\\ 
\textcolor{gray}{\small{THIS IS A TEMPORARY TITLE PAGE \\ It will be replaced for the final print.}}\\

{\centering \large Dissertation}\\ 

{\centering \large der Mathematisch-Naturwissenschaftlichen Fakultät}\\ 
{\centering \large der Eberhard Karls Universität Tübingen}\\ 
{\centering \large zur Erlangung des Grades eines}\\ 
{\centering \large Doktors der Naturwissenschaft}\\ 
{\centering \large (Dr. rer. nat.)}\\ 
\vspace{2cm}
{\centering \large vorgelegt von}\\ 
{\centering \large M. Eng. Johannes Meyer}\\ 
{\centering \large aus Haselünne, Deutschland}\\ 
\vspace{5cm}

{\centering \large Tübingen}\\ 
{\centering\large 2022}\\ 
    
\vfill

\end{titlepage}

\chapter*{Acknowledgements}
\markboth{Acknowledgements}{Acknowledgements}
\addcontentsline{toc}{chapter}{Acknowledgements}
I was fortunate to write my thesis in an inspiring environment within the µTAS and Optical Microsystems Group of the Department of Advanced Technologies and Microsystems at the Corporate Sector Research and Advance Engineering Division of Robert Bosch GmbH as well as within the Human-Computer Interaction (former Perception engineering) Group of the Department of Computer Science at the University of Tübingen. I am grateful to have been able to conduct my research at this exciting intersection of industry and academia, as I have learned a lot from both worlds over the past three years.\\     

I would like to thank my supervisor Dr. Thomas Schlebusch from the Bosch corporate research department, for his continuous support over the last three years. He always had an open ear for my problems and was open to any kind of discussion. I would like to thank him for his patient and positive attitude, which inspired and motivated me to always strive for the highest goals during the thesis.\\

I would also like to thank Prof. Dr. Enkelejda Kasneci, Head of the Human-Computer Interaction Group within the University of Tübingen, for accepting me as Ph.D. candidate, and guiding me through my doctoral study and integrating me into her Human-Computer Interaction Group.\\

I also want to thank the members of the examination committee Prof. Dr. Andreas Zell, Prof. Dr. Andreas Schilling and Jun.-Prof. Dr. Michael Krone for examining my thesis and serving in the committee. \\

Further, I like to thank the current and past members of the eye-tracking team Dr. Stefan Gering, Dr. Ahmad Mansour, Thomas Buck, and Dr. Andreas Petersen, for sharing their expertise and the fruitful collaboration within the team. In addition, I like to thank Tobias Wilm and Dr. Carsten Reichert for sharing their knowledge within the field of optics. I enjoyed the discussions and the daily exchange with you.  \\

During this thesis, I have been fortunate to have many fruitful collaborations. In particular, I would like to thank Dr. Jochen Hellmig and Dr. Hans Spruit from Trumpf Photonic Components for their collaboration on the gaze gesture topics and John Fischer and Dr. Christian Nitschke from Bosch Sensortec for their collaboration and discussions on the scanned laser topics.\\

Furthermore, I had the opportunity to work with highly motivated bachelor's and master's students during the supervision of their theses. Many thanks to Alexander Zimmer, Adrian Frank, Mario Weckerle, Sarah Dagne, and Michael Mühlbauer for their commitment and dedication to this project. It was a pleasure working with you. \\

I would also like to gratefully acknowledge Dr. André Kretschmann and Dr. Anne Serout from the microsystem technologies department for allowing me to work in their department with extensive freedom on this exciting topic. Thank you, Anne, for proofreading my manuscripts over the last three years. \\

Writing a Ph.D. thesis during a global pandemic was not an easy task at all. Therefore, I also want to thank my former roommates Philip, David, Patrick, Natalie, Laura, and José and my friends Christian, Daniel, Patrick, René, Michael, Maik, Markus, Carmen, Hendrik, Jason, and Marcel, who supported me aside of the work and reminded me of the joy of life. \\

Also, I want to thank Prof. Dr. Gerd von Cölln, who encouraged me already during my master's study to pursue a Ph.D. \\

Further, I am very grateful to Itis for proofreading the thesis. \\

Finally, I want to thank my parents, Karin and Hans-Jörg, and my brothers Daniel and Christian for their understanding and continuous support throughout my life.\\

\bigskip
 
\noindent\textit{Renningen, \today}
\hfill J.~M.


\cleardoublepage
\chapter*{Abstract}
\markboth{Abstract}{Abstract}
\addcontentsline{toc}{chapter}{Abstract} 
After the introduction of smartphones and smartwatches, \ac{ar} glasses are considered the next breakthrough in the field of wearables. While the transition from smartphones to smartwatches was based mainly on established display technologies, the display technology of \ac{ar} glasses presents a technological challenge. Many display technologies, such as retina projectors, are based on continuous adaptive control of the display based on the user's pupil position. Furthermore, head-mounted systems require an adaptation and extension of established interaction concepts to provide the user with an immersive experience. Eye-tracking is a crucial technology to help \ac{ar} glasses achieve a breakthrough through optimized display technology and gaze-based interaction concepts. Available eye-tracking technologies, such as \ac{vog}, do not meet the requirements of \ac{ar} glasses, especially regarding power consumption, robustness, and integrability. To further overcome these limitations and push mobile eye-tracking for \ac{ar} glasses forward, novel laser-based eye-tracking sensor technologies are researched in this thesis. The thesis contributes to a significant scientific advancement towards energy-efficient mobile eye-tracking for \ac{ar} glasses.\\
In the first part of the thesis, novel scanned laser eye-tracking sensor technologies for \ac{ar} glasses with retina projectors as display technology are researched. The goal is to solve the disadvantages of \ac{vog} systems and to enable robust eye-tracking and efficient ambient light and slippage through optimized sensing methods and algorithms.\\
The second part of the thesis researches the use of static \ac{lfi} sensors as low power always-on sensor modality for detection of user interaction by gaze gestures and context recognition through \ac{har} for \ac{ar} glasses. The static \ac{lfi} sensors can measure the distance to the eye and the eye's surface velocity with an outstanding sampling rate. Furthermore, they offer high integrability regardless of the display technology.\\
In the third part of the thesis, a model-based eye-tracking approach is researched based on the static \ac{lfi} sensor technology. The approach leads to eye-tracking with an extremely high sampling rate by fusing multiple \ac{lfi} sensors, which enables methods for display resolution enhancement such as foveated rendering for \ac{ar} glasses and \ac{vr} systems. The scientific contributions of this work lead to a significant advance in the field of mobile eye-tracking for \ac{ar} glasses through the introduction of novel sensor technologies that enable robust eye tracking in uncontrolled environments in particular. Furthermore, the scientific contributions of this work have been published in internationally renowned journals and conferences.

\begin{otherlanguage}{german}
\cleardoublepage
\chapter*{Zusammenfassung}
\markboth{Zusammenfassung}{Zusammenfassung}
\addcontentsline{toc}{chapter}{Zusammenfassung} 
Nach der Einführung von Smartphones und Smartwatches gelten Augmented Reality (AR)-Brillen als der nächste Durchbruch im Bereich der Wearables. Während der Übergang von Smartphones zu Smartwatches weitgehend auf etablierten Displaytechnologien beruhte, stellt die Displaytechnologie von AR-Brillen eine technologische Herausforderung dar. Viele Display-Technologien, wie z. B. Retina-Projektoren, basieren auf einer kontinuierlichen adaptiven Steuerung des Displays in Abhängigkeit der Pupillenposition des Nutzers. Weiterhin erfordern kopfgetragene Systeme eine Anpassung und Erweiterung etablierter Interaktionskonzepte, um dem Nutzer ein immersives Erlebnis zu ermöglichen. In beiden Fällen stellt Eye-Tracking eine Schlüsseltechnologie dar, um AR-Brillen durch optimierte Displaytechnologie und blickbasierte Interaktionskonzepte zum Durchbruch zu verhelfen. Verfügbare Eye-Tracking-Technologien, wie z.B. die Video- Okulographie (VOG), erfüllen die Anforderungen von AR-Brillen insbesondere in Bezug auf Stromverbrauch, Robustheit und Integrierbarkeit nicht. Um diese Einschränkungen zu beheben und mobiles Eye-Tracking für AR-Brillen weiter voranzutreiben, werden in dieser Arbeit neuartige laserbasierte Eye-Tracking Sensortechnologien erforscht. Die Beiträge dieser Arbeit tragen  zu einem bedeutenden wissenschaftlichen Fortschritt in Richtung energieeffizientes mobiles Eye-Tracking für AR-Brillen bei.\\
Im ersten Teil der Arbeit werden neuartige gescannten Laser Eye-Tracking Sensortechnologien für AR-Brillen mit Retina-Projektoren als Displaytechnologie erforscht. Ziel ist es, die Nachteile von VOG-Systemen zu lösen und energieeffizientes, Umgebungslicht- sowie gegenüber Verrutschen der Brille robustes Eye-Tracking durch die Einführung von optimierten Messmethoden und Algorithmen zu ermöglichen.\\
Der zweite Teil der Arbeit erforscht den Einsatz von statischen Laser Feedback Interferometrie (LFI) Sensoren als stromsparende kontinuierlich verfügbare Sensormodalität für die Detektion von Benutzerinteraktion auf Basis von Blickgesten sowie Kontexterkennung durch die Erkennung von menschlichen Aktivitäten für AR Brillen. Die statischen LFI-Sensoren sind in der Lage, den Abstand zum Auge und die Geschwindigkeit der Augenoberfläche mit einer herausragenden Abtastrate zu messen. Weiterhin weisen sie unabhängig von der Displaytechnologie eine hohe Integrierbarkeit auf.\\
Im dritten Teil der Arbeit wird basierend auf der statischen LFI-Sensortechnologie ein modellbasierter Eye-Tracking-Ansatz erforscht. Der Ansatz führt durch die Fusion mehrerer LFI-Sensoren zu einem Eye-Tracking System mit einer äußerst hohen Abtastrate, was Methoden zur Verbesserung der Displayauflösung wie z.B. foveated rendering für AR-Brillen und Virtual Reality (VR) Systeme ermöglicht.\\
Die wissenschaftlichen Beiträge dieser Arbeit tragen durch die Einführung neuartiger Eye-Tracking Sensortechnologien, die insbesondere robustes Eye-Tracking in unkontrollierten Umgebungen ermöglichen, zu einem wesentlichen Fortschritt auf dem Gebiet des mobilen Eye-Trackings für AR Brillen bei. Die wissenschaftlichen Beiträge dieser Arbeit wurden in international renommierten Fachzeitschriften und Konferenzen veröffentlicht.

\end{otherlanguage}


\cleardoublepage
\listoffigures
\addcontentsline{toc}{chapter}{List of Figures} 
\cleardoublepage
\listoftables
\addcontentsline{toc}{chapter}{List of Tables} 
\cleardoublepage
\chapter*{List of Abbreviations}
\addcontentsline{toc}{chapter}{List of Abbreviations} 
\begin{acronym}[LOPOCV]
\acro{ar}[AR]{Augmented Reality}
\acro{adc}[ADC]{Analog Digital Converter}
\acro{am}[AM]{Amplitude Modulation}
\acro{asic}[ASIC]{Application Specific Integrated Circuit}

\acro{cad}[CAD]{Computer Aided Design}
\acro{cie}[CIE]{International Commission on Illumination}
\acro{cmos}[CMOS]{Complementary Metal-Oxid Semiconductor}
\acro{cnn}[CNN]{Convolutional Neural Network}
\acro{cogain}[COGAIN]{Communication by Gaze Interaction}
\acro{cnn}[CNN]{Convolutional Neural Network}

\acro{dac}[DAC]{Digital Analog Converter}
\acro{dc}[DC]{Direct Current}
\acro{dbr}[DBR]{Distributed Bragg Reflector}

\acro{eog}[EOG]{Electro Oculography}
\acro{eel}[EEL]{Edge Emitting Laser}

\acro{fmcw}[FMCW]{Frequency Modulated Continuous Wave}
\acro{fov}[FOV]{Field of View}
\acro{fsm}[FSM]{Finite State Machine}
\acro{fft}[FFT]{Fast Fourier Transform}

\acro{har}[HAR]{Human Activity Recognition}
\acro{hci}[HCI]{Human Computer Interaction}
\acro{hoe}[HOE]{Holographic Optical Element}
\acro{hmd}[HMD]{Head Mounted Device}
\acro{hmm}[HMM]{Hidden Markov Model}

\acro{iec}[IEC]{International Electrotechnical Commission}
\acro{imu}[IMU]{Inertial Measurement Unit}
\acro{ir}[IR]{Infra Red}
\acro{ipd}[IPD]{Intra Pupillary Distance}

\acro{lfi}[LFI]{Laser Feedback Interferometry}
\acro{lidar}[LIDAR]{Light Detection and Ranging}
\acro{led}[LED]{Light Emitting Diode}
\acro{lopocv}[LOPOCV]{Leave one Participant out Cross Validation}

\acro{mems}[MEMS]{Micro-Electro-Mechanical System}

\acro{npu}[NPU]{Neural Processing Unit}

\acro{pog}[POG]{Point of Gaze}
\acro{psog}[PSOG]{Photo Sensor Oculography}
\acro{pccr}[PCCR]{Pupil Center Corneal Reflection}

\acro{rgb}[RGB]{Red Green Blue}
\acro{ransac}[RANSAC]{Random sample consensus}

\acro{snr}[SNR]{Signal to noise ratio}

\acro{tpu}[TPU]{Tensor Processing Unit}
\acro{tia}[TIA]{Trans Impedance Amplifier}

\acro{ui}[UI]{User Interface}

\acro{vr}[VR]{Virtual Reality}
\acro{vog}[VOG]{Video Oculography}
\acro{vac}[VAC]{Vergence Accomodation Conflict}
\acro{vcsel}[VCSEL]{Vertical Cavity Surface Emitting Laser}

\end{acronym}
\setcounter{page}{0}

\cleardoublepage
\pdfbookmark{\contentsname}{toc}
\tableofcontents

%

\setlength{\parskip}{1em}

\mainmatter
\cleardoublepage
\chapter{List of Publications}
The research conducted within this thesis was published in renowned international peer-reviewed conferences (such as ETRA or CHI) and high-impact journals (such as IMWUT or IEEE Sensors) and paves the way for novel energy-efficient and highly integrated sensor solutions for human-computer interaction as well as eye-tracking within \ac{ar} glasses. The full-text publications are included in the appendix of the thesis. Furthermore the technology developed in this thesis lead to granted patents\cite{Meyer_Pat_1, Meyer_Pat_2}.

\subsection*{Accepted Articles}
\begin{enumerate}[label=\Roman*.]
	\item \textbf{Johannes Meyer}, Tobias Wilm, Reinhold Fiess, Thomas Schlebusch, Wilhelm Stork,  Enkelejda Kasneci. "A holographic single-pixel stereo camera eye-tracking sensor for calibration-free eye-tracking in retinal projection AR glasses". In 2022 Symposium on Eye Tracking Research and Applications (2022) \faIcon{trophy} Best Short Paper
	\item \textbf{Johannes Meyer}, Adrian Frank, Thomas Schlebusch,  Enkelejda Kasneci. "U-HAR: A Convolutional Approach to Human Activity Recognition Combining Head and Eye Movements for Context-Aware Smart Glasses". Proc. ACM Hum.-Comput. Interact. 6 (2022)  
	\item \textbf{Johannes Meyer}, Thomas Schlebusch, Enkelejda Kasneci. "A Highly Integrated Ambient Light Robust Eye-Tracking Sensor for Retinal Projection AR Glasses Based on Laser Feedback Interferometry". Proc. ACM Hum.-Comput. Interact. 6 (2022)  
	\item \textbf{Johannes Meyer}, Adrian Frank, Thomas Schlebusch, Enkelejda Kasneci. "A CNN-based Human Activity Recognition System Combining a Laser Feedback Interferometry Eye Movement Sensor and an IMU for Context-aware Smart Glasses". Proc. ACM Interact. Mob. Wearable Ubiquitous Technol. 5 (2021) 
	\item \textbf{Johannes Meyer}, Thomas Schlebusch, Hans Spruit, Jochen Hellmig, Enkelejda Kasneci. "A compact low-power gaze gesture sensor based on laser feedback interferometry for smart glasses". Proc. SPIE 11788, Digital Optical Technologies 2021 (2021) 
	\item \textbf{Johannes Meyer}, Thomas Schlebusch, Hans Spruit, Jochen Hellmig, Enkelejda Kasneci. "A Novel Gaze Gesture Sensor for Smart Glasses Based on Laser Self-Mixing". In Extended Abstracts of the 2021 CHI Conference on Human Factors in Computing Systems (2021) 
	\item \textbf{Johannes Meyer}, Thomas Schlebusch, Wolfgang Fuhl, Enkelejda Kasneci. "A novel camera-free eye tracking sensor for augmented reality based on laser scanning". In IEEE Sensors Journal (2020) \faIcon{trophy} VDE GMM Publication Award 
	\item \textbf{Johannes Meyer}, Thomas Schlebusch, Hans Spruit, Jochen Hellmig, Enkelejda Kasneci. "A novel-eye-tracking sensor for ar glasses based on laser self-mixing showing exceptional robustness against illumination". In ACM Symposium on Eye Tracking Research and Applications (2020) 
	\item \textbf{Johannes Meyer}, Thomas Schlebusch, Thomas Kuebler, Enkelejda Kasneci. "Low Power Scanned Laser Eye Tracking for Retinal Projection AR Glasses". In ACM Symposium on Eye Tracking Research and Applications (2020)  
\end{enumerate}

\subsection*{Submitted Articles}
\begin{enumerate}[label=\Roman*.]
	\item \textbf{Johannes Meyer}, Stefan Gehring,  Enkelejda Kasneci. "Static Laser Feedback Interferometry based Gaze Estimation for Wearable Glasses". Submitted to IEEE Transactions on Systems, Man, and Cybernetics: Systems (2022)   
\end{enumerate}

\subsection*{Granted Patents}
\begin{enumerate}[label=\Roman*.]
	\item Andreas Petersen, Thomas Schlebusch, \textbf{Johannes Meyer}, Hans Spruit, Jochen Hellmig. "Method for detecting a gaze direction of an eye". US 11,435,578 B2,  (2022)   
\end{enumerate}


\section{Scientific Contribution}

This section summarizes this thesis's contributions and most essential results to enable energy-efficient mobile eye-tracking for \ac{ar} glasses. Contributions within this thesis focus mainly on the mobile eye-tracking sensor technology field. They are split into three parts, which are summarized in the following.

\subsubsection*{Scanned Laser Eye Tracking} The main contribution in this part of the thesis is a series of novel methods to a scanned laser eye-tracking sensor system in which 2D images of the eye are captured via a single-pixel camera sensor. The scanned laser beam is directed by an \ac{hoe}, integrated into the glasses lens, to the eye, resulting in a broad coverage of the user's \ac{fov}. The single-pixel sensor used to capture the 2D image is fully integrated into the frame temple, leading to outstanding integrability of the sensor system. With redirection of the laser beam over the \ac{hoe}, the virtual camera observes the eye from a frontal perspective through the glasses lens, which leads to a superior perspective compared to mobile \ac{vog}-systems. To solve the limitations of \ac{vog}-systems in terms of robustness to ambient light and power consumption, the integration of an \ac{lfi}-sensor is proposed as the coherent detection scheme of the \ac{lfi} sensor is immune to ambient light. In addition, the \ac{lfi} sensor detects a bright pupil signal, which eliminates the pupil segmentation stage known from \ac{vog} algorithms, reducing the power consumption due to the algorithm's execution on an embedded platform. The sensor operates robustly regardless of iris color or partially occluded pupils, e.g., by the eyelid or eyelashes, leading to an outstanding pupil detection and thus eye-tracking robustness compared to \ac{vog} systems. Finally, an approach to limit the degradation of gaze accuracy due to glasses slippage known from VOG systems is presented. The \ac{hoe} is spatially multiplexed to generate images from two perspectives, leading to a highly integrated virtual single-pixel stereo camera system. The stereo camera system enables 3D reconstruction of the pupil disc, which significantly reduces the calibration effort of the system and thus improves usability, especially for consumer \ac{ar} glasses, as well as solves the glasses slippage issue, which is mandatory for everyday consumer devices. To summarize contributions within this part of the thesis, novel scanned laser eye-tracking sensor approaches are derived, which solve well-known limitations of \ac{vog} systems concerning sensor integration, power consumption, pupil signal robustness, ambient light robustness as well as slippage robustness. Therefore, this part of the thesis paves the way for integrating eye-tracking sensor technology in consumer-grade retinal projection AR glasses.

\subsubsection*{Static LFI Human Computer Interaction} In this second part of the thesis, static \ac{lfi} sensors are introduced as a novel sensing modality in a near-to-eye setting. The sensors allow measuring the distance between the sensor and the eye and the eye's surface velocity with an outstanding update rate of 1\,kHz by operating the laser sensors in a \ac{fmcw} modulation scheme. Compared to \ac{vog} systems, the static \ac{lfi} sensors require only a fraction of the power consumption. Therefore, they enable novel applications for consumer-grade AR glasses, which always require eye-tracking. As the first application, gaze gesture interaction for hands-free control of the \ac{ui} is presented. Due to the high update rate, a negative latency between gaze gesture execution and classification of the gaze gesture is achieved, allowing rendering systems to react to user inputs before it is finished. The second application presents context awareness for \ac{ar} glasses by recognition of human activities. Within this application, eye- and head movements were fused to recognize a wide range of activities from the physical as well as the cognitive domain. Two data sets are collected during this part of the thesis, one data set with a system containing the \ac{lfi} sensor and an \ac{imu} sensor, and one data set containing a \ac{vog} sensor and an \ac{imu} sensor. The second data set is published as an additional contribution to emphasize research within the area of context-aware \ac{ar} glasses. 

\subsubsection*{Static LFI Eye Tracking} The third and final part of this thesis, six static \ac{lfi}-sensors are combined into a high-speed gaze estimation sensor system. The approach outperforms \ac{vog} systems with its outstanding update rate of 1 \,kHz by magnitudes while consuming less power, showing higher integrability, and being robust to ambient light. Therefore the sensor technology introduced in this part of the thesis enables applications such as foveated rendering or saccadic endpoint prediction for mobile AR and VR devices. Furthermore, since the sensors work independently of the display technology, this approach contributes to mobile eye-tracking in general and thus is applicable for \ac{ar}-glasses with, e.g., waveguide or micro-LED displays as well as \ac{vr}-glasses. The essential contribution besides the sensor system is the sensor fusion algorithm used to estimate the user's gaze from the LFI sensor readings. A hybrid model-based sensor fusion algorithm that combines a geometric model of the eye tailored to the \ac{lfi} sensor modalities with a machine learning approach is proposed. The algorithm is robust to glasses slippage. Furthermore, the system works purely based on distance and velocity information. It thus does not capture privacy-related information such as images from the iris region, thus paving the way for privacy-preserving high-speed gaze estimation for \ac{ar} as well as \ac{vr}.


\cleardoublepage
\cleardoublepage
\chapter{Introduction}
\label{Ch:Introduction}
The launch of the first iPhone in 2010 opened the market for smart wearables, such as smartwatches and smart headphones. After the success of smartphones and smartwatches, the next smart wearables to be considered are smart glasses. The term smart glasses summarizes different types of head-worn intelligent wearables, ranging from simple audio glasses with integrated speakers to \ac{ar} glasses with lightweight, transparent \ac{hmd}. Unlike \ac{vr}, whose display technology has mainly been adopted from smartphones, \ac{ar} glasses require novel display technologies such as retinal projection displays to provide high visual comfort while minimizing the system's weight. The main disadvantage of retinal projection displays is the small eye box, as there is only a single exit pupil from which light must hit the retina to display an image. Therefore mobile eye-tracking is a crucial technology for retinal projection \ac{ar} glasses to steer the exit pupil based on the pupil position in a closed loop.

Part one of the thesis presents novel sensor technology approaches to enable mobile eye-tracking within retinal projection \ac{ar} glasses. To better understand the integration of the presented sensor technology approaches, the building blocks and the system design of a retinal projection display are introduced in \Cref{sec:retinal_projection_ar_glasses}. To further motivate the research towards novel sensor technologies for mobile eye-tracking, the subsequent section, \Cref{sec:Eye_Tracking_for_AR_glasses}, gives an overview of eye-tracking applications for \ac{ar} glasses. The applications and the literature requirements for mobile eye-tracking sensors are derived in \Cref{subsec:Requirements_for_eye_tracking_sensors_in_AR_glasses}. 

In order to place the contributions of this thesis in state-of-the-art, \Cref{METsec:Mobile_eye_tracking_sensor_technologies} gives an overview of mobile eye-tracking sensor technologies with a focus on established \ac{vog} sensor technology and algorithms. In addition to \ac{vog} sensor systems, related scanned laser eye-tracking sensors, and other emerging non-intrusive mobile eye-tracking sensor approaches are introduced. Finally, the limitations of the different sensor technologies are discussed in the section.

Part two and three of this thesis's main contribution introduces the \ac{lfi} sensor technology as an eye-tracking sensor. For this purpose, the sensing principle of the sensor is introduced in \Cref{LFIsec:Laser_feedback_interferometry}  together with a brief description of the building blocks of the sensor system.

\section{Retinal Projection AR Glasses}
\label{sec:retinal_projection_ar_glasses}
\begin{figure}[ht]
	\centering
	\includegraphics[width=0.5\linewidth]{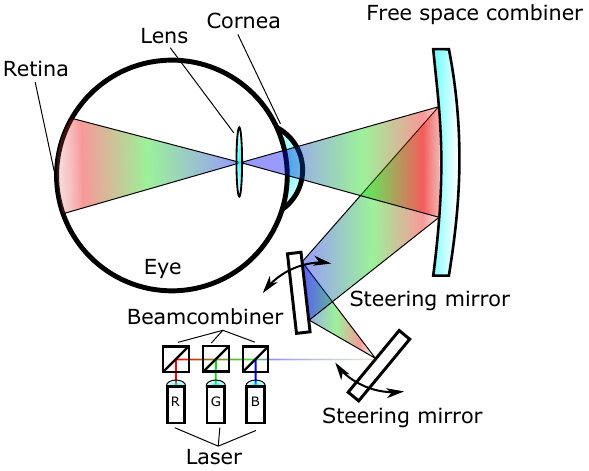}
	\caption{Working principle of a retinal projection \ac{hmd}.  }
	\label{INT:Ret_proj_AR_glasses}
\end{figure}

Although the retinal projection display technology itself was already described by Viirre et al. \cite{viirre1998virtual} back in 1998, the technology was first introduced by Sugawara et al. in 2017 \cite{8006062, 10.1117/12.2295751} in the domain of \ac{ar} glasses. \Cref{INT:Ret_proj_AR_glasses} shows a sketch of a retinal projection \ac{hmd}. 

Retinal projection \acp{hmd} writes an image directly onto the human eye's retina. The image's pixels are created by a set of \ac{rgb} laser diodes, which are combined using a beam combiner and collimated to a tiny light source. The optical power of the laser diodes is modulated in their amplitude at a high frequency to control the color and brightness of the individual pixels. A set of two one-dimensional steering mirrors or a single two-dimensional steering mirror are used to deflect the light source vertically and horizontally to form a two-dimensional image. The two-dimensional image is combined to form a uniform ray bundle, the so-called exit pupil, using a free space combiner. The exit pupil is deflected towards the eye, and after all the beams crossed in the eye's lens, an image is built up at the retina \cite{10.1117/12.2295751}. Current retinal projection systems use \ac{mems} micro mirrors to direct the laser beam and \acp{hoe} as free-space combiners because they provide high visual comfort. In addition, system efficiency is very high as almost all the light from the laser diodes is directed to the human eye, reducing thus power consumption compared to waveguide combiners. The direct projection onto the retina also results in a focus-free image projection with a large \ac{fov}, which is independent of the accommodation state of the eye lens. Despite these advantages, the tiny exit pupil is a significant drawback of retinal projection \acp{hmd}. A small deviation of the eye's position, e.g., due to a slipping of the glasses or a large rotation of the eyeball, leads to a complete disappearance of the projected image, as the laser beams do not enter the pupil anymore. To solve this limitation, the exit pupil must be adjusted according to the current pupil position. This limitation motivates using mobile eye-tracking sensors for \ac{ar} glasses to track the pupil and steer the exit pupil in a closed loop to ensure high visual comfort.
\section{Eye Tracking for AR glasses}
\label{sec:Eye_Tracking_for_AR_glasses}
The term eye-tracking in AR glasses summarizes different approaches to recognizing a user's visual state. These approaches mainly rely on representing the \ac{pog} in a corresponding coordinate system. For example, in the exit pupil control approach, the pupil position is mapped in head-fixed coordinates onto the display coordinate system to control the exit pupil and ensure that the image is projected correctly onto the retina.      


\begin{figure}[h!]
	\centering
	\includegraphics[width=0.6\linewidth]{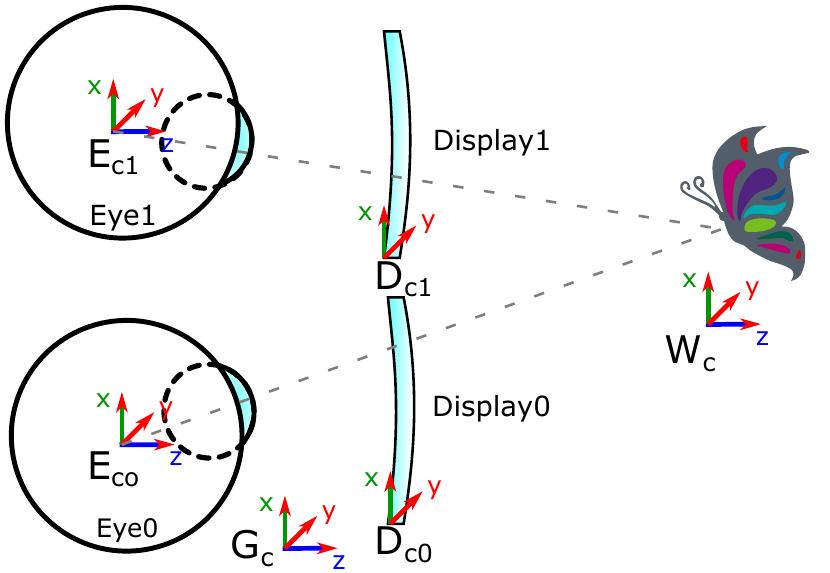}
	\caption[Overview of different Coordinate spaces]{Overview of the different coordinate spaces in a mobile eye-tracking \ac{ar} glasses }
	\label{MET:Coordinate_spaces}
\end{figure}

\Cref{MET:Coordinate_spaces} gives an overview of the different coordinate systems in mobile eye-tracking AR glasses. Each eye has its local coordinate system $E_{c0}, E_{c1}$ typically originated at the eye's rotational center. These coordinate systems are referred to as the head-fixed coordinate system, and the gaze vector per eye is described as a head-fixed gaze vector. The second coordinate system $G_c$ originated in the glasses frame. It is loosely coupled to the eye's coordinate system $E_C$ determined by the positioning of the glasses on the head. The correspondence between the coordinate systems is individual for each user as the \ac{ipd} and the head geometry vary across different users. Therefore dependent on the application, a mapping between the individual coordinate systems is necessary.

\textbf{2D eye tracking} applications  map the \ac{pog} from the corresponding eye coordinate system $E_c$ to the corresponding display coordinate system $D_c$. The \ac{pog} is described by the intersection of the individual gaze vector with the 2D display plane.

\textbf{3D eye tracking} applications map the eye coordinate system of both eyes ($E_{c0}, E_{c1}$) to the glasses coordinate system $G_c$ and thus the \ac{pog} is described by the intersect of both gaze vectors.

\textbf{3D gaze tracking} applications map the \ac{pog} to the world coordinate systems $W_c$ by adding a reference sensor to the glasses linked to the glasses coordinate system $G_c$.


\subsection{Eye Movement Pattern Recognition}

Eye movement pattern recognition applications rely on relative eye movement or position sequences, and therefore, a complete description of the relationship between $G_c$ and $E_c$ is not required. Eye movement pattern recognition is mainly used to derive contextual information of the user by detection of specific eye movement patterns. Most of the applications in this context belong to gaze-based \ac{hci} \cite{majaranta2014eye}, such as gaze-based control of the \ac{ui} of the AR glasses \cite{Drews2007, 8368051, Bace2020}, gaze-based \ac{har} \cite{5444879,10.5555/1735835.1735842, DAR_Scanpath}, expertise identification \cite{10.1145/3204493.3204550, 10.1145/3379155.3391320, kubler2015automated, kasneci2017eye, 10.1371/journal.pone.0251070, hosp2020eye, ReichETRA2022}, or user identification \cite{10.1145/3313831.3376840,JaegerECML2019, lohr2022eye}. In addition, medical features can be derived from a sequence of pupil dilation measurements, which can be used to estimate the user's cognitive load \cite{krejtz2018eye, appel2019predicting, bozkir2019person, appel2018cross, appel2021cross} or to detect mental disorders \cite{leigh2015neurology}. 

\subsection{2D Eye Tracking}
2D eye-tracking applications cast the individual gaze vector to its corresponding display coordinate system and thus describe the \ac{pog} in 2D display coordinates. 2D eye-tracking applications are used to enhance the display quality based on the users \ac{pog} on the display. Applications are mainly divided into approaches to improving the image resolution and applications to increase the displays \ac{fov}. Foveated rendering is a common application enabled by 2D eye-tracking to improve image resolution. Foveated rendering adapts the display resolution based on the \ac{pog} on the display by rendering content with high image quality around the \ac{pog} on the display. This dynamic rendering scheme increases the image quality of AR systems or reduces the displays power consumption while keeping the image resolution \cite{Kaplanyan:2019:DNR:3355089.3356557, Kim:2019:FAD:3306346.3322987, 9005240}. Further 2D eye-tracking applications like pupil duplication \cite{Lin:20} or exit pupil steering \cite{Ratnam:19, Xiong:21}  are used to increase the displays \ac{fov} of  \ac{ar} glasses \cite{Xiong2021}. In pupil duplication systems, the exit pupil is duplicated to cover a wide range of pupil positions to project display content across a large \ac{fov}. To prevent image degradation, e.g., by double images formed by two exit pupils reaching the retina, 2D eye-tracking is required to disable not matching exit pupils. In pupil steering systems, content is projected through a single exit pupil into the eye. This exit pupil is steered, e.g., by a mirror in a way that the exit pupil of the display follows the pupil to provide image content over a wide \ac{fov}. To correctly steer the exit pupil, 2D eye-tracking information is necessary.
\subsection{3D Eye Tracking}
In 3D eye-tracking applications, the gaze vectors of both eyes are calculated and mapped to the glasses coordinate system $G_c$ to estimate the \ac{pog} based on the intersect of both gaze vectors. 3D eye-tracking allows the estimation of the depth of focus of a user, which allows to estimate the depth of field of AR content accurately and thus enables gaze-contingent stereo rendering \cite{10.1145/3414685.3417820}. Furthermore, 3D eye-tracking is used to solve the \ac{vac} of head-mounted displays by estimation of the  vergence \cite{mlot20163d} to allow adaption of the display focus plane e.g., in a multi-focal display \cite{Zhan2020}, or an auto-focus display \cite{8458263}.   
\subsection{3D Gaze Tracking}
In 3D gaze tracking applications, the \ac{pog} is mapped to the world coordinate system $W_c$. The relationship between $G_c$ and $W_c$ is known by an additional visual sensor such as a world camera sensor or \ac{lidar} sensor attached to the glasses. By casting the \ac{pog} to $W_c$ and gathering additional information by the world camera sensor, contextual information such as the object (e.g., the butterfly) the user is fixating on can be derived. The contextual information can be used, i.e., for advertisement \cite{MEINER2019445}. In addition, 3D gaze tracking enables applications like gaze adaptive \ac{ar}. Gaze adaptive augmented reality summarizes approaches to adapt the projected content of AR glasses based on the \ac{pog} and an object in the real world \cite{10.1007/978-3-030-85623-6_32}.

\subsection{Requirements for Mobile Eye Tracking Sensors in AR glasses}
\label{subsec:Requirements_for_eye_tracking_sensors_in_AR_glasses}
As discussed previously, mobile eye-tracking enables a wide range of applications in \ac{ar} glasses and is therefore mobile eye-tracking sensors are a key technology for \ac{ar} glasses \cite{kress2020optical}.

However, the adaption of eye-tracking into \ac{ar} glasses adds new requirements to eye-tracking sensors such as robust operation for a large portion of the population e.g. with different head and face geometries or different iris colors, robust operation under a variety of lightning conditions as well as robust operation during different scenarios while wearing the glasses e.g. while performing physical activities like cycling \cite{Komogortsev21}. 

Derived from the application, which the eye-tracking sensor signal should support, further requirements arise for the sensor update rate and gaze accuracy \cite{Komogortsev21}. In addition, the lightweight system design of \ac{ar} glasses further adds requirements to the sensor integration to be highly integrated into the space constraint glasses frame without interfering with the user's \ac{fov} \cite{kress2020optical}. 

Finally, like other wearable devices, \ac{ar} glasses must operate for a whole day. Therefore a low power consumption of the eye-tracking sensor is mandatory as battery capacities are limited due to space and weight constraints of the glasses \cite{kress2020optical}. These different requirements are addressed in more detail during the next subsections.     


\subsubsection{Gaze Accuracy}
The gaze accuracy of an eye-tracking sensor describes the average angular offset between a fixation location and the corresponding location of the reference target of fixation \cite{Kassner:2014:POS:2638728.2641695}. The required accuracy is given by the application the eye-tracking sensor needs to fulfill e.g. gaze gesture recognition applications have rather low accuracy requirements. In contrast, display enhancement applications like foveated rendering or exit pupil steering require a gaze accuracy of 1$^\circ$ \cite{Kim:2019:FAD:3306346.3322987}.   

\subsubsection{Power Consumption}
AR glasses are battery-powered like other wearables, and the weight of the glasses is directly related to the wearing comfort, which is why the battery capacity is limited. For this reason, the power consumption of the eye-tracking sensor must be as low as possible. Hong et.al. \cite{10.1117/12.2322657} stated that the power consumption of eye-tracking sensors for battery-powered \ac{ar}- and \ac{vr} glasses should not exceed 100\,mW.   
\subsubsection{Update rate}
The update rate describes the number of successive measurements of the eye pose per second. A high update rate above 300\,Hz is required to precisely reconstruct the eye position during fast eye movements \cite{Juhola1985}, which is especially required for display enhancement applications like foveated rendering or exit pupil switching \cite{10.1117/12.2322657}.  
\subsubsection{Robust Operation}
AR glasses are everyday devices where eye-tracking functionality needs to operate robustly for various users and in various environmental conditions \cite{Fuhl2016, holmqvist2022eye}. They need to cover a wide range of users with their eye and face properties and thus be robust to different iris color \cite{hessels2015consequences}, eyelashes covering the eye \cite{schnipke2000trials}, mascara \cite{schnipke2000trials}, partially occluded pupils by the eyelids \cite{schnipke2000trials}, as well as specular reflections on the eye surface caused by contact lenses or the glasses lens. Furthermore, the eye-tracking sensors are exposed to uncontrolled ambient lighting conditions ranging from bright sunlight to darkness \cite{majaranta2014eye}. 
\subsubsection{Sensor Integration}
The eye-tracking sensor must be integrated into the temple of the spectacle frame or the spectacle lens without restricting the user's field of vision and thus not distracting the user. From an integration point of view, the sensor should be fully integrated into the temple so that ideally, no components are integrated into the glasses lens, as components in the glasses lens require more difficult wiring via the glasses hinge. A low number of required components facilitates the integration of the eye-tracking sensor into the glasses, which is particularly advantageous for a high-volume product such as AR glasses.

\section{Mobile Eye Tracking Sensor Technologies}
\label{METsec:Mobile_eye_tracking_sensor_technologies}
To motivate this thesis, an overview of state-of-the-art eye-tracking sensor technology concerning the requirements introduced in \Cref{subsec:Requirements_for_eye_tracking_sensors_in_AR_glasses} is given. Before diving into the details of eye-tracking sensor technology, a brief introduction to anatomical and optical features of the eye is given to ease understanding of the working principle of the different approaches.

\subsection{Anatomical and Optical Features of the Eye}
\label{subsec:Anatomical_optical_features}
\begin{figure}[h!]
	\centering
	\includegraphics[width=\linewidth]{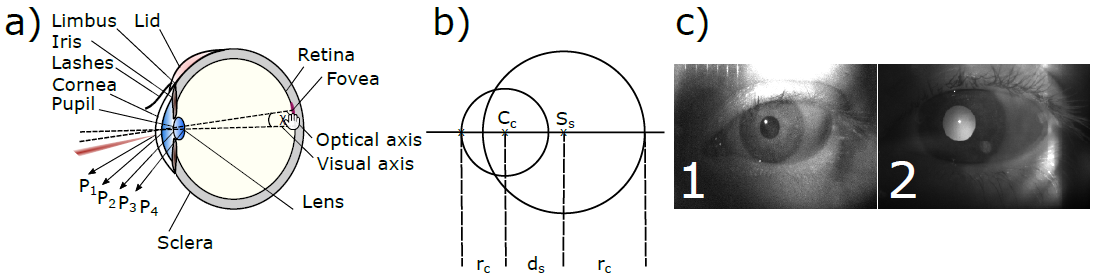}
	\caption[Anatomical and optical features of the eye]{a) Section through the human eye to describe the anatomical structure in a simplified form. b) Emsley's reduced eye model according to \cite{doi:10.1111/cxo.12352} c) Visualizing the bright- and dark pupil effect in the presence of \ac{ir} illumination.  }
	\label{MET:anatomy}
\end{figure}
\Cref{MET:anatomy} a) shows a section through the human eye to describe the anatomy and physiology in a reduced form. The human eye consists of the retina, a light-sensitive layer of tissue that converts incident light into a two-dimensional image, and an optical system that focuses the incident light onto the retina using an adjustable lens. Inside the retina, the fovea is embedded, a region with a high density of cones responsible for sharp central vision. The fovea has an individual angular offset $\chi$ of \textasciitilde 5$^\circ$ between the visual and the optical axis of the eye \cite{gross2008human}.  

 The optical system of the eye consisting of the lens and the pupil, is protected by the transparent cornea. The eye's depth of focus is adjusted by a set of muscles attached to the lens. The iris covers these muscles. In the iris, the pupil is centered. The pupil serves as an aperture of the optical system to control the amount of ambient light entering the optical system. The human eye can operate in a high dynamic range by variation of the pupil diameter between 7.5\,mm in the dark and 2\,mm in bright sunlight. The outer structure of the iris, also referred to as limbus, connects the iris with the white eyeball, the sclera \cite{gross2008human}.  
 
 The normal vector of the pupil plane or the limbus plane directly correlates with the eye's visual axis and thus with the user's gaze. Therefore limbus and pupil features are used in most imaging applications to inferring gaze information from images.
 
 The advantage of the limbus feature is that the diameter does not vary in the presence of varying ambient light and that, especially for dark-colored eyes, the limbus is characterized by a stark contrast to the white sclera, which eases its detection in images. The main drawback of the limbus as a feature of the human eye is that it is often partially occluded by the eyelid or lashes. As the pupil is centered inside the iris, it is less prone to occlusion. Especially with dark eye colors, the contrast between iris and pupil is rather low, which hamper pupil detection in images. To overcome this drawback, the dark- and bright pupil effect is used to increase its contrast. \Cref{MET:anatomy} c) shows images of an eye taken by an \ac{ir} camera sensor while the eye was illuminated with active \ac{ir} illumination. Suppose the illumination source is off-axis w.r.t the camera sensors. In that case, the pupil appears dark (\Cref{MET:anatomy} c) 1) while the pupil appears bright when the axis of the camera sensor and \ac{ir} illumination are aligned (\Cref{MET:anatomy} c) 2) \cite{Duchowski2017}. Besides this effect, the use of active \ac{ir} illumination has several advantages as \ac{ir} illumination is invisible to the human eye and therefore does not influence the pupil diameter while improving pupil detection based on images in low lighting conditions.

Aside from illuminating the eye, \ac{ir} illumination sources are also used to create bright spots on the eye's surface, also known as glints, which are used as a feature in various eye-tracking sensor technologies to infer the user's gaze. To illustrate the occurrence of glints, \Cref{MET:anatomy} a) shows a red laser beam that hits the cornea as well as the lens of the eye. At each boundary surface of the different tissues characterized by a change in refractive index $n$, a portion of the incident light is reflected according to the Fresnel equation, resulting in four reflections ($P_1$-$P_4$), also called Purkinje images. The first reflection $P_1$ is brightest due to the refractive index jump at the cornea and is further referred to as a glint.

Finally, especially model-based eye-tracking methods incorporate the geometrical structure of the eye. A commonly used eye model e.g., by Gustrin et al., is the Emsley's reduced eye shown in \Cref{MET:anatomy} b). The eye model is a first-order approximation based on a sclera sphere and a cornea sphere with its centers $S_c$ and $C_c$ and its corresponding radii $r_s$ and $r_c$. An offset $ d_s$ shifts the centers of the spheres. According to Guestrin et.al. \cite{1634506},  the cornea sphere rotates around the sclera sphere with an offset $d_s$ of 6.1\,mm and the radius of the cornea sphere $r_c$ is 7.8\,mm while the curvature of the sclera sphere $r_s$ is 12\,mm.

 According to \cite{doi:10.1111/cxo.12352} the Emsley's reduced eye, which is shown in \Cref{MET:anatomy} b) is widely used to describe the human eye in a first-order approximation not including optical properties. The eye model is constructed based on two spheres describing the sclera with its center at $S_c$ and the cornea $C_c$. According to Guestrin et.al. \cite{1634506},  the cornea sphere rotates around the sclera sphere with an offset $d_s$ of 6.1\,mm and the radius of the cornea sphere $r_c$ is 7.8\,mm while the curvature of the sclera sphere $r_s$ is 12\,mm.

\subsection{Videooculography Mobile Eye Tracking}
\label{MET:Gaze_estimation}
State of the art in mobile eye-tracking systems are \ac{vog} systems. They rely on IR-sensitive camera sensors attached to the eyeglass frame to capture eye images to improve the image quality and allow operation in low lighting conditions, \ac{vog} systems are equipped with IR \acp{led} to illuminate the eye region. The user's gaze is estimated from the appearance of the eye in the captured images utilizing the anatomical and optical features as introduced in \Cref{subsec:Anatomical_optical_features}. 

In general \ac{vog} systems incorporate a per-user calibration to gather a mapping between the different coordinate systems, capturing camera images to extract features, and a gaze estimation step \cite{8003267}. Feature extraction from images and the gaze estimation step are active areas of research, and thus a brief overview of state-of-the-art methods is given in the upcoming Sections \cite{4770110, 8003267}.

\subsection{2D Regression based Approach}
\label{MET:2D_regression}
A standard method of estimating the gaze is using a 2D regression-based approach is the \ac{pccr} method. In this method, a set of \ac{ir} \acp{led} is used to produce glints on the surface of the cornea as described in \Cref{MET:anatomy} a). Afterward, images are captured from the eye, and the glint centers, as well as the pupil center, are extracted. For each glint, the glint vector between the pupil center and the corresponding glint center is calculated \cite{1634506}. This set of glint vectors is used as input features for a 2D regression approach to map the input features to a gaze vector to 2D display coordinates $D_c$ using a polynomial mapping function and a set of calibration markers within the display coordinate space \cite{1634506}.

\subsection{3D Model based Approach}
\label{MET:3D_model_based_approach}
In model-based approaches a 3D model of the eye (e.g. \Cref{MET:anatomy} b)) is constructed based on the observations of features of the eye. Dependent on the utilized features, either a corneal reflection model or a geometric eye model is used to construct an eye model and derive the gaze vector.

\subsubsection{Corneal Reflection Model}

\begin{figure}[ht]
	\centering
	\includegraphics[width=0.4\linewidth]{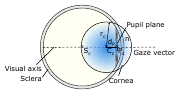}
	\caption[Corneal reflection eye model]{Eye model to calculate the \ac{pog} using corneal reflection methods according to \cite{Hennessey2006}. The model parameters are the radius of the cornea $r_c$, the distance between the pupil plane and the center of the cornea $d_p$ and the refractive index $n$ of the aqueous humor. }
	\label{MET:corneal_reflection_model}
\end{figure}
The corneal reflection model estimates the point of gaze-based on the vector between the cornea center $C_c$ and the center of the pupil $P_c$ based on observations of the pupil center and glints created by \ac{ir} \acp{led} through an IR camera sensor. Under the assumption that the glasses frame is a rigid body, the pose of the \ac{ir} \acp{led} w.r.t. the camera center is fixed. Furthermore, the refractive index $n$ of the cornea as well as the radius of the cornea $r_c$, and the distance between the cornea center and the pupil plane $d_p$ are fixed optical and geometric properties of the human eye, which can be derived e.g. from Emsley's reduced eye model as presented in \Cref{MET:anatomy} b).

Based on the known geometry of the system and the known optical and geometric properties of the human eye,  Hennessey et al. \cite{Hennessey2006} showed that two glints are sufficient to construct an over-determined system of equations to calculate the corneal center $C_c$ in 3D space. To construct the gaze vector, they extract the pupil center $P_c$ as a second reference point from the camera image. To increase robustness of the pupil extraction they apply the bright pupil effect as shown in  \Cref{MET:anatomy} c) by adding an additional on-axis \ac{ir} \ac{led} to their system \cite{Hennessey2006}. 
\subsubsection{Geometric Eye Model}
\label{MET:Geometric_eye_model}
\begin{figure}[ht]
	\centering
\includegraphics[width=.4\linewidth]{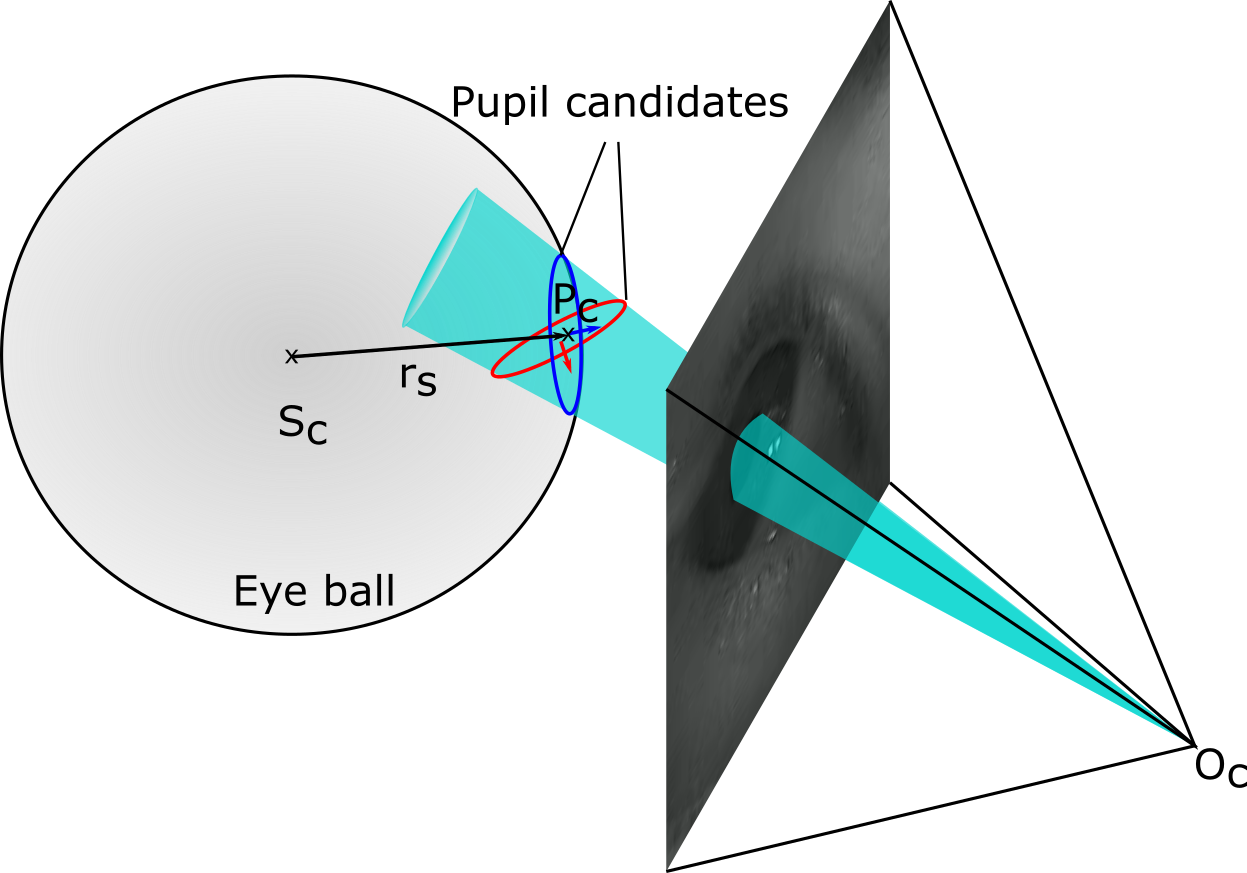}
\caption[Geometric eye model]{Construction of a geometric eye model by  projection of the pupil as cone from the 2D image plane to the 3D space according to \cite{Swirski2013}.}
\label{MET:geometric_eye_model}
\end{figure}

Geometric eye models use a single-camera \ac{vog} system without additional glint \acp{led}. The 3D eye model is derived from a sequence of pupil observations in the 2D image plane. 

Considering a calibrated camera sensor and assuming a pinhole camera model, in a first step the 2D projection of the pupil in the image coordinate space is unprojected to its 3D representation by constructing a cone using the 2D pupil ellipse as a base and the focal point of the pinhole camera model as the vertex, as shown in \Cref{MET:geometric_eye_model}. The cone is propagated towards the eye in the 3D space. Considering a fixed pupil diameter, two possible pupil candidates (red and green) along the cone remain valid solutions, leading to a disambiguity. By adding additional observations from the sequence of images and adding the constraint that the normal vector of the pupil must be directed towards the camera sensor, the disambiguity can be resolved. From a set of observed pupils and their corresponding unprojections, the center of the eye $S_c$ in 3D space is estimated assuming a fixed eyeball radius $r_s$. Based on the eyeball center and the pupil observations, the gaze vector is calculated as the vector from the eyeball center towards the pupil center \cite{Swirski2013}. To further improve the accuracy of geometric eye models, Dierkes et al. \cite{10.1145/3204493.3204525} extend the initial geometric eye model by Swirski et al. \cite{Swirski2013} to handle refraction effects to estimate the pupil contour correctly.



\subsection{Pupil Detection Pipeline}
\label{MET:Subsecpupil_detection_pipeline}
\begin{figure}[ht]
	\centering
	\includegraphics[width=0.6\linewidth]{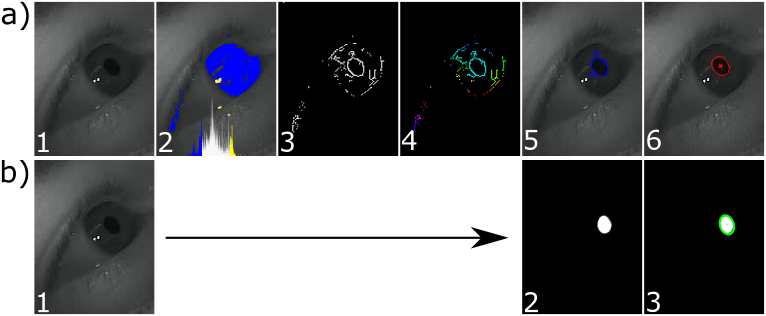}
	\caption[Exemplary pupil detection pipeline for VOG systems]{Exemplary pupil detection pipeline of different \ac{vog} systems. a) shows a pupil detection pipeline comparable to the pupil labs approach using classic computer vision methods \cite{Fuhl2016} while b) shows the pupil detection approach using a CNN model for pupil segmentation as proposed by \cite{YIU2019108307} }
	\label{MET:Pupil_Detection_Pipeline}
\end{figure}

A key challenge for all mentioned gaze vector estimation approaches is a robust extraction of pupil features from camera images, either the center of the pupil, e.g., for 2D regression approaches, or the entire pupil contour. Thus the pupil ellipse, e.g., for the geometric eye model approach. \Cref{MET:Pupil_Detection_Pipeline} a) visualizes the different steps of the pupil detection pipeline using classic computer vision methods while \Cref{MET:Pupil_Detection_Pipeline} b) visualizes the pupil detection pipeline of deep learning-based pupil detection approaches. 

Classic computer vision-based pupil detection methods perform several steps to extract the pupil features. In the first step, a coarse pupil region is estimated, and the captured image is cropped to the pupil region to reduce distortions, e.g., by eyelashes \cite{10.1145/2168556.2168585, Kassner:2014:POS:2638728.2641695}. Afterward, the pupil candidate pixels are estimated, e.g., by using histogram-based methods. The pupil candidate pixels are either the dark pixels if dark pupil tracking is used or the bright pixels if the bright pupil effect is exploited to increase the contrast between pupil and iris as described in \Cref{subsec:Anatomical_optical_features}. In the next step, possible pupil edges are extracted from an image, e.g., by using the canny edge detector as shown in \Cref{MET:Pupil_Detection_Pipeline} a) 3 \cite{Fuhl2016}. Next, the detected edges are converted into pupil contours using, for example, connected component analysis. Then, the pupil contour candidates are filtered based on a set of extracted features such as straightness, the intensity value, or elliptical features such as ellipse aspect ratio or ellipse outline contrast of the pupil contour candidates \cite{fuhl2016else, santini2018pure,santini2018purest, fuhl2018bore}. Finally, based on this extracted pupil, the most likely pupil contour is selected as shown in \Cref{MET:Pupil_Detection_Pipeline} a) 5. Next, the candidate pupil pixels are used to estimate the pupil ellipse, using, e.g., the \ac{ransac} algorithm, and calculate the pupil center as shown in \Cref{MET:Pupil_Detection_Pipeline} a) 6.

To get rid of handcrafted features e.g., for pupil candidate ellipse selection, and thus improve the robustness of extracting pupil features from images, most recent pupil detection approaches apply machine-learning methods \cite{fuhl2016pupilnet, fuhl2017pupilnet, fuhl2019applicability, YIU2019108307, chaudhary2019ritnet, kothari2021ellseg}. \Cref{MET:Pupil_Detection_Pipeline} b) shows the pupil detection pipeline of the \textit{DeepVOG} method \cite{YIU2019108307}. Most of the classical pupil detection steps are replaced by a \ac{cnn} model to segment the pupil and robustly estimate the pupil contour afterward. The segmented pupil contour is used as an input to a geometric eye model, e.g., by Swirski et al.. Kim et al. \cite{10.1145/3290605.3300780} replace the geometric eye model and present \textit{NVGaze}, an end-to-end gaze estimation approach including gaze vector regression.   Most recently, Harsimran et al. \cite{10.1145/3530797} proposed to further incorporate learned eye model parameters to improve the accuracy of end-to-end gaze estimation solutions.     


\subsection{Technological Challenges of VOG Systems for AR Glasses}
\label{MET:technological_challenges_VOG}
\begin{figure}[ht]
	\centering
	\includegraphics[width=0.8\linewidth]{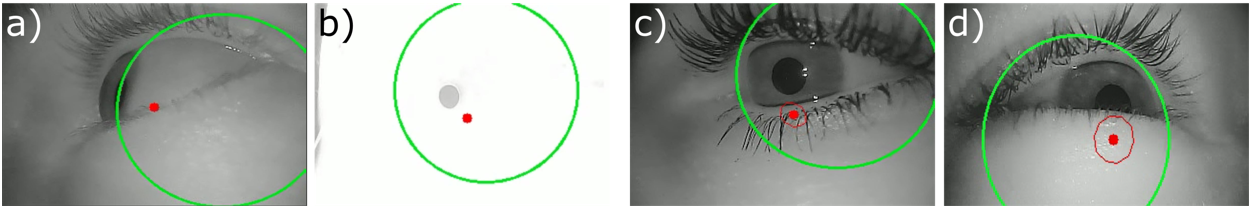}
	\caption[Visualization of technological challenges of VOG systems]{Visualization of technological challenges of \ac{vog} systems for mobile eye-tracking. a) Limited \ac{fov} coverage due to highly off-axis camera integration. b) Loss of pupil detection due to ambient light and limited dynamic range of camera sensors c) False pupil contour edge detection caused by mascara d) False pupil detection due to partial pupil occlusion by the lower eyelid. The images were taken with a Pupil Core eye tracker.  }
	\label{MET:Challenges_VOG_Systems}
\end{figure}
\Cref{MET:Challenges_VOG_Systems} highlights some of the technological challenges of \ac{vog} systems for use in unconstrained mobile eye-tracking scenarios, which arise from the sensing technology (\ac{cmos} image sensors) and the pupil detection. The following discusses the main technological challenges of \ac{vog} systems and possible solutions.

\subsubsection{Power Consumption}
A state of the art \ac{ir} \ac{cmos} image sensor (Python300 series \cite{OnSemi2018}) with a in \ac{vog} systems commonly used image resolution of 0.3\,MP, consumes $\approx$ 90\,mW at a frame rate of 120\,Hz. As \ac{cmos} image sensors mainly consist of digital logic, the power consumption scales linearly with the frame rate. It thus will further increase as according to Juhola et al. \cite{Juhola1985} for detecting saccadic eye movements, at least 300\,fps are required. To reduce the sensor's power consumption, Mayberry et al. \cite{10.1145/2594368.2594388} proposed to capture only a subset of pixels selected by a neural network and infer gaze information from the particular subset.   

 Additional power consumption is added by the pupil detection algorithm as well as the gaze estimation algorithm to \ac{vog} systems as these algorithms need to be executed on an embedded system on the glasses. Furthermore, to achieve a robust pupil detection and thus a high gaze estimation accuracy, pupil detection algorithms robust against errors due to mascara or partially occluded pupils, as shown in \Cref{MET:Challenges_VOG_Systems} c) and d), are mandatory. These robustness requirements lead to the need for machine learning-based pupil detection algorithms, as discussed in \Cref{MET:Subsecpupil_detection_pipeline}, which significantly impacts the power consumption of \ac{vog} systems. Approaches to reduce the power consumption added by the algorithms are either in the direction of using optimized hardware like \acp{tpu} or \acp{npu}  to execute the algorithms \cite{Bringmann_CNN_acc, garofalo2020pulp} or by optimization of the pupil detection models as proposed by Fuhl et al. \cite{fuhl2020tiny}.  
\subsubsection{Ambient Light Robustness} 
\Cref{MET:Challenges_VOG_Systems} b) illustrates the effects of bright ambient light, e.g., sunlight during outdoor activities, on \ac{vog} systems. Due to the limited dynamic range of \ac{cmos} image sensors, overexposure makes pupil detection and thus gaze estimation impossible \cite{Fuhl2016} and has been considered as a limiting factor in many outdoors studies \cite{10.1371/journal.pone.0087470, fuhl2016evaluation, santini2018art}. According to Geissler et al. \cite{8460800} this effect also appears for glint-based methods as the glints are not reliably detected in bright sunlight.  

\subsubsection{Camera Sensor Integration} 

A highly off-axis integration of the camera sensor into the eyeglass frame restricts the \ac{fov} in which the pupil can be reliably detected. Therefore, gaze estimation deteriorates if the camera sensor and the eye's optical axis are far apart, as shown in \Cref{MET:Challenges_VOG_Systems} a). Narcizo et al. \cite{vision5030041} shows that the highest gaze estimation accuracy of \ac{vog} systems is achieved if the camera sensor is centered in front of the eye. However, this camera sensor position infers with the user's \ac {fov} and therefore leads to user distraction and reduces the comfort of \ac{vog} systems. To overcome user distraction by camera integration and still cover a wide \ac{fov}, Tonsen et al. \cite{10.1145/3130971} propose using several low-resolution camera sensors arranged around the lens frame and fuse information of all cameras for gaze estimation.

A second aspect to consider is the size of the camera sensors and imaging optics in front of the image sensors, which limit sensor integration options and increase the weight of \ac{ar} glasses, which also results in reduced user comfort \cite{koulieris2019near}.  
\subsubsection{Glasses slippage} 
\label{MET:Glasses slippage}

Especially during physical activities, glasses tend to slip on the user's head, resulting in a change in the relationship between the head-fixed coordinate system and the glasses coordinate system and thus the camera coordinate system. Since the relationship between the head-fixed coordinate system and the glasses coordinate system in 2D regression-based methods is learned during initial calibration, as described in \Cref{MET:2D_regression}, the 2D regression-based methods are prone to glasses slippage as the learned relationship between the coordinate systems mismatches after slippage. Thus the gaze estimation is erroneous \cite{10.1145/3314111.3319835}, \cite{niehorster2020impact}.

Geometric model-based approaches, as described in \Cref{MET:Geometric_eye_model}, derive the relationship between the head-fixed coordinate system and the glasses coordinate system from a set of observations of the pupil ellipse. Therefore they are, to some extent, vulnerable to slippage as the coordinate system mismatch needs to be detected, and a new mapping needs to be learned from new observations \cite{niehorster2020impact}.  

To approach the degradation of gaze accuracy due to glasses slippage, several methods have been applied to estimate the relationship between the coordinate systems, e.g., by reconstruction from landmarks like eye corners \cite{Fabricio2017} or the eyelids \cite{6595992}. More recently, Santini et al. \cite{10.1145/3314111.3319835} proposed the \textit{Get a Grip} method to extract slippage robust features for geometric model-based gaze estimation approaches to minimize the impact of slippage after calibration. 

Other solutions to improve this technological challenge are proposed by Kohlbecher et al. \cite{10.1145/1344471.1344506}, and Tobii \cite{TobiiPro2021}, which address the problem by adding additional camera sensors to construct a stereo camera setup to reconstruct the pupil in 3D space.
\subsubsection{Update Rate}
The update rate or frame rate of \ac{vog} system is limited by the trade-off between image resolution, system bandwidth, and power consumption of the camera sensor and the pupil detection pipeline. Therefore update rates above 120\,Hz is rare for highly integrated mobile eye-tracking sensors. Limitations in update rate further have a negative influence on the system latency.
\subsubsection{Calibration}
Depending on the chosen gaze estimation method and the hardware equipment of a \ac{vog} system, a more or less complex calibration procedure is required to achieve a high gaze accuracy of the system. 2D regression-based methods require one calibration per session, limiting the user experience and adoption of \ac{vog} systems for \ac{ar} glasses \cite{MORIMOTO20054}. For model-based approaches, either glint-based models or geometric models, the calibration is reduced to one calibration per user since only the offset between the visual and the optical axis of the user's eye has to be calibrated once \cite{Calibme}.

Calibration-free approaches e.g., as proposed by Tonsen et al. \cite{tonsen2020high},  avoid calibrating the system at the cost of lower gaze accuracy.

\subsection{Scanned Laser Mobile Eye Tracking Technologies}
\label{MET:Scanned_Laser_ET_Approaches}
\begin{figure}[ht]
	\centering
	\includegraphics[width=0.8\linewidth]{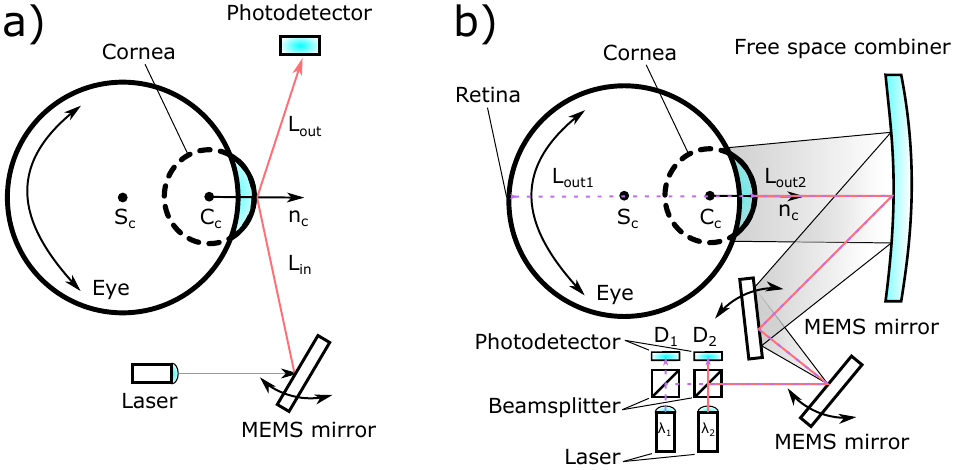}
	\caption[System concepts of scanned laser mobile eye-tracking approaches]{System concepts of scanned laser mobile eye-tracking approaches. a) Corneal reflection tracking method with an off-axis alignment of laser and photo diode b) Retinal reflection tracking method with an on-axis alignment of laser sources and photo diodes  }
	\label{MET:Scanned_Laser}
\end{figure}
Due to the technological challenges of \ac{vog} systems as summarized in \Cref{MET:technological_challenges_VOG}, particularly in the areas of power consumption and update rate, laser eye-tracking approaches have emerged in recent years. These methods are characterized by using a scanning unit, e.g., a set of \ac{mems} micro mirrors to deflect a laser beam towards the eye and a photodetector to detect reflected \ac{ir} light to estimate the gaze vector. In general, the approaches can be divided into corneal reflection tracking methods and retinal tracking methods. \Cref{MET:Scanned_Laser} visualize the system design of both approaches.    
\subsubsection{Corneal Reflection Tracking}
Corneal reflection tracking methods utilize the glint feature similar to \ac{vog} systems. Instead of a divergent light source and a 2D array of photodetectors (e.g. a \ac{cmos} sensor), they scan a focused laser beam in a 2D pattern over the surface of the eye and use a single photodetector to detect the corneal reflection, which occurs for a mirror poistion \cite{7181058}. \Cref{MET:Scanned_Laser} a) shows a corneal reflection tracking method as proposed by Sarkar et al. \cite{7181058}. They integrated a resonant 2D \ac{mems} micro mirror into the glasses frame to scan an \ac{ir} laser beam with a scan frequency of 5\,kHz over the cornea and track the corneal reflection with a photodiode integrated into the nose pad. Suppose the photodiode detects a peak due to a corneal reflection. In that case, the corresponding mirror deflection angles are captured and used as input features for a regression-based approach to estimate the gaze vector \cite{7863402}. As regression-based methods tend to be vulnerable to slippage, as discussed in \Cref{MET:Glasses slippage}, Sarkar et al. added three additional photodiodes to their system in more recent work reconstructed the position and orientation of the cornea from four corneal reflections. This approach led to an accuracy of $\approx$ 1 $^\circ$ at an output rate of 500\,Hz \cite{Mindlink2021}. Instead of additional photodiodes, Holmqvist et al. \cite{10.1145/3379155.3391330} added a stereo camera to correct for glasses slippage. This hybrid eye tracker combines the high update rate of a corneal reflection \ac{mems}-based tracker and the absolute reference of a stereo camera setup.  
\subsubsection{Retinal Reflection Tracking}
Retinal reflection tracking methods utilize the high reflectivity of the retina under \ac{ir} illumination, which also causes the bright pupil effect, as mentioned in \Cref{MET:anatomy}, to construct a feature that somehow corresponds to the pupil center. To detect this feature, a collimated \ac{ir} laser beam is scanned in a 2D pattern over the eye's surface. If the laser beam hits the pupil, it enters the eye and gets backscattered from the retina, leading to a strong on-axis backscattering of \ac{ir} light $L_{out1}$. This is detected by a photodetector, which is integrated into the \ac{ir} laser beam path \cite{EyeWay2021}. \Cref{MET:Scanned_Laser} b) shows a retinal tracking method proposed by Greenberg et al. \cite{EyeWay2021} according to \cite{EyewayPatent2021}. They used a set of two lasers with different wavelengths $\lambda_1$ and $\lambda_2$ to simultaneously track the bright pupil response from retinal center $L_{out1}$, corresponding to the pupil center, and the glint on the vertex point of the cornea $L_{out2}$. To detect the two reflections, two photodiodes $D_1$ and $D_2$ are integrated into the system on-axis w.r.t. to the laser beam via a beamsplitter. Under the assumption of a known and fixed system geometry, the gaze vector can be estimated from the detected pupil center and the vertex point \cite{EyeWay2021}.    
\subsubsection{Technological Challenges}
While scanned laser eye-tracking approaches solve issues of \ac{vog} systems regarding the update rate and reduce the power consumption, technological challenges remain mainly in sensor integration. For example, the final approach of Sarkar et al. \cite{Mindlink2021} requires extensive wiring for the four photodiodes across the glasses hinge while the approach of Greenberg et al. \cite{EyeWay2021} adds complexity due to the high amount of required optical components. In addition, ambient light could lead to an accuracy degradation of the system, especially for cornea reflection tracking methods, since the photodiodes are mounted outside the temple and are directly exposed to ambient light. However, according to Sarkar et al. and Greenberg et al., both methods are on par with \ac{vog} systems in terms of slippage robustness.

\subsection{Other Mobile Eye Tracking Technologies}
Aside from \ac{vog} systems and scanned laser eye-tracking systems described in \Cref{MET:Gaze_estimation} and \Cref{MET:Scanned_Laser_ET_Approaches} respectively, several other eye-tracking sensor technologies have been introduced in the past. The following \ac{psog} and neuromorphic camera-based eye-tracking methods are introduced as they fit to some extent the requirements of AR glasses as mentioned in \Cref{subsec:Requirements_for_eye_tracking_sensors_in_AR_glasses}. Other mainly intrusive approaches such as the sclera search coil technology \cite{4322822} or \ac{eog} \cite{10.1145/1409635.1409647} are not described here as they are not applicable for unobtrusive mobile eye-tracking.  

\subsubsection{Photosensor Oculography}
\ac{psog} systems, initially introduced by Torok et al. \cite{doi:10.1177/000348945106000402}, use a set of photodetectors and IR illumination sources to estimate gaze by examining the scattering behavior and reflectivity of ocular tissues,  particularly of the sclera and iris. The IR illumination sources are arranged around the glasses lens to illuminate different parts of the ocular tissue. Similarly, photodetectors are arranged around the glasses lens to receive the backscattered light of the corresponding illuminated part. The measured reflectivity is used as an input feature for 2D regression-based eye-tracking methods \cite{8307473}. Since this technology is based on 2D regression methods, it has similar weaknesses in terms of glasses slippage as the 2D regression methods of \ac{vog} systems. Rigas et.al. \cite{8067485} propose combining a camera sensor at a low sampling rate with a \ac{psog} system to address the limitation of glasses slippage. Katrychuk et al. address the limitations of the \ac{psog} system concerning glasses slippage by simulating the \ac{psog} outputs of a set of 15 pairs of \ac{ir} illumination sources and corresponding photodetectors aligned in a 2D array over the entire eye region. They used the simulated data combined with machine learning methods to correct for glasses slippage after an initial system calibration \cite{Katrychuk:2019:PSE:3314111.3319821}.

While \ac{psog} systems provide a high update rate and low power consumption due to their simplicity, the sensor integration into an \ac{ar} glasses system is a technical challenge. In addition, the \ac{psog} systems are sensitive to ambient light because the reflected light detected by the photodetectors, and thus the detected ambient light, is directly related to gaze estimation. Finally, a \ac{psog} system requires a per session calibration because a 2D regression-based method is used to estimate gaze, which also induces a reduction of gaze estimation accuracy due to glasses slippage. 
\subsubsection{Neuromorphic Camera}
Neuromorphic cameras, also called event-based cameras, are a new bio-inspired camera sensor technology that mimics the behavior of the human retina to overcome the limitations of classic \ac{cmos} camera sensors. Instead of capturing images with a fixed global shutter, pixels in neuromorphic camera sensors act independently to brightness changes and send out digital events. As the events operate asynchronously, event cameras reach equivalent update rates above 10\,kHz while consuming only a fraction of power (about 10\,mW on die level) compared to \ac{cmos} sensors. Furthermore the dynamic range of neuromorphic camera sensors is very high (above 120\,dB) compared to \ac{cmos} sensors (~60\,dB) \cite{9138762}.

From a sensor point of view, event cameras overcome drawbacks with respect to power consumption, update rate, and ambient light robustness of classic \ac{vog} systems as discussed in \Cref{MET:technological_challenges_VOG}. Angelopoulos et al. \cite{9389490} were the first to adopt this technology to the eye-tracking domain. They used a classic camera sensor to initialize a geometric eye model consisting of a parametric ellipse, a parabola, and a circle model for the pupil, eyelid, and a single corneal glint. Afterward, the events of the event camera are used to interpolate the geometric model with an update rate of 10\,kHz between successive full camera frames. Finally, a 2D regression-based method is proposed to estimate the gaze vector with the pupil and the glint model features as input features. Stoffregen et al. \cite{Stoffregen_2022_WACV} advances the approach of Angelopoulos et al. by removing the classic camera sensor and relying solely on an event camera sensor combined with a set of \ac{ir} \acp{led} to produce a circle of glints on the cornea. Toggling the \ac{ir} \acp{led} with a predefined pattern leads to an event pattern in the neuromorphic camera sensor. These events correspond to corneal glints, which can be tracked over time to estimate the center $c_c$ as well as the orientation of the cornea in space using a regression-based approach. 

While neuromorphic cameras solve some limitations of classic camera sensors used in most \ac{vog} systems, especially in the area of power consumption and update rate, algorithm approaches to robustly estimate the gaze vector from a set of events are still not fully solved. Additionally, the same technological challenges from \ac{vog} systems arise with respect to sensor integration as discussed in \Cref{MET:technological_challenges_VOG}.

\subsection{Summary on Eye Tracking Sensor Technologies}
\label{MET:subsec:Comparison}
 The different eye-tracking sensor technologies are rated according to the requirements for mobile eye-tracking sensors for AR glasses as summarized in \Cref{subsec:Requirements_for_eye_tracking_sensors_in_AR_glasses} to summarize the different technology approaches introduced within this part of the thesis.

Since most of the works lack an accurate description of the system-level power consumption, as the power consumption of the gaze estimation algorithms is usually not considered, power consumption is compared on sensor level including all components required to track a single eye. To allow a fair comparison between the setups, the power consumption is set similar to \cite{Stoffregen_2022_WACV} of IR \acp{led} to 5\,mW and the power consumption of photodiodes including the TIA to 15\,mW, which is reasonable for state of the art integrated circuit solution like the THS4567 \cite{THS4567}.    

The sensor integration requirement is evaluated based on the number of components that need to be integrated into the glasses lens as they restrict the user's field of view and require more complex wiring than components integrated into the frame temple. In addition, the robustness against ambient light is compared by the dynamic range of the receiver circuit (single photodiode circuit or \ac{cmos} sensor).

As a robust operation in the presence of glasses slippage is mandatory for AR glasses, only approaches that tend to be robust against glasses slippage are taken into account in the following comparison.
\subsubsection{VOG system}
The latest TobiiPro Glasses 3 is chosen as a state-of-the-art reference to compare other \ac{vog} systems. The eye-tracking hardware consists of a stereo camera setup and eight \ac{ir} \acp{led}, which are integrated into the glasses lens. The setup allows slippage robust gaze estimation with up to 100\,Hz and a gaze accuracy of 0.6$^\circ$ \cite{TobiiPro2021}. 

As no information regarding the power consumption of the camera sensors is available, the power consumption is estimated from the Python 300 \ac{cmos} camera sensor \cite{OnSemi2018} by scaling the dynamic power linearly according to the frame rate. In addition, we scale down the dynamic power consumption linearly by the pixel difference to match the sensor size of the cameras used by the Tobii 3 glasses. For a single camera sensor, this leads to 
\begin{equation}
\begin{split}
	P_{Camera} &= P_{static} + P_{dynamic} \cdot f \cdot pixel_{diff} \\ 
	&= 50\,mW + 0.3312\frac{mW}{Hz} * 100 Hz * \frac{240\times 960}{640 \times 480} = 74.85\,mW.
\end{split}
\end{equation}
As two cameras are required per eye and in addition 8 \acp{led} are used to illuminate the eye and generate glints, in total  10 components are integrated into the glasses lens.This sums up to a total power consumption of 189.69\,mW, not including the gaze tracking algorithm. The ambient light robustness of \ac{vog} systems is derived from the Python 300 \ac{cmos} sensor which states a dynamic range of 60\,dB \cite{OnSemi2018}.

\subsubsection{Scanned Laser System}
The Adhawk Mindlink \cite{Mindlink2021} is chosen as a reference for scanned laser eye-tracking systems because it is the only system that is robust against slippage and achieves product maturity. The hardware consists of a 2D \ac{mems} micro mirror, which is integrated into the glasses nose pad, and four photodiodes integrated into the glasses lens frame. The power consumption is split between the power consumption of the micro mirror (15\,mW \cite{7181058}) and the four photodiodes, including the associated TIA, resulting in estimated total power consumption of 75\,mW, not including the gaze tracking algorithm. Adhawk reports a gaze accuracy below 1$^\circ$ and an update rate of 500\,Hz of their scanned laser eye-tracking system. According to Adhawk \cite{Mindlink2021} robust tracking in the presence of sunlight is possible thus, the ambient light robustness exceeds the ambient light robustness of \ac{vog} systems. This is mainly because scanned laser systems rely on coherent light sources with a small bandwidth variation around the central wavelength, which allow efficient filtering of ambient light on the detector side e.g. by using narrow-band optical filters in front of the photodiode as mentioned by Sarkar et. al. \cite{7181058}.  

\subsubsection{PSOG system}
As a \ac{psog} eye-tracking sensor reference system, the latest work of Katrychuk et al. \cite{Katrychuk:2019:PSE:3314111.3319821} is considered. They achieve a slippage robust design by using an array of 15 photodiodes integrated into the glasses lens and at least one IR LED as an illumination source to illuminate the eye region. This leads to 16 components that consume power on a sensor hardware level of 230\,mW. The authors reported a gaze accuracy of 1.09$^\circ$ with a maximum update rate of 1\,kHz using a \ac{cnn}. The dynamic range of \ac{psog} systems of $\approx$ 30\,dB is derived from the reference \ac{tia} \cite{THS4567} gain bandwidth product at a frequency of 1\,kHz.    

\subsubsection{Neuromorphic Camera}
Recently, Stoffregen et al. \cite{Stoffregen_2022_WACV} presented a neuromorphic camera eye-tracking system. They used ten blinking \ac{ir} \acp{led} integrated into the eyeglass frame to generate corneal glints, leading to a corneal glint event stream. Based on the event stream, the cornea is tracked at an update rate of 1\,kHz. By reconstructing the cornea from the glint events, the authors can perform slippage robust gaze estimation with a total power consumption of the hardware components of 85\,mW. Since the authors only report the pixel error (~0.5\,pixel) and not gaze accuracy directly, this is derived from a comparable work by Angelopoulos et al. \cite{9389490}, which achieved a gaze accuracy of 0.46$^\circ$. According to Gallego et al. \cite{9138762} the dynamic range of the neuromorphic camera technology is 120\,dB. 
\subsubsection{Conclusion} 
\begin{figure}[ht]
	\centering
	\includegraphics[width=0.8\linewidth]{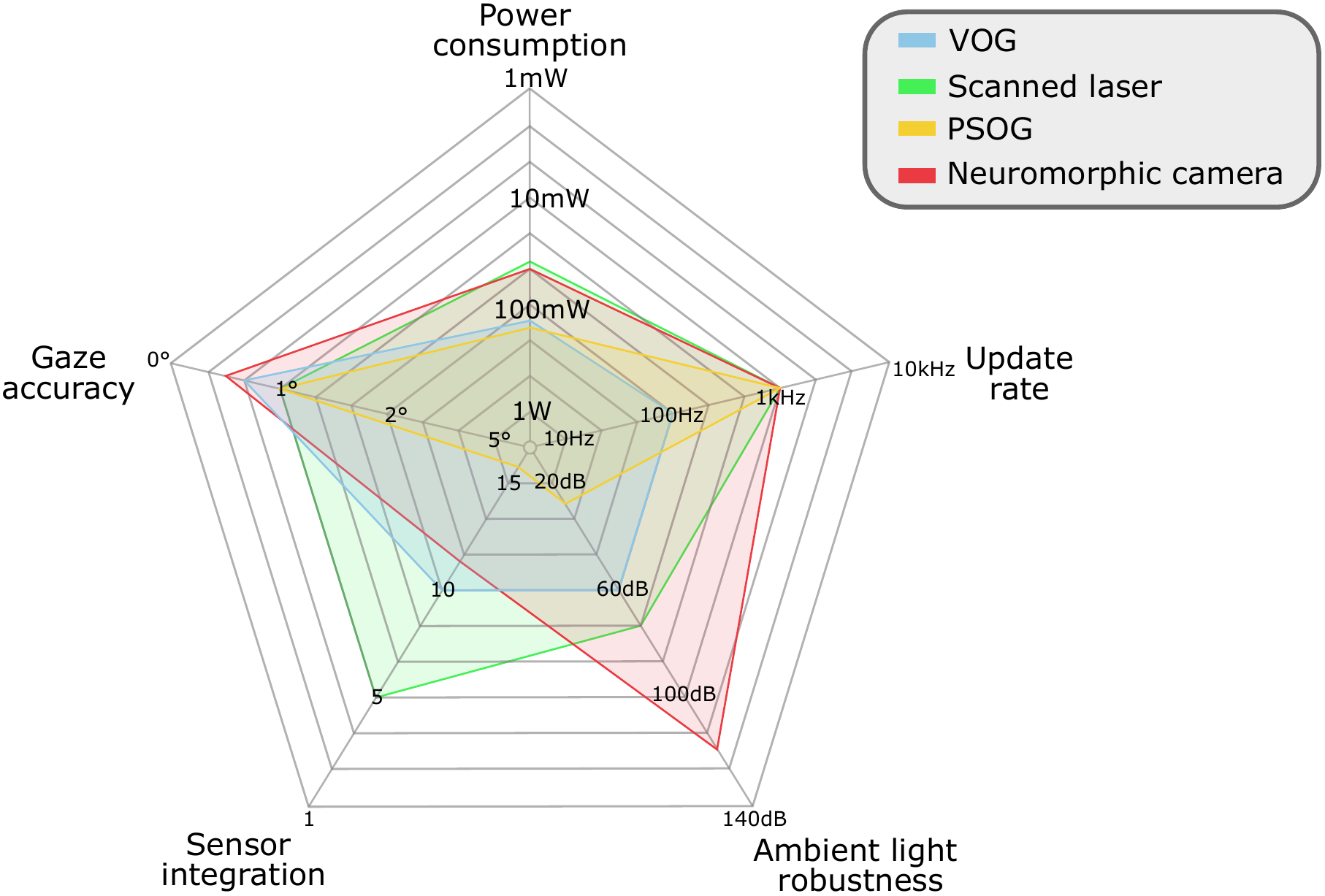}
	\caption{Radar chart to compare different eye-tracking sensor technologies for AR glasses}
	\label{MET:Comparisson}
\end{figure}

The radar chart in \Cref{MET:Comparisson} summarizes the different eye-tracking sensor technologies with respect to the key requirements for mobile eye-tracking sensors for \ac{ar} glasses. A large coverage of the chart corresponds to a better fulfillment of the requirements.
As the power consumption is limited to the hardware power consumption, excluding the power consumption of tracking algorithms, this will increase, especially for systems with a high update rate. To get a sense of the impact of gaze estimation algorithms on system power consumption, the authors of the \ac{psog} system \cite{Katrychuk:2019:PSE:3314111.3319821} estimated the power consumption of their \ac{cnn} model optimized for low power consumption between 70\,mW to 5.75\,W. Therefore this is the most critical requirement as current systems already have a significant power consumption on the hardware level.

The scanned laser technology has the best coverage among all eye-tracking sensor technologies compared in the chart. The technology has advantages with respect to sensor integration and power consumption.

\section{Laser Feedback Interferometry}
\label{LFIsec:Laser_feedback_interferometry}
As AR glasses with retinal projection are already equipped with an integrated laser scanner and scanned laser eye-tracking sensors seem to fit best the requirements for AR glasses, this thesis focuses on laser-based eye-tracking sensors. In particular, the use of \ac{lfi} sensors for eye tracking is investigated. To better understand the contributions in the domain of \ac{lfi} sensors of this thesis, the basic functionality of semiconductor lasers and, in particular, the measurement method of \ac{lfi} will be introduced. The following Sections are mainly based on the ground laying work of Coldren et.al. \cite{Coldren2012} about diode lasers as well as the works of Michalzik et.al. \cite{Michalzik2013} about \acp{vcsel} and Taimre et.al. \cite{Taimre:15} about \ac{lfi}.  

\subsection{Semiconductor Laser}
\label{LFIsubsubsec:Basic_principles_of_semiconductor_lasers}
The main component of semiconductor lasers is the gain medium with a crystalline structure. In this medium, electrons can occupy several energy levels, so-called energy bands. Similar to other semiconductor devices, these structures are divided into a valence band with energy level $E_v$ and a conduction band with energy level $E_c$. If the electrons are not excited, all electrons are in the valence band. However, suppose an external current is applied to the active medium. In that case, electrons get excited and travel to the conduction band, which results in holes in the valence band and free electrons in the conduction band \cite{Coldren2012}. 

\begin{figure}[h!]
	\centering
	\includegraphics[width=0.35 \textwidth]{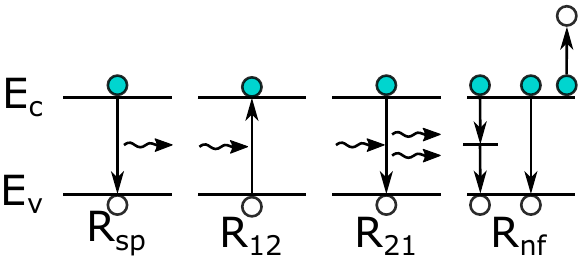}
	\caption[Different electron transitions in semiconductors]{Different electron transitions that are important, emphasizing those that involve the absorption or emission of photons  \cite{Coldren2012}.}
		\label{LFIim:energy_bands}
\end{figure}

\Cref{LFIim:energy_bands} shows different electron transitions between the valence band and the conduction band, which are essential for the absorption or emission of photons. 

The four recombination or generation mechanisms of photons are spontaneous recombination $R_{sp}$, stimulated generation $R_{12}$, stimulated recombination  $R_{21}$ and non-radiative recombination $R_{nf}$. 

$R_{sp}$ describes spontaneous recombination of an electron in the conduction band with a hole (missing electron) in the valence band. They do not contribute to a coherent emission \cite{Coldren2012}. 

$R_{12}$ describes photon absorbance. A photon is absorbed by an electron in the valence band and stimulates the generation of an electron in the conduction band \cite{Coldren2012}.

$R_{21}$ describes a photon that perturbs the system stimulating recombination of an electron and a hole and simultaneously generating a new photon. This state is also called stimulated emission and is the effect that allows the laser to operate and output coherent light \cite{Coldren2012}.

The last effect is non-radiative recombination, which means energy is dissipated as heat in the semiconductor crystal structure \cite{Coldren2012}. These losses are material dependent and can be summarized as internal material loss $\alpha_{i}$ \cite{Coldren2012}.

\subsection{Vertical Cavity Surface Emitting Lasers}

\begin{figure}[h!]
	\centering
	\includegraphics[width=0.5 \textwidth]{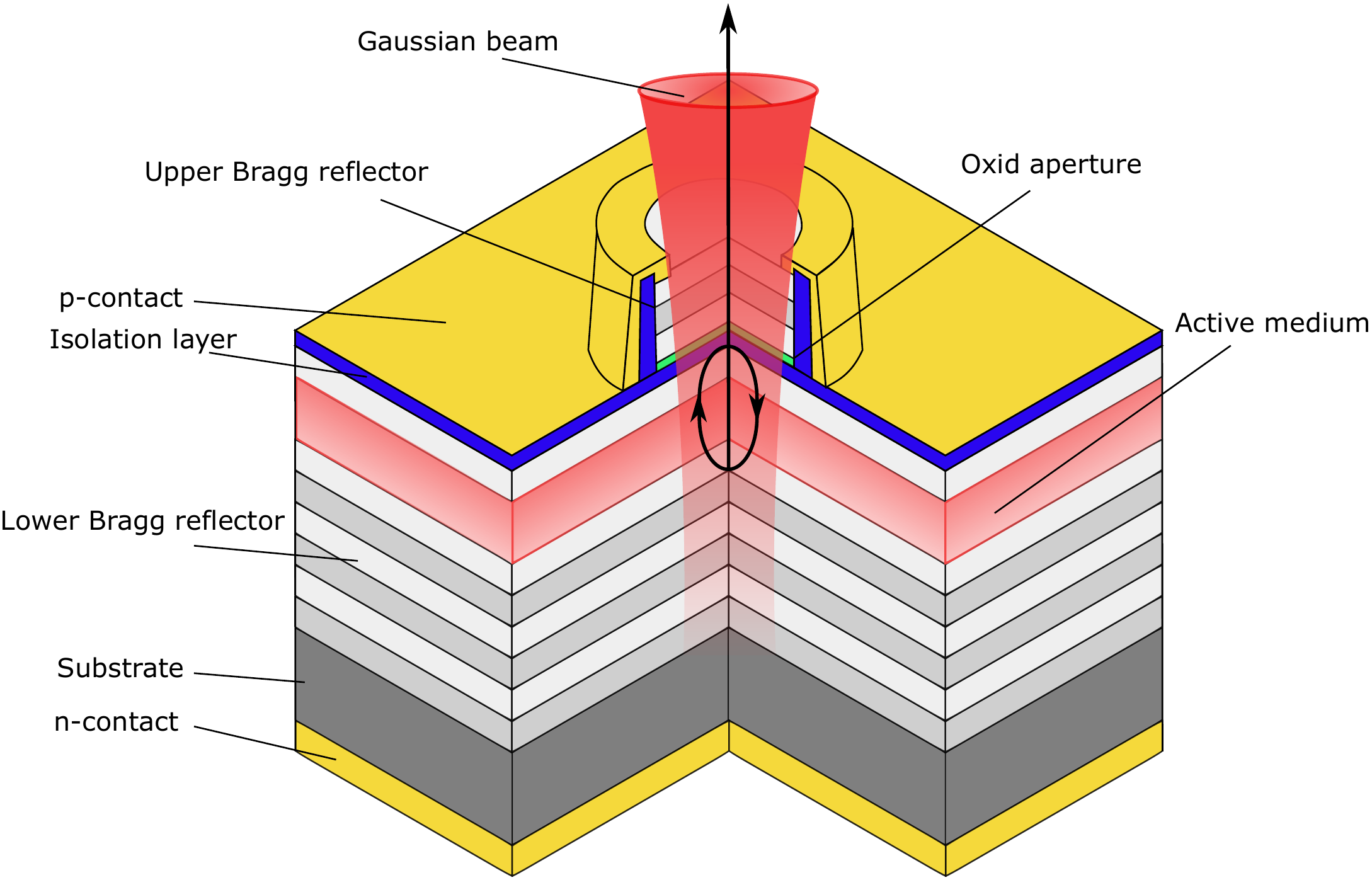}
	\caption[Schematics of a vertical-cavity surface-emitting lasers]{Schematics of a vertical-cavity surface-emitting lasers illustrating the structure and the basic operation of an \ac{vcsel}  \cite{Michalzik2013}}
		\label{LFIim:VCSEL_structure}
\end{figure}

\Cref{LFIim:VCSEL_structure} shows the basic structure of an \ac{vcsel}.
The excitation current flows from the p-contact through the active medium towards the n-contact of the semiconductor surpassing the resonator of the \ac{vcsel}. The resonator builds up from two \acp{dbr} on top and bottom of the active medium. Inside the resonator, the emission and recombination process occurs, which finally leads to a coherent Gaussian laser beam emission. The diameter and the beam's divergence angle are controlled by the oxide aperture applied on top of the active medium. Compared to \acp{eel}, the \acp{vcsel} has a lower threshold current and a high-power conservation efficiency $\eta_i$, which allows a low power operation of the laser, a circular beam profile with lower divergence angles, which allows a simple and smaller optical design. \acp{vcsel} further can be designed to work in a single transversal mode condition to emit only a single-mode electromagnetic wave, which is beneficial for an \ac{lfi} sensor \cite{Michalzik2013}. 
\subsubsection{Distributed Bragg Reflectors}

\begin{figure}[h!]
	\centering
	\includegraphics[width=0.2 \textwidth]{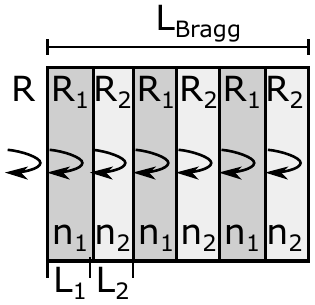}
	\caption{Schematics of a \ac{dbr} mirror of a VCSEL \cite{Coldren2012}.}
		\label{LFIim:Bragg_Mirrors}
\end{figure} 
The \acp{dbr} in \Cref{LFIim:VCSEL_structure} consist of an alternating sequence of high- and low refractive index layers ($n_1, n_2$) with thickness ($L_1, L_2$) of one quarter of the material wave length as shown in \Cref{LFIim:Bragg_Mirrors}. With these conditions, the reflection of each reflector pair will add up to an overall effective reflectivity $R$ of the mirror.

If the oscillating electric field $E(z,t)$ propagates along the z-axis inside the resonator, it is reflected by one of the two mirrors. Losses occur inside the mirrors as the mirrors are no ideal reflectors and a part of the field is transmitted and coupled out of the cavity. 
The mirror losses of a \ac{dbr} are calculated according to Michalzik et. al. \cite{Michalzik2013} by 
\begin{equation}
\label{LFIequ:Mirror_losses}
\alpha_{m} = \frac{1}{L_{Bragg}} ln\left[ \frac{1}{\sqrt{R_1 R_2}} \right] .
\end{equation}

\subsubsection{Optical Resonator}
\label{LFIsubsubsec:Optical_cavity}
\begin{figure}[h!]
	\centering
	\includegraphics[width=0.4 \textwidth]{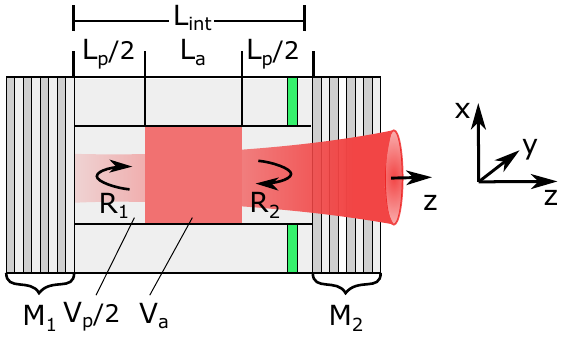}
	\caption{Schematics of the resonator of a diode laser \cite{Coldren2012}.}
		\label{LFIim:Optical_resonator}
\end{figure} 
The optical resonator shown in \Cref{LFIim:Optical_resonator} is the main component of a laser. It can be divided into two parts, an active volume $V_a$ with length $L_a$ including the active medium and a passive volume $V_p$ with length $L_p$. Together both volumes form the whole resonator volume $V$. The fraction between the active volume and the overall volume is described by the confinement factor $\Gamma$. In the active volume $V_a$ photons are generated, which propagate with the group velocity $v_g$ along the z-axis of the resonator until they hit one of the two \acp{dbr} ($M_1, M_2$) and are back reflected leading to a harmonic oscillating field $E(z,t)$ inside the resonator. The group velocity inside the resonator is given by 
\begin{equation}
\label{LFI:equ:group_velocity}
	v_g = \frac{c_0}{\overline{n}}
\end{equation}
with $c_0$ describes the speed of light and $\overline{n}$ the effective refraction index of the materials \cite{Coldren2012}. 

The harmonic oscillating field $E(z,t)$ inside the resonator is described by
\begin{equation}
\label{LFIequ:electrical_wave}
E(z,t) = E_0 e^{-j\omega t - \tilde{\beta}z}
\end{equation}
with the angular frequency  $\omega$ of the electric field $E$ and the complex propagation constant $\tilde{\beta}$.  The angular frequency $\omega$ of $E$ in \Cref{LFIequ:electrical_wave} is determined by the laser wavelength $\lambda$ and the speed of light as follows
\begin{equation}
\label{LFIequ:angular_wavelength}
	\omega = 2 \pi f = 2 \pi \frac{\lambda}{c_0}.
\end{equation}
The complex propagation constant is according to Coldren et. al. \cite{Coldren2012} given by
\begin{equation}
\label{LFIequ:complex_gamma}
	\tilde{\beta} = \beta + \frac{1}{2} j (\Gamma g - \alpha_i)
\end{equation}
where $\beta$ summarizes the real part of $\tilde{\beta}$ and the modal gain $g$. $\alpha_{i}$ summarizes the internal losses as described in \Cref{LFIsec:Laser_feedback_interferometry}. Besides the gain, there are internal losses $\alpha_{i}$ inside the resonator which are mainly related to non-radiative recombination $R_{nf}$ of charge carries \cite{Coldren2012}.

The real part $\beta$ is composed of the laser wavelength $\lambda$ and the effective refractive index $\overline{n}$ of the active medium 
\begin{equation}
\label{LFIequ:real_prop_const}
	\beta = \frac{2 \pi \overline{n}}{\lambda}.
\end{equation} 
The propagation of a mode inside the resonator requires that stimulated emission of photons exceed the internal losses $\alpha_{i}$ as well as the mirror losses $\alpha_{m}$. At this point the modal gain has reached $g_{th}$, the threshold gain. At this condition the electric field $E(z,t)$ will replicate itself after a round trip through the resonator which yields $E(z = 2L_{int},t) = E(z=0,t)$ \cite{Coldren2012}. Inserting these boundaries and the \ac{dbr} reflectivity $R_1$ and $R_2$ into \Cref{LFIequ:electrical_wave} together with \Cref{LFIequ:complex_gamma} gives the threshold condition which needs to be fulfilled for stable oscillation of $E(z,t)$ inside the resonator
\begin{equation}
\label{LFIequ:laser_condition}
		\sqrt{R_1} \cdot \sqrt{R_2}\cdot  e^{-j\beta_{th} L +  (\Gamma g_{th} - \alpha_{i}L)} = 1.
\end{equation}

The subscript $th$ denotes that \Cref{LFIequ:laser_condition} defines $\beta$ and $g$ at threshold operation and above. The amplitude condition is therefore the real part of \Cref{LFIequ:laser_condition} given by
\begin{equation}
\label{LFIequ:amplitude}
	\sqrt{R_1} \cdot \sqrt{R_2} e^{(\Gamma g_{th} - \alpha_{i})L} = 1.
\end{equation} 
Resolving \Cref{LFIequ:amplitude} for the threshold gain leads to
\begin{equation}
\label{LFIequ:gain_threshold}
	g_{th} = \frac{1}{\Gamma} \left(\alpha_{i} + \frac{1}{L_{int}} ln\left(\frac{1}{\sqrt{R_1} \cdot \sqrt{R_2}}\right)\right)
\end{equation}
where $L_{int}$ denotes the length of the resonator. Recalling \Cref{LFIequ:Mirror_losses}, \Cref{LFIequ:gain_threshold} can be rewritten to
\begin{equation}
\label{LFIequ:threshold_gain}
	\Gamma g_{th} = \frac{1}{v_g \tau_p} = v_g (\alpha_{i} + \alpha_{m})
\end{equation}
where $\tau_p$ is referred to as photon decay lifetime containing all losses inside the resonator. In \Cref{LFIequ:threshold_gain} the losses are described in a more general way by the photon lifetime $\tau_p$, leading to
\begin{equation}
\label{LFIequ:photon_lifetime}
\frac{1}{\tau_p} = \alpha_{i} + \alpha_{m}.
\end{equation}
The photon lifetime therefore describes the mean lifetime of a photon traveling inside the resonator until the internal losses stops it \cite{Coldren2012}. 

The duration a photon takes to fulfill a full round trip inside the resonator is given by the laser round trip time $\tau_{int}$. The round-trip time is calculated from the group velocity and the cavity length 
\begin{equation}
	\label{LFIequ:Laser_Round_trip_time}
	\tau_{int} = \frac{2 \cdot L_{int}}{v_g}.
\end{equation}

The length of the cavity further defines the wavelength of the laser. This can be shown by analyzing the complex part of \Cref{LFIequ:laser_condition} which states the following condition
\begin{equation}
e^{-2j\beta_{th} L_{int}} = 1.
\end{equation} 
A solution which fulfills this condition is according to Coldren et. al. \cite{Coldren2012} 
\begin{equation}
\label{LFIequ:phasecondition}
\beta_{th} L_{int} = m \pi
\end{equation}
with $m$ describing the longitudinal mode number of the laser, which is 1 for a single mode laser used in this work. Inserting \Cref{LFIequ:real_prop_const} into \Cref{LFIequ:phasecondition} the wavelength of the laser is given by
\begin{equation}
\label{LFIequ:cavity_wave_length}
\lambda = \frac{2}{m}\overline{n}L_{int}.
\end{equation}

\subsubsection{Threshold Current and Optical Power}
One important parameter of a \ac{vcsel} is the threshold current. If this current is applied to the \ac{vcsel}, the laser condition, which is described in \Cref{LFIequ:laser_condition}, is fulfilled and stimulated emission takes place.

The threshold current is given by
\begin{equation}
	I_{th} = \frac{q V}{\eta_i \tau_{sp}} \cdot N_{th}
\end{equation}
with $q$ describing the elementary charge and $\tau_{sp}$ the spontaneous emission rate which does not contribute to the stimulated emission of photons, as described in \Cref{LFIsubsubsec:Basic_principles_of_semiconductor_lasers} \cite{Coldren2012}. At threshold the carrier density  $N_{th}$ inside the cavity reaches its maximum and a further increase of the current $I$ will not further increase the carrier density. 

Based on this observation, above threshold only the photon density $N_p$ will increase and contribute to the emission of photons
\begin{equation}
	N_p = \frac{\eta_i (I - I_{th})}{q \cdot v_g \cdot g_{th} \cdot V}
\end{equation}
which is equal to an increase of optical power. This leads us to the basic equation to describe the optical power of a laser 
\begin{equation}
\label{LFIequ:optical_power}
	P_0 = \eta_d \frac{h\cdot v}{q}\left( I-I_{th}\right)
\end{equation}
with the differential efficiency
\begin{equation}
\label{LFIequ:differential_efficency}
	\eta_d =\frac{\eta_i \cdot \alpha_{m}}{\alpha_{m} + \alpha_{i}}.
\end{equation}
\Cref{LFIequ:optical_power} is only valid if both mirrors are symmetric and have the same reflectivity $R$. Otherwise a correction factor $F_2$ is applied, yielding
\begin{equation}
	P_{02} = F_2 \cdot \eta_d \cdot \frac{hv}{q} \left(I-I_{th}\right)
\end{equation}
with 
\begin{equation}
\label{LFIequ:Symmetry_missmatch_factor}
	F_2= \frac{T_2}{\left(1-\sqrt{R_2}\right)^2 + \sqrt{ \frac{R_2}{R_1} } \cdot \left(1-\sqrt{R_1}\right)^2 }.
\end{equation}
\subsection{Semiconductor Laser under Feedback}
\label{LFIsubsec:Semiconductor_laser_under_feedback}
\begin{figure}
	\centering
	\includegraphics[width = 0.5\linewidth]{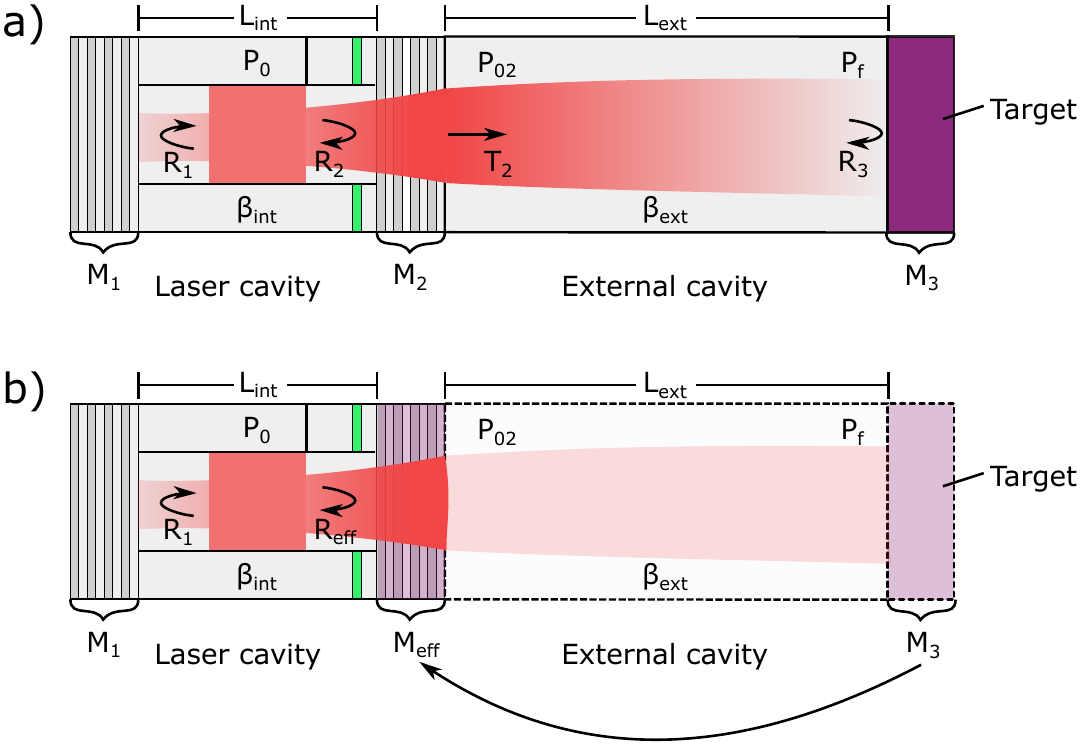}
	\caption[Three mirror model]{a) shows the active laser cavity coupled with a passive external cavity formed by an external measurement target. The external target is modeled as an additional Mirror $M_3$ with a reflectivity $R_3$. b) shows the three-mirror model according to Coldren et. al. \cite{Coldren2012}. In this model the out-coupling mirror $M_2$ is replaced by an effective mirror $M_{eff}$, which describes the coupling of $M_2$ and $M_3$ obtained by an S-parameter analysis. }
	\label{LFIim:three_mirror_model}
\end{figure}

\ac{lfi} sensors are \acp{vcsel} working on the self-mixing phenomena. Self-mixing describes an operation mode of a laser where a part of the emitted light is back injected into the laser cavity and thus leads to direct feedback to the laser source \cite{1072281}. The operation of a laser under feedback can be described by extending the laser cavity as shown in \Cref{LFIim:Optical_resonator} by addition of an external cavity with an additional mirror $M_3$, which scatters back light into the laser cavity, as shown in \Cref{LFIim:three_mirror_model} a). This leads to the well-known three-mirror model of a semiconductor laser under feedback. A schematic of the three-mirror model is shown in \Cref{LFIim:three_mirror_model} b). An analytical description of the physical behavior of a laser under feedback was introduced by Lang and Kobayashi \cite{1070479}. 

The cavity of the laser  with length $L_{int}$, propagation constant $\beta_{int}$ and the two mirrors $M_1$ and $M_2$ with reflectivity $R_1$ and $R_2$ are coupled with an external passive cavity with length $L_{ext}$, propagation constant $\beta_{ext}$ and a mirror $M_3$ with a reflectivity $R_3$ \cite{Coldren2012}.

To simplify the further analysis of the system, an effective mirror model is introduced. The mirrors $M_2$ and $M_3$ of the laser cavity and the external cavity are replaced by an effective mirror $M_{eff}$ with an effective reflectivity $R_{eff}$. Based on the S-parameter analysis, the reflectivity of the effective mirror is according to Coldren \cite{Coldren2012} given by 
\begin{equation}
\label{LFI:equeffective_mirror}
	\sqrt{R_{eff}} = \frac{\sqrt{R_2} + (T_2) \cdot \sqrt{R_3} \cdot e^{-j 2 \cdot\tilde{\beta}_{ext}\cdot L_{ext}}}{1 + \sqrt{R_2} \cdot \sqrt{R_3} \cdot e^{-j 2 \cdot\tilde{\beta}_{ext}\cdot L_{ext}}}.
\end{equation}
Considering a weak reflectivity of the external mirror $R_3 << 1$ the effective mirror reflection, \Cref{LFI:equeffective_mirror} can be reduced to
\begin{equation}
\label{LFI:equ_effective_mirror2}
	\sqrt{R_{eff}} = \sqrt{R_2} + (T_2) \cdot \sqrt{R_3} \cdot e^{-j 2 \cdot\tilde{\beta}_{ext}\cdot L_{ext}}.
\end{equation}
Thus, a change of feedback due to a change of the reflectivity $\Delta R$ can impact both the amplitude and the phase of the effective mirror. A change of external reflectivity is expressed by $\Delta R = R_{eff} - R_2$. Inserted into \Cref{LFI:equ_effective_mirror2}, the impact of changing reflectivity can be described by a complex vector, which can, according to Coldren et.al. \cite{Coldren2012},  be partitioned into a real in-phase part $\Delta R_{ip}$
\begin{equation}
\label{LFIequ:Real_part_effective_reflection_index}
	\Delta R_{ip} = (T_2) \cdot \sqrt{R_3} \cdot \cos \left(2 \cdot \beta_{ext} \cdot L_{ext}\right)
\end{equation}
and a imaginary quadrature part
\begin{equation}
\label{LFIequ:Imag_part_effective_reflection_index}
\Delta R_{q} = -(T_2) \cdot \sqrt{R_3} \cdot \sin \left(2 \cdot \beta_{ext} \cdot L_{ext}\right).
\end{equation}

Recalling \Cref{LFIequ:phasecondition}, the term $2 \cdot \beta_{ext} \cdot L_{ext}$ describes the external phase $\phi_{ext}$ of the external cavity and thus equals $\omega_{ext} \cdot \tau_{ext}$.

The in-phase component \Cref{LFIequ:Real_part_effective_reflection_index} affects the laser by changing the reflectivity of the second mirror and therefore the mirror losses $\Delta \alpha_i$ given by \Cref{LFIequ:Mirror_losses}. This leads to a new photon life time (\Cref{LFIequ:photon_lifetime})
\begin{equation}
	\frac{1}{\tau_p^{'}} = \frac{1}{\tau_p} + v_g \cdot \Delta \alpha_{m} = \frac{1}{\tau_p} + \frac{v_g}{L_{int}} \frac{\sqrt{R_1 \cdot \Delta R_{ip} }}{\sqrt{R_1 \cdot R_2}}.
\end{equation}
By plugging in the laser round trip time, given in \Cref{LFIequ:Laser_Round_trip_time}, and the in-phase component of the effective reflectivity, \Cref{LFIequ:Real_part_effective_reflection_index} leads to
\begin{equation}
	v_g\cdot \Delta \alpha_i = -2 \cdot T_2 \cdot \sqrt{\frac{R_3}{R_2}} \cdot \frac{1}{\tau_{int}} \cdot \cos\left(2 \cdot \beta_{ext} \cdot L_{ext}\right) = -2 \cdot k_f \cdot \cos\left(2 \cdot \beta_{ext} \cdot L_{ext}\right).
\end{equation}

\begin{equation}
\label{LFIequ:Coupling_strength}
	k_f = T_2 \cdot \sqrt{\frac{R_3}{R_2}} \cdot \frac{1}{\tau_{int}}
\end{equation}
describes the coupling strength between the laser cavity and the external cavity and influences the signal quality.

The cosine modulation of the photon life time leads also to a modulation of the threshold gain $g_{th}$ ,introduced in \Cref{LFIequ:threshold_gain}, and therefore to a shift in threshold carrier density $N_{th}$, which leads to a variation in the threshold current $I_{th}$ and therefore a modulation of the optical power (\Cref{LFIequ:optical_power}). In addition the differential efficiency $\eta_d$ (\Cref{LFIequ:differential_efficency}) and the symmetry correction factor $F_2$ given by \Cref{LFIequ:Symmetry_missmatch_factor} are modified by the change in mirror reflectivity. This effects lead to a new optical power equation for a perturbed laser \cite{Coldren2012}
\begin{equation}
\label{LFIequ:power_modulation}
	P_f = P_{02}\left(1 - m \cdot \cos \left(2 \cdot \beta_{ext} \cdot L_{ext}\right)\right)
\end{equation}
which is equal to the more common notation of the equation
\begin{equation}
\label{LFIequ:power_under_feedback_freq}
	P_f\left(\omega_{ext} \right) = P_{02} \left(1 - m \cdot \cos \left(\omega_{ext} \cdot \tau_{ext} \right)\right)
\end{equation} 
as $2 \cdot \beta_{ext} \cdot L_{ext} $ equals $  \omega_{ext} \tau_{ext} $ as well as $ \phi_{ext}$.
The modulation factor $m$ describes the power variation of the perturbed laser due to coupling effects between the two cavities for robust operation of the laser above the threshold \cite{Coldren2012}
\begin{equation}
\label{LFIequ:modulation_depth}
	m = 2 \cdot k_f \cdot \tau_{p} (\frac{\eta_i}{\eta_d} -1) + k_f \tau_{ext} (1- F_2) ( \frac{1 + R_2}{T_2}).
\end{equation}

Variation of the in-phase component $\Delta R_{ip}$ affects the threshold level and the optical power of the laser leading to an \ac{am} of the laser. The quadrature component $\Delta R_q$ in contrast modifies the round-trip phase angle $\phi_{eff}$ at mirror $R_2$ 
\begin{equation}
	\phi_{eff} = \frac{\Delta R_q}{R_2}
\end{equation}
modifying the resonant wavelength of the cavity given by \Cref{LFIequ:cavity_wave_length} and thus leading to an angular frequency shift, which is given according to Coldren et.al. \cite{Coldren2012} by 
\begin{equation}
	\Delta \omega_\phi = -k_f \cdot \sin \left(2 \cdot \beta_{ext} \cdot L_{ext}\right).
\end{equation} 
Besides the round trip phase angle, an additional change of the angular frequency is introduced by the change of the threshold carrier density $N_{th}$ described by Henry's line-width enhancement factor $\alpha$  \cite{1071522}
\begin{equation}
\Delta \omega_N = -\alpha k_f \cdot \cos \left(2 \cdot \beta_{ext} \cdot L_{ext}\right).
\end{equation} 

Adding both frequency shifts $\Delta \omega_\phi$ and $\Delta \omega_N$ using the trigonometric identity, the total frequency shift $\Delta \omega$  is given by
\begin{equation}
\label{LFIequ:omega_equation}
\Delta \omega = -k_f \cdot \sqrt{1 + \alpha^2}  \cdot sin(\phi_{ext} + \arctan(\alpha)).
\end{equation}
The angular frequency shift describes the difference of the angular frequency $\omega_0$, experienced by the laser during the round trip through the external cavity without optical feedback. The angular frequency $\omega_{ext}$ corresponds to the actual frequency response including optical feedback \cite{Taimre:15}. As linkage between phase and angular frequency is given by $\phi = \omega \cdot \tau$,   $\Delta \omega$ can be decomposed as follows
\begin{equation}
	\Delta \omega = \omega_{ext} - \omega_{0} = \frac{\phi_{ext}}{\tau_{ext}} - \frac{\phi_0}{\tau_{ext}} = \left(\phi_{ext}-\phi_0\right) \frac{1}{\tau_{ext}}.
\end{equation}  
With this expression and the introduction of Acket's feedback parameter $C$ \cite{1072281}
\begin{equation}
\label{LFIequ:Ackets_feedback_parameter}
C = k_f \sqrt{1+\alpha^2} \tau_{ext}
\end{equation}
\Cref{LFIequ:omega_equation} can be rephrased, and the well-known excess phase equation is obtained
\begin{equation}
\label{LFIequ:excess_phase_equation}
\phi_{ext} - \phi_0 + C \cdot \sin \left(\phi_{ext} + \arctan \left(\alpha \right)\right) = 0
\end{equation}
which is the fundamental equation to describe the \ac{lfi} sensing principle found in most of the literature dealing with \ac{lfi} sensors e.g., by Giuliani et.al. \cite{Giuliani2002} or Taimre et.al. \cite{Taimre:15}. As \Cref{LFIequ:excess_phase_equation} is a transcendental equation, a stable phase contingent operation of the laser is only possible if $C<1$ otherwise more than one possible solution of \Cref{LFIequ:excess_phase_equation} exists, which leads to phase jumps and thus an unstable behavior of the laser \cite{Taimre:15}. Thus, in the further descriptions a stable operation of the system in the low feedback regime ($C<1$) is assumed.

\subsection{Modulation of Semiconductor Lasers under Feedback}
Modulation of the coupled cavity is required to observe a change of phase response and thus measure physical quantities. In general, modulation of the coupled cavity can be divided into modulation of the internal cavity and modulation of the external cavity. Modulation of the internal cavity affects the effective mirror $M_{eff}$ of the coupled system by changing the output mirror $M_2$ of the laser by variation of the internal laser parameters. Modulation of the external cavity affects the effective mirror $M_{eff}$ by variation of the external cavity mirror $M_3$ by variation of the parameters of the external cavity. In the following, both modulation schemes and the corresponding measurement quantities are discussed.
\subsubsection{External Cavity Modulation}
Based on \Cref{LFIequ:power_modulation}, the modulation of optical power is strongly dependent on the cosine term and therefore to $\omega_{ext}\cdot \tau_{ext} = \phi_{ext}$, which is linked via \Cref{LFIequ:excess_phase_equation} with the phase $\phi_0$ of the unperturbed laser. Examining \Cref{LFIequ:excess_phase_equation}, two parameters influence $\phi_{ext}$. One is Acket's feedback parameter $C$ and the other is the phase $\phi_0$ of the unperturbed laser. The linewidth enhancement factor $\alpha$ as third possible parameter is omitted, as according to Taimre et al. \cite{Taimre:15} this parameter can modeled as constant.

Analyzing the remaining parameters, the most straightforward way to modulate $\phi_{ext}$ is a modulation of the round trip phase $\phi_0$ of the unperturbed laser \cite{Taimre:15}. Recalling the correspondence between phase $\phi_0$ and the angular frequency $\omega_{0}$ ($\phi_0 = \omega_{0} \cdot \tau_{0}$) as well as \Cref{LFIequ:cavity_wave_length} and the relationship between wavelength and frequency $\lambda = c_0 / f$, the following equation shows which parameters affect the round trip phase $\phi_0$ of the unperturbed laser
\begin{equation}
\label{LFIequ:distance_by_phase}
	\phi_0 = 2 \cdot \pi \cdot f_0 \cdot \tau_{ext} = \frac{4 \cdot \pi \cdot n_{ext} L_{ext}}{\lambda}.
\end{equation}

Considering operation in free space $(n_{ext} =1)$ and a constant wavelength $\lambda$ of the laser, the round-trip phase is directly proportional to the length of the external cavity $\phi_0 \approx L_{ext}$. A periodic variation of $\Delta L_{ext}$ of at least $\lambda / 2$ leads to a periodic variation of the optical power $P_f$. Differentiating \Cref{LFIequ:distance_by_phase} with respect to time $t$ leads to 
\begin{equation}
\label{LFIequ_phase_derivation}
\frac{d \phi_0}{dt} = \frac{4 \pi}{\lambda} \frac{dL_{ext}}{dt}.
\end{equation}   
As the derivation of the phase with respect to time equals the frequency and the derivation of the distance with respect to the time equals the velocity $v$ of displacement in beam direction. This leads to the well-known Doppler equation
\begin{equation}
\label{LFIequ:in_plane_doppler}
f_d = \frac{2 \cdot |v_T| }{\lambda}
\end{equation}
which equals a Doppler shift by the superposition of the incident wave and the back scattered wave. \Cref{LFIequ:in_plane_doppler} is limited to target movements which are transversal to the propagation axis denoted by $v_T$. To overcome this limitation, the Doppler frequency equation is expressed in a more general way
\begin{equation}
\label{LFIequ:doppler_equation}
f_d = \frac{2 \cdot |v| \cdot \cos(\Psi) }{\lambda}
\end{equation}
where $\Psi$ denotes the angle between the wave vector $k$ and the velocity vector $v$ \cite{10.1117/12.775131}.
To measure absolute velocity, the internal parameters of the cavity need to be modulated as well.
\subsubsection{Internal Cavity Modulation}
The modulation of internal cavity parameters of a laser under feedback is carried out by a modulation of the drive current of the laser \cite{Michalzik2013}. A variation in drive current affects the internal cavity by two different mechanisms. First, as the current increases, the cavity temperature rises, which affects the effective cavity length $L_{int}$ and the gain curve in the material. Secondly, increasing the current increases the carrier and photon density in the material, which influences the material's refractive index $n_{int}$. This second effect is also referred to as the plasma effect, which tends to be the dominating effect at high modulation frequencies \cite{256171}. Recalling \Cref{LFIequ:cavity_wave_length}, both effects influence the wavelength of the laser. For low modulation frequencies, the temperature effect dominates the modulation of the cavity \cite{Nordin2004} which leads to 
\begin{equation}
\label{LFIequ:wavelength modulation}
	\lambda(t) = 2 \cdot \overline{n} \cdot L_{int}(t).
\end{equation} 
Recalling \Cref{LFIequ:distance_by_phase} and plugging in the variation of the wave length leads to a modulation of $\phi_0$ and therefore a modulation of the optical output power $P_f$. The modulation of the wavelength can be interpreted in \Cref{LFIequ:distance_by_phase} as a slight movement of the laser towards the target, if the driving current is increased and a movement away from the target as the drive current is reduced respectively. One important observation is the proportionality between drive current modulation, active cavity modulation and the modulation of the wavelength
\begin{equation}
\label{LFIequ:correspondence_modulation}
	\frac{d \lambda}{dt} \sim \frac{d L_{int}}{dt} \sim \frac{dI}{dt}.
\end{equation}
Plugging in \Cref{LFIequ:wavelength modulation} into \Cref{LFIequ:distance_by_phase} leads to a modulation of $\phi_s$ even if $L_{ext}$ does not vary, which allows the measurement of the distance $L_{ext}$ towards still standing targets. 

Time differentiating  \Cref{LFIequ_phase_derivation} with respect to the wavelength $\lambda$ as a modulation of the drive current leads to a variation of the wavelength and gives the following equation
\begin{equation}
\label{LFIequ:distance_frequency}
f_b = \frac{2 \cdot L_{ext}}{\lambda^2} \cdot \frac{d \cdot \lambda}{dt}.
\end{equation}
with $f_b$ as beat frequency \cite{NORGIA201731}. The beat frequency is directly proportional to the target distance \cite{NORGIA201731} .

\subsubsection{Simultaneous Measurement of Distance and Velocity}
\label{LFIsubsubsec:Simultaneous_measurement_of_distance_and_velocity}
As both frequencies, $f_d$ and $f_b$ lead to a modulation of the optical power $P_f$ either induced by a moving target or by a change of the wavelength caused by the laser drive current modulation, both effects superimpose and thus cannot be observed independently. Therefore, a known current modulation scheme with two distinguishable states is required to resolve this singularity. A widely applied modulation scheme with two states is a triangular modulation as used in \ac{fmcw} \ac{lidar} systems \cite{8067701}.

The drive current of the laser is modulated by a symmetric \ac{dc}-free triangle signal with a frequency $f_{mod}$, a period duration $T_{mod}$ and an amplitude $I_{mod}$.
 
\begin{figure}[ht]
 	\centering
 	\includegraphics[width = 0.55\linewidth]{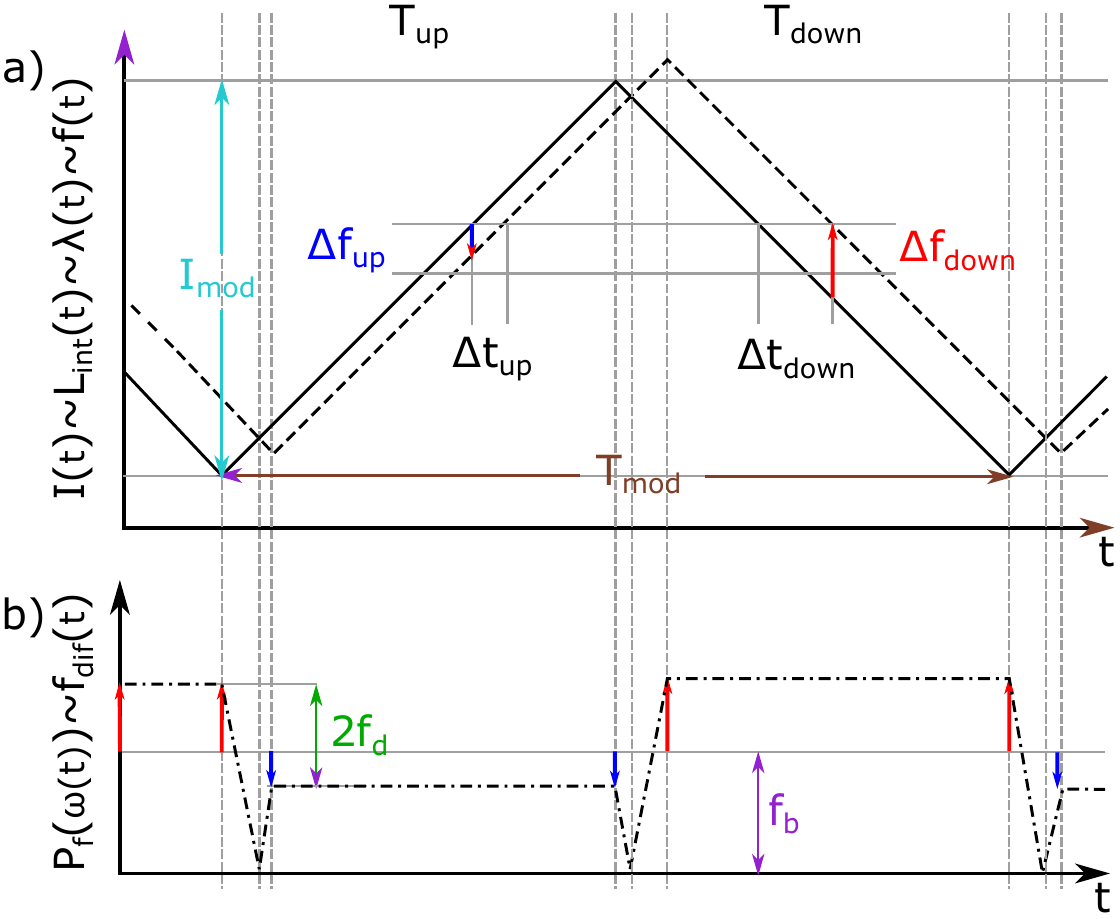}
 	\caption[Effect of  triangular modulation on time and frequency domain]{a) Effect of triangular modulation of the wavelength $\lambda$ by the laser's drive current on the transmitted signal (solid triangular line) and the back-reflected echo signal (dashed line) from the target. b) the resulting frequency inside the laser cavity as a sum of transmitted and back-reflected signal. }
 	\label{LFIim:triangle_doppler}
\end{figure}

\Cref{LFIim:triangle_doppler} a) shows the effect of triangular modulation of the wavelength $\lambda$ by the drive current of the laser on the transmitted signal (solid triangular line) and the back-reflected echo signal (dashed line) from the target. The echo is delayed by $\tau_{ext}$ and shifted by the Doppler frequency. $f_{up}$ corresponds to the difference frequency measured during the up ramp, and $f_{down}$ corresponds to the difference frequency measured during the down ramp of the triangular signal. The amplitude of the triangular signal is controlled by the modulation current $I_{mod}$ and the duration of a triangular wave by $T_{mod}$. \Cref{LFIim:triangle_doppler} b) shows the resulting frequency inside the laser cavity as the sum of transmitted and back-reflected signal. The mean resulting frequency corresponds to the distance-related beat frequency $f_b$ and the difference between up ramp frequency $f_{up}$ and the down ramp frequency $f_{down}$ corresponds to twice the frequency shift induced by the Doppler effect and thus to the target velocity. The sum frequency finally modulates the optical power inside the laser cavity according to \Cref{LFIequ:power_under_feedback_freq}

The current modulation leads to a change in the internal cavity length, and thus the wavelength of the cavity as described by \Cref{LFIequ:correspondence_modulation}. This change leads to a modulation of the frequency of the optical wave transmitted from the laser. With a delay of $\tau_{ext}$, corresponding via \Cref{LFIequ:Laser_Round_trip_time} to the target distance, the modulated wave is back-reflected from the target, and the echo signal is back injected into the laser cavity. Due to the velocity of the target, the optical wave is shifted by the Doppler frequency. Thus the cavity acts as an optical mixer constructing the sum and difference of both wave frequencies\cite{NORGIA201731}. \Cref{LFIim:triangle_doppler}  b) shows the frequency difference of both wave frequencies. While the laser operates in the Terahertz Regime, the optical mixer converts the difference signal to the Megahertz regime, which eases signal processing. As depicted in \Cref{LFIim:triangle_doppler}, the mean of both frequencies $f_{up}$ and $f_{down}$ corresponds to the beat frequency $f_b$ and half of the difference between them to the Doppler frequency $f_d$. 

Due to the symmetric shift of the signal wave during a full modulation swing $T_{mod}$, the distance related frequency $f_b$ is given by
\begin{equation}
\label{LFIequ:tri_distance_frequency}
	f_b = \frac{f_{up} + f_{down}}{2}
\end{equation}
and the target velocity related frequency $f_d$ is given by
\begin{equation}
\label{LFIequ:tri_doppler_frequency}
	f_d = \frac{f_{up} - f_{down}}{2}.
\end{equation}  
By measuring the frequencies $f_{up}$ and $f_{down}$ during the corresponding triangle ramps $T_{up}$ and $T_{down}$, the Doppler- and beat-frequencies can be calculated. Recalling \Cref{LFIequ:distance_frequency}, the distance to the target can be calculated from the laser wavelength $\lambda$, the triangle parameters and the drive current corresponding wavelength modulation constant $ d \lambda /dt$. In addition, the target velocity can be calculated by plugging in \Cref{LFIequ:tri_doppler_frequency} into \Cref{LFIequ:doppler_equation} if the angle of incidence $\psi$ is known.
\subsection{LFI Sensor}
\begin{figure}[h!]
	\centering
	\includegraphics[width = 0.55\linewidth]{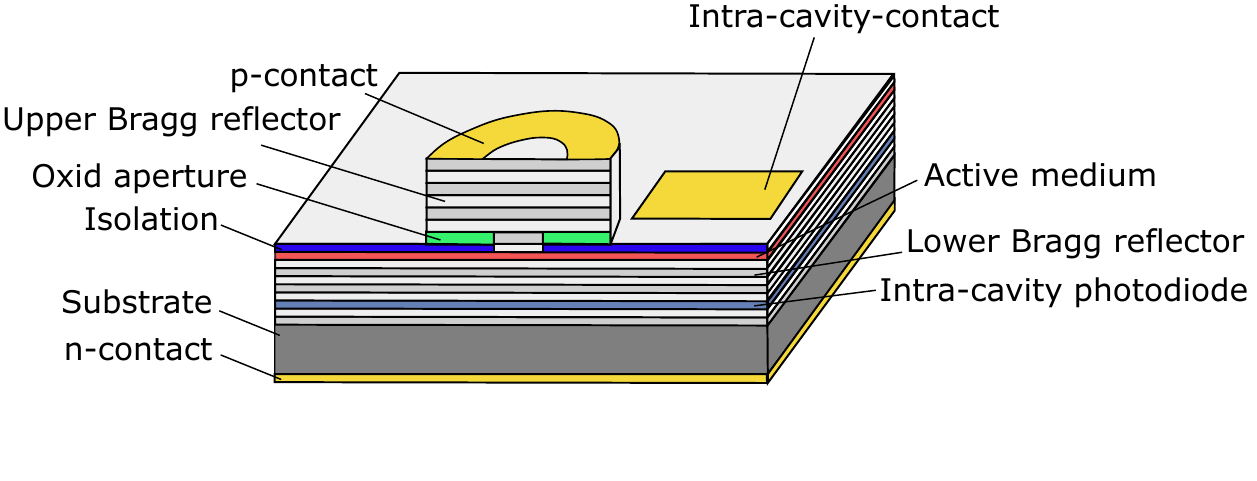}
	\caption{Integration of a photodiode into the lower \ac{dbr} of the laser cavity \cite{grabherr2009integrated}.} 
	\label{LFIim:ViP_Sensor}
\end{figure}

As described in the previous sections, the optical power $P_f$ of the laser under feedback is modulated based on an external target, which back reflects laser light into the laser cavity as described by \Cref{LFIequ:power_under_feedback_freq}. Therefore it is required to measure the varying optical power inside the resonator of the laser cavity. A suitable solution is to integrate an intra-cavity photodiode into the lower \ac{dbr} of the cavity as shown in \Cref{LFIim:ViP_Sensor} \cite{Taimre:15}. A detailed description of the integration of a so-called monitoring diode is given by Grabherr et al. \cite{grabherr2009integrated}.

\begin{figure}[ht]
	\centering
	\includegraphics[width=0.8\linewidth]{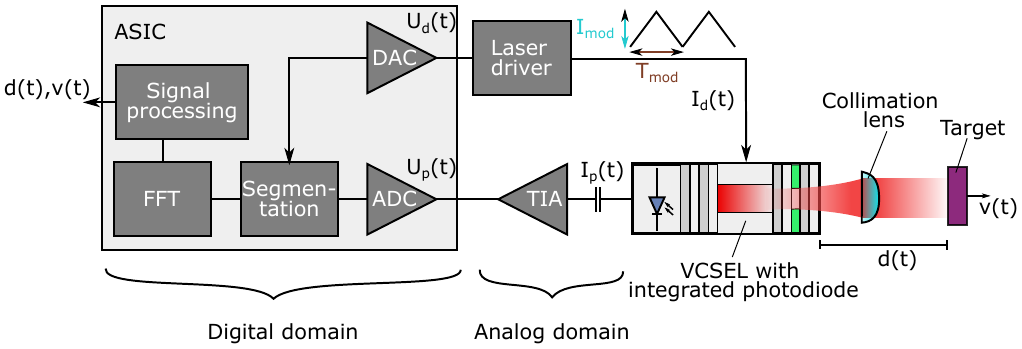}
	\caption[Block diagram of an LFI sensor]{Block diagram of an \ac{lfi} sensor with modulation source (\ac{dac} and laser driver) and analog frontend (\ac{tia}) for to capture the photodiode signal. Finally the measurement quantities (distance, velocity) are extracted in the digital domain \cite{Taimre:15}.}
	\label{LFIim:bsbsmisensor}
\end{figure}

Besides the sensor component itself, additional electrical components like a laser driver circuit, a receiver circuit, and a digital circuit are required to build up an \ac{lfi} sensor to measure distance and velocity as described in \Cref{LFIsubsubsec:Simultaneous_measurement_of_distance_and_velocity}. \Cref{LFIim:bsbsmisensor} shows a block diagram of an \ac{lfi} sensor, including the sensing component as well as the circuitry to operate an \ac{lfi} sensor.

The circuitry can be divided into analog and digital domains. The digital domain, which could be realized by an \ac{asic}, generates a triangular modulation signal $U_d(t)$, e.g., by an \ac{dac}. This signal is fed into a laser driver circuit, which converts the voltage signal into a current modulation signal $I_d(t)$. The \ac{vcsel} with integrated photodiode emits frequency-modulated light, which is collimated using, e.g., a collimation lens. The light travels along the optical axis of the laser until it hits the target at a distance $d(t)$, which moves with a velocity $v(t)$. A portion of the light is back-reflected and back injected into the cavity where both the local optical wave and the back-reflected optical wave interfere. The photodiode, integrated into the back \ac{dbr}, measures the power modulation, which follows the inference signal according to \Cref{LFIequ:power_under_feedback_freq}. The resulting current signal $I_p(t)$ is highpass filtered to suppress the \ac{dc} power of the laser and amplified and converted into a voltage $U_p(t)$ by a \ac{tia}. This voltage signal is digitized by an \ac{adc} and segmented into voltage segments $U_{p,up}(t)$ and $U_{p,down}(t)$ which either belong to the up-ramp or the down-ramp of the current modulation triangle. From both segments the peak frequencies $f_{up}$ and $f_{down}$ are calculated. With these frequencies $f_d$ and $f_b$ are calculated using \Cref{LFIequ:tri_doppler_frequency} and \Cref{LFIequ:tri_distance_frequency}. Afterwards, the target velocity $v(t)$ in beam axis and the target distance $d(t)$ are calculated by applying \Cref{LFIequ:doppler_equation} and \Cref{LFIequ:distance_frequency}.

\section{Conclusion}
Mobile eye-tracking sensors are a key sensing technology for \ac{ar} glasses, especially for retinal projection glasses, as discussed in \Cref{sec:retinal_projection_ar_glasses}. Moreover, they are the enabling technology for a wide range of applications from \ac{hci} applications over display enhancement applications to medical and well-being applications as introduced in \Cref{sec:Eye_Tracking_for_AR_glasses}, adding significant value to consumer \ac{ar} glasses. 
The integration of eye-tracking sensors into the domain of consumer graded \ac{ar} glasses sets high requirements in the area of gaze accuracy, robustness, sensor integration, power consumption as well as sensor update rate, which are discussed in detail within \Cref{subsec:Requirements_for_eye_tracking_sensors_in_AR_glasses}. 

A detailed analysis of established state-of-the-art \ac{vog} eye-tracking sensor technology shows that \ac{vog} systems fulfill requirements regarding accuracy but are limited with regard to pupil detection and ambient light robustness, sensor integration, power consumption, and sensor update rate as discussed in \Cref{MET:Challenges_VOG_Systems}. Therefore, emerging eye-tracking sensor technologies are analyzed within \cref{METsec:Mobile_eye_tracking_sensor_technologies}. The summary of different eye-tracking sensor approaches in \cref{MET:Comparisson} shows that scanned laser eye-tracking sensors have the best fulfillment of all requirements. 

Therefore, contributions within the first part of the thesis focus on closing the gap between the high gaze accuracy of established \ac{vog} sensors and the advantages of scanned laser eye-tracking sensor technology with respect to sensor integration and power consumption. In addition, known technological limitations like ambient light robustness, glasses slippage, and user calibration are addressed within the contributions in this part of the thesis.

As contributions in the first part of the thesis focus on integrating scanned laser eye-tracking sensors into retinal projection \ac{ar} glasses, the second and third part of the thesis focus on an alternative novel static laser approach, which enables mobile eye-tracking independent of the display technology. As potential technology \ac{lfi} is identified due to the small size of the sensors and the capability to measure distance and rotational velocity with an outstanding update rate. Therefore this capture closes with a detailed description of this sensing technology in \Cref{LFIsec:Laser_feedback_interferometry} to ease understanding of the contributions within the second and third part of the thesis.

The major contributions this thesis makes to enable energy-efficient mobile eye-tracking for \ac{ar} glasses through optical sensor technology are discussed in the next chapter.

\cleardoublepage
\chapter{Major Contributions}
\begin{figure}[h!]
    \centering
    \includegraphics[width = \textwidth]{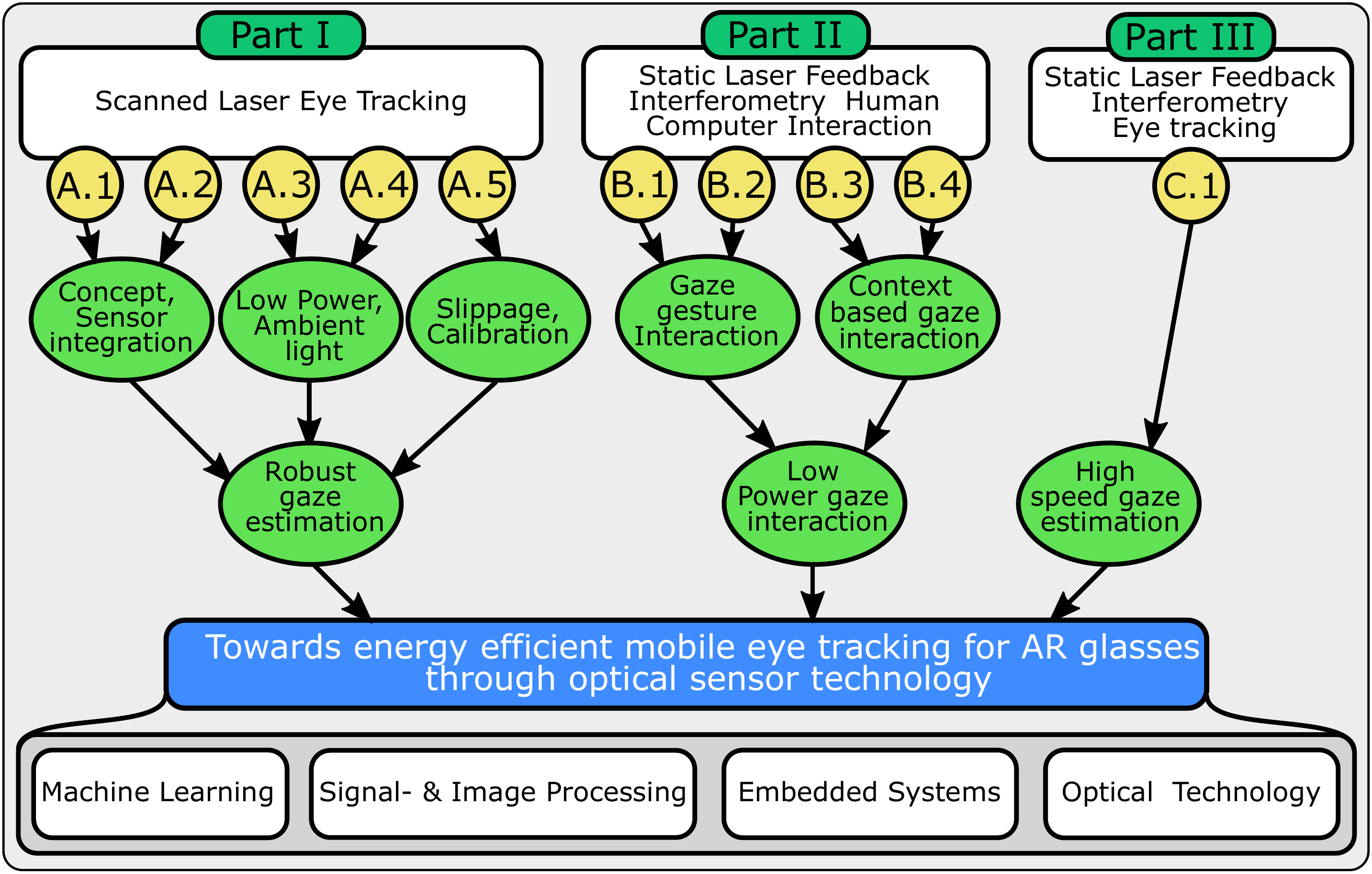}
    \caption{Overview and relationship of contributions in the thesis.}
    \label{CON:fig:Overview_Contribution}
\end{figure}
\FloatBarrier
\Cref{CON:fig:Overview_Contribution} summarizes the contributions of this thesis. The major contributions are split into the three parts \textit{Scanned Laser Eye Tracking}, \textit{Static LFI for Human Computer Interaction for AR glasses}, and \textit{Static LFI Eye Tracking}.

The first part of the thesis focuses on replacing \ac{vog} systems for retinal projection \ac{ar} glasses through a scanned laser eye-tracking approach. 

Contribution in the second part \textit{Static LFI Human-Computer Interaction for AR Glasses} of the thesis focuses on low power gaze \ac{hci} through static \ac{lfi} sensors. 

To enable low-power gaze interaction with \ac{ar} glasses, the publications B.1 and B.2 introduce static \ac{lfi} sensors operating with the Doppler principle to measure rotational eye velocities. Furthermore, based on the measured eye velocities, a gaze gesture recognition algorithm is derived to enable a gaze gesture-based control of the glass's user interface.

Contribution within the third part \textit{Static LFI Eye Tracking} focuses on fusing multiple static \ac{lfi} sensors to develop a high-speed eye-tracking senor.

This thesis is based on a rich set of fundamentals from machine learning and signal- and image processing over embedded systems to optical technology, covering various aspects from computer science and electrical engineering.  

The following sections summarize each part and its contributions regarding motivation, methods, and the achieved results. At the end of the chapter, a summary of the individual contribution is drawn.
\section{Scanned Laser Eye Tracking}
The contributions \cite{10.1145/3379157.3391995,Meyer2020e,9149591,10.1145/3530881,10.1145/3517031.3529616} focuses on scanned laser eye-tracking sensor approaches for retinal projection AR glasses in order to replace \ac{vog} systems and overcome the technological challenges of \ac{vog} systems as outlined in \Cref{MET:technological_challenges_VOG}. The first two publications focus on the sensor integration concept, the second and third publication on ambient light robustness and minimization of power consumption, and the final publication on robustness against glasses slippage and reduction of calibration effort.
\subsection{Scanned Laser Eye Tracking for Retinal Projection AR Glasses}
\label{MO:Scanned_Laser}
 This subsection is based on the publications \ref{APP:A1} \textit{Low Power Scanned Laser Eye Tracking for Retinal Projection AR Glasses} and \ref{APP:A2} \textit{A novel camera-free eye-tracking sensor for augmented reality based on laser scanning} in \Cref{APP:Scanned_Laser_Eye_Tracking}.
\subsubsection{Motivation}
Retinal projection AR glasses are emerging as potential near-eye display technology with several advantages compared to other near-eye display technologies like high contrast and low power consumption. However, the main drawback of the retinal projection display technology is the single eye box due to the tiny exit pupil, limiting the effective eye box. Therefore, an eye box expansion mechanism like exit pupil steering is necessary to increase the eye box size. Furthermore, introducing an eye box expansion mechanism requires robust eye-tracking; thus, eye-tracking sensor technology is mandatory for the success of retinal projection \ac{ar} glasses.   

 Integrating state-of-the-art \ac{vog} eye-tracking sensors into lightweight \ac{ar} glasses is challenging due to the large dimensions of camera optics and the required optical axis of the camera sensor w.r.t the visual axis of the eye. Furthermore, the power consumption of camera sensors limits the operation time of battery-constrained \ac{ar} glasses. Therefore, the possibility of integrating a scanned laser eye-tracking system into a retinal projection \ac{ar} glass as a replacement of \ac{vog} sensors is investigated in these publications to solve these limitations. 
\subsubsection{Methods}
For this purpose, an \ac{ir} laser was integrated into the \ac{rgb} laser projection module of a retinal projection \ac{ar} glasses system. The \ac{ir} laser beam shares the same beam path as the \ac{rgb} laser beams, and thus the existing \ac{mems} micro mirror scanner used for image projection can be reused to scan the \ac{ir} laser beam over the surface of a holographic free space combiner. By adding a additional optical function for the \ac{ir} wavelength to the holographic beam combiner, the \ac{ir} light is deflected towards the eye such that a 2D area on the surface of the eye is illuminated. 

An external single-pixel photodetector was integrated into the glasses frame temple to generate a grey scale image from the illuminated area on the eye. The photodetector captured the intensity of the back scattered \ac{ir} light for each scan position ($\alpha$,$\beta$). Combining the single-pixel detector and the free space combiner yields a virtual single-pixel grey scale camera. As the laser beams are off-axis w.r.t the photodetector, dark pupil images are captured by the single-pixel camera. This allows the application of state-of-the-art \ac{vog} algorithms for pupil tracking and gaze estimation. 
\subsubsection{Results}
Reusing the existing \ac{mems} scanner leads to the comparable low power consumption of the proposed sensor of roughly 11\,mW mainly driven by the photodetector circuitry and the \ac{ir} laser. In addition, the holographic free-space combiner of the single-pixel camera enables virtual observation of the eye from a centered perspective through the glasses lens while not obscuring the user's \ac{fov}. This is a superior perspective for \ac{vog} algorithms as the pupil can be observed in a wide range of pupil positions. Finally, the system's output allows reusing state-of-the-art \ac{vog} algorithms. Therefore, the proposed system achieves similar accuracy as state-of-the-art \ac{vog} systems and immediately benefits from advancements in \ac{vog} algorithm developments. The achieved gaze accuracy of the system was 1.31\,$^\circ$, and the precision 0.01\,$^\circ$. The theoretical achievable single-pixel gaze angle resolution estimated from the geometric system analysis was 0.28\,$^\circ$. The deviation between this theoretical gaze accuracy and the achieved gaze accuracy is related to the optical geometry of the laboratory setup, which leads to a large pixel size compared to the glasses geometry and a defocusing of the laser beam and thus to a reduction of the image contrast.

\subsection{Scanned LFI Eye Tracking for Retinal Projection  AR Glasses}
\label{MO:Scanned_LFI}
This subsection is based on the publications \ref{APP:A3} \textit{A novel-eye-tracking sensor for \ac{ar} glasses based on laser self-mixing showing exceptional robustness against illumination} and \ref{APP:A4} \textit{A Highly Integrated Ambient Light Robust Eye-Tracking Sensor for Retinal Projection AR Glasses Based on Laser Feedback Interferometry} in \Cref{APP:Scanned_Laser_Eye_Tracking}.
\subsubsection{Motivation}
Aside from power consumption and sensor integration, \ac{vog} systems further suffer from a signal loss in the presence of ambient light and a high-power consumption through the required image processing steps for pupil segmentation. In addition, other artifacts such as partly occluded pupils by the lashes or the eyelid and wearing mascara reduces the accuracy of \ac{vog} systems.

As AR glasses are everyday devices, robust operation and thus robust eye-tracking under various lighting conditions are mandatory. In addition, a similar robust gaze estimation for various pupil occlusions and mascara is required to achieve a great user experience for retinal projection systems with pupil steering. 
\subsubsection{Methods}
A similar optical architecture as proposed by the publications discussed in \Cref{MO:Scanned_Laser} is chosen. Only the \ac{ir} laser sensor inside the \ac{rgb} laser projection module is replaced with an \ac{lfi} sensor. The advantage of the \ac{lfi} sensor is its coherent measurement principle, which is robust against external lighting. This aspect is investigated in the paper \ref{APP:A3} by exposing the sensor to various ambient lighting sources. 

In \ref{APP:A4} the \ac{lfi} sensor is integrated into an \ac{ar} glasses prototype system in a laboratory setup to show the miniaturization potential as well as to investigate the system performance. In addition, a highly transparent \ac{ir} \ac{hoe} as free space combiner was fabricated to redirect the \ac{ir} light of the \ac{lfi} sensor towards the eye.

Compared to the external off-axis single-pixel photodetector used in the previous approaches the \ac{lfi} sensor consists of an \ac{ir} laser and a on-axis photodetector integrated into the back \ac{dbr} of the laser. Therefore the photodetector is perfectly aligned with the illumination source, emphasizing the detection of the pupil through the bright pupil effect. Furthermore, the bright pupil response reduces the complexity of pupil segmentation algorithms as the pupil contour is immediately derived from the detected signals of the \ac{lfi} sensor. As the bright pupil signal is corrupted by speckling, a multivariate Gaussian fitting approach was proposed to derive the pupil contour.
\subsubsection{Results}
The introduction of the \ac{lfi} sensing modality led to a robust pupil detection method that is immune to ambient light at eye-safe exposure limits and provides robust pupil signals even for partly occluded pupils. Furthermore, as only a bright pupil signal is detected, the approach is robust against lashes and mascara. The robust signals from the pupil further reduce the computational complexity of pupil segmentation algorithms and thus reduce the overall system power consumption. A laboratory experiment with 15 participants unveiled a gaze accuracy of 1.674\,$^\circ$ and a precision of 0.945\,$^\circ$ of the proposed system. 

\subsection{A Holographic Single-Pixel Stereo Camera Eye Tracking Sensor for Retinal Projection AR Glasses}
This subsection summarizes the publication \ref{APP:A5} \textit{A holographic single-pixel stereo camera eye-tracking sensor for calibration-free eye-tracking in retinal projection AR glasses}
 in \Cref{APP:Scanned_Laser_Eye_Tracking}.
\subsubsection{Motivation}
A significant challenge for \ac{vog} eye-tracking sensors in everyday devices such as \ac{ar} glasses is the degradation of gaze accuracy due to glasses slippage \cite{niehorster2020impact}. The reduction in gaze accuracy when the glasses slip is mainly because gaze estimation algorithms assume a known static mapping function $f$ between glasses- and the eye coordinate system, which is stationary over time. This mapping function $f$ is derived from an initial system calibration performed by the user. If the glasses slip, $f$ is no longer valid, and a recalibration is required \cite{10.1145/3314111.3319835}. Frequent recalibration by the user leads to a poor user experience. 
A possible solution to obtain $f$ without a calibration step is to use a stereo camera setup, e.g., used in the Tobii Pro glasses \cite{TobiiPro2021}. However, this requires the integration of two camera sensors per eye, which drastically increases the power consumption of \ac{vog} systems and makes them impractical for lightweight \ac{ar} glasses.
\subsubsection{Methods}
In this publication, a virtual single-pixel holographic stereo camera is porposed to achieve calibration-free, slippage-robust eye-tracking for the previously proposed scanned laser eye-tracking approaches, while keeping the advantgeous sensor integration and low power consumption. The main idea to create a stereoscopic perspective of the user's eye in a near-eye display relies on the spatial multiplexing of the \ac{hoe}, which is used as a free space combiner in retinal projection systems. 

By scanning over the spatial multiplexed \ac{hoe}, two images of the eye from two perspectives generated by two optical functions embedded in the spatial multiplexed \ac{hoe} are captured sequentially. By applying a cone reconstruction algorithm, the pupil disk can be reconstructed from the two images in space with respect to the origin of the virtual camera. Therefore, the mapping $f$ is derived from each pair of images. As the reconstruction algorithm relies only on a set of algebraic equations and omits demanding computational operations, a power-efficient implementation on \ac{ar} glasses is reasonable. A spatial multiplexed \ac{hoe} was fabricated using a holographic wavefront printer to assess the gaze accuracy of the proposed system experimentally.
\subsubsection{Results}
With the fabricated spatial multiplexed \ac{hoe} the robustness of the proposed approach was investigated using a laboratory setup. As a result, a gaze accuracy of 1.35$^\circ$ and a precision of 0.02$^\circ$ over a \ac{fov} of 30$^\circ$ was achieved. The resolution of the individual camera sensors was 110\,px $\times$ 240\,px with a frame rate of 60\,Hz. In addition, the reconstruction of the pupil disc in space using an artificial eye model was shown.

\section{Static LFI Human-Computer Interaction for AR Glasses}
The last part of the thesis focused on eye-tracking methods tailored to retinal projection \ac{ar} glasses. However, aside from retinal projection display technology, other near-eye-display technologies like wave-guide displays or micro-\ac{led} displays exist. In addition, audio glasses such as the Amazon Echo frames \cite{amazon2019} are emerging to the market without a near-eye display. Therefore, gaze-based interaction concepts based on static \ac{lfi} sensors are investigated through the contributions \cite{meyer11788compact,Meyer2021,CNN_Meyer_2021, 10.1145/3530884} in this part of the thesis to enable gaze-based \ac{hci} concepts for \ac{ar} glasses independent of the display technology. 
\subsection{A Novel Gaze Gesture Sensor for Smart Glasses Based on Laser Feedback Interferometry}
This subsection is based on the publications \ref{APP:B1} \textit{A Novel Gaze Gesture Sensor for Smart Glasses Based on Laser Self-Mixing}
 and  \ref{APP:B2} \textit{A compact low-power gaze gesture sensor based on laser feedback interferometry for smart glasses} in \ref{APP:Static_LFI_HCI}.
\subsubsection{Motivation}
Various \ac{hci} input methods such as push buttons or capacitive sliders have been adapted from existing smart wearable technologies such as mobile phones to the field of smart glasses. The main drawback is that they do not allow hands-free interaction with the glasses, which reduces the immersion of \ac{ar} glasses and limits the user experience, especially during cycling or driving. Aside from device interaction purposes, low-power always-on input methods like a push-button on a smartphone are used to control the display to save power and thus extend battery life. Since the display in \ac{ar} glasses like the smartphone accounts for the tremendous power demand, implementing a low power always-on input modality is also beneficial from a power-saving perspective. 
\subsubsection{Methods}
Static \ac{lfi} sensors are investigated to achieve low power always-on gaze gesture control for \ac{ar} glasses. Static \ac{lfi} sensors are capable of measuring the eye's rotational velocity as well as the distance between the sensor and the eye through a triangular modulation of laser wavelength, leading to a near range \ac{fmcw} \ac{lidar}. Due to the small size, high robustness against ambient light, and low power consumption, the sensor perfectly fits the requirements of AR glasses. From measured raw eye velocities, eye movement directions into the four quadrants up, down, left, and right were derived using a decision tree to classify four gaze symbols. In addition, blinks were classified from the distance measurement signal yielding a fifth gaze symbol. Finally, aside from classifying individual gaze symbols, an \ac{fsm} was proposed to model gaze gestures as a sequence of individual gaze symbols.
\subsubsection{Results}
A laboratory setup was built up to evaluate the proposed static \ac{lfi} gaze gesture sensor. Velocity and distance features from two participants performing in a total of 162 gaze symbols were captured together with a camera sensor used to derive ground truth labels on a single sample level. As a result, the proposed gaze symbol classifier achieved a macro-F1-score of 93.44\,\% on a single sample scale. Furthermore, as the static \ac{lfi} sensors are capable of measuring distance and velocity features at a sampling rate of 1\,kHz a gaze gesture was recognized before the user finished the eye movement belonging to the gaze gesture, leading to a negative latency of about 100\,ms. This allowed the user interface to start rendering interaction-dependent content even before the user finished the gaze gesture input.

\subsection{A CNN-based Human Activity Recognition System for context-aware Smart Glasses}
This subsection is based on the publications \ref{APP:B3} \textit{A CNN-based Human Activity Recognition System Combining a Laser Feedback Interferometry Eye Movement Sensor and an IMU for Context-aware Smart Glasses} and \ref{APP:B4} \textit{U-HAR: A Convolutional Approach to Human Activity Recognition Combining Head and Eye Movements for Context-Aware Smart Glasses}  in \Cref{APP:Static_LFI_HCI}.
\subsubsection{Motivation}
State-of-the-art \ac{ar} glasses rely on interaction concepts derived from existing wearable technology like push buttons, capacitive sliders, hand gesture control, or voice control. However, these interaction methods rely on active interaction with the glasses, distracting the user and limiting the immersion of the \ac{ar} glasses. To resolve this issue and enable true user immersion for \ac{ar} glasses adding context awareness through recognition of human activities is proposed to control the user interface without active interaction. While the recognition of physical activities through a body-worn motion sensor or the recognition of cognitive activities through  an eye-tracking sensor, are well established in research, the combination of both sensor modalities is less well explored. The combination of both sensors allows recognition of a rich set of human activities to infer a rich set of contextual information during everyday activities. 
\subsubsection{Methods}
The input of the \ac{har} system used to derive contextual information is eye- and head movement trajectories over time. The head movement trajectories are detected by an \ac{imu} sensor attached to the glasses frame. In contrast, the trajectories of the eyes in \ref{APP:B3} are detected by static \ac{lfi} sensors, as also used in the previous section. In \ref{APP:B4}, the static \ac{lfi} sensors are replaced by a commercially available \ac{vog} sensor to highlight the operation of the \ac{har} system independent of the eye movement sensor, thus enabling context awareness through \ac{har} e.g., \ac{vr} glasses with already integrated camera-based eye-tracking sensors.

To recognize human activities from the captured eye- and head movement trajectories, a windowing approach was chosen to split raw data into slices of 30\,s from which features as input of a \ac{har} classifier are derived. Furthermore, a \ac{cnn} model was proposed to automate feature extraction. A \ac{cnn} model allows the extraction of features on different time scales and reduces the input data size and thus the required system memory. Compared to other network architectures, \ac{cnn} models further require a comparatively small number of parameters, which is beneficial primarily for integrating the \ac{har} system into memory-constrained embedded systems. In \ref{APP:B4}, the number of model parameters was further optimized by adopting a U-Net like \ac{cnn} model for the \ac{har} task. 

To personalize the \ac{har} classifier to each user and thus reduce false classification, transfer learning was applied to adapt the decision boundaries of a pretrained \ac {cnn} model.

Finally, an experimental setup was the build-up to collect head- and eye-movement data from a set of participants, which performed the seven activities cycling, walking, talking, reading, typing, solving, and watching media. The activities were chosen to represent activities spanning from cognitive activities like reading to varied activities like talking involving both head and eye movements, as well as physical activities like cycling.
\subsubsection{Results}
By using state-of-the-art \ac{cnn} models for activity classification and the adaption of the decision boundaries through transfer learning an macro F1-score of 88.15\,\% and 86.59\,\% was achieved for \ref{APP:B3} and \ref{APP:B4} respectively. The evaluation was carried out by using leave-one-participant-out cross-validation \ac{lopocv}. Both \ac{har} systems outperformed existing works on a more challenging set of activities. They showed generalization across users by evaluating the \ac{har} systems of both works on a set of 15 and 20 participants, respectively. In addition, the importance of head- and eye movements for different activities was highlighted through an ablation study in both publications. The study supports the initial idea that the combination of both sensor modalities enhances \ac{har} across various activities and thus enables context awareness through \ac{har} for \ac{ar} glasses.

\section{Static LFI Eye Tracking}
The second part of this thesis introduced static \ac{lfi} sensors for \ac{hci} e.g., to recognize gaze gestures from measured relative eye movements. In this part of the thesis, the initial approach of static \ac{lfi} sensors is extended to achieve full eye-tracking with static \ac{lfi} sensors. The main advantages of the approach are the high update rate, the ambient light robustness, and a, compared to low power consumption in a \ac{vog} system. In addition, the proposed system operates independent of the display technology and thus can be used in \ac{ar} and \ac{vr} glasses.

This part is based on the publication \ref{APP:C1} \textit{Static Laser Feedback Interferometry Gaze estimation} in \Cref{APP:Static_LFI_Eye_Tracking}.

\subsection{Motivation}
\ac{vog} systems are limited in their update rate due to limitations of camera technology as well as the exponential increase of power consumption with respect to the update rate due to the camera sensor itself as well as the image processing required to infer gaze information from images  In addition, \ac{vog} systems suffer from obtrusive sensor integration, loss of gaze estimation accuracy in dynamic lightning conditions and degradation of gaze accuracy in the presence of glasses slippage  To overcome these limitations, this publication extends the gaze gesture approach. It introduces a model-based eye-tracking approach optimized for static \ac{lfi} sensors.
\subsubsection{Methods}
Like \ac{vog} model-based eye-tracking approaches, a geometric model of the human eye is used to link \ac{lfi} measurements to the pose of the eye. For this purpose, a measurement setup was built up, and the human eye of a participant was scanned with a static \ac{lfi} sensor to derive the geometric eye model from \ac{lfi} distance measurements. The experiment showed that most light of the IR laser beam is backscattered from the sclera, iris, and retina. Thus, the transparent optical components like the lens or the cornea can be omitted in the geometric eye model.

A static \ac{lfi} simulation tool was introduced to generate measurement trajectories of a multi-static \ac{lfi} configuration and avoid human error. \ac{lfi} sensors were modeled as point sources, and the human eye model was derived from a 3D scan of a human eye. The derived 3D model is rotated in a 3D \ac{cad} program based on an input eye movement trajectory ($\theta, \phi$). At each trajectory point, the measured distance and velocity for each modeled \ac{lfi} sensor are calculated from the 3D model. In addition, a sensor noise model is added to the simulation tool to consider sensor noise.

An algorithm consisting of four stages was proposed to estimate the gaze vector from a set of static \ac{lfi} distance and velocity measurements. In the first stage, the part of the eye (none, sclera, iris, retina) hit by an individual \ac{ir} laser beam of an \ac{lfi} sensors were classified by using the distance measure and the distance difference as input features for an \ac{hmm} classifier. In the second stage, the eyeball center is estimated via trilateration from the distance measures of at least three \ac{lfi} sensors. The use of trilateration allows for continuous reestimation of the eyeball center. This reestimation adds slippage robustness to the proposed method. In the third stage, the gaze angle of the eye is estimated through continuous integration of measured rotational velocities. To compensate for errors introduced by sensor noise and an unknown initial pose of the eye frequent absolute eye pose estimation is added as a final stage to the gaze estimation algorithm. Distance measurements are fitted to the derived geometric eye model during this step. 
\subsection{Results}
A laboratory setup was built up to characterize distance and velocity in a range of 20\,mm - 30\,mm and 0\,$^\circ/s$ - 500\,$^\circ/s$ to set up the sensor noise model required for simulation of multi \ac{lfi} laser sensor distance and velocity measure trajectories. A distance noise of 68.66\,$\mu$m, and a velocity noise of 2.5 $^\circ$/s was measured. In addition, the classification accuracy of the proposed \ac{hmm} model to classify the part of the eye hit by the laser was evaluated, and a macro F1-score of 93.33\,\% was reported. Finally, an actual eye trajectory, including different eye movements from fixations to saccades, was captured by an \ac{vog} system and fed into the simulation tool to generate distance and velocity measurement trajectories for six static \ac{lfi} sensors.  Based on the generated trajectories the gaze estimation algorithm was evaluated. A gaze accuracy of 1.79\,$^\circ$ was achieved.

\section{Summary}
\begin{figure}[h!]
    \centering
    \includegraphics[width=0.8\linewidth]{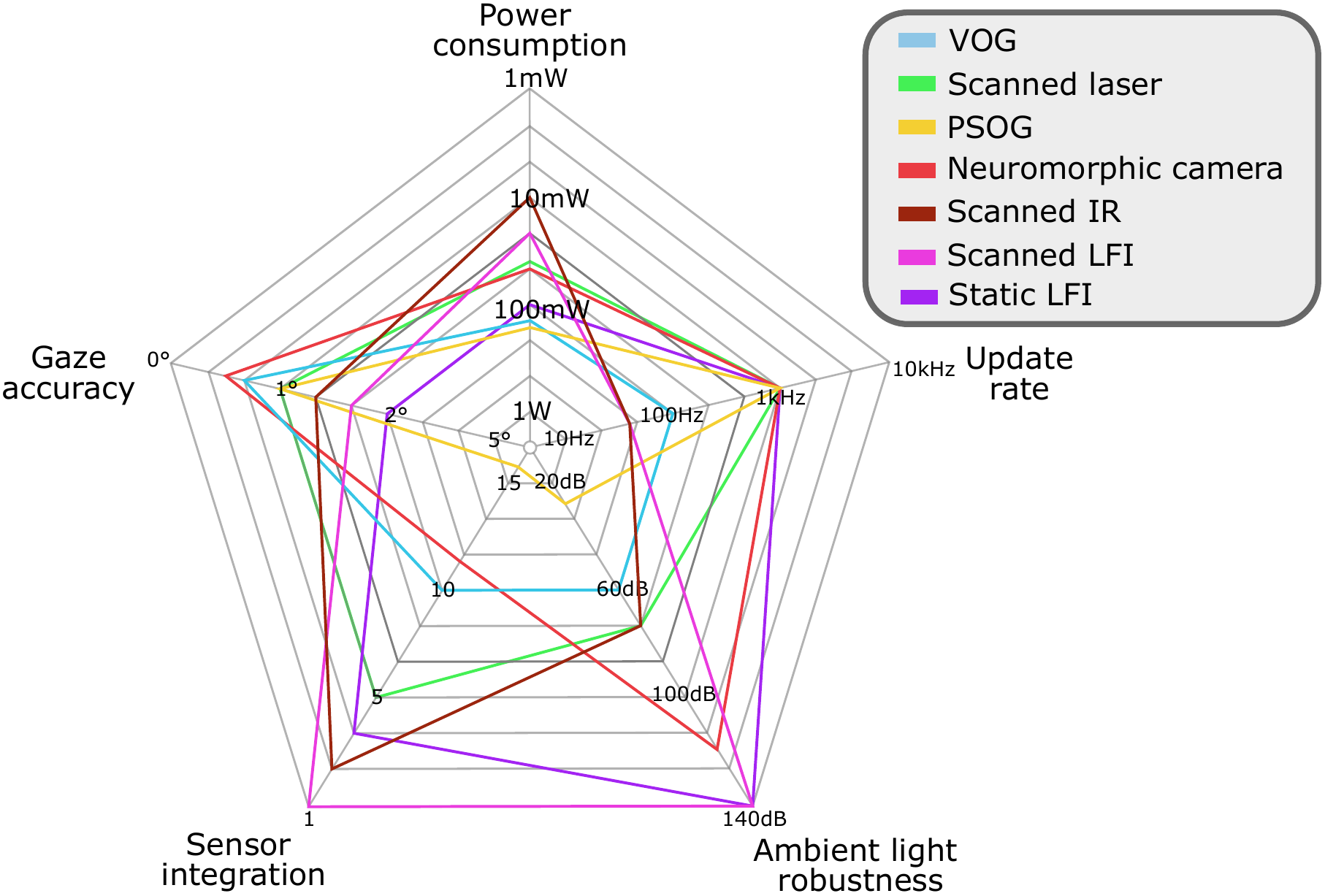}
    \caption[Comparison of the presented three developed eye-tracking sensor approaches]{Comparison of the presented three developed eye-tracking sensor approaches compared to the state-of-the-art eye-tracking sensor technologies.}
    \label{CON:fig:Overview_Contribution_all}
\end{figure}

To summarize the contributions made through this thesis, the radar chart from \Cref{MET:subsec:Comparison} is advanced by the eye-tracking sensor approaches contributed by this thesis. The resulting radar chart is shown in \Cref{CON:fig:Overview_Contribution_all}.
\FloatBarrier
The following contributions within this thesis are summarized for each part in the next three subsections.
\subsection{Scanned Laser Eye Tracking}
The first part of the thesis contributes a scanned laser eye-tracking sensor technology for integration into retinal projection \ac{ar} glasses. The approach closes the gap between established \ac{vog} systems by capturing 2D images of the eye with a highly integrated single-pixel camera setup.

Within the \textbf{publications \ref{APP:A1} and \ref{APP:A2}} the \textit{Scanned IR} approach is presented, which rely on a 2D scanner and an off-axis photodetector to capture dark pupil images. The main advantages are the high sensor integration and the redirection of the light via an \ac{hoe} to create a superior perspective of the virtual camera compared to state-of-the-art \ac{vog} sensors. Therefore, these publications address the technological challenges of \textbf{camera sensor integration} as well as \textbf{power consumption}. 

Within the \textbf{publications \ref{APP:A3} and \ref{APP:A4}} the \textit{Scanned LFI} approach is presented, which extends the previous publications of this part of the thesis by using an \ac{lfi} sensor and its integrated in-axis photodiode to capture bright pupil images. This approach's main advantage is exploiting the red-eye effect to capture the bright pupil directly. Therefore, the sensor segments the pupil in the captured image and shows a robust pupil signal independent of the iris color or pupil occlusion through lids or lashes, which further reduces the algorithm power consumption compared to \ac{vog} systems. In addition, the sensor operates coherently and thus is immune to ambient light. Within these two publications, the technological challenges of \textbf{ambient light robustness} and \textbf{power consumption} are addressed.

Within the final \textbf{publication \ref{APP:A5}} of this part of the thesis, a methodology is contributed to achieve slippage robustness and minimize the calibration effort of the previously introduced approaches \textit{Scanned IR} and \textit{Scanned LFI} by spatial multiplexing of the \ac{hoe} to capture the eye from two perspectives. This results in a single-pixel stereo camera system, which allows reconstruction of the pupil disk in 3D from a single pair of images. Therefore this publication addresses the technological challenges of \textbf{glasses slippage} and \textbf{calibration}.

\Cref{CON:fig:Overview_Contribution_all} summarizes the \textit{Scanned IR} and \textit{Scanned LFI} approaches with respect to the requirements derived in \Cref{subsec:Requirements_for_eye_tracking_sensors_in_AR_glasses}. Especially the \textit{Scanned LFI} method shows superiority w.r.t sensor integration and ambient light robustness compared to all investigated technologies. In addition, the scanned laser sensor approaches achieve a comparable low power consumption. The theoretical system analysis in \ref{APP:A2} further shows that from a geometric point of view, a maximum gaze accuracy of 0.28$^\circ$ is achievable in a near-eye setting. Therefore, improvements regarding gaze accuracy are expected in the future.

\subsection{Static LFI Human-Computer Interaction for AR Glasses} 
In the second part of the thesis, static \ac{lfi} sensor technology for gaze interaction applications is contributed. The sensor modality allows deriving the unique features \textit{distance} towards the eye and the eye's rotational \textit{velocity} independent of the display technology with an outstanding sample rate of 1\,kHz.

Within the \textbf{publications \ref{APP:B1} and \ref{APP:B2}} gaze gesture interaction with this sensor modality is investigated. As the sensors consume only a fraction of the power of \ac{vog} sensors, they can be used as always on gaze gesture interaction sensor, e.g., for low power system wake up my gaze. Due to the high sample rate, it is further possible to classify a gaze gesture before the user finishes its execution leading to a negative sensor latency, which relaxes the system rendering constraints.

Within the \textbf{publications \ref{APP:B3} and \ref{APP:B4}} the combination of gaze and head movement interaction are investigated to achieve context-awareness for \ac{ar} glasses. Context-awareness is derived through \ac{har} based on eye and head movements. For the collection of eye movements, the \ac{lfi} sensor is used as a power-efficient alternative to \ac{vog} sensors. During an experimental study, data for seven activities is collected, and a classification approach to fuse head- and eye movement data with an \ac{cnn} model is contributed. The same experiment is also made with an \ac{vog} sensor to show the generalization of the approach to existing \ac{ar} glasses with \ac{vog} sensors integrated. The collected second dataset is published to emphasize research within context-aware \ac{ar} glasses through \ac{har} as an additional contribution to this part of the thesis.

\subsection{Static LFI Eye Tracking}

The main contribution of the \textbf{publication \ref{APP:C1}} in this part of the thesis is a novel high-speed eye-tracking sensor approach based on highly integrated, ambient light robust static \ac{lfi} sensors. Compared to \ac{vog} systems, the \ac{lfi} sensors did not resolve the eye's surface in 2D as they have no spatial perception. This reduces the amount of captured data, which allows a higher update rate while reducing power consumption. With the introduction of the multi \ac{lfi} sensor fusion algorithm based on a geometric model and spatial perception of the sensor is achieved by integration of observations over time. This allows obtaining absolute gaze estimation for static \ac{lfi} sensors, including slippage robustness. Therefore within this publication the technological challenges of \textbf{power consumption}, \textbf{glasses slippage}, \textbf{update rate} and \textbf{ambient light robustness} are addressed.

In \Cref{CON:fig:Overview_Contribution_all} the \textit{Static LFI} eye-tracking sensor technology developed within this part of the thesis is summarized with respect to the derived requirements for eye-tracking sensor technology for \ac{ar} glasses within \Cref{subsec:Requirements_for_eye_tracking_sensors_in_AR_glasses}. The sensor technology achieves comparable update rates to emerging neuromorphic camera eye-tracking approaches while outperforming other sensor technologies regarding ambient light robustness.

\cleardoublepage
\chapter{Discussion}
While the introduction of mobile eye-tracking for everyday devices enables new applications and opportunities to improve the immersion of \ac{ar} glasses, e.g., through gaze-controlled interaction, it poses some risks to the user. The main risks are the medical impact of long-term \ac{ir} exposure to the human eye and the impact on the user's privacy due to everyday eye-tracking. These aspects will be discussed in the following sections.
\section{Health Impact of long-term IR Radiation on the Human Eye}
The human eye is exposed to \ac{ir} illumination regularly as the heat transfer from the sun to the earth occurs in the \ac{ir} regime. While the human organism can handle this natural \ac{ir} exposure. Recently, a rising number of artificial \ac{ir} light sources like \ac{lidar} sensors on cars or \ac{ir} flooding illumination and \ac{ir} dot projectors used for face recognition on smartphones are introduced. Finally, \ac{vog} eye-tracking systems rely on \ac{ir} LEDs, e.g., to illuminate the eye region or produce glints on the cornea. While \ac{lidar} sensors and \ac{ir} illumination used for face recognition on smartphones emit artificial \ac{ir} light only for a relatively short duration, state-of-the-art \ac{vog} systems with flooding \ac{ir} illumination for AR glasses expose the eye with artificial \ac{ir} illumination on a daily basis over a long period of time \cite{COGAIN2008}.

To address the potential risk of \ac{ir} illumination of \ac{vog} systems, the research network Communication by Gaze Interaction (COGAIN) \cite{COGAIN2008} initiated a study to explore safety issues in eye-tracking by \ac{ir} illumination. The study identified three potential optical radiation hazards sources through \ac{ir} illumination. The first source of the potential hazard to the eye is a thermal hazard to the cornea. This hazard mainly applies to \ac{ir} illumination in the IR-B band ranging from 1400\,nm - 1\,mm wavelength as the tissue of the cornea has high transparency for \ac{ir} wavelength in the IR-A band ranging from 780\,nm - 1400\,nm not absorbing much \ac{ir} energy. The second potential hazard source is a retina tissue thermal burn. According to Vos et al. \cite{RetinalBurn2005} the risk of a thermal burn of the retina exists primarily in the range from 400\,nm - 1400\,nm and thus including the IR-A band at which \ac{vog} \ac{ir} illumination sources are emitting. Finally, it is known from workers dealing with hot materials, e.g., molten glass over a long period, that there is a higher risk of developing cataract due to the intense \ac{ir} radiation, especially from the IR-A band \cite{Cataract1994}.

As a result of the study by \acs{cogain} the \ac{cie} issued in 2021 a technical report on the optical safety of eye trackers applied for extended duration \cite{CIE2021}. The authors analyzed the different sources of potential radiation hazards for \ac{ir} illumination. They pointed out that the most limiting exposure criterion is the \ac{ir} exposure limit, leading to cataracts in long-term exposure conditions. The study further analyzed different state-of-the-art \ac{vog} systems and showed that the emitted dose of \ac{ir} radiation for all tested \ac{vog} systems was below the \ac{ir} exposure limit which increases the risk to develop cataract \cite{CIE2021}.  

The scanned laser systems, as well as the static \ac{lfi} systems introduced in this thesis, rely on coherent \ac{ir} illumination from laser diodes with a compared to \acp{led} slight divergence angle. Therefore the authors in \cite{CIE2021} pointed out that the long-term exposure limits for laser illumination sources need to be derive from the \ac{iec} 60825-1 norm \cite{IEC2014}. To protect participants' health during the experiments with the proposed static and scanned systems for all experimental setups, a risk assessment according to \ac{iec} 60825-1 was made. For the experimental setups in this thesis, the most restricting exposure limits, which assume a daily dose of \ac{ir} radiation with a duration of >\,8h, were used to rate the eye safety.  

The exposure limit for the scanned systems with geometry as described in \cref{APP:A2} is according to \cite{IEC2014} 9.25\,mW for a wavelength of 820\,nm and exposure duration of >\,8h. Compared to the exposure limit of 677\,$\mu$W  for a static laser of the same wavelength is because according to \cite{IEC2014} scanning systems are rated as pulsed systems. The scanned \ac{ir} system introduced in \ref{APP:A1} and \ref{APP:A2} used a \ac{ir} laser with an optical power of 150\,$\mu$W on the eye surface and the scanned \ac{lfi} systems introduced in \ref{APP:A3} and \ref{APP:A4} used an \ac{ir} laser with an optical power of 142\,$\mu$W on the eye surface. Therefore, both scanned laser systems introduced in this thesis stay below the exposure limit of a static \ac{ir} laser, ensuring eye safety even in fault conditions, e.g., when the \ac{mems} micro scanner stops scanning the laser beam. The scanned \ac{lfi} approach introduced in \ref{APP:A3} and \ref{APP:A4} further shows a robust signal in the presence of low-level illumination. While other \ac{vog} systems tend to increase the optical power of their \ac{ir} flooding illumination in low lighting conditions, the scanned \ac{lfi} system did not require to increase in the optical power.

The exposure limit for the static \ac{lfi} approaches discussed in the publications \ref{APP:Static_LFI_HCI} and \ref{APP:Static_LFI_Eye_Tracking} also stay at their operating wavelength of 850\,nm below the exposure limit of 778 $\mu$W  for each sensor. In addition, the mechanical design of the demonstrators and laboratory setups built during this thesis ensured that the combination of multiple \ac{lfi} sensors did not violate exposure limits. Therefore, the static \ac{lfi} approaches introduced in this thesis did not pose any hazard to the participants.

From a product development perspective, it cannot be ensured that for different head geometries or in the case of glasses slippage, the laser beams of multiple \ac{lfi} sensors did not hit the same spot and thus violate the exposure limits. Therefore, time multiplexing of the individual sensors is required to ensure eye safety for a product. Time multiplexing does not harm the performance of the sensors as the current update rate of 1\,kHz can be preserved by slight adjustments on the triangular modulation rate of the individual static \ac{lfi} sensors.

To finally rate the eye safety of an \ac{ar} glasses, it is further required to ensure that the exposure limit of all visible and invisible light and laser sources combined stays below the exposure limits. Therefore, carefully considering eye safety on a system level is mandatory.

\section{Privacy Concerns of Mobile Eye Tracking}
Privacy concerns in the domain of mobile eye-tracking can be divided into concerns regarding user authentication and biometric identification as well as information obtained through the linkage between gaze- and environment information. 

\subsection{User Authentication and Biometric Identification}
The advancement of sensor technology for the acquisition of gaze information inevitably leads to the extraction of features from the eye usable for user authentication and occular-based biometric identification \cite{10.1145/3313831.3376840}.

Ocular-based biometric identification relies mainly on high-quality images of the user's eye to derive structural features of the iris or on high-quality scans of the retina to derive structural features such as the retina's blood vessels for retina recognition \cite{BiometricBook}. These methods allow the identification of a single user and are therefore reliable identification methods for security-critical applications \cite{BiometricBook}. 

The scanned laser eye-tracking method developed in this thesis can capture images of the eye, and the scanned \ac{lfi} method can capture images of the retina. 

According to \cite{IEC19794} an image resolution of at least 200 pixels across the iris and a \ac{snr} of at least 40\,dB is necessary to gather enough information from the iris for biometric identification. Both aspects are not fulfilled by the scanned laser method as introduced in \ref{APP:A1} and \ref{APP:A2}.  

Retinal recognition systems rely on high-quality retina images to derive individual user features from the blood vessels \cite{BiometricBook}. However, the scanned \ac{lfi} method introduced in \ref{APP:A3} and \ref{APP:A4} is not capable of resolving individual blood vessels of the retina as the interference signal of the \ac{lfi} sensors are superimposed by speckle patterns. 

In contrast, user authentication approaches perform recognition of a user in a closed set of N users, which is compared to biometric identification as less complex \cite{todorov2007mechanics}. Therefore, user authentication using eye-tracking features like number and duration of fixations, eye movement velocities, and amplitudes or blink frequency can be used to identify users with a certain accuracy \cite{10.1145/2904018}. 

The high update rate of 1\,kHz together with the capability to direct measure eye velocities, makes the static \ac{lfi} method introduced in \ref{APP:B1}  a beneficial sensing modality to support gaze-based user authentication for \ac{ar} glasses \cite{PRASSE20202088}. Therefore this aspect must be considered during development in the future.

\subsection{Linkage between Gaze- and Environment Information}
The energy-efficient eye-tracking methods presented in this thesis enable ubiquitous eye-tracking in everyday scenarios. However, the gaze information that is thus available in a previously impossible quantity poses an additional risk to the user's privacy, especially if gaze information is linked to environmental information, e.g., derived from a world camera. The linkage between the environment and gaze information enables applications like target advertisement \cite{hun_kyung_jae_2018} as it can be used to derive private information like user's interest, attention, or other sensitive information like gender, age, or race \cite{10.1145/2638728.2641688,Kroger2020}. 

Steil et al. \cite{10.1145/3314111.3319913} presented a potential solution to decouple environment and gaze information. The authors used a shutter in front of the world camera sensor, which is closed in privacy concerning situations like talking or reading messages on the smartphone. To recognize privacy concerning situations, they used the eye movement patterns captured by an \ac{vog} system. The publications \ref{APP:B3} and \ref{APP:B4} of this thesis can be used to improve the recognition of privacy concerning situations by adding context awareness to allow the user to define privacy concerning situations in which the world camera is disabled. 

In addition, the system design can restrict the use of gaze information. For example, retina projection systems require gaze information for display enhancement methods like exit pupil steering or eye box switching. However, these applications are realized within a closed control loop inside the projection system. Therefore, it is not required to output gaze information outside the projection system. 

Finally, the publications \ref{APP:B1} and \ref{APP:B2} of this thesis show that relative eye velocities are sufficient to interact with the  \ac{ar} glasses interface without complete gaze information. 

\subsection{Gaze based User State Estimation}
Besides biometry and the linkage between gaze- and environment information, the user's gaze reveals additional user-sensitive information, which can be exploited on a large scale in an everyday eye-tracking setting. This information can be summarized as the user's state, containing the user's emotion \cite{lim2020emotion}, cognitive load \cite{10.1145/3313831.3376394,krejtz2018eye}, user's attention \cite{8082802} as well as medical features \cite{leigh2015neurology}. 

This information can be used either in a user-beneficial way, e.g., through early detection of medical issues, or in a non-beneficial way, e.g., through target advertising based on the user's emotion or attention. 

Features from which the user's state is derived are pupil diameter variations, the user's scan path, the duration and number of fixations, (micro) saccades, and blink information. Although the scanned laser methods measure the pupil diameter variation with a low update rate and thus emotion recognition is rather complicated, the static \ac{lfi} methods can resolve the eye movement-related features with a high update rate.
Thus, the static \ac{lfi} approaches pose a potential risk for gaze-based user state estimation. A potential miss-use of the sensors needs to be prevented, e.g., by adding a differential privacy approach \cite{bozkir_diff_priv_2020}.

\cleardoublepage
\chapter{Outlook}
Energy-efficient and robust eye-tracking is a key technology required for the success of \ac{ar} glasses. Furthermore, eye-tracking is crucial for novel hands-free interaction concepts with the glasses and furthermore enables advanced near-eye display technologies like retinal projection. The contributions of this thesis address the limitations of state-of-the-art eye-tracking sensor technology for \ac{ar} glasses, and different approaches are proposed to solve individual short comes of existing eye-tracking sensors. Furthermore, the optical sensing technology approaches for eye-tracking provided in this thesis can be advanced in several ways, which will be discussed in the following.
\section{Scanned Laser Eye Tracking}
\begin{figure}
    \centering
    \includegraphics[width=\textwidth]{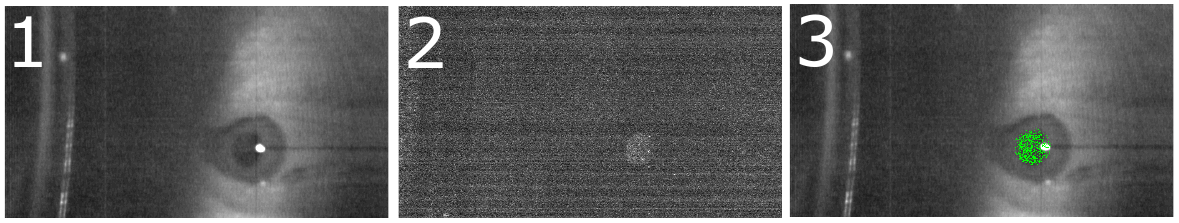}
    \caption[Combination of scanned laser and scanned lfi eye-tracking]{1) Scanned laser image of the eye taken by external off-axis photodetector, 2) Scanned \ac{lfi} image taken by the laser cavity integrated on-axis photodetector, 3) Scanned laser image with scanned \ac{lfi} image overlaid in green}
    \label{CON:fig:Scanned_LBS}
\end{figure}
The main difference between the scanned laser eye-tracking sensor approach described in \ref{APP:A1} and \ref{APP:A2} and the scanned \ac{lfi} eye-tracking approach described in \ref{APP:A3} and \ref{APP:A4} is the position of the photodetector.  By combining both sensing methods,  the bright as well as the dark pupil effect, can be captured in the same coordinate space as shown in \Cref{CON:fig:Scanned_LBS}. This approach yields a dual bright- and dark pupil eye tracker. The pupil response signal from both sensors can be combined to increase the signal strength and improve pupil detection accuracy. In addition, the scanned laser image can be used to track landmarks on the eye to compensate for glasses slippage.

An alternative to compensate for slippage is shown in \ref{APP:A5} where a single-pixel stereo camera approach based on laser scanning is introduced. This approach works for both the scanned laser and the scanned \ac{lfi} approaches. The two perspectives of the stereo camera are created by spatial multiplexing of the \ac{hoe} in the horizontal dimension. Furthermore, by spatial multiplexing of the \ac{hoe} in the vertical dimension as well, a multi-camera setup can be realized. This can be used to increase further the covered \ac{fov} of the scanned laser eye-tracking sensors or to increase the robustness and accuracy of gaze estimation.
\section{Static LFI for Human-Computer Interaction for AR Glasses}
The distance and velocity features of the static \ac{lfi} sensors used for gaze gesture recognition in \ref{APP:B1} and \ref{APP:B2} and \ac{har} in \ref{APP:B3} are derived from interference frequencies measured by the photodetector integrated into the laser cavity. The peak frequencies are derived from a distance spectrum obtained by an \ac{fft} of the captured interference signal recorded during a rising and falling triangular ramp. 
\begin{figure}
    \centering
    \includegraphics[width=\textwidth]{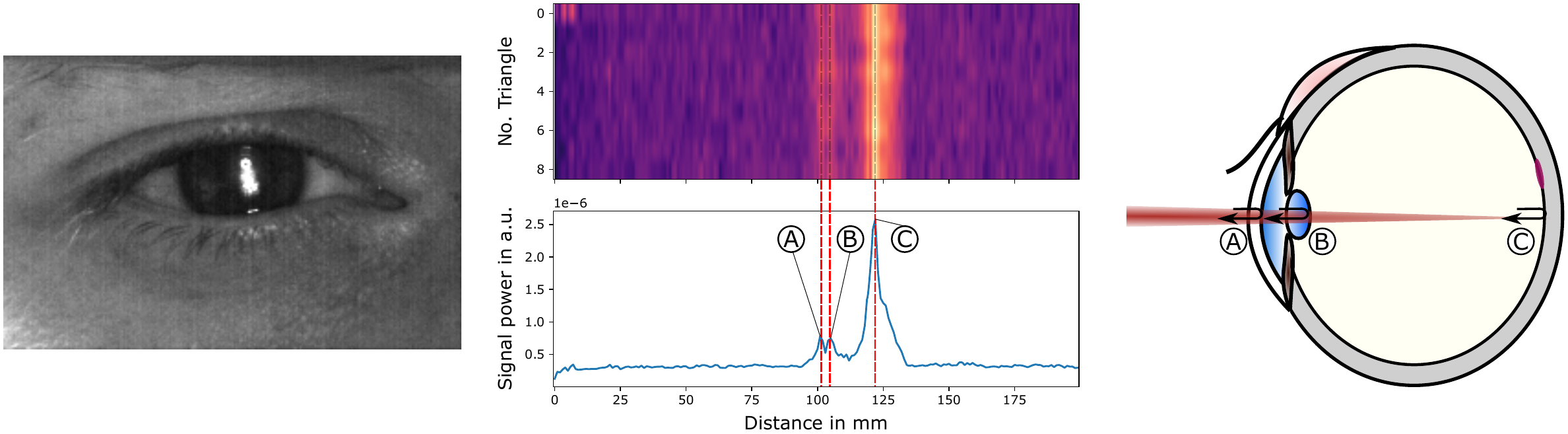} 
    \caption[Medical features in LFI spectra]{Left: Single static \ac{lfi} sensor \ac{ir} laser spot on the pupil of a participant captured with an IR camera, Center: Two subplots of the raw distance spectra (top) for a sequence of triangular modulation patterns and the averaged distance spectra (bottom) with three signal peaks A-C in the spectra. The frequencies on the x-axis are converted into corresponding distance measures. Right: Section of the human eye with the \ac{lfi} laser beam penetrating the cornea and the lens and finally hitting the retina. At each change of the refraction index, a part of the emitted light is back-reflected (A, B), and the remaining light is back-reflected from the retina (C).  }
    \label{CON:fig:Static_LFI_medical}
\end{figure}
\Cref{CON:fig:Static_LFI_medical} shows in the center the distance spectra measured by a static \ac{lfi} sensor hitting the pupil of a participant perpendicularly (white spot on the pupil in the left image). The upper subplot of the figure shows that frequencies of multiple interference's appear in the spectra. The lower subplot shows the signal strength of the spectra calculated for each frequency bin averaged over ten consecutive distance spectra obtained from 10 triangular modulation slopes. From the frequencies and the known \ac{lfi} sensor parameters, the corresponding distance can be calculated, and the individual peaks in the spectra (A-C) can be mapped to different parts (cornea, lens, retina) of the eye. From this observation, a further research direction is the applicability of the static \ac{lfi} sensor for medical or well-being applications in everyday devices. Especially if the sensor is integrated into AR glasses, this might allow early detection of biomarkers, which hint at eye disease like a cataract.
\section{Static LFI Eye Tracking}
The gaze estimation accuracy of the static \ac{lfi} eye-tracking method is particularly limited by the cases where the \ac{lfi} laser beams hit the sclera, as in this case, the gaze position is estimated by integrating velocity measurements, and an integration error builds up until measurement data can eliminate the integration error from the iris. One possible solution to increase the accuracy of the static \ac{lfi} eye-tracking approach is to combine it with a camera sensor operating at a low frame rate. This combines the high update rate and ambient light robustness of \ac{lfi} sensors with the high accuracy of camera-based eye-tracking systems while reducing the overall system's power consumption. Research towards hybrid eye-tracking systems through multi-sensor fusion offers the possibility to overcome the limitations of \ac{vog} systems such as limited sensor update rate and high-power consumption and is therefore of great importance for mobile eye-tracking for AR glasses.

\addtocontents{toc}{\vspace{\normalbaselineskip}}
\cleardoublepage
\bookmarksetup{startatroot}

\appendix
\markboth{Appendix}{Appendix}
\addcontentsline{toc}{chapter}{Appendix}
\chapter{Scanned Laser Eye Tracking}
\label{APP:Scanned_Laser_Eye_Tracking}
This chapter includes the publications \cite{10.1145/3379157.3391995, Meyer2020e,9149591, 10.1145/3530881, 10.1145/3517031.3529616}: \blfootnote{Publications are included with minor template modifications. Original versions are available via the digital object identifier at the corresponding publishers. Publications 1,3-5 are \copyright 2020 ACM and \copyright 2022 ACM respectively, and included with relevant permission. Publication 2 is \copyright 2020 IEEE and reprinted, with permission, from 2.  In reference to IEEE copyrighted material which is used with permission in this thesis, the IEEE does not endorse any of University of Tübingen's products or services. Internal or personal use of this material is permitted. If interested in reprinting/republishing IEEE copyrighted material for advertising or promotional purposes or for creating new collective works for resale or redistribution, please go to \url{http://www.ieee.org/publications_standards/publications/rights/rights_link.html} to learn how to obtain a License from RightsLink. If applicable, University Microfilms and/or ProQuest Library, or the Archives of Canada may supply single copies of the dissertation. }

\begin{enumerate}[label=\arabic*.]

	\item \textbf{Johannes Meyer}, Thomas Schlebusch, Thomas Kuebler, Enkelejda Kasneci. "Low Power Scanned Laser Eye Tracking for Retinal Projection AR Glasses". In ACM Symposium on Eye Tracking Research and Applications (2020), \url{https://doi.org/10.1145/3379157.3391995}  
	\item \textbf{Johannes Meyer}, Thomas Schlebusch, Wolfgang Fuhl, Enkelejda Kasneci. "A novel camera-free eye tracking sensor for augmented reality based on laser scanning". In IEEE Sensors Journal  (2020), \url{https://doi.org/10.1109/JSEN.2020.3011985} 
	\item \textbf{Johannes Meyer}, Thomas Schlebusch, Hans Spruit, Jochen Hellmig, Enkelejda Kasneci. "A novel-eye-tracking sensor for ar glasses based on laser self-mixing showing exceptional robustness against illumination". In ACM Symposium on Eye Tracking Research and Applications   (2020), \url{https://doi.org/10.1145/3379156.3391352} 
	\item \textbf{Johannes Meyer}, Thomas Schlebusch, Enkelejda Kasneci. "A Highly Integrated Ambient Light Robust Eye-Tracking Sensor for Retinal Projection AR Glasses Based on Laser Feedback Interferometry". Proc. ACM Hum.-Comput. Interact. 6  (2022), \url{https://doi.org/10.1145/3530881}  
	\item \textbf{Johannes Meyer}, Tobias Wilm, Reinhold Fiess, Thomas Schlebusch, Wilhelm Stork,  Enkelejda Kasneci. "A holographic single-pixel stereo camera eye-tracking sensor for calibration-free eye-tracking in retinal projection AR glasses".  In 2022 Symposium on Eye Tracking Research and Applications (2022), \url{https://doi.org/10.1145/3517031.3529616} \faIcon{trophy} Best Short Paper  
	
\end{enumerate}

\section{Low Power Scanned Laser Eye Tracking for Retinal Projection AR Glasses}
\label{APP:A1}
\subsection{Abstract}
Next generation AR glasses require a highly integrated, high-re\-so\-lu\-tion near-eye display technique such as focus-free retinal projection to enhance usability. Combined with low-power eye-tracking, such glasses enable better user experience and performance. This research work focuses on low power eye tracking sensor technology for integration into retinal projection systems. In our approach, a MEMS micro mirror scans an IR laser beam over the eye region and the scattered light is received by a photodiode. The advantages of our approach over typical VOG systems are its high integration capability and low-power consumption, which qualify our approach for next generation AR glasses.

\subsection{Research objectives}
The number of commercially available augmented reality (AR) glasses has largely increased in the last years. A new shift in the commercial domain is the integration of eye-tracking sensors into AR glasses to increase user experience by adding new functions and introducing new ways of interaction with systems around us \cite{7308147}. Examples for interactions with systems around us are driver assistance systems \cite{8082802}, human computer interaction (HCI) \cite{6227433} or smart home control \cite{s19040859}.

Additionally, the performance and resolution of AR applications can be increased by foveated imaging, where high resolution content is only projected sharply on the projection area corresponding to the fovea and lower-resolution content is used in peripheral regions. It is obvious that this technique is dependent on eye-tracking and can significantly reduce the required processing power for image rendering compared to a full-resolution rendering in the full field of view (FOV) \cite{Kim:2019:FAD:3306346.3322987}. A recent implementation of foveated imaging for VR applications was presented by Tobii \cite{Tobii2019}.

In this research work we focus on the integration of low power eye tracking sensors in AR-Glasses utilizing retinal projection as near-eye display technique. Retinal projection displays are based on direct illumination of the eye's retina using an ultra low-power eye-safe laser projector. Unique aspects of this display technology are a high degree of integratability into a frame temple, providing outstanding design opportunities, as well as near focus-free image projection, providing a sharp projection independent on the user's accommodation state \cite{Lin2017}.
\begin{figure}[!ht]
	\centering
	\includegraphics[width=0.4\linewidth]{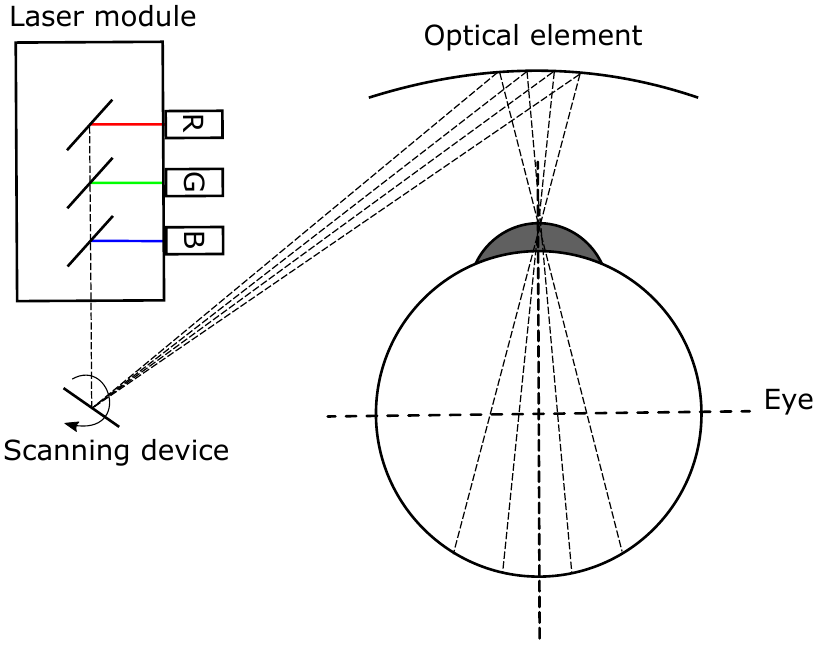}
	\caption{Principle structure and optical path of a retinal projection system. }
	\label{A1im:ret_scanner_basics}
\end{figure}
\FloatBarrier
\Cref{A1im:ret_scanner_basics} illustrates the basic structure and the main components of a retinal projection system. A laser module combines the laser beams of diode laser emitters for red, green and blue wavelengths. The combined beam is led to a scanning device which deflects the laser beam two-dimensionally onto a holographical optical element (HOE). This element redirects the beam towards the retina of the eye, where it forms a projected image.
For most retinal projection systems, eye-tracking is crucial as the scanned beam has to follow the pupil's position to enter the eye \cite{Jang:2017:RAR:3130800.3130889}.

\subsection{Hypothesis and problem statement}
Current eye-tracking sensors for low cost, commercially available systems, e.g. Pupil Labs \cite{Kassner:2014:POS:2638728.2641695}, use video-based oculography (VOG). These eye-tracking sensors rely on a set of infrared (IR) emitters to illuminate the eye and an IR camera to capture eye images. The gaze direction is then determined by image processing algorithms. 
While camera-based VOG systems are well-established and perform at high accuracy, there is only little potential for significant reduction of power consumption. Further, feasible orientation of the camera modules limits integration into a AR glasses frame and often interferes with the user's view \cite{10.1117/12.2322657,Katrychuk:2019:PSE:3314111.3319821}.

To solve the mentioned issues of current eye tracking technology new eye tracking sensor concepts with focus on a low power consumption and a high integratability are required. Furthermore, these sensors should be unobtrusive to ensure the user's comfort. 

\subsection{Approach and Methods}
The concept we investigated is the integration of an eye tracking sensor into a retinal projection AR glasses by exploiting the existing laser projection unit and the HOE required for the display of the glasses.

This concept extends previous scanning laser eye tracking sensor approaches by \cite{Brother2007} and \cite{7863402}. The main difference is the use of a laser scan path over the full region of the eye. This allows us to capture a full image of the user's eye and the use of robust slippage invariant state of the art VOG algorithms to extract the pupil position. The integration of the eye tracking sensor into a retinal projection display is shown in \Cref{A1im:ret_scanner_basics}.

\begin{figure}[h]
	\centering
	\includegraphics[width=0.4\linewidth]{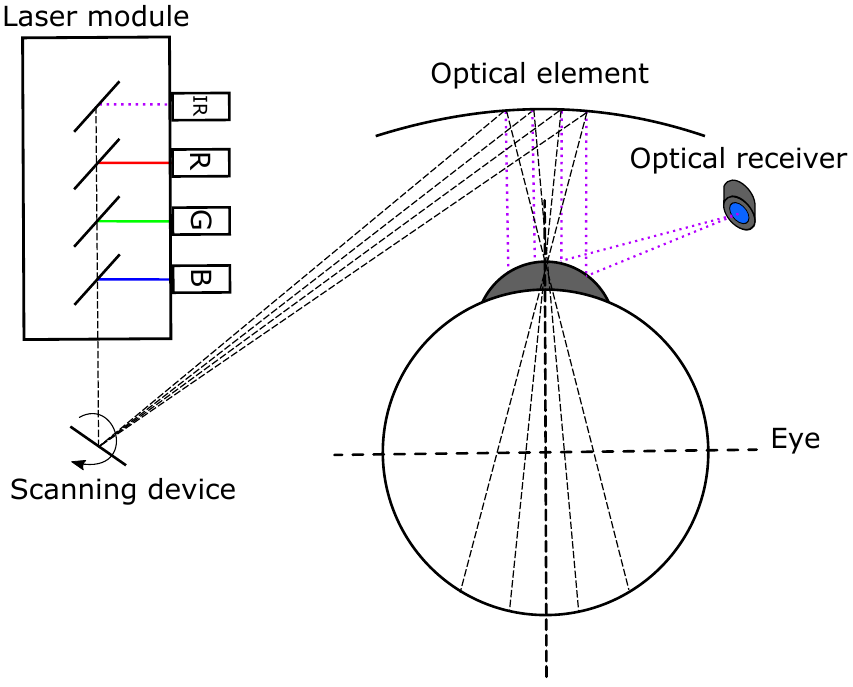}
	\caption{Principle structure and optical path of a retinal projector with integrated eye-tracking capabilities. An IR laser is integrated into the RGB laser module and an optical receiver is placed near the eye.}
	\label{A1im:ret_eyetracker}
\end{figure}
\FloatBarrier

 Beside the components of the retinal projector the added components are an IR laser, an optical receiver and an additional laser beam deflection function of the optical element specific to the IR wavelength of the eye-tracking laser. The integration of the new components into a retinal projection system are shown in \Cref{A1im:ret_eyetracker}.
The laser projection unit consists of a laser module and a scanning device. The laser projection module projects IR laser light (purple dotted line) onto the optical element. The optical element redirects the laser beam with a defined optical function (e.g. parabolic mirror) onto the surface of the eye. Based on the varying IR reflectivity of the different eye regions, the laser beam is scattered with a different intensity from the surface of the eye. The reflected light is measured by an optical receiver.

\subsection{Preliminary results}
\begin{figure}[ht]
	\centering
	\includegraphics[width=0.4\linewidth]{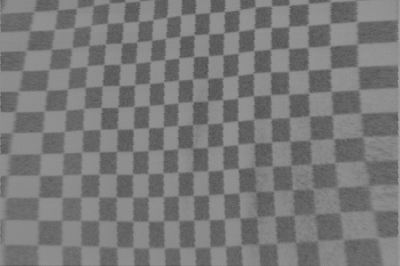}
	\caption{The image of a $5\,mm$ chess pattern in the eye-tracking region of the laboratory setup.}
	\label{A1im:image_calibarion}
\end{figure}
\FloatBarrier
\begin{figure}[ht]
	\centering
	\includegraphics[width=0.4\linewidth]{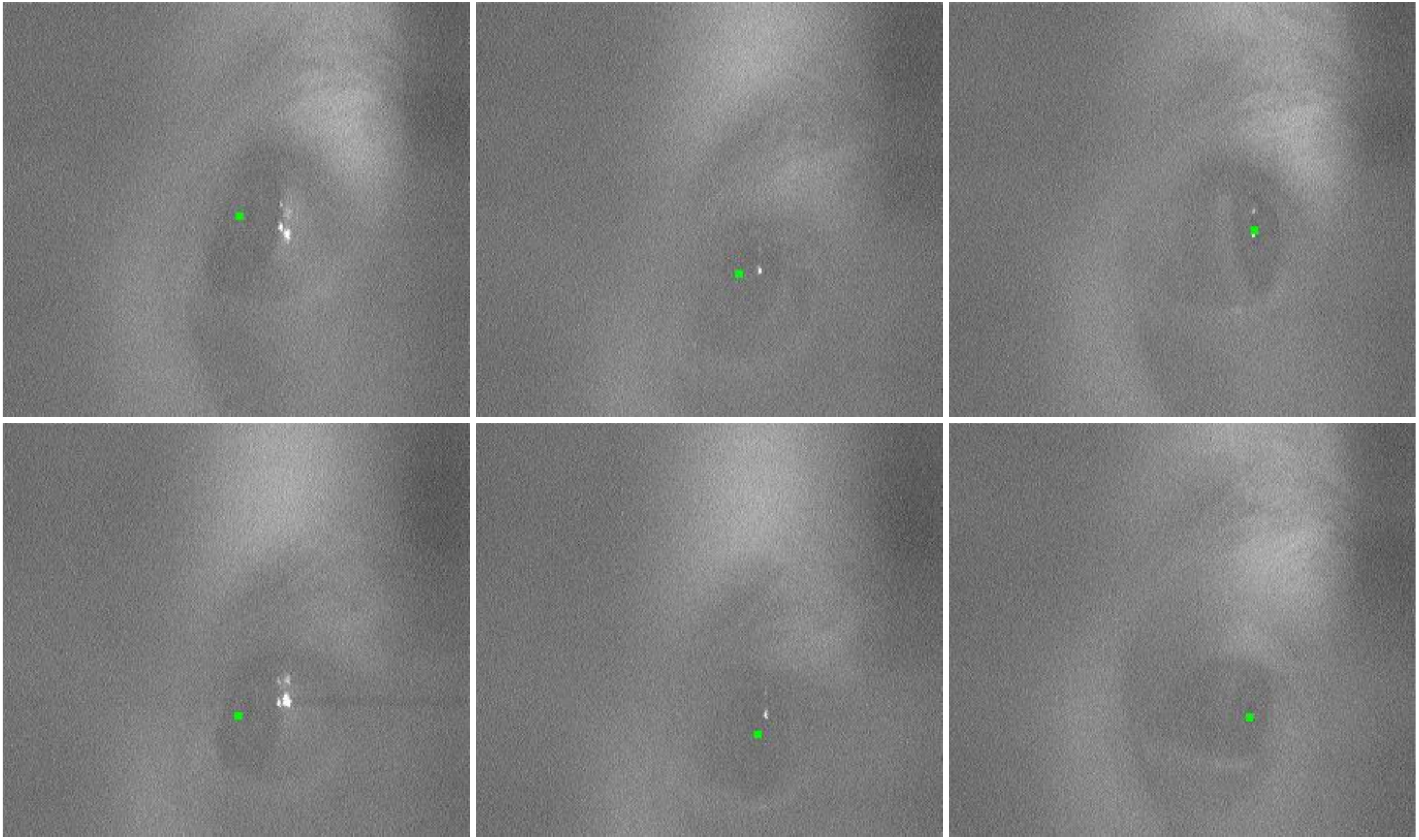}
	\caption{Sample images captured with the proposed eye-tracking approach. The pupil position is detected by a state of the art CBF pupil detection algorithm \cite{Fuhl:2018:CCB:3204493.3204559}.}
	\label{A1im:Augen}
\end{figure}
\FloatBarrier
To evaluate the feasibility of such a eye tracking sensor concept we created a laboratory setup by using  similar components as for the retinal projection glasses demonstrator shown in \Cref{A1im:ret_scanner_basics}. We integrated an IR laser into the projection module and used a semi transparent mirror to redirect the scanned laser beam towards the eye region. To measure the backscattered light from the eye region we used a separate photodiode and place it close to the eye region.

We ensured that the laboratory setup is a class 1 laser system according to IEC 60825-1 \cite{IEC2014} and therefore does not pose any medical hazard to the eye. The emitted IR laser power towards the eye is less than $150\,\mu$W.

The power consumption of the proposed eye-tracking sensor is estimated roughly at $11\, $mW using off-the-shelf components. This estimation does not include the power budget of the MEMS micro mirror, which is already included in the power budget of the retinal display. \cite{7863402} report that state of the art VOG eye-tracking sensors consume more than $150 \, $mW of power, which is significantly higher than our sensor approach. Compared to the scanned laser approach by \cite{7863402}, a similar power consumption is achieved.

To evaluate the eye tracking sensor concept we carried out two experiments. One experiment evaluates the spatial resolution of the proposed eye tracking sensor. In this experiment we placed a 5mm chess pattern in the scan region of the eye tracking sensor. \Cref{A1im:image_calibarion} shows the result of the experiment. The image is constructed by synchronized measurements of the backscattered light captured by the photodiode and measurements of the deflection angles of the micro mirror.

In the second experiment we evaluated the feasibility of capturing a human eye with the proposed eye tracking sensor and extracted pupil positions from the captured images using state of the art VOG algorithms. \Cref{A1im:Augen} shows the application of an VOG algorithm for pupil position extraction. To extract the pupil position we used the circular binary features (CBF) pupil detection algorithm by \cite{Fuhl:2018:CCB:3204493.3204559}. The estimated pupil center positions are marked by a green dot in \Cref{A1im:Augen}. The result is very promising and proves the feasibility to apply standard VOG algorithms to our low-power eye tracking data. This enables robust slippage invariant pupil extraction while minimizing development effort.

\subsection{Broader Impact}
The field of low power eye tracking sensors for AR glasses is an emerging area of research~\cite{Katrychuk:2019:PSE:3314111.3319821}. With the increase of AR glasses eye tracking sensors will become state of the art wearable sensors. This impacts additional research fields in the psychological or medical domain e.g. long term supervision of the user's behaviour to detect early signs of medical diseases or mental disorders. 

The demand for a low power consumption also impacts the research field of eye tracking algorithm design. New algorithm concepts will focus on low power consumption and high robustness outside the laboratory while keeping a sufficient gaze angle resolution to fulfill the requirements of AR glasses applications.

\subsection{Future work}

We presented a novel eye-tracking sensor for integration into our existing retinal projection AR glasses prototype. Currently, the viability of the approach has been demonstrated under laboratory conditions only. The head-worn demonstrator of the retinal projection system currently contains only red, green and blue projection lasers, as shown in \Cref{A2im:ret_scanner_basics}. The next step of our work is focused on integrating the eye-tracking sensor into this demonstrator and evaluation of the performance under real-world conditions, e.g. in the presence of various illumination conditions. For this purpose, some improvements of the photodiode circuit are considered. In addition we want to carry out additional experiments on the laboratory setup to estimate the achievable gaze angle resolution for the AR glasses.

Another challenging task is the integration of pupil tracking algorithms into the head-mounted demonstrator under the constraint of limited computational resources. In particular, a compromise between the required gaze angle resolution for AR glasses applications and the required computational effort to achieve this resolution needs to be examined in more detail.

\newpage

\section{A Novel Camera-Free Eye Tracking Sensor for Augmented Reality based on Laser Scanning}
\label{APP:A2}
\subsection{Abstract}
Next generation AR glasses require a highly integrated, high-re\-so\-lu\-tion near-eye display technique such as focus-free retinal projection to enhance usability. Combined with low-power eye-tracking, such glasses enable better user experience and performance. We propose a novel and robust low-power eye-tracking sensor for integration into retinal projection systems. In our approach, a MEMS micro mirror scans an IR laser beam over the eye region and the scattered light is received by a photodiode. The advantages of our approach over typical VOG systems are its high integration capability and low-power consumption, which qualify our approach for next generation AR glasses. In here along with the technical components, we present a mathematical framework to estimate the achievable gaze angle resolution of our approach. We further show the viability of the proposed eye-tracking sensor based on a laboratory setup and discuss power consumption and gaze angle resolution compared to typical eye-tracking techniques.

\subsection{Introduction}
\label{A2sec:Introduction}
The number of commercially available augmented reality (AR) glasses has largely increased in the last years. A new shift in the commercial domain is the integration of eye-tracking sensors into AR glasses to increase user experience by adding new functions and introducing new ways of interaction with systems around us \cite{7308147}. Examples for interactions with systems around us are driver assistance systems \cite{8082802,6476020}, human computer interaction (HCI) \cite{7464272,6227433} or smart home control \cite{s19040859,BONINO2011484}.

Additionally, the performance and resolution of AR applications can be increased by foveated imaging, where high resolution content is only projected sharply on the projection area corresponding to the fovea and lower-resolution content is used in peripheral regions. It is obvious that this technique is dependent on eye-tracking and can significantly reduce the required processing power for image rendering compared to a full-resolution rendering in the full field of view (FOV) \cite{Kim:2019:FAD:3306346.3322987,Kaplanyan:2019:DNR:3355089.3356557}. A recent implementation of foveated imaging for VR applications was presented by Tobii \cite{Tobii2019}.

Current eye-tracking sensors for low cost, commercially available systems, e.g. Pupil Labs \cite{Kassner:2014:POS:2638728.2641695}, use video-based oculography (VOG). These eye-tracking sensors rely on a set of infrared (IR) emitters to illuminate the eye and an IR camera to capture eye images. The gaze direction is then determined by image processing algorithms. 
While camera-based VOG systems are well-established and perform at high accuracy, there is only little potential for significant reduction of power consumption. Further, feasible orientation of the camera modules limits integration into a smart glasses frame and often interferes with the user's view \cite{10.1117/12.2322657,Katrychuk:2019:PSE:3314111.3319821}.

These drawbacks limit the use of VOG based eye-tracking sensors for battery powered AR glasses in consumer applications. The requirements of AR glasses in this segment are low-power consumption and a high degree of integration into the frame temple  \cite{Katrychuk:2019:PSE:3314111.3319821}.
Additionally, high stability and reliability are required for everyday use in the wild. A main problem that arises in the outside world is however reduced sensor performance due to artificial or natural light sources \cite{Fuhl2016}.

To meet these requirements, we propose a novel camera-free eye-tracking sensor based on retinal projection AR glasses. Our main contribution is a new miniaturized low-power eye-tracking sensor approach using state of the art VOG algorithms for integration into retinal projection AR glasses.

Retinal projection is a very promising near-eye display technique for AR glasses. It is based on direct illumination of the eye's retina using an ultra low-power eye-safe laser projector. Unique aspects of this display technology are a high degree of integratability into a frame temple, providing outstanding design opportunities, as well as near focus-free image projection, providing a sharp projection independent on the user's accommodation state \cite{Lin2017}.
\begin{figure}[h]
	\centering
	\includegraphics[width=0.4\linewidth]{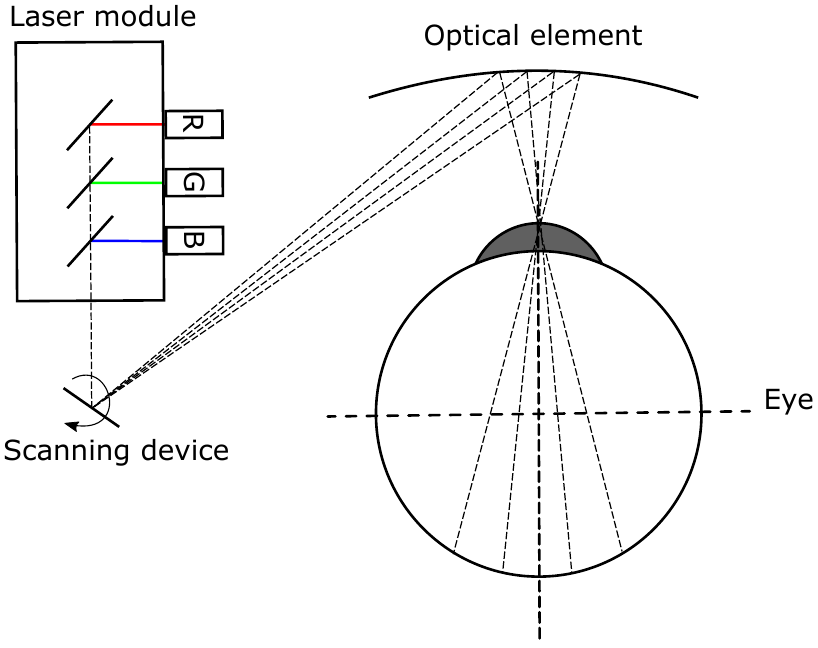}
	\caption{Principle structure and optical path of a retinal projection system. }
	\label{A2im:ret_scanner_basics}
\end{figure}
\FloatBarrier
\Cref{A2im:ret_scanner_basics} illustrates the basic structure and the main components of a retinal projection system. A laser module combines the laser beams of diode laser emitters for red, green and blue wavelengths. The combined beam is led to a scanning device which deflects the laser beam two-dimensionally onto a holographical optical element (HOE). This element redirects the beam towards the retina of the eye, where it forms a projected image \cite{10.1117/12.2295751}.
For most retinal projection systems, eye-tracking is crucial as the scanned beam has to follow the pupil's position to enter the eye \cite{Jang:2017:RAR:3130800.3130889}.

We propose an extension of these retinal projection systems to fully integrate eye-tracking. More specifically, we place an additional IR laser diode in the laser module and a tiny photodiode close to the joint in the frame temple. This resulting setup can be compared to photo sensor oculography (PSOG). PSOG exploits the varying IR reflectivity of different regions of the eye such as sclera, iris, pupil etc. \cite{8307473}.

Exploiting the scanning device and HOE already available for the projection system, this enables us to significantly simplify the PSOG setup by using only one IR diode laser and one photodiode for the whole tracking region. At the same time, spatial resolution is significantly increased. Moreover, system output are greyscale images suitable for processing using state of the art VOG algorithms.

The remaining of the paper is organized as follows. The next section discusses the state-of-the-art with regard to eye-tracking based on laser scanning. Section 3 presents our eye-tracking technique and its integration into retinal projection AR glasses as well as a mathematical framework to analyse the theoretical spatial and gaze angle resolution achievable by our technique. The evaluation in Section 4 shows the expected viability and resolution of the proposed eye-tracking sensor by an laboratory setup. Based on this  setup the resolution for an head worn sensor is estimated utilizing the mathematical framework derived in Section 3. Afterwards a comparison between our sensor approach and VOG eye-tracking sensors is performed. Section 5 concludes this work and gives an outlook to our future research activities.

\subsection{Related Work}
\label{A2sec:Related Work}
First scanned lasers for retina imaging appeared in the clinical section. Webb et al. introduced a scanning laser opthalmoscope (SLO), which works according to a process of scanning a laser beam over the retina surface to capture an image of the retina \cite{Webb:80}. The captured image was used for medical diagnosis of eye diseases. Eye motions during the scan process lead to distortions of the retina images. To resolve this issue, the SLO method was later extended by an eye-tracking system, as for example in \cite{Sheehy2012, STEVENSON201698} who presented a binocular tracking scanning laser ophthalmoscope (TSLO). The beam of a super luminescent diode was collimated and deflected by a $15\,$kHz resonant scanner horizontally and by a $30\,$Hz mirror galvanometer scanner vertically onto the eye. A photomultiplier tube and a beam splitter were used to detect the reflections of the eye. The photomultiplier tube was sampled based on the position of the mirrors to extract an image. Afterwards, the eye position was calculated offline on a host device. For this techniques, \cite{STEVENSON201698} reported a possible tracking speed of $366\,$Hz at a resolution of $0,003^\circ$. Such medical scanned laser eye-tracking approaches are characterized by high image resolution, however only during fixation phases of the eye. Furthermore, the optical setups are expensive, large and only suitable for laboratory use. To reduce the size of the optical setups \cite{10.1117/1.JBO.22.5.056006} and \cite{DuBose:18} used micro-electro-mechanical systems (MEMS) micro mirrors as scanning devices.   

A different scanned laser approach was described by \cite{Brother2007}. They integrated an IR laser into an RGB laser module to track the eye position. The IR laser beam was deflected onto a polygon mirror for horizontal deflection and a galvanometer scanner for vertical deflection. The scanned laser beam was then guided by a prism onto the surface of the eye. To capture the reflections of the eye, a semi-transparent mirror was placed in front of the user's eye region. The mirror directs the reflected light from the surface of the eye onto a collective lens on top of a photodiode. Based on the position of both scanning mirrors, the photodiode was sampled to capture an image of the eye. This setup was however rather large and was based on multiple mirrors which interfere with the user's view. Therefore, such eye-tracking setups can hardly be integrated into AR glasses.

To reduce the power consumption and size of scanned laser eye-tracking sensors,  \cite{7181058} introduced MEMS micro mirrors to scan a laser beam vertically and horizontally across the eye-tracking region. Based on this improvement, \cite{7863402} integrated the technology into AR glasses. In their setup, a two dimensional resonant MEMS micro mirror and an IR laser were placed on the inside of the frame temple. The MEMS micro mirror deflected the light towards the surface of the eye. To receive reflections from the eye's surface, a photodiode was placed on the nose bridge of the glasses. To track the horizontal eye position, a linear trajectory was scanned by the MEMS micro mirror over the surface of the eye. The photodiode received the reflected IR light from the surface of the cornea. The maximum light intensity detected by the photodiode was then used to estimate the horizontal eye position. Afterwards, the vertical position of the eye was estimated by a hill climbing algorithm based on the amplitude shifts of the photodiode output in horizontal scanning direction. This approach achieves an angular resolution of approximately 1$^\circ$ at a temporal resolution of $3300\,$Hz and a power consumption of less than $15\,$mW  \cite{7863402}.

The main draw back of this approach is the vulnerability to shifts of the glasses. Even small movements of the glasses on the user face lead to significant drifts and therefore inaccuracy of the eye position estimation \cite{7863402}.

Our approach combines the advantages of robust camera-based eye-tracking systems and the low-power consumption and small size of MEMS scanned laser eye-tracking sensors. Additionally, we integrate an optical element into the spectacle lens to obtain an eye-tracking sensor invisible to the user and therefore do not disturb the users view.

\subsection{Scanned laser eye-tracking}
\label{A2sec:MEMS scanned laser eye tracking}
\begin{figure}[h!]
	\centering
	\includegraphics[width=0.4\linewidth]{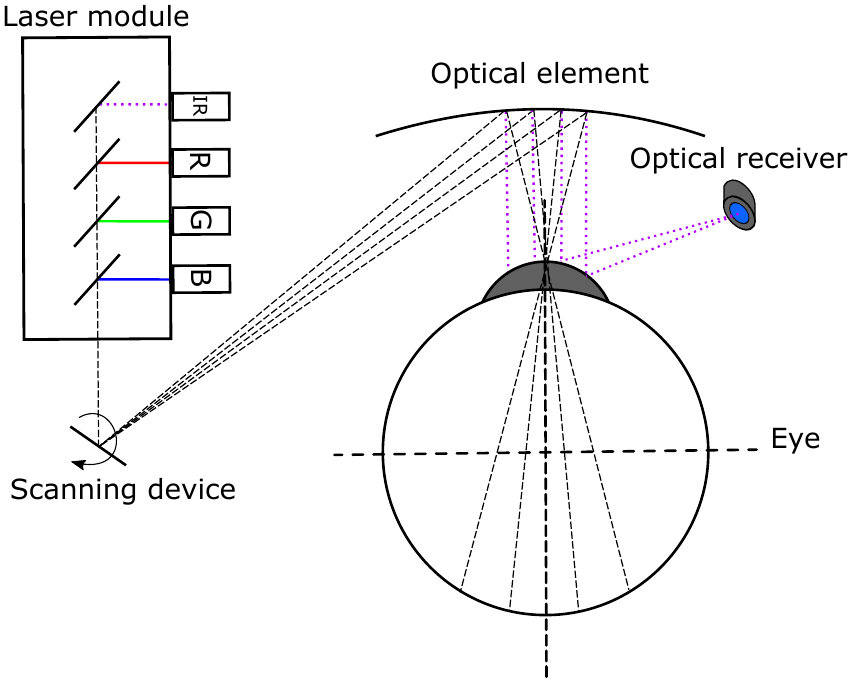}
	\caption{Principle structure and optical path of a retinal projector with integrated eye-tracking capabilities. An IR laser is integrated into the RGB laser module and an optical receiver is placed near the eye.}
	\label{A2im:ret_eyetracker}
\end{figure}
\FloatBarrier
The proposed eye-tracking sensor is based on a retina projector for AR glasses as described in \cite{10.1117/12.2295751}. The main components of the system are shown in \Cref{A2im:ret_scanner_basics}. Additional components are an IR laser, an optical receiver and an additional laser beam deflection function of the optical element specific to the IR wavelength of the eye-tracking laser. The integration of the new components into a retinal projection system are shown in \Cref{A2im:ret_eyetracker}.
The laser projection unit consists of a laser module and a scanning device. The laser projection module projects IR laser light (purple doted line) onto the optical element. The optical element redirects the laser beam with a defined optical function (e.g. parabolic mirror) onto the surface of the eye. Based on the varying IR reflectivity of the different eye regions, the laser beam is scattered with a different intensity from the surface of the eye. The reflected light is measured by an optical receiver.

The components of the sensor are described in more detail in the following.
\subsubsection{Laser projection module}
\label{A2subsec: Laser projection module}
\begin{figure}[h!]
	\centering
	\includegraphics[width=0.5\linewidth]{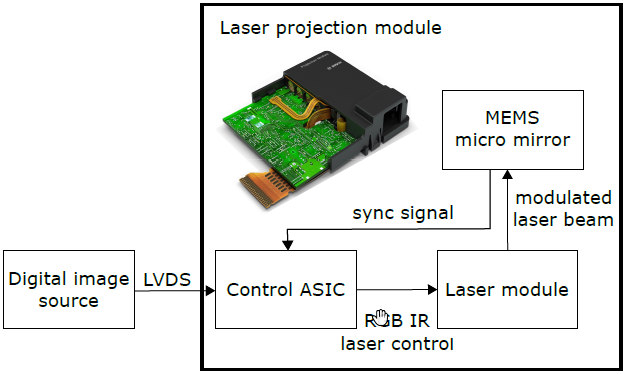}
	\caption{Block diagram of the laser projection module. Based on incoming frames, the integrated laser module and the MEMS micro mirrors are controlled by a control ASIC to project the frames onto a surface.}
	\label{A2im:BML100P}
\end{figure}

\Cref{A2im:BML100P} shows a block diagram of a laser projection module. 
\FloatBarrier
The main component of the projection module is the control application specific integrated circuit (ASIC). The ASIC is fed with a digital image via a low voltage differential signaling (LVDS) interface. 

Based on the incoming image stream, the lasers are modulated by the RGB values of each image pixel. The IR laser remains in an active state as long as eye-tracking remains active. The modulated and collimated RGB laser beam and the IR laser beam are directed onto the MEMS micro mirrors.

The micro mirror module consists of a fast axis MEMS micro mirror for the horizontal scan direction and a slow axis MEMS micro mirror for the vertical scan direction. The horizontal scan mirror is driven at resonance frequency and  oscillates sinusoidal.  The vertical mirror is actuated non resonantly by an external magnetic force. The start of a new line scan is reported by sync signals to the control ASIC.

\subsubsection{Holographic optical element} 
\label{A2subsec: Holographic optical element}
\begin{figure}[h]
	\centering
	\includegraphics[width=0.5\linewidth]{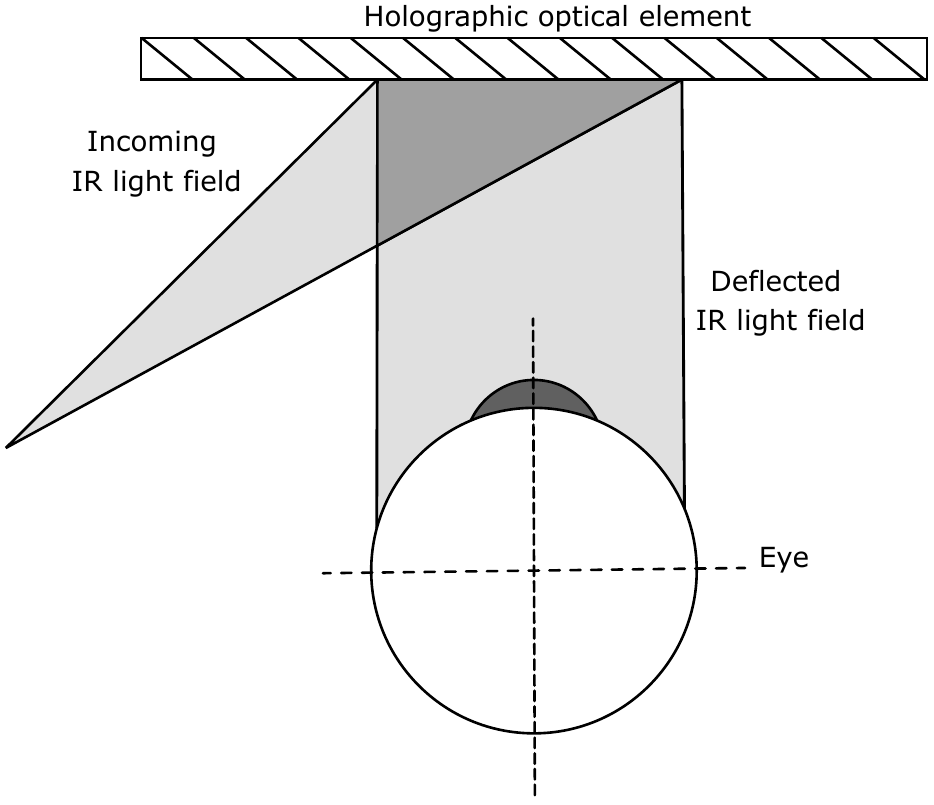}
	\caption{Description of the function of an holographic optical element (HOE). The Bragg structure inside the HOE diffracts the incoming IR light along parallel lines towards the eye to form a rectangular region in the surface of the eye.}
	\label{A2im:HOE}
\end{figure}

An HOE is integrated into the AR spectacle lens to deflect the incoming scanned IR laser light field of the laser projection module towards the surface of the eye as shown in \Cref{A2im:HOE}. 

\FloatBarrier
The HOE is realized by recording a mirror function into a photo polymer material. For a more detailed description of the recording process and integration of optical functions into HOEs using Bragg structures, we refer to \cite{10.1117/12.2309788}.

The main advantage of HOEs over other optical elements like semi-transparent mirrors is the high selectivity and optical transparency. Ideally, the optical function is only active for the recording wavelength \cite{Jang:2017:RAR:3130800.3130889}. 

The HOE function controls the size and contour of the region which is scanned by the IR laser on the eye. For the scanned laser eye-tracking sensor, this scanned surface is referred to as eye-tracking region. 

\subsubsection{Receiving photodiode circuit}
\label{A2subsec: Receiving photodiode circuit}
\begin{figure}[h]
	\centering
	\includegraphics[width=0.4\linewidth]{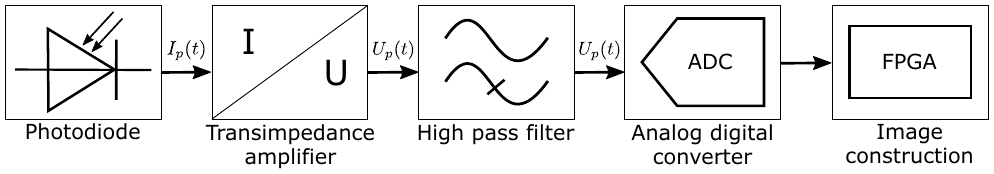}
	\caption{System block diagram of the receiving photodiode circuit. }
	\label{A2im:AFE}
\end{figure}

To capture a reflection map of the eye-tracking region a similar receiving photodiode circuit as known from previous approaches like PSOG or the scanned laser approach by \cite{7863402} is used.\Cref{A2im:AFE} shows the system block diagram  of the receiving photodiode circuit.
\FloatBarrier
The main component is a photodiode which is sensitive to IR light. To increase the resolution and convert the photodiode current $I_p(t)$ into a measurable voltage $U_p(t)$, a transimpedance amplifier (TIA) circuit is used. The voltage $U_p(t)$ is AC coupled via a high pass filter to remove low frequency disturbances like ambient light. The amplified and filtered signal is then converted into the digital domain by an analogue to digital converter (ADC). A Field Programmable Gate Array (FPGA) converts the digitized samples into an image utilizing the sync signals of the laser projection module. To capture an image of the eye, the photodiode current is sampled at fixed time intervals by the ADC.

\begin{figure}[h]
	\centering
	\includegraphics[width=0.4\linewidth]{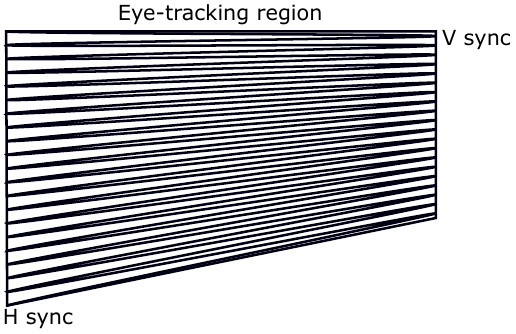}
	\caption{Simplified sinusoidal scan path of the laser beam over the eye-tracking region. The sync signals are used to reconstruct an image based on the samples captured by the ADC.}
	\label{A2im:scan_path}
\end{figure}
\FloatBarrier
The start of a new frame is indicated by the vertical synchronization signal V sync. With the rise of this signal, the ADC starts to sample the photodiode signal. The start of a new row of the frame is indicated by the horizontal synchronisation signal H sync. With these two signals, a reflectivity image of the surface of the eye is reconstructed by the FPGA.

\Cref{A2im:scan_path} shows a simplified scan path (much less lines) of the laser beam in the eye-tracking region. The characteristic bounding box results from the geometric setup as depicted in \Cref{A2im:dimensions}.

The resolution of the image, and therefore the pixel size, is dependent on the bandwidth of the AFE, the sample rate of the ADC, the time-dependent sinusoidal angular scan speed of the laser projection system and the geometric setup.

\subsubsection{Spatial resolution}
\label{A2subsec:Spatial resolution}
State of the art camera-based eye-tracking approaches utilize pupil edge detection methods to extract the pupil location and calculate the gaze direction \cite{santini2018pure}. For this purpose, a sufficient spatial resolution in the captured eye region is required. The spatial resolution in VOG eye-tracking approaches is limited by the camera resolution and image distortions due to lens misalignments. 

In the scanned laser eye-tracking sensor approach, pixel size and image distortion are related to the scan frequency of the MEMS micro mirrors, the optical path length and the performance of the photodiode circuit \cite{Folks2005}.

Horizontal distortions occur due to a sinusoidal-like velocity pattern of the laser spot along the scan path shown in \Cref{A2im:scan_path}. Combined with equitemporal sampling of the photodiode circuit, this leads to barrel distortions \cite{mi6111450}.

The spatial resolution is characterized by analysing the image data of a high contrast linear or square edged test pattern with known properties in the eye-tracking region captured by the photodiode circuit. The known properties of the test pattern and the resulting number of pixels in the image are used to calculate pixel sizes at various positions in the image. With this, image distortions due to varying optical path length and the sinusoidal angular velocity pattern are quantified.

\subsubsection{Gaze angle resolution}
\label{A2subsec:Gaze_angle_resolution_teo} 
A gaze angle is described by the pupil center position and eye rotation angles. With rotation of the eye, the pupil center, and therefore, the pupil edges are rotated as well. To assess the gaze angle resolution of the proposed scanned laser eye-tracking sensor, a mapping from spatial resolution to angular resolution is required. It is dependent on the geometrical system design, the laser projection unit, the sample rate of the photodiode circuit, the HOE function, and the rotation angle of the eye. For the calculations, a parallel deflection function of the HOE as described in \Cref{A2im:HOE} is assumed.

\begin{figure}[h]
	\centering
	\includegraphics[width=0.5\linewidth]{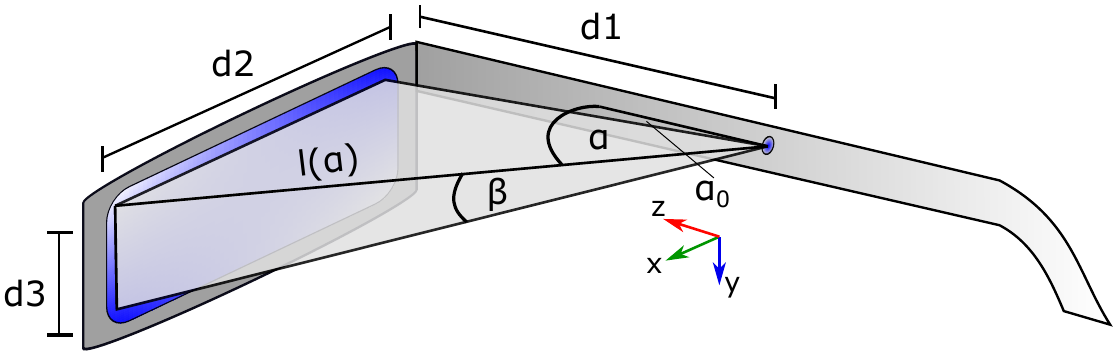}
	\caption{Integration of the proposed scanned laser eye-tracking sensor into AR glasses. The HOE is integrated into the spectacle lens and the laser projection module is integrated into the frame temple.}
	\label{A2im:dimensions}
\end{figure}

\Cref{A2im:dimensions} shows the geometrical dimensions of the AR glasses setup. 
\FloatBarrier
The laser projection module is integrated into the right frame temple with a distance $d_1$ to the hinge between frame temple and spectacle lens. The angles $\alpha$ and $\beta$ describe the horizontal and vertical scan angles of the laser projection module. The distance $d_2$ denotes the width of the HOE inside the spectacle lens. The distance $d_3$ describes the height of the HOE.

As mentioned earlier, pixel sizes vary throughout the eye tracking region. They are dependent on the oscillation frequencies $f_h$ and $f_v$ as well as the maximum scan angles $\alpha_{max}$ and $\beta_{max}$ of the horizontal and vertical micro mirrors, respectively. In addition, the sample speed $f_s$ of the ADC and the distance $d_1$ are key parameters.

The x-coordinate of a point on the HOE can be expressed by $d_1$ and $\alpha$:
\begin{equation}
	x(\alpha) = d_1 \cdot \tan(\alpha).
\end{equation}

The angle $\alpha$ changes sinusoidal in time. For the geometry shown in \Cref{A2im:dimensions}, this is
\begin{equation}
\alpha(t)= \frac{\alpha_{max}}{2} \cdot \sin(\omega_h\cdot t-\frac{\pi}{2}))+\frac{\alpha_{max}}{2} + \alpha_0.
\label{A2Equ:alpha}
\end{equation}

$\alpha_0$ denotes the horizontal offset angle between frame temple and the right edge of the HOE integrated in the spectacle glasses.
The pixel size $s_x$ along the horizontal direction can be expressed as the difference between a first horizontal position at angle $\alpha$ and a second position at an angle incremented by $\Delta \alpha$
\begin{equation}
	s_x = x(\alpha + \Delta \alpha) - x(\alpha).
	\label{A2equ:x_pixel_res}
\end{equation} 
$\Delta \alpha$ denotes the angle increment between two consecutive samples of the ADC and is dependent on the sample speed $f_s$ of the ADC. Therefore,
\begin{equation}
	\Delta \alpha = \alpha(t+\frac{1}{f_s}) - \alpha(t) .
\end{equation}
This can be used to eliminate $t$ in \Cref{A2Equ:alpha}:
\begin{equation}
\Delta \alpha =\alpha - \frac{\alpha_{max} \cos{\left(\frac{2 \pi f_h}{f_s} + \operatorname{acos}{\left( \frac{2\alpha_0 - 2 \alpha + \alpha_{max}}{\alpha_{max}} + 1 \right)} \right)}}{2} + \frac{\alpha_{max}}{2}.
\end{equation}
Now, the horizontal pixel size $s_x$ from \Cref{A2equ:x_pixel_res} can be expressed by the position on the HOE given by $\alpha$ and known system constants only:
 \begin{equation}
\resizebox{\linewidth}{!}{
	$s_x(\alpha) = - d_1 \cdot \left(\tan{\left(\alpha \right)} + \tan{\left(\alpha_0 + \frac{\alpha_{max}}{2} - \frac{\alpha_{max} \cdot \left(\cos{\left(\frac{\omega_h}{f_s} + \operatorname{acos}{\left( \frac{2 \alpha_0 - 2 \alpha- \alpha_{max}}{\alpha_{max}} \right)} \right)} - 1\right)}{2} \right)}\right).$}
\label{A2equ:horizontal_pixel_size}
\end{equation}

The vertical axis is operated in a triangular-like fashion with a linearly increasing $\beta(t)$ until $\beta_{max}$ and a linear flyback to $\beta=0$. As apparent from \Cref{A2im:dimensions}, the optical path length increases with increasing deflection of the micro mirrors. With this, the vertical position is dependent on the length of the projection of the optical path between micro mirror and HOE onto the xz-plane. This can be expressed as
\begin{equation}
	l(\alpha) = \frac{d_1}{\cos(\alpha)}.
\end{equation}
The resulting y-coordinate is then calculated using $\beta$:
\begin{equation}
	y(\alpha, \beta) = l(\alpha) \cdot \tan(\beta).
\end{equation}
Analogy to the horizontal case, the pixel size $s_y$ along the vertical direction is expressed as the difference between a first vertical position defined by $\alpha$ and $\beta$ and a second position at an angle incremented by $\Delta \beta$. Here, $\Delta \beta$ is the linear angle increment between two consecutive horizontal scan lines and can be expressed as
\begin{equation}
	\Delta \beta = \frac{\beta_{max}}{N_h}
\end{equation}
with the number of horizontal lines $N_h$ scanned over the eye-tracking region. $N_h$ is calculated by 
\begin{equation}
	N_h = \frac{f_h}{f_v \cdot k_v}.
\end{equation}
The constant $k_v < 1$ describes the fraction of forward scan to flyback time which is required to drive the horizontal mirror back to the starting position.

With this, the vertical pixel size $s_y$ is
\begin{equation}
	s_y(\alpha, \beta) = \frac{y(\alpha, \beta + \Delta \beta) - y(\alpha, \beta)}{\cos(\beta)}.
	\label{A2equ:vertical_pixel_size}
\end{equation}
The division by $\cos(\beta)$ results in a projection of the pixel size vector perpendicular to the optical path onto the plane of the HOE.

Now, pixel dimensions can be expressed dependent on mirror positions only (given by $\alpha$ and $\beta$). To calculate gaze angle resolution, this spatial resolution has to be mapped to eye rotational resolution.

\begin{figure}[h]
	\centering
	\includegraphics[width=0.35\linewidth]{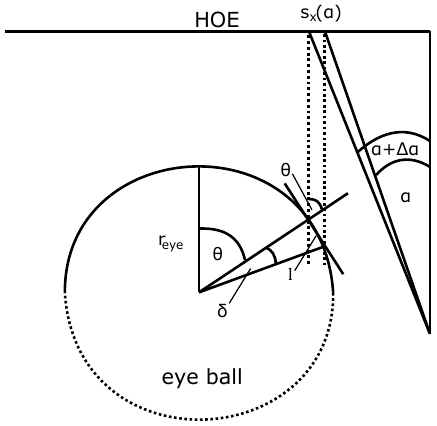}
	\caption{Mapping of horizontal eye rotation to spatial resolution on the HOE. A parallel deflection function of the HOE is assumed.}
	\label{A2im:eye_gaze_res}
\end{figure}
\FloatBarrier
For calculation, we assume a simplified eye ball model as shown in \Cref{A2im:eye_gaze_res} with radius $r_{eye}$. It is rotated by $\theta$ in the horizontal plane and $\phi$ in vertical direction around its center. This rotation causes the pupil edge to propagate through different pixels along the reconstructed image. Neglecting methodologies like subpixel interpolation, eye rotation is only noticeable by the eye-tracking system if the pupil edge propagates from one pixel to an adjacent one. Thus, we define an infinitesimal angular increment $\delta$ of eye rotation that keeps the pupil edge in the current pixel. In this view, $\delta$ is the position-dependent angular resolution of our eye-tracking system.

Based on \Cref{A2im:eye_gaze_res}, $\delta_h$ for horizontal gaze angle resolution is calculated by
\begin{equation}
	\delta_h(\theta) = \operatorname{atan}\left(\frac{s_x(\alpha(\theta))}{r_{eye} \cdot \cos(\theta)}\right)
	\label{A2equ:gamma_theta}
\end{equation}
with $\alpha(\theta)$ as mapping function to map an eye rotation angle $\theta$ to a horizontal angle $\alpha$. With the assumption that the eye-tracking region, and therefore the HOE, is centered over the eye, the mapping function $\alpha(\theta)$ is
\begin{equation}
	\alpha(\theta) = \operatorname{atan}\left(\frac{\frac{d_2}{2} - r_{eye} \cdot \sin(\theta)}{d1}\right).
\end{equation}
The vertical gaze angle resolution $\delta_v$ in analogy to \Cref{A2equ:gamma_theta} is
\begin{equation}
\delta_v(\phi) = \operatorname{atan}\left(\frac{s_y (\alpha(\theta), \beta(\phi))}{r_{eye} \cdot \cos(\phi)}\right)
\label{A2equ:gamma_phi}
\end{equation} 
with 
\begin{equation}
\beta(\phi) = \operatorname{atan}\left(\frac{\frac{d_3}{2} - r_{eye} \cdot \sin(\phi)}{d1}\right).
\end{equation}

The overall gaze angle resolution is calculated by the euclidean distance of the horizontal and vertical gaze angle resolution
\begin{equation}
	\delta(\theta, \phi) = \sqrt{\delta_v(\theta)^2 + \delta_h(\phi)^2}.
	\label{A2equ:gaze_resolution}
\end{equation}

\subsection{Evaluation}
\label{A2sec:Evaluation}
\begin{figure}[h]
	\centering
	\includegraphics[width=0.5\linewidth]{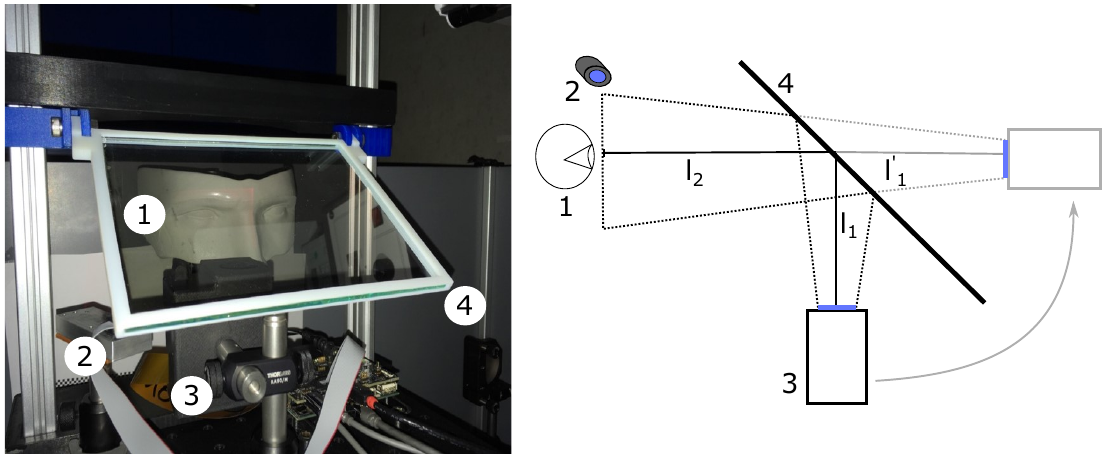}
	\caption{Laboratory setup to evaluate the accuracy and precision of the proposed eye tracking sensor.}
	\label{A2im:lab_setup_2}
\end{figure}

To determine the gaze angle resolutions of the proposed head worn eye-tracking sensor, a laboratory setup is used. It is shown in \Cref{A2im:lab_setup_2}. 
\FloatBarrier
The subject \textcircled{1} is placed in a distance of $l_2 = 90\,mm$ in front of an semitransparent mirror \textcircled{4} to align the eye tracking region with the subjects eye. The semitransparent mirror mimics the HOE with a parallel beam deflection function as shown in \Cref{A2im:HOE}. The mirror is used to deflect the scanned IR laser beam of the laser projection module \textcircled{3}. The laser module is placed in a distance of $l_1 = 30mm$ towards the semi-transparent mirror. The photodiode circuit \textcircled{1} is directly orientated towards the subjects eye to receive as much backscattered IR light as possible and thus improve sensitivity. The mirror allows the laser projection module to be virtually rotated so that it points directly at the eye tracking region without disturbing the user's view. In addition, artefacts of the captured images caused by eyelids and eyelashes are reduced, which improves the robustness of the sensors.

\Cref{A2tab:eye_tracking_params} shows the geometrical and electrical properties of the the laboratory setup. The overall distance $d_1$ is the sum of $l_1$ and $l_2$.
\begin{table}[h]
	\caption{Electrical and geometrical properties of the laboratory setup.}
	\label{A2tab:eye_tracking_params}
	\resizebox{\columnwidth}{!}{%
	\begin{tabular}{cccccccccc}
		\hline
		$\alpha_0$& $\pm \alpha_{max}$ &$\pm \beta_{max}$	&$f_s$  &$f_h$  &$f_v$  &$k_v$  &$d_1$  &$d_2$  &$d_3$  \\ 
		
		0$^\circ$ &	15$^\circ$&9$^\circ$  &22 MHz  &21kHz  &60Hz  &0.83  &120mm  &69mm  &39mm  \\\hline
	\end{tabular}
}
\end{table}

The laboratory setup is a class 1 laser system according to IEC 60825-1 \cite{IEC2014} and therefore does not pose any medical hazard to the eye. The emitted IR laser power towards the eye is less than $150\,\mu$W. Laser class 1 would allow $670\,\mu $W for a steady beam, even significantly more in scanned operation. Laser safety is therefore ensured for all single error cases as required by IEC 60825-1 as well.

The power consumption of the proposed eye-tracking sensor is estimated roughly at $11\, $mW using off-the-shelf components. The main components that affect power consumption are the TIA and the ADC. With a higher degree of integration, e.g. by a custom ASIC, further power reductions are expected.

\cite{7863402} report that state of the art VOG eye-tracking sensors consume more than $150 \, $mW of power, which is significantly higher than our sensor approach. Compared to the scanned laser approach by \cite{7863402}, a similar power consumption is achieved.

\subsubsection{Spatial resolution}
\label{A2subsec:Spatial resolution eval}
To prove the mathematical framework in \Cref{A2subsec: Gaze angle resolution}, the spatial resolution as described in \Cref{A2subsec:Spatial resolution} is calculated by placing a chess pattern of defined size in the eye-tracking region and compared with the theoretical spatial resolution. The experimental spatial resolution is determined based on the known properties of the chess pattern image. The number of pixels per chess field is counted and divided by the known length of a chess field to determine the pixel size. 

\begin{figure}[h]
	\centering
	\includegraphics[width=0.5\linewidth]{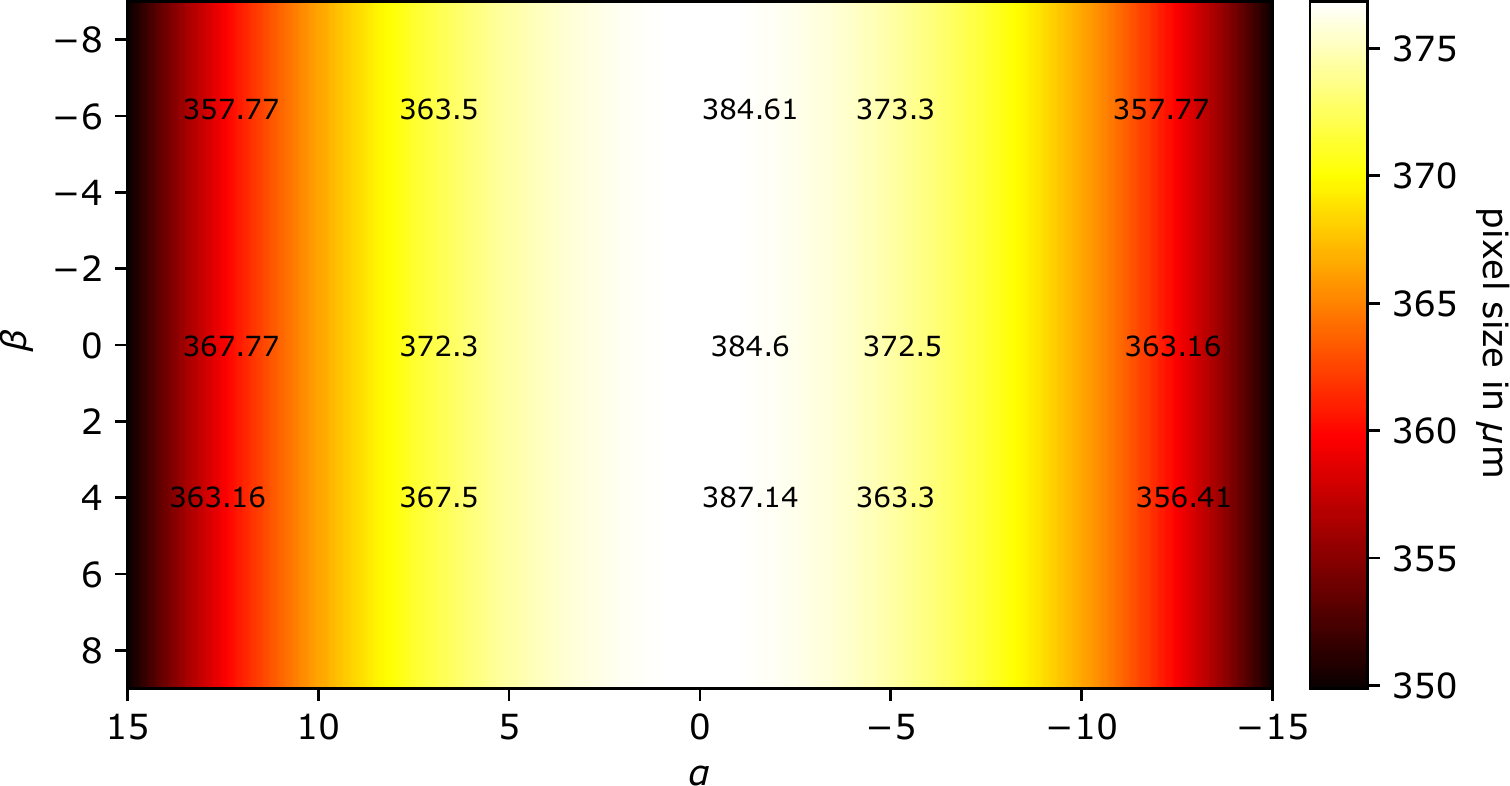}
	\caption{Calculated pixel size for the horizontal direction based on \Cref{A2equ:horizontal_pixel_size} with the parameters of \Cref{A2tab:eye_tracking_params}. In addition, the measured horizontal pixel sizes from the chess pattern are annotated. }
	\label{A2im:pixel_size_bx}
\end{figure}

\begin{figure}[h]
	\centering
	\includegraphics[width=0.4\linewidth]{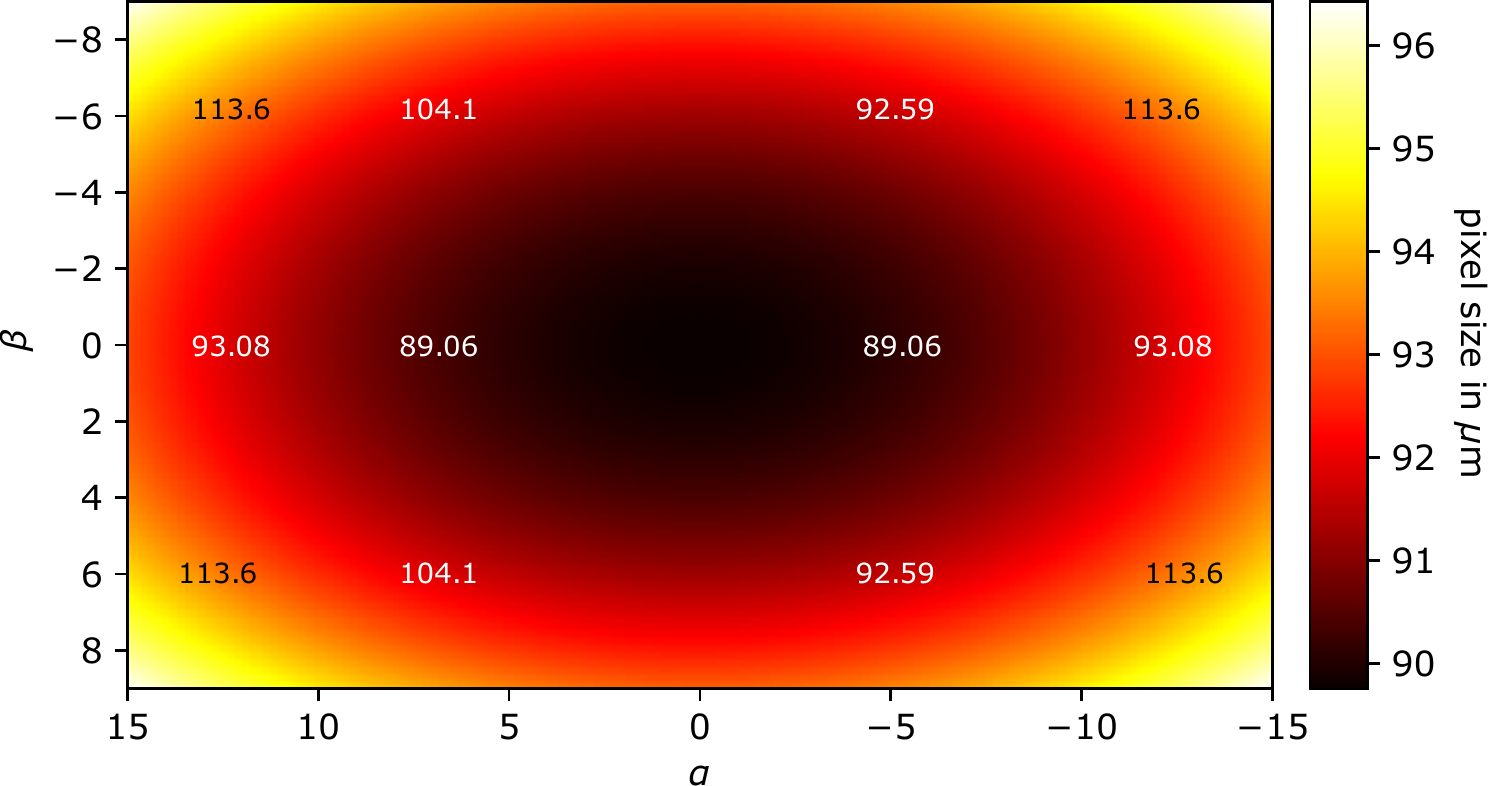}
	\caption{Calculated pixel size for the vertical direction based on \Cref{A2equ:vertical_pixel_size} with the parameters of \Cref{A2tab:eye_tracking_params}. In addition, the measured vertical pixel sizes from the chess pattern are annotated.}
	\label{A2im:pixel_size_by}
\end{figure}

\Cref{A2im:pixel_size_bx} and \Cref{A2im:pixel_size_by} show the resulting theoretical and experimental spatial resolutions, as color surface and annotated numbers respectively.

\FloatBarrier
The spatial resolution in horizontal direction shown in \Cref{A2im:pixel_size_bx} is dominated by the non-linear horizontal scan speed of the MEMS micro mirror described by \Cref{A2Equ:alpha}. The sinusoidal change of velocity results in smaller pixels in the left and right edge regions of the eye tracking region. In the centre of the eye tracking region, the peak velocity is reached and thus increasing pixel size. In consequence, the sensor resolution is higher towards the left and right sides of the eye tracking region. 

\FloatBarrier
For vertical direction the pixel sizes increases for increasing angles $\alpha$ and $\beta$, as shown in \Cref{A2im:pixel_size_by}. This effect is mainly caused by increasing optical path length of the laser beam for increasing angles $\alpha$ and $\beta$. This effect is superimposed by the effect of the sinusoidal change of velocity, which leads to additional distortions in the horizontal direction. 

\subsubsection{Experimental gaze angle resolution}
To estimate the achievable gaze angle resolution a experiment with the laboratory setup is performed. A subject sits approximately 0.5m away from a chart with visual markers $M$ and fixate the markers. For each marker a set of $N$ images is taken. The markers are placed on the chart to cause eye rotation angles $\theta$ and $\phi$ in the range of $\pm 20^\circ$ and of $\pm$10$^\circ$. 

\begin{figure}[h]
	\centering
	\includegraphics[width=0.5\linewidth]{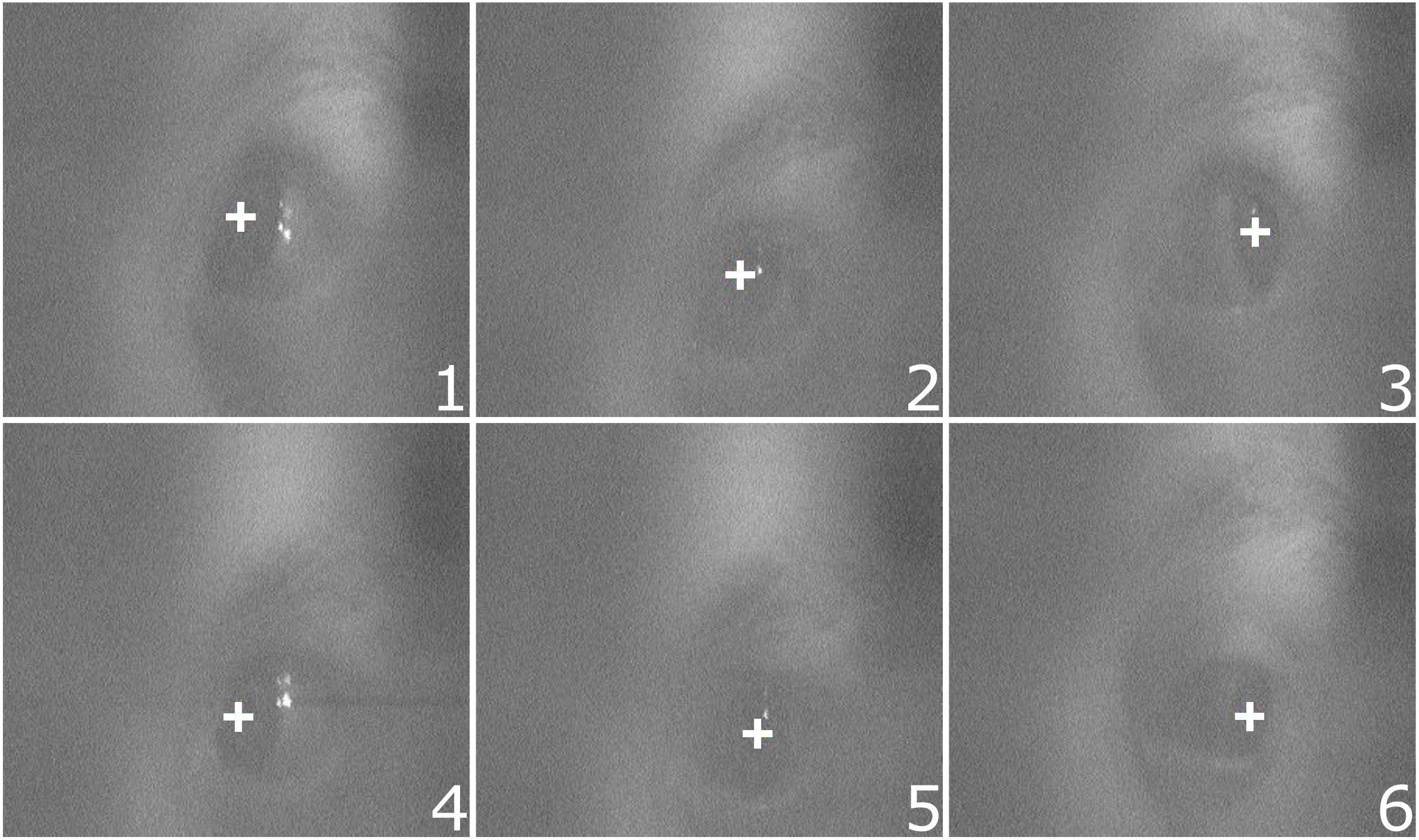}
	\caption{Images of a subject fixating different markers on a chart. The images are captured with the proposed eye-tracking approach. The pupil position is detected by a state of the art VOG pupil detection algorithm \cite{Fuhl:2018:CCB:3204493.3204559}. The numbers indicate which marker of the chart is fixated by the subject.}
	\label{A2im:Augen}
\end{figure}

\Cref{A2im:Augen} shows a subject fixating different markers. The images are captured with the laboratory setup.
\FloatBarrier
To extract the pupil positions we use the state of the art VOG circular binary features (CBF) pupil detection algorithm by \cite{Fuhl:2018:CCB:3204493.3204559}. The estimated pupil center positions are marked with a green dot in \Cref{A2im:Augen}. The result is very promising and proves the feasibility to apply state of the art VOG algorithms to our low-power eye tracking data. This enables slippage robust pupil extraction for the scanned laser eye tracking sensor with minimal algorithm design effort. In addition the sensor benefits directly from advances in VOG algorithms. 

To calculate the resolution and precision of the sensor we perform the standard 9 point chart marker based calibration method similar to \cite{Kassner:2014:POS:2638728.2641695}. For each calibration marker $M_c$ $N$ images are captured and the corresponding pupil coordinates $P_c$ are calculated using the CBF algorithm. With the known calibration coordinates and the related pupil coordinates a second order polynomial function 
\begin{equation}
x_c = a_0 + a_1 x_c + a_2 y_c + a_3 x_c y_c + a_4 x_c^2 + a_5 y_c^2
\end{equation}
for x coordinates and
\begin{equation}
y_c = b_0 + b_1 x_c + b_2 y_c + b_3 x_c y_c + b_4 x_c^2 + b_5 y_c^2 	
\end{equation}
for y coordinates is fitted using a least mean square optimizer to estimate the coefficients $a$ and $b$. 

\begin{figure}[h]
	\centering
	\includegraphics[width=0.6\linewidth]{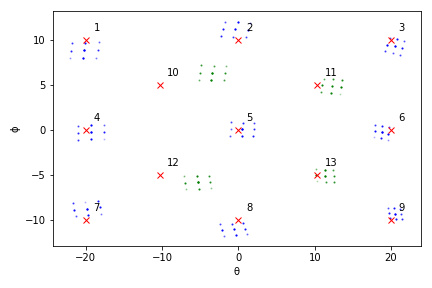}
	\caption{Result of the accuracy and precision experiment with calibration markers $M_c$ (1-9) and test markers $M_t$ (11-13) and the corresponding pupil coordinates $P$. The arrow indicates the error between corresponding marker coordinates and estimated pupil coordinates.}
	\label{A2im:calibration_results}
\end{figure}

In addition to the calibration markers $M_c$ test markers $M_t$ are placed on the calibration chart. These markers are used to estimate the spatial accuracy and precision of the proposed eye tracking sensor. \Cref{A2im:calibration_results} shows the results of the experiments in gaze angle coordinate space.  
\FloatBarrier
To estimate the resolution based on the experiment the average angular distance between the position of the test markers $M_t$ and a set of corresponding locations of fixations $P_t$ is calculated for every test marker
\begin{equation}
Res(M_t, P_t) = \frac{1}{N} \sum_{e=0}^{N} \sqrt{\left(x_{M_t} - x_{P_t}\right)^2 + \left(y_{M_t} - y_{P_t}\right)^2 }.
\end{equation}   
In addition the precision is estimated by the Root Mean Square (RMS) of successive samples of fixations $P_e$ for a given test marker $M_t$. It is calculated by 
\begin{equation}
Pre(P_t) = \sqrt{\frac{1}{N}\sum_{e=0}^{N-1}\left(x_{P_t} - x_{{P_t}+1}\right)^2 + \left(y_{P_t} - y_{{P_t}+1}\right)^2 }
\end{equation}
for every test marker. The resulting mean resolution of the test markers is $1.31^\circ$ and the precision is $0.01^\circ$ for this setup.

A possible source of error, which reduces the achieved gaze accuracy, is that the head of the test person is not fixed during the experiment. In addition, a large part of the captured images covers the face around the eye. The pupil information is therefore only contained in a small subset of pixels in the center of the image.



\subsubsection{Theoretical Gaze angle resolution}
\label{A2subsec: Gaze angle resolution}
\begin{figure}[h]
	\centering
	\includegraphics[width=0.5\linewidth]{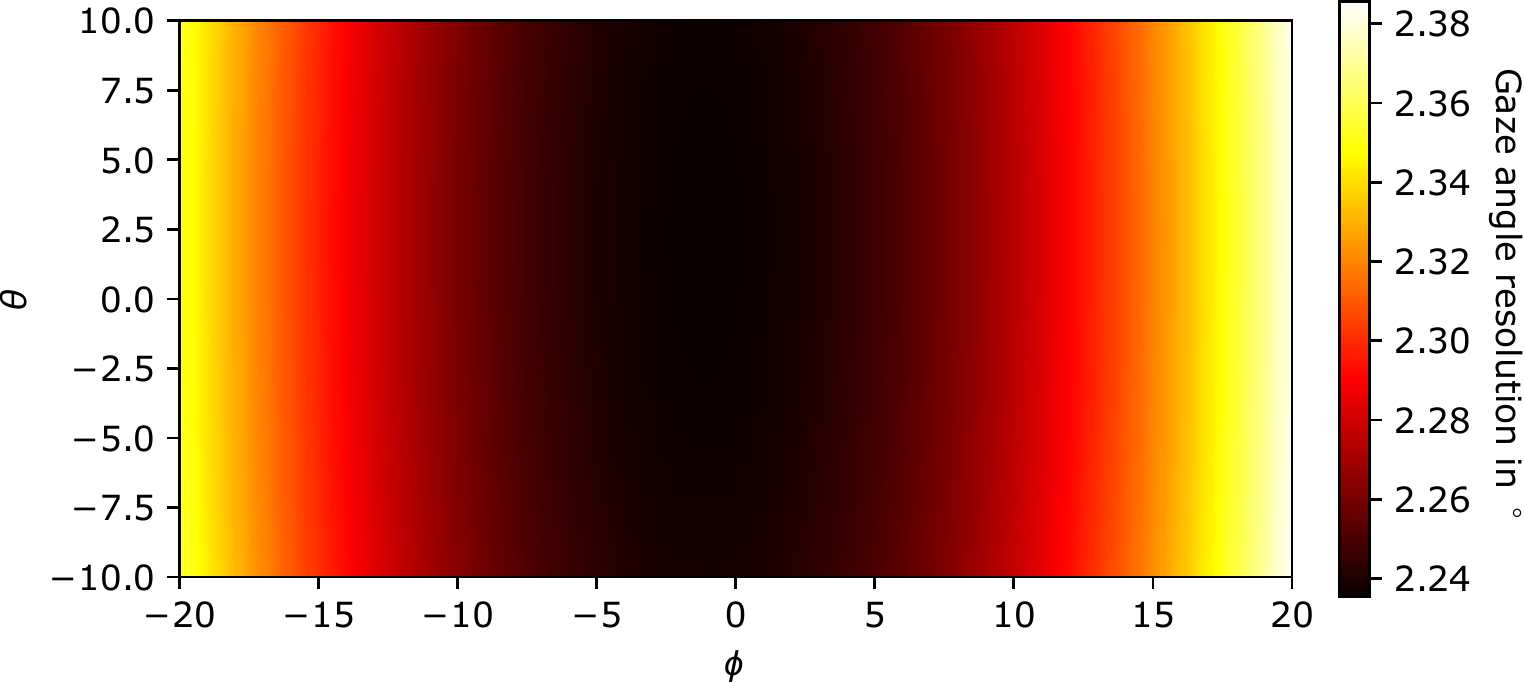}
	\caption{Calculated gaze angle resolution of the proposed eye tracking sensor based on the laboratory setup based on \Cref{A2equ:gaze_resolution} using the geometry of the Emsley's reduced eye model.}
	\label{A2im:Gaze_angle_res_lab}
\end{figure}

In addition to the experimental estimation of the gaze resolution of the proposed eye tracking sensor the the theoretical gaze angle resolution is estimated by \Cref{A2equ:gaze_resolution}. The distance $r_{eye}$ in \Cref{A2equ:gaze_resolution} is derived from the Emsley's reduced eye model. Based on this model, the distance $r_{eye}$ between the iris and the center point of the eye ball is $9.77\, $mm. The result is shown in \Cref{A2im:Gaze_angle_res_lab}.

\FloatBarrier
Recalling \Cref{A2im:eye_gaze_res}, $\delta$ is much smaller for given $s_x(\alpha)$ close to $\theta \approx 0, \phi \approx 0$. To some extent, this compensates the lower spatial resolution in the center of the eye tracking region as shown in \Cref{A2im:pixel_size_bx}. The theoretical mean single-pixel gaze angle accuracy of the laboratory setups is around 2.3\,$^\circ$, which is significant lower as the experimental estimated gaze angle resolution. Thus, our simplified mathematical approach neglecting the effect of pupil tracking algorithms can be understood as an upper boundary estimation for gaze angle resolution.

\begin{figure}[h]
	\centering  
	\includegraphics[width=0.5\linewidth]{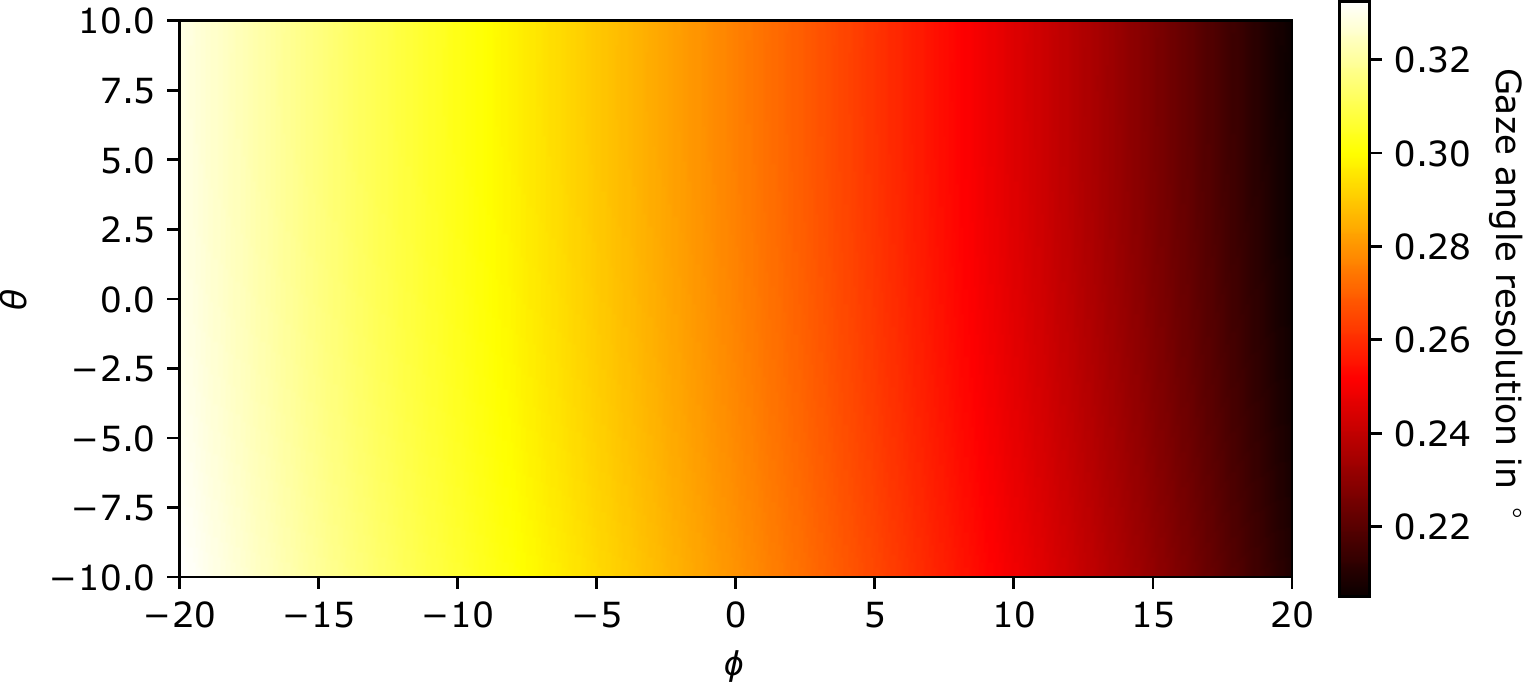}
	\caption{Calculated gaze angle resolution of the proposed eye tracking sensor for the glasses geometry based on \Cref{A2equ:gaze_resolution} using the geometry of the Emsley's reduced eye model.}
	\label{A2im:Gaze_angle_res_glasses}
\end{figure}

Based on this assumption the theoretical gaze angle resolution of the head worn eye tracking sensor is calculated using the electrical and geometric parameters shown in \Cref{A2tab:eye_tracking_params_glasses}. The main differences between the laboratory setup and the proposed head worn demonstrator are the reduced distance $d_1$ towards the HOE, the offset angle $\alpha_0$ and the maximum angles for $\alpha_{max}$ and $\beta_{max}$. Due to the glasses geometry $\alpha$  is in a range between $0$ and $\alpha_{max}$ an $\beta$ is in a range between $0$ and $\beta_{max}$. 
\begin{table}[h]
	\caption{Electrical and geometrical properties of the AR glasses setup.}
	\label{A2tab:eye_tracking_params_glasses}
	\resizebox{\columnwidth}{!}{%
	\begin{tabular}{cccccccccc}
		\hline
		$\alpha_0$& $\alpha_{max}$ &$\beta_{max}$	&$f_s$  &$f_h$  &$f_v$  &$k_v$  &$d_1$  &$d_2$  &$d_3$  \\ 
		
		15$^\circ$ &	30$^\circ$&18$^\circ$  &22 MHz  &21kHz  &60Hz  &0.83  &22.5mm  &22.5mm  &7.3mm  \\\hline
	\end{tabular}
}
\end{table}
The theoretical gaze angle resolution of the proposed head worn eye tracking sensor is shown in \Cref{A2im:Gaze_angle_res_glasses}. 
\FloatBarrier
The offset angle $\alpha_0$ adds an increasing optical path length with increasing angles $\alpha$ which leads to an decrease in resolution for extreme eye rotational angles especially towards the lower left edge, as apparent from \Cref{A2im:Gaze_angle_res_glasses}. The theoretical mean single-pixel gaze angle accuracy for this setup is around 0.28\,$^\circ$ and is therefore a upper boundary estimation of the gaze angle resolution. A further increase in gaze angle resolution by the use of an pupil tracking algorithms is expected. 

Compared with the scanned laser approach of \cite{7863402}, the calculated gaze angle resolution of the proposed laser based eye-tracking sensor for the glasses geometry is higher especially for relevant eye rotation angles around the center. Compared to state of the art VOG eye-tracking sensors like \cite{Kassner:2014:POS:2638728.2641695}, our approach is capable to achieves higher gaze angle resolution. Based on this results a less complex pupil tracking algorithm can be used to reduce the computational effort and therefore power consumption, while keep comparable gaze angle resolutions to VOG sensors.

\subsubsection{Limitations}
The temporal resolution of the eye-tracking sensor is limited by the frame rate of the retinal projection system, which currently is $60\, $Hz. To overcome this limitation, a faster micro mirror could be used. However, as a collimated laser beam requires sufficient micro mirror aperture, miniaturization of the micro mirror is restricted. Therefore, frequencies up to about $120 \, $Hz are technically feasible. 

The low-power consumption is achieved because components already contained in the projection system such as the micro-mirror module are not considered in the eye tracking power budget. This is valid as long as the sensor is a replacement for an VOG sensor in a retinal projection system, for example. For stand-alone application of the eye tracker, contribution of the micro-mirror module to the power budget has to be considered.

\subsection{Conclusion}
We presented a novel eye-tracking sensor for integration into our existing retinal projection AR glasses prototype. 
Compared to VOG eye-tracking sensors, our technology achieves a significant reduction in power consumption at comparable gaze angle resolution and full integration into a frame temple. Besides the photodiode in the frame temple, the sensor is completely invisible to the user. Furthermore, our approach is used in combination with state of the art robust VOG eye-tracking algorithms. 

The achievable gaze angle resolution is evaluated experimentally based on a laboratory setup. It shows comparable gaze angle resolution in comparison with VOG sensors. In addition the theoretical gaze angle resolution of the proposed head worn eye tracking sensor is calculated, which leads to an even better gaze angle resolution as the laboratory setup.  

Based on these results the next step is to integration of the proposed eye tracking sensor into our AR glasses demonstrator to perform experiments and evaluate the sensor under real conditions, e.g. in the presence of various illumination conditions.

Additionally, the eye tracking algorithms used must be transferred to an embedded platform and integrated into the demonstrator.

\newpage

\section{A Novel -Eye-Tracking Sensor for AR Glasses Based on Laser Self-Mixing Showing Exceptional Robustness Against Illumination}
\label{APP:A3}
\subsection{Abstract}
The integration of eye-tracking sensors in next-generation AR glasses will increase usability and enable new interaction concepts. Consumer AR glasses emphasize however additional requirements to eye-tracking sensors, such as high integratability and robustness to ambient illumination. We propose a novel eye-tracking sensor based on the self-mixing interference (SMI) effect of lasers. In consequence, our sensor as small as a grain of sand shows exceptional robustness against ambient radiation compared to conventional camera-based eye trackers. In this paper, we evaluate ambient light robustness under different illumination conditions for video-based oculography, conventional scanned laser eye tracking as well as the SMI-based sensor.

\subsection{Introduction}
\label{A3sec:Introduction}
Next generation AR glasses are already integrating eye-tracking sensors to enable new gaze-based interaction concepts with the glasses and the surroundings \cite{8409769, Salehifar2019, doi:10.1111/cgf.13654}. In addition, new rendering schemes to reduce the power consumption and increase the image quality during image projection like foveated imaging are introduced \cite{Kim:2019:FAD:3306346.3322987, Maimone:2017:HND:3072959.3073624}. These rely mainly on eye-tracking for pupil position estimation \cite{Jang:2017:RAR:3130800.3130889}.

The state of the art eye-tracking sensor technology is video oculography (VOG). VOG sensors use infrared (IR) illumination and an IR camera to capture images of the eye surface and determine the pupil position utilizing image processing techniques. Such technology is meanwhile well established and provides high accuracies \cite{10.1117/12.2322657}. 

To perform best, VOG systems require high-contrast, disturbance-free eye images. In real world conditions, variations in illumination affect image quality and in the end lead to a low pupil detection ratio \cite{Fuhl2016}. To improve the detection ratio, recent algorithms using deep neural networks are applied which introduce additional computational effort and high power consumption for real-time applications \cite{Kim:2019:NAD:3290605.3300780}. 

To overcome the susceptibility of VOG sensors to ambient light, we propose a novel eye-tracking sensor for AR glasses based on lasers featuring the self-mixing interference (SMI) effect \cite{10.1117/12.775131}. Our main contribution is the successful application of the SMI effect for an exceptionally robust eye-tracking sensor as well as systematic comparison to competing technologies. We use two micro-electro-mechanical system (MEMS) micro mirrors to scan the IR laser beam of a vertical cavity surface emitting laser (VCSEL) with a cavity-integrated photodiode across the eye region. Strong pupil signals are based on the high reflectivity of the retina in the infrared wavelength range.

The remaining of the paper is organized as follows: Section 2 discusses previous work in the area of MEMS scanned laser eye-tracking sensors. In Section 3, we explain the underlying SMI VCSEL technology and introduce an index for disturbance robustness. The evaluation in Section 4 compares the performance of the selected three technologies under different artificial illumination conditions up to direct exposure to an intense camera flash. Section 5 concludes this work and gives a brief outlook to future work.


\subsection{Related Work}
\label{A3sec:Related Work}
MEMS scanned laser eye-tracking sensor concepts categorize into two classes that differ in the way of scanning the eye surface as well as in the algorithm for pupil position estimation.

Sarkar et al. \cite{7181058} introduced small MEMS micro mirrors to scan a laser beam vertically and horizontally across the eye's surface. The micro mirror and the IR laser were attached to the frame temple of the glasses and a photodiode was integrated close to the nose pads \cite{7863402}. The laser beam was scanned along a straight line over the eye. The micro mirror angle at which the photodiode observed a corneal glint was found to be linked to the horizontal pupil position. For the vertical pupil position, a hill climbing algorithm based on the amplitude shifts of the photodiode output in horizontal scanning direction was used. An optical bandpass filter in front of the photodiode reduced susceptibility to ambient radiation. Sakar et al. \cite{7863402} reported an angular gaze resolution of approximately 1$^\circ$ with a temporal resolution of $3300\,$Hz at a power consumption of less than $15\,$mW. Recently, a further implementation of this sensor concept was presented by \cite{Osram2019}.

One drawback of this approach is a high susceptibility to movements of the glasses movements of the glasses while being worn. As mirror scan angles at glint occurrence are directly linked to pupil positions, recalibration is required each time the glasses move \cite{7863402}.

To approach shift-invariance, \cite{Brother2007} used a different scan path of the IR laser beam. Their setup was based on two micro mirrors for horizontal and vertical deflection to scan the IR laser beam across the entire eye region. An external photodiode captured the reflectivity of different regions of the eye that could be used to reconstruct an image of the eye region. Then, state of the art VOG image processing algorithms were used to extract pupil positions. Again, an optical long-pass filter reduced susceptibility to ambient radiation. However, only radiation of shorter wavelength can be blocked by such a filter and interference with infrared sources is possible.

Both prior works used an external photodiode to record laser beam reflectivity in the eye region. Our approach exploits the SMI effect to record such a reflectivity map at outstanding robustness to ambient radiation as the laser cavity itself is a very narrow bandpass and the SMI effect relies on the laser's coherence length.

\subsection{Methodology}
\label{A3sec:Methodology}
\begin{figure}[h]
	\centering
	\includegraphics[width=0.3\linewidth]{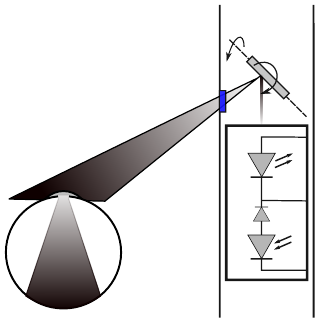}
	\caption{Setup of the proposed eye-tracking sensor integrated in a frame temple}
	\label{A3im:sensor_setup}
\end{figure}
\FloatBarrier
Figure \ref{A3im:sensor_setup} shows the IR VCSEL with integrated photodiode and a MEMS micro mirror integrated in a frame temple. The MEMS micro mirror deflects the laser beam of the IR VCSEL towards the eye. By using a two-axes mirror, a rectangular region covering the eye is scanned.

At a specific mirror scan angle, the IR laser beam enters the eye through the pupil and is retro-reflected towards the laser cavity. The retina acts as retro-reflector similar to VOG eye-tracking methods based on the bright eye effect \cite{Duchowski2017}. This results in a strong interference inside the laser cavity, which is observable at the integrated photodiode output. Aligning this photodiode output as a grey-value pixel with the mirror scan angles, a two dimensional reflectivity image of the scan region is formed. Then, pupil extraction can be done in a similar way to the VOG sensors. Configuring the laser power in such a way that strong SMI interference is only maintained for retinal reflections. This significantly enhances the image contrast and much simpler image processing algorithms can be applied.

\subsubsection{Self-mixing interference laser sensor}
\label{A3subsec:Self-mixing interference laser sensor}
The proposed eye-tracking sensor relies on a VCSEL semiconductor laser. The VCSEL's technology has several advantages over edge emitting lasers (EEL). They are characterized by a very low threshold current and thus a low power consumption. Furthermore, they have small dimensions and can easy be manufactured and tested, enabling low cost devices \cite{Michalzik2013}.
 
\begin{figure}[h!]
	\centering
	\includegraphics[width=0.2\linewidth]{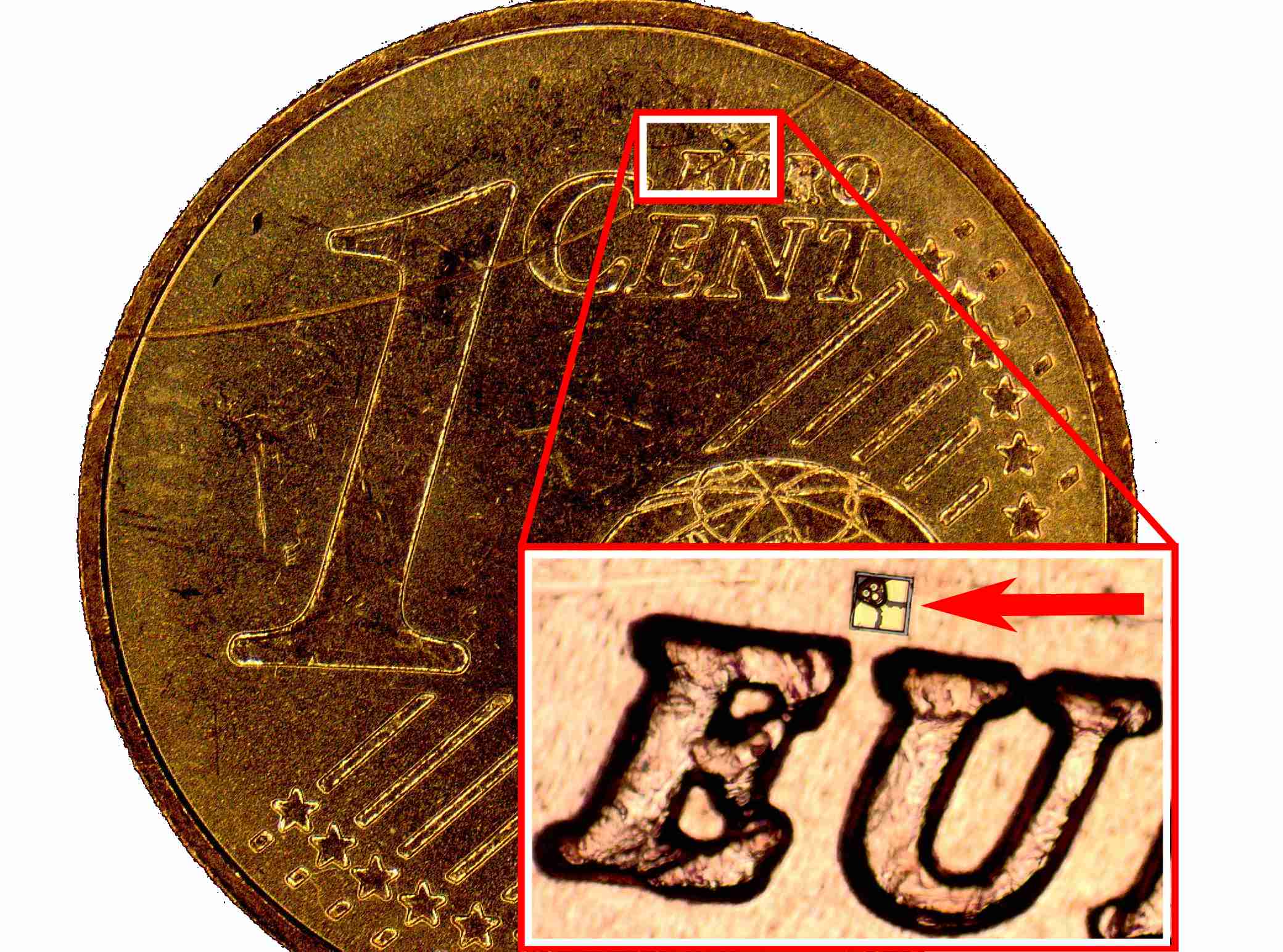}
	\caption{Size of the sensor in comparison to a Euro cent coin. The size of the sensor is approximately 180x180 $\mu$m.}
	\label{A3im:sensor_size}
\end{figure}

Due to the small size of the aperture, a circular emission profile and low beam divergence is achieved. This allows the use of small optical components, which results in a high integratability in the glasses frame temple \cite{Michalzik2013}. The size of the sensor without external optical components is shown in Figure \ref{A3im:sensor_size}.  

The SMI effect relies on backscattering of a fraction of the coherent laser light into the laser cavity. Inside the cavity, it interferes with the lasing field and modulates the amplitude and frequency of the lasing field. The cavity serves as an optical mixer and amplifier as well as an optical filter due to the narrow optical bandwidth \cite{Michalzik2013}.

To obtain information about the reflectivity of the structure that causes backscattering, the optical power of the modulated laser field is recorded by a photodiode. In our sensor, the photodiode is directly integrated into the bottom reflector of the cavity yielding a tiny sensor module \cite{grabherr2009integrated}. 

\FloatBarrier


\subsubsection{Image quality metric}
VOG pupil detection algorithms rely on the recognition of features such as the pupil's edges. To achieve a high pupil detection rate however, a high contrast and a noise-free image with suitable resolution is required \cite{Fuhl2016}. To take these requirements into account, the structural similarity index (SSIM) is used as an image quality metric to describe the robustness of an eye-tracking sensor against disturbance by ambient light.

The SSIM expresses image quality degradation as a combination of structural, illuminance and contrast distortions of a disturbed image $F$ relative to a reference image $R$  \cite{10.1117/1.1748209}. It is calculated as
\begin{equation}
	SSIM(R,F) = \frac{(2\mu_R \mu_F + C_1) (2 \sigma_{RF} + C_2) }{(\mu_R^2 + \mu_F^2 + C_1) (\sigma_R \sigma_F +C_2)}.
	\label{A3equ:SSMI}
\end{equation}
Here, $\mu$ and $\sigma$ describe the mean illuminance and standard deviation of illuminance, respectively, which can be interpreted as an estimation of the image contrast. The cross correlation $\sigma_{RF}$ between $R$ and $F$ expresses structural distortions and the constants $C_1$ and $C_2$ are required to avoid the denominator becoming zero.

The index is calculated using a local $N \times N$ sliding window over the whole frame to consider local distortions. Then, the arithmetic mean of all $m$ window positions represents an overall quality metric, the mean structural similarity index (MSSIM):
\begin{equation}
	MSSIM(R,F) = \frac{1}{m}\sum_{j=1}^{j} SSIM(R_j,F_j) .
	\label{A3equ:MSSIM}
\end{equation}


\subsection{Evaluation}
\label{A3sec:Evaluation}

\begin{figure}[h]
	\centering
	\includegraphics[width=0.5\linewidth]{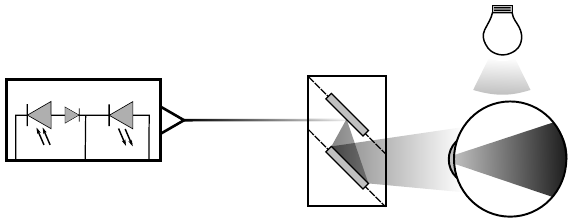}
	\caption{Laboratory setup to evaluate the disturbance immunity of the proposed eye-tracking sensor against disturbances induced by artificial light sources.}
	\label{A3im:lab_setup}
\end{figure}
\FloatBarrier
Figure \ref{A3im:lab_setup} shows the laboratory setup of the proposed eye-tracking sensor. The laser beam of the VCSEL is directed to the micro mirror module, which consists of two MEMS micro mirrors. Each micro mirror deflects the laser beam in one orthogonal direction resulting in a two-dimensional scan region covering the eye and its surroundings. Synchronization signals indicate the start of a new frame as well as the start of a new horizontal line. These signals are temporal synchronized with the output of the VCSEL's integrated photodiode and used to reconstruct a spatial mapping of the retro-reflectivity sensed by the SMI effect. 

\subsubsection{Ambient light sources}
\label{A3subsec:Artificial light sources}
\begin{figure}[h]
	\centering
	\includegraphics[width=0.65\linewidth]{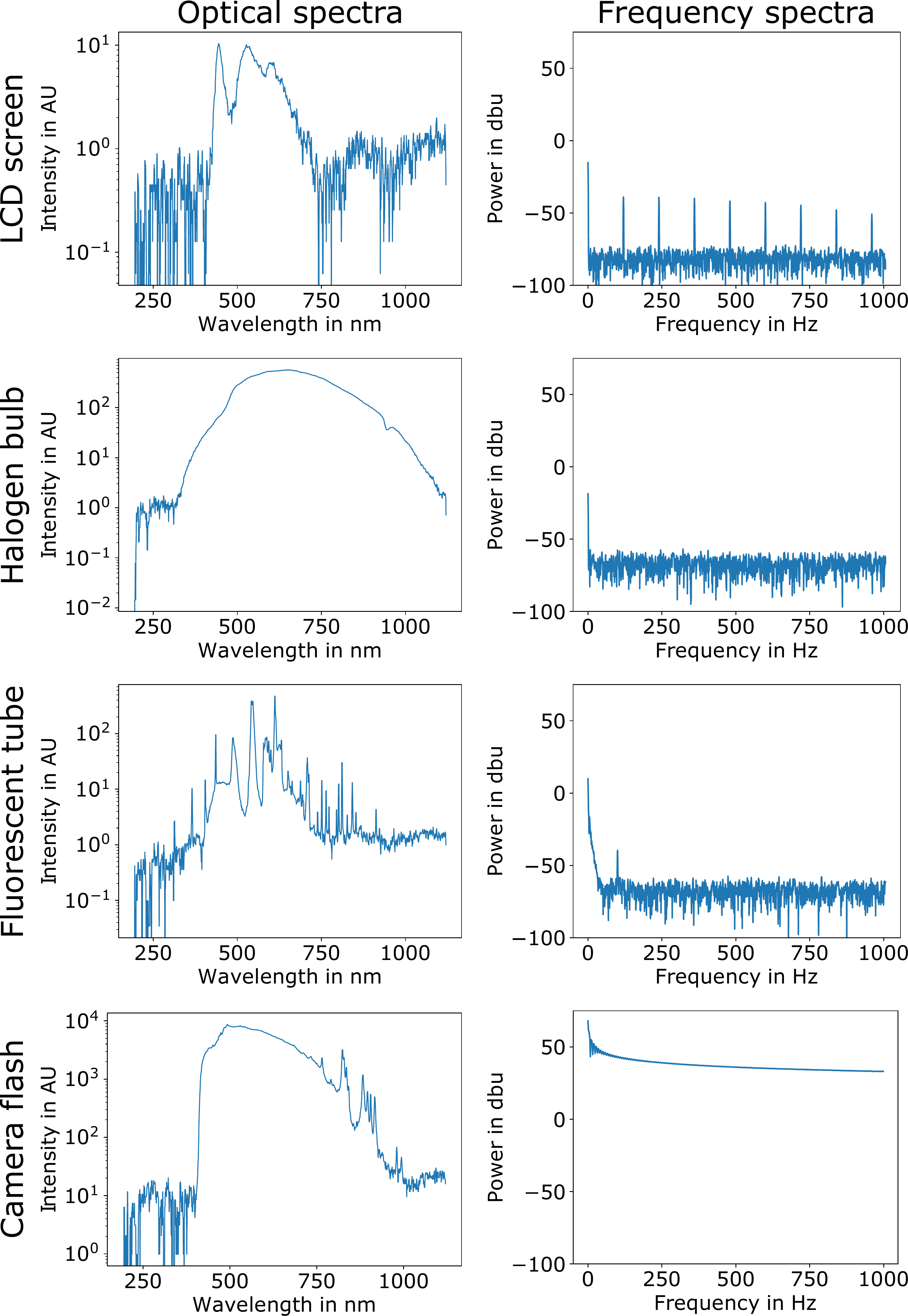}
	\caption{Optical and temporal characterization of the light sources used to disturb the three eye-tracking sensors.}
	\label{A3im:artifical_light_sources}
\end{figure}

To evaluate the robustness against ambient radiation of the proposed eye-tracking sensor, it is exposed to different light sources. To evaluate even extreme illumination scenarios, high illuminance, modulated light sources and light sources with broad wavelength spectra are used. Figure \ref{A3im:artifical_light_sources} shows the optical \footnote[1]{Measured with Ocean Optics HR4000} and frequency spectra \footnote[2]{Measured with Ultra fast photodiode UDP-200-SP}  of the light sources considered in the experiment.  
\FloatBarrier
To assess ambient sources relevant to data glasses, we considered a liquid crystal display (LCD) (watching TV or using PC), a fluorescent tube as well as a halogen bulb (indoor illumination) and a photoflash \footnote[3]{Canon Speedlite 580EX II} as very strong disturbance in terms of transient and spectral bandwidth.  Measurements in direct sunlight are not carried out, as no uniform conditions can be guaranteed for all measurements. For this reason the halogen lamp is used, which imitates the wavelength spectrum and intensity of the sun on a bright day.
\subsubsection{Disturbance immunity}
\label{A3subsec:Robustness measurements}
For each light source, we evaluate three different eye-tracking approaches: a system based on a scanned IR laser comparable to the system of \cite{Brother2007}, the proposed eye-tracking sensor based on the SMI effect and a commercial VOG sensor by Pupil Labs.

For each system, test cards optimized to the underlying technology are used to obtain a high resolution and contrast-rich image. Therefore, a chess pattern is used for the scanned IR laser sensor and the VOG sensor. For the SMI sensor a black coloured test chart with circular retro-reflectors of different diameters is used, mimicking pupils of different diameters. 
\begin{figure}[h]
	\centering
	\includegraphics[width=.7\linewidth]{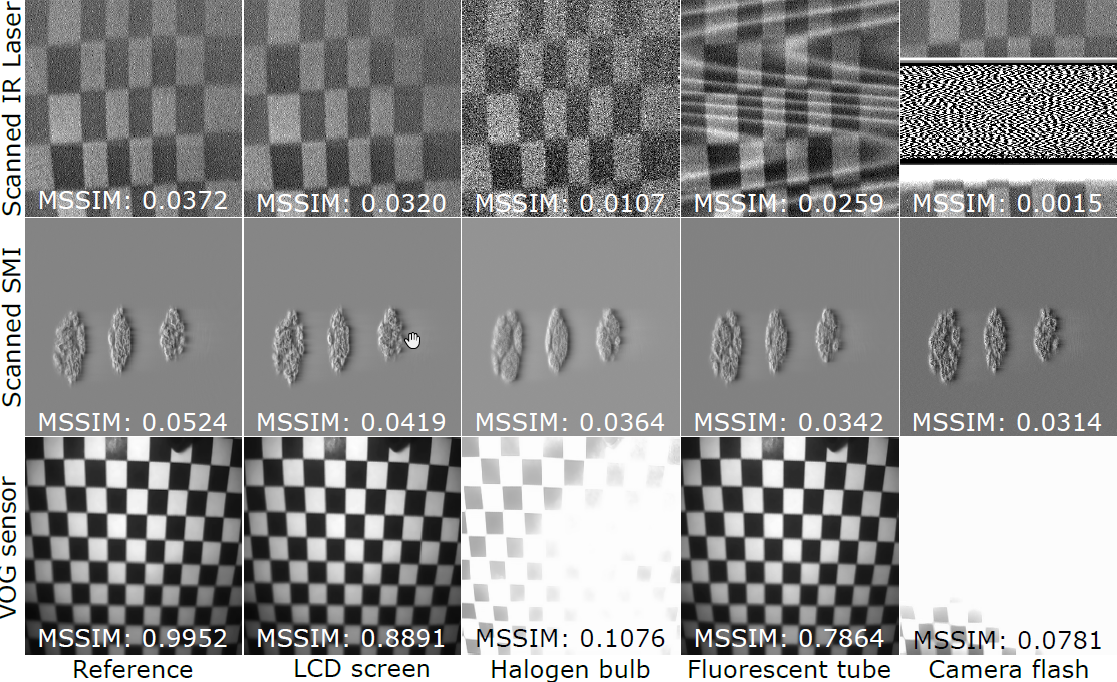}
	\caption{Results of the disturbance evaluation for the three eye-tracking sensors.}
	
	\label{A3im:images_with_MSSMI}
\end{figure}
\FloatBarrier
First, the eye-tracking sensors are placed in front of their respective test chart to capture a high quality disturbance free reference image $R$ without any external illumination. Afterwards, the measurement is repeated for each of the four chosen light source to capture a disturbed image $F$. Figure \ref{A3im:images_with_MSSMI} shows the resulting images $F$ for each eye-tracking sensor and illumination condition.


Afterwards, the MSSIM is calculated by Figure \ref{A3equ:SSMI} and Figure \ref{A3equ:MSSIM} with $C_1 = C_2 = 0.26 $ and $N=9$ similar to \cite{1284395}. For differentiation of intra-sensor image quality variations to ambient light disturbance, the MSSIM of two consecutive reference images (one as $R$ and one as $F$) is calculated as a representation of sensor noise. The MSSIM is included as annotations for every experiment in Figure \ref{A3im:images_with_MSSMI} with a low MSSIM indicates a low repeatability within one row.The scanned IR laser sensor shows an overall low MSSIM due to Gaussian noise of the photodiode and a reduced dynamic range due to currently low amplification in the analog frontend. With rising illumination, noise increases and affects the MSSIM. Further, periodic flickering of the fluorescent tube and the high illumination change by the photoflash produce strong interference in the images, thus leading to poor MSSIM. 

The scanned SMI setup shows a low reference MSSIM due to noise induced by speckling on the targets surface as well. Image noise is considered a minor impact for eye-tracking using SMI signals, however directly affects the MSSIM. However, exceptional robustness to ambient radiation is obvious from the images, as other disturbances like periodic flickering and high illumination changes (photo flash) are almost invisible in the recorded images.

\begin{figure}[h]
	\centering
	\includegraphics[width=0.5\linewidth]{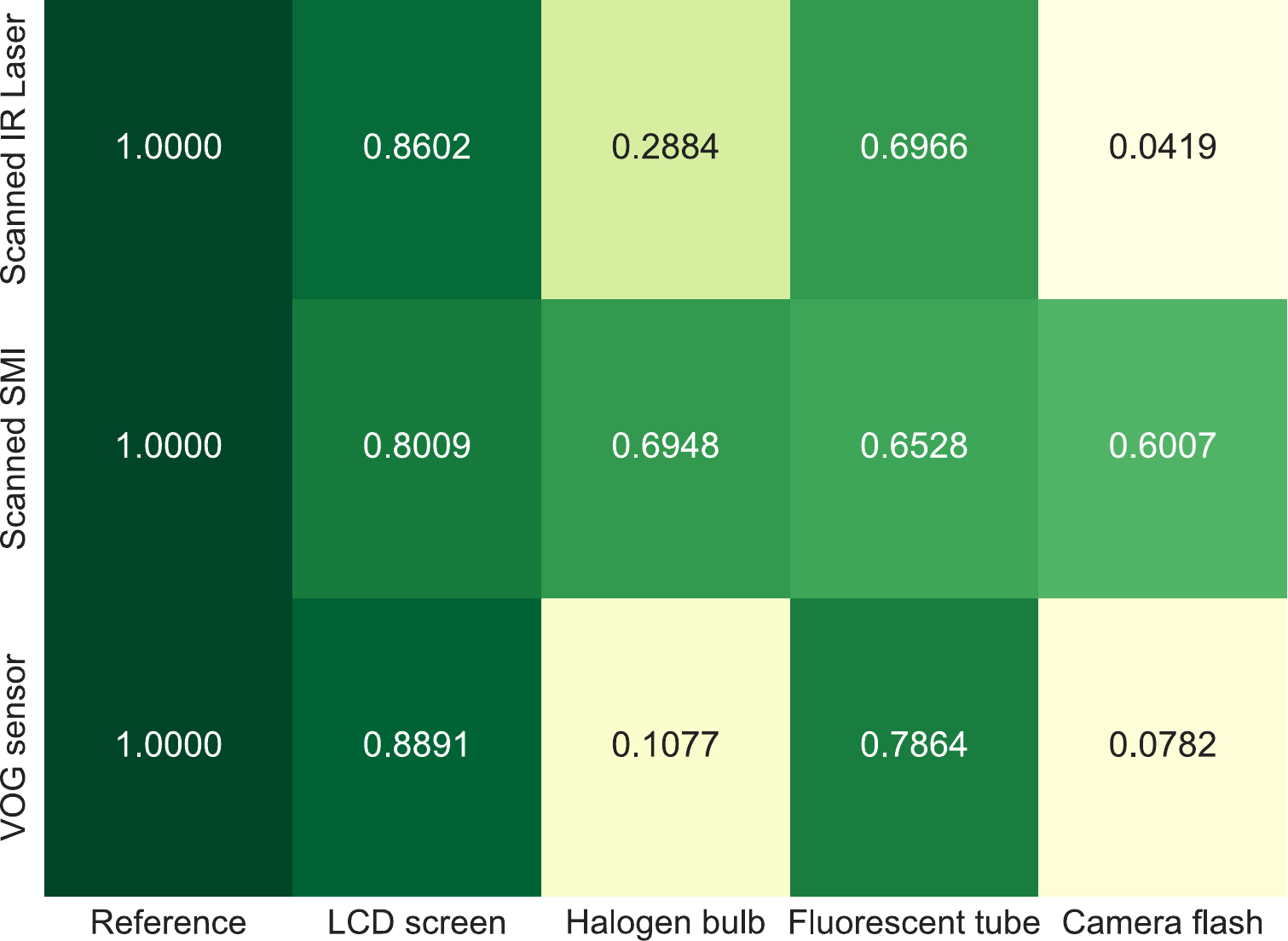}
	\caption{Visualization of the MSSIM-n to compare the eye-tracking sensors for the different illumination conditions.}
	\label{A3im:MSSMI}
\end{figure}

The VOG sensor as our eye-tracking Gold Standard reference is characterized by an overall high MSSMI and very good image quality. Even periodic flickering of fluorescent lights is effectively suppressed in the sensor. Only the strong illumination change of the photo flash strongly affects the image quality and leads to an low MSSIM. To compare these three very different eye-tracking sensor principles, the MSSIM is normalized by dividing the respective MSSIM by the MSSIM of the corresponding reference frame. This way, repeatability issues are suppressed to emphasize on ambient light robustness. Figure \ref{A3im:MSSMI} shows the resulting normalized MSSIM (MSSIM-n) with high values indicating high image quality.
\FloatBarrier
Concluding, the proposed scanned SMI sensor is very robust in terms of ambient light suppression seen in overall high MSSMI-n indices. Although overall imaging quality is poor compared to recent VOG systems, the technology is well suited for eye-tracking. Exceptional image contrast is not required as the sensor principle exploits the retro-reflectivity of the retina to separate the pupil from surrounding eye regions. This way, edge- or template-based pupil detectors are not required. Based on this feature, the pupil position is detectable with less computational effort. In addition rotations of the eye in relation to the sensor may result in partial shadowing of the retina by the sclera. This leads to distortions of the image of captured retro-reflections, which can be used as additional features to estimate the gaze angle. For extreme eye rotations, the optical path of the laser beam can be deflected via a parabolic mirror to reduce large angles of incidence and cover the whole eye region.

\subsection{Conclusion}
\label{A3sec:Conclusion}
We presented a novel eye-tracking sensor based on MEMS micro mirrors and a laser featuring the SMI effect. Compared to VOG sensors and scanned laser eye-tracking with separate photodiode, exceptional robustness to ambient radiation is shown. Paired with the sensor's low power consumption and its tiny size, it is a promising technology for fully integrated always-on ubiquitous eye-tracking in next generation AR glasses.

An important next step is to repeat the experiment with the human eye instead of test charts, especially as there exists only little data about the SMI effect from human retinal reflections. Furthermore, the pupil detection accuracy and precision of the proposed eye-tracking sensor will be evaluated. For outdoor use, robustness to sunlight should be analysed, although no relevant degradation is expected following to the halogen bulb and photo flash experiments.

One promising increase the pupil-iris contrast further sensor is the modulation of the VCSEL wavelength by current modulation. Similarly to frequency modulated continuous wave (FMCW) radar systems, this introduces depth resolution of the sensor.  However, for enhanced performance, modifications of pupil detection algorithms are required to exploit these new sensor capabilities.

\newpage

\section{A Highly Integrated Ambient Light Robust Eye-Tracking Sensor for Retinal Projection AR Glasses Based on Laser Feedback Interferometry}
\label{APP:A4}
\subsection{Abstract}
Robust and highly integrated eye-tracking is a key technology to improve resolution of near-eye-display technologies for augmented reality (AR) glasses such as focus-free retinal projection as it enables display enhancements like foveated rendering. Furthermore, eye-tracking sensors enables novel ways to interact with user interfaces of AR glasses, improving thus the user experience compared to other wearables. In this work, we present a novel approach to track the user's eye by scanned laser feedback interferometry sensing. The main advantages over modern video-oculography (VOG) systems are the seamless integration of the eye-tracking sensor and the excellent robustness to ambient light with significantly lower power consumption. We further present an algorithm to track the bright pupil signal captured by our sensor with a significantly lower computational effort compared to VOG systems. We evaluate a prototype to prove the high robustness against ambient light and achieve a gaze accuracy of 1.62\,$^\circ$, which is comparable to other state-of-the-art scanned laser eye-tracking sensors. The outstanding robustness and high integrability of the proposed sensor will pave the way for everyday eye-tracking in consumer AR glasses.

\subsection{Introduction}
\label{A4sec:Introduction}
Robust and highly integrated eye-tracking sensors are a key technology to improve resolution of display technologies like focus-free retinal projection for augmented reality (AR) glasses e.g. by enabling display enhancement methods like foveated rendering  \cite{Kim:2019:FAD:3306346.3322987, 9005240, Kaplanyan:2019:DNR:3355089.3356557}. Furthermore eye-tracking allows to steer the exit pupil increasing the display's field of view (FOV)  \cite{Kim:2019:FAD:3306346.3322987, EyeWay2021} of AR glasses. In addition to display enhancement techniques, eye-tracking sensors enable novel ways to seamlessly interact with the user interface of AR glasses \cite{DUCHOWSKI2018b, Meyer2021, meyer11788compact}, improving thus the user experience.

\begin{figure}
	\centering
	\includegraphics[width=0.8\textwidth]{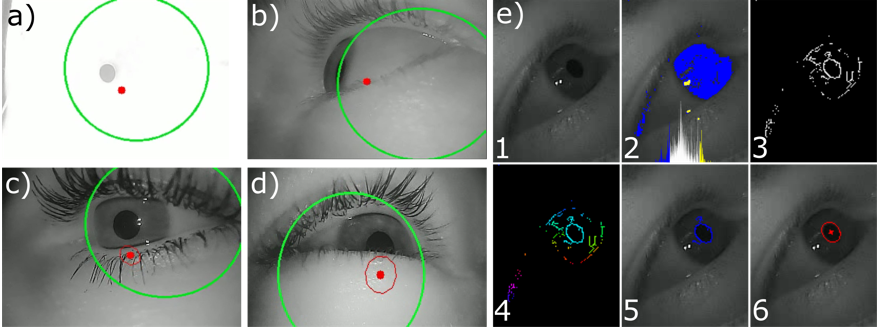}
	\caption{Key challenges of VOG based eye-tracking sensors is a robust detection of the pupil, which is limited due to a) limited dynamic range of camera sensors to operate under a wide range of ambient illumination settings e.g. in bright sun light, b) detection of the pupil over the whole field of view due to the high off-axis integration of camera sensors, c) false pupil detection e.g. due to mascara, other disturbances, dirt on the lens and d) false detection due to partly occluded pupils by lashes or eye lids. Furthermore the pupil detection is rather computational complex as several image processing steps are required to extract the pupil as shown in e)}
	\label{A4fig:teaser}
\end{figure}

Video oculography (VOG)camera sensors are the state-of-the-art in mobile eye-tracking, tracking either the pupil in the 2D image and estimate gaze using a geometric 3D eye model \cite{Swirski2013}, or track the pupil and corneal reflections from additional infrared (IR) LEDs and use a regression-based approach to determine gaze direction, as shown by \cite{1634506}. In both cases, the key to robust eye tracking is robust detection and tracking of the pupil under a variety of conditions, which is, as shown in \Cref{A4fig:teaser}, not always the case with current VOG systems.

A well-known issue with state-of-the-art VOG sensors is the limited dynamic range of camera sensors, leading to a loss of the pupil signal in presence of varying ambient light or in bright sun light \cite{10.1371/journal.pone.0087470, Fuhl2016}, as illustrated in \Cref{A4fig:teaser} a).

Furthermore the high off-axis integration of camera sensors in current VOG systems, as illustrated in \Cref{A4fig:teaser} b) \cite{tonsen2020high, Kassner:2014:POS:2638728.2641695}, leads to a loss of pupil detection especially if the gaze vector points away from the cameras optical axis, which allows robust eye-tracking only in a part of the user's FOV \cite{vision5030041}. This problem is solved by adding more camera sensors to cover a larger field of view, such as proposed by \cite{TobiiPro2021} or \cite{10.1145/3130971}. However, this leads to more complex sensor integration as well as higher power consumption stem from additional sensors and higher complexity eye-tracking algorithms.

Additional cases which leads to a false pupil detection are due to the wearing of mascara \cite{10.1145/3379155.3391330}, as false edges are considered as pupil edges by the pupil detection algorithm, illustrated in \Cref{A4fig:teaser} c). A similar case is shown in \Cref{A4fig:teaser} d) where a part of the pupil is occluded by the eyelid, which also leads to a false pupil detection \cite{Fuhl_Hospach_Kasneci_2017, Fuhl2016}. This issues is addressed by more advanced pupil detection algorithms e.g. by using convolutional neural networks like PupilNet \cite{fuhl2017pupilnet}, RITNet \cite{chaudhary2019ritnet} or the Deep VOG approach by \cite{YIU2019108307}. The main drawback of these advanced algorithms is increased demand of processing power which increases the power consumption of VOG eye-tracking systems.  

Finally, VOG algorithms require several steps of image processing to extract pupil features from camera images, as shown in \Cref{A4fig:teaser} e), which illustrates the processing steps of the VOG algorithm proposed by  \cite{Kassner:2014:POS:2638728.2641695}. There are several variants of the algorithm with optimization of individual steps of the pupil detection pipeline to improve detection accuracy and robustness, e.g. ELSE \cite{fuhl2016else}, PURE \cite{santini2018pure} or PUREST \cite{santini2018purest}. The increased robustness is accompanied by higher computational requirements. In recent years this issue is addressed by pupil detection algorithms which are optimized with respect to computational requirements and latency e.g. by \cite{fuhl2020tiny} or \cite{fuhl20211000}.

The presented disadvantages of VOG sensors and the corresponding power consuming eye-tracking algorithms indicate that the sensing technology itself puts some hurdles for eye-tracking sensor integration into AR glasses. To enable robust eye-tracking and overcome these limitations, we introduce a low power eye-tracking sensor approach using laser feedback interferometry (LFI) sensing technology to integrate eye-tracking capability into retinal projection AR glasses.

The LFI sensor is composed of a tiny vertical cavity surface emitting laser (VCSEL), operating at the infrared (IR) regime. In addition, a photodetector is integrated into the laser cavity using semiconductor processes. The small sensor size enables high integration into the frame temple of AR glasses. Integration of the photodetector enables the LFI sensing method, a coherent sensing method, leading to a high robustness against ambient light as most light stemming not from the lasers own radiation is suppressed \cite{Michalzik2013}. Thus the sensor is capable to robustly operate in presence of ambient light \cite{Meyer2020e}.

To solve the sensor integration problem and the high-off-axis integration of VOG systems, we further propose to integrate the LFI IR laser sensor into a retinal projection AR glasses system which consists of a micro-electro-mechanical system (MEMS) micro mirror based laser scanner and a holographic optical element (HOE) to steer the laser beam towards the eye.

Furthermore, we exploit the unique sensing modality of the LFI sensor and propose a low-power pupil detection and tracking algorithm by exploiting the characteristic bright pupil signal.

Our contribution is three fold:

(i) We propose an highly integrated eye-tracking sensor approach for retinal projection AR glasses based on an ambient light robust LFI sensor. By combining the LFI sensor with a highly transparent IR HOE and a MEMS micro mirror we further solve the highly-off-axis sensor integration. In addition, the eye tracker is invisible to the user, as it is fully integrated into the frame temple. 

(ii) We propose an algorithm optimized to detect and track the pupil based on the characteristically bright pupil signal captured by the LFI sensor and 

(iii) We evaluate the resulting gaze accuracy of the proposed algorithm and the ambient light robustness of the proposed sensor experimentally in a prototype setup.

Compared to the work of \cite{9149591} we switch from an IR laser with external photodiode to the LFI sensor with integrated photodiode and further show the high integratability into a glasses frame. In addition we manufacture the high transparent IR HOE, which is mandatory for the system.

Compared to the work of \cite{Meyer2020e} we apply the LFI sensor to human eyes and proof the proposed bright pupil effect. We further evaluate gaze accuracy in a human study with 20 participants and further propose an power saving algorithm for pupil detection algorithm.   

In the upcoming Section the state-of-the-art w.r.t. scanned IR laser eye-tracking sensors is discussed. Afterwards, in \Cref{A4sec:Methodology}, we introduce the proposed eye-tracking sensor and describe briefly the system components. Furthermore, we describe the underlying sensing principle of the LFI sensor technology as well as the origin of the observed bright pupil pattern. In addition, we describe our algorithm to detect and track the pupil. In \Cref{A4sec:Evaluation}, we describe our setup used to evaluate the gaze accuracy and compare it to a VOG system. Further, we show the robustness against artificial light. Finally, we compare our results with other state-of-the-art scanned IR laser eye-tracking approaches, discuss the applicability for AR glasses w.r.t. power consumption, sensor integration, glasses slippage and system latency, and finally draw a conclusion from our work.

\subsection{Related Work}
\label{A4sec:Related Work}

One of the first works which address scanned IR laser eye-tracking technology for AR glasses was introduced by \cite{7181058}. The authors used a 2D MEMS  mirror to scan the beam of an laser operating in the IR regime in a 2D pattern over the eye's surface. The photodiode, which receives backscattered light, was integrated close to the nosepad while the scan unit consisting of the IR laser and the 2D MEMS mirror were integrated into the glasses frame temple. The photodiode detects corneal reflections originating from the eye's surface \cite{7863402}. To obtain the horizontal gaze angle, the MEMS mirror scan angle under which a corneal reflection was detected by the photodiode is captured. To further obtain the vertical gaze angle \cite{7863402} proposed a hill climbing algorithm using the  the photodiode amplitude variation as feature. To address ambient light robustness, an optical bandpass filter was applied to the front of the photodiode. The authors reported a gaze resolution of  $\approx$ 1$^\circ$ with an update rate of $3300\,$Hz while their system consumes less than $15\,$mW power.

A major drawback of their method is the vulnerability to glasses slippage. As gaze angles are directly linked to the MEMS mirror scan angles via calibration, the system requires calibration after occurrence of slippage of the glasses \cite{7863402}. 

To achieve slippage robustness, the authors most recently released MindLink \cite{Mindlink2021}, which incorporates five photodiodes attached around the spectacle frame and a 2D MEMS micro mirror placed in the nose pad of the glasses. With this improved setup, the authors reported a gaze accuracy of <1$^\circ$ over a FOV of 40$^\circ$ x 25$^\circ$ and achieved an update rate of 500\,Hz.

\cite{9149591} approach slippage robustness by scanning an IR laser beam with a 2D scan path over the surface of the eye. The scan path was formed using two 1D MEMS mirrors for vertical and horizontal deflection. Backscattered light from the eye is measured by a photodiode, which is placed in the frame temple. The measured intensity variation over both the horizontal and the vertical scan angles is used to construct a gray scale image of the eye's surface. By applying a state-of-the-art VOG algorithm \cite{Fuhl:2018:CCB:3204493.3204559}, they achieved a gaze accuracy of 1.31$^\circ$ with an $60\,Hz$ update rate. The authors further reported a power consumption of 11\,mW and estimated a theoretical resolution of 0.28$^\circ$ with an improved optical design. To increase robustness against ambient light they propose to use optical filters in front of their photodiode circuitry, similar to \cite{7863402}.  

Most recently, EyeWay Vision \cite{EyeWay2021} released a scanned IR laser based eye-tracking sensor to steer the exit pupil for their retinal projection AR glasses. In a previous evaluation of the prototype system by \cite{10.1145/3379155.3391330} a gaze accuracy of 1.72$^\circ$ at a sampling rate of the corneal reflection signals of 4000\,Hz was reported. For absolute eye-tracking accuracy and to compensate translation movements of the eye with respect to the glasses e.g. due to slippage, a stereo camera with a sample rate of 120\,Hz was added to the laboratory setup.

All above-mentioned related approaches used a photodiode to capture back reflected light of an IR laser, which was scanned in a 2D pattern over the surface of the eye. Sarkar et. al.\cite{7863402, Mindlink2021} and EyeWay Vision \cite{10.1145/3379155.3391330} focus on glint features from the cornea, the limbus or the retina to estimate the gaze direction while \cite{9149591} reconstruct a gray scale image and extract the dark pupil from the image by applying state-of-the-art VOG algorithms. 

All methods have drawbacks with respect to ambient light robustness, which are addressed by protecting the photodiode with optical filters from ambient light. Furthermore, the glint feature based approaches by Sarkar et. al. and EyeWay Vision tend to have issues with slippage. To address this issue they either add an reference sensors or additional photodiodes, which adds to the overall power budget of these systems. Furthermore the approaches of \cite{9149591} and \cite{10.1145/3379155.3391330} are validated only in a laboratory setup and the sensor integration is not fully solved.

In our approach, we address the issue of ambient light robustness and sensor integration by using the LFI sensor technology. We further follow the path of \cite{9149591} and use a 2D scan pattern to reconstruct gray scale images to extract the bright pupil feature. With this approach, we address the issue of a high power consumption by exploiting the bright pupil effect to directly detect the pupil in an image to reduce computational complexity.

\subsection{Scanned laser feedback interferometry}
\label{A4sec:Methodology}

\begin{figure}[h]
	\centering
	\includegraphics[width=0.3\linewidth]{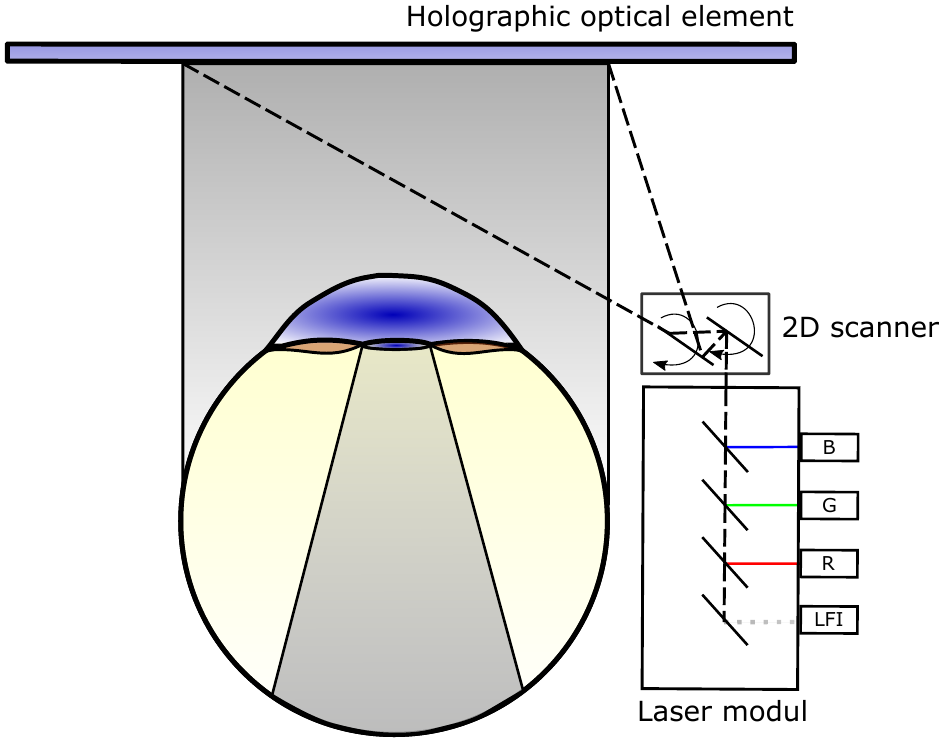}
	\caption{The LFI sensor added to the RGB module and shares the same optical path as the visible light. The holographic optical element acts as a wavelength selective mirror and redirects the scanned laser pattern to the eye's surface.}
	\label{A4im:laser_projector}
\end{figure}

\Cref{A4im:laser_projector} illustrates the integration of the LFI sensor into the retinal projection AR glasses to form a scanned LFI eye-tracking sensor.
The LFI sensor component is added to the RGB module, integrated in the glasses frame temple. The IR laser of the VCSEL is coupled via a prism into the beam path of the visible light of the RGB lasers and the combined beam is scanned via a MEMS mirror module over the HOE surface. The HOE acts as a wavelength selective mirror which parallelised the incoming beam pattern and redirects it towards the eye region.

The HOE is recorded into a photopolymer (Bayfol HX TP photopolymer) by constructing a reference wavefront and an imaging wavefront and expose the photopolymer with both wavefronts.  As our photopolymer is only active for visible light, we recorded the HOE at a wavelength of 650\,nm with an angular offset such that if the HOE is played back at 850\,nm under an different angle the desired wavefront is reconstructed. \cite{Wilm2021} and \cite{10.1117/1.OE.60.8.085101} gave a detailed description of the recording HOEs using photopolymer with an angular offset.

\begin{figure}[h]
	\centering
	\includegraphics[width=0.25\linewidth]{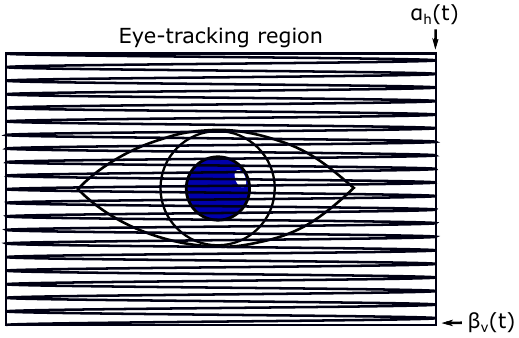}
	\caption{Scan pattern of the laser beam over the eye's surface.}
	\label{A4im:scan_pattern}
\end{figure}

The MEMS mirror module contains two 1D MEMS mirrors to scan in a 2D pattern over the HOE.  The horizontal mirror scans in a sinusoidal pattern, while the vertical MEMS mirror is non resonantly actuated using an electrodynamic driver to steer the sinusoidal pattern vertically over the HOE. With the known geometry and the mirror deflection angles $\alpha_h(t)$ and $\beta_v(t)$, the corresponding intersection point of the laser beam with the HOE can be calculated. For a detailed description of the geometry and the image generation we refer to \cite{9149591}. The scan pattern is illustrated in \Cref{A4im:scan_pattern}.
HOEs are characterized by a high wavelength selectivity and optical transparency allowing integrating them invisible to the user into the glasses lenses \cite{Wilm2021}.
\subsubsection{Laser feedback interferometry}	
The key element in our scanned eye-tracking approach is the LFI sensor. LFI is a widely applied interferometry sensing method \cite{Taimre:15}, which is used e.g. to measure displacement or velocity of solid targets. Recently, LFI sensors have also been applied to AR glasses e.g. for gaze gesture recognition \cite{meyer11788compact, Meyer2021} and human activity recognition \cite{CNN_Meyer_2021}. This works address the applicability of static LFI sensors for a near-eye setting. 
\label{A4sec:Laser Feedback interferometry}
\begin{figure}[ht]
	\centering
	\includegraphics[width=0.85\linewidth]{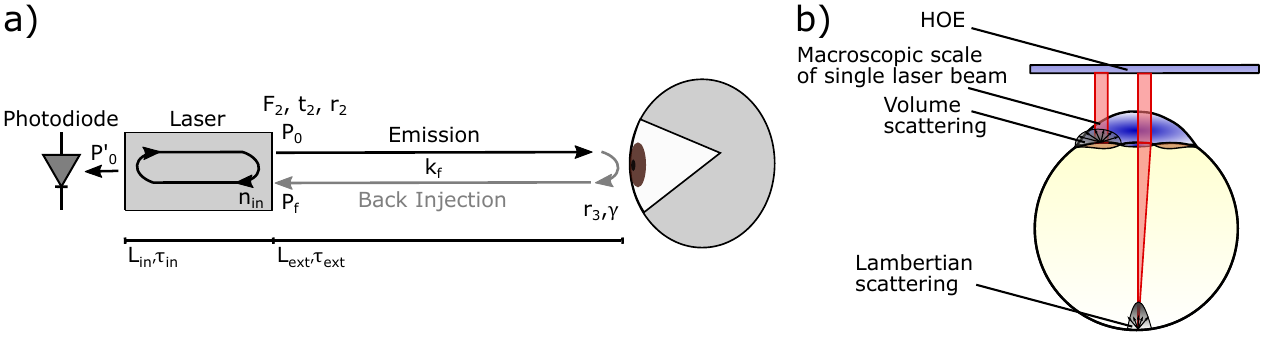}
	\caption{a) LFI sensing scheme modeled by the well known Coupled-cavity model. Emitted light from the laser is backscattered from the eye's surface and backinjected into the cavity. The photodiode integrated into the back mirror monitors the optical power inside the cavity, which changes based on variation of the feedback path \cite{meyer11788compact, Meyer2021}. \b) Macroscopic scale of the laser beam hitting the outer surface (sclera, iris) of the laser or the retina of the eye.}
	\label{A4im:coupled_cavity_model}
\end{figure}

To describe the basic sensing method of LFI sensors, the coupled-cavity model as shown in \Cref{A4im:coupled_cavity_model} a) is used. The laser with its cavity length $L_{int}$ and laser round trip time $\tau_{int}$ emits coherent light with an optical power $P_0$ towards the eye's surface. The laser hits the eye ball with an incident angle $\gamma$ and dependent on the reflectivity and absorption, summarized by $r_3$, and the scattering behavior of the tissue, a fraction of the emitted power $P_f$ is backinjected in the cavity of the laser. $\tau_{ext}$ describes the round-trip time of the laser to cross the distance $L_{ext}$. $\tau_{ext}$ is given by the speed of light $c_0$ and the refractive index $n_{ext}$ inside the external cavity \cite{Taimre:15}.


The backinjected wave interferes inside the cavity with the locally oscillating wave, which results in a optical power modulation of the laser
\begin{equation}
\label{A4equ:modulated_power}
P_0'= P_0 \left( 1 + m \cdot \cos \left(\phi_{fb}\right)\right).
\end{equation} 
The feedback power $P_0'$ relies on the laser's optical power $P_0$, the modulation depth $m$ and variations of the feedback phase $\phi_{fb}$. The photodiode inside the Bragg reflector measures a tiny fraction of the optical modulated feedback power $P_0'$ \cite{Taimre:15}.

While scanning the surface of the eye region, two effects influence the modulated feedback power $P_0'$. The first effect is an amplitude modulation due to varying reflectivity $r_3$ and scattering behavior of the different parts of the eye, which influence the modulation depth $m$. According to \cite{Coldren2012} $m$ is given in a second order approximation  by 
\begin{equation}
\label{A4equ:mod_depth}
m = 2 \cdot k_f \cdot \tau_p \cdot \left(\frac{\eta_i}{\eta_d}-1\right) + k_f \cdot \tau_{int} \cdot\left(1-F_2\right) \left(\frac{1+r_2^2}{t_2^2}\right).
\end{equation}

The feedback rate $k_f$ describes the normalized reflected field injection rate, $\tau_p$ the photon lifetime and the fraction of $\eta_i$ and $\eta_d$ the differential efficiency between pump efficiency and quantum efficiency of the cavity. $F_2$ describes the fraction of total power which is coupled out of the front mirror of the laser cavity. The mirror is further described by its transmitivity $t_2$ and $r_2$. The feedback rate can be rewritten with respect to the three-mirror model by 
\begin{equation}
k_f = \frac{t_2^2 \sqrt{\frac{P_0}{P_f}}}{r_2} / \tau_{int} \qquad with \qquad \sqrt{\frac{P_0}{P_{f}}} \propto \sqrt{\frac{r_3}{r_2}}.
\end{equation}
Considering a constant transmitivity $t_2$ and reflectivity $r_2$ of the front mirror and a constant output of laser power $P_0$, the coupling factor is mainly affected by a variation of the power of backscattered light $P_f$ due to an increase of reflectivity $r_3$ of the target and scattering behavior as shown in \Cref{A4im:coupled_cavity_model} b). 

\Cref{A4im:coupled_cavity_model} b) shows the macroscopic scale of a single laser beam reflected by the HOE for two deflection angles of the micro mirror. The left beam position describes the beam hitting the outer tissue of the eye (sclera, iris) where volume scattering effects dominate the overall scattering and thus a rather low portion of light is backinjected into the laser cavity.  The right beam position describes the beam hitting the retina. In this case the lens of the eye focuses the laser beam onto the retina and as the retina surface is dominated by Lambertian scattering \cite{Donnelly2008TheAH}, a large portion of light is back scattered. This effect is also referred as red eye effect or bright pupil effect, which is varying in severity across different human eyes \cite{10.1145/507072.507099}.

The second effect which influences the modulated feedback power $P_0'$ is given by the modulation of the feedback phase $\phi_{fb}$ due to speckling effects. Speckling describes the additive superposition of several backscattered signal components with random amplitude and phase. The sum of this components leads to a random modulation of the phase $\phi_{fb}$ of the back injected light and in particular whether constructive or destructive interference dominates the signal \cite{Goodman2020}. With respect to the eye this effect is well known from optical coherence tomography (OCT) imaging, where the signal from the retina is characterized by dark an bright speckle patterns \cite{10.1117/1.429925}.

\subsubsection{Bright pupil detection}
\label{A4subsec:Bright_Pupil_algorithm}
\begin{figure}[h]
	\centering
	\includegraphics[width=0.85\linewidth]{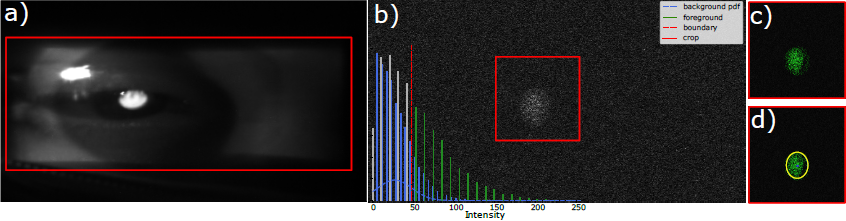}
	\caption{a) Image of the scan area (bright area inside red box) on a person's eye taken with an IR camera looking directly through the HOE from the outside. b) Background: Modulated feedback power $P_0' $ measured by the integrated photodiode of the LFI sensor over the scan area. Foreground: Histogram of the retinal area pixel intensity distribution (green and grey) and the non-retinal area  distribution (blue). c) Segmented bright  retinal area pupil pixels from b) using the intensity boundary (red dashed line). d) Multivariate Gaussian fit of the retinal area pixels in c) with pupil center in blue and pupil contour in yellow.}
	\label{A4im:pupil_extraction_pipeline}
\end{figure}
\FloatBarrier
\Cref{A4im:pupil_extraction_pipeline} a) shows the region of the eye, scanned by the LFI sensor. An IR camera looking directly through the HOE from the outside towards the eye. The pupil appears bright when the IR laser beam hits the retina during the 2D scan, also known as the bright pupil effect. This supports our assumption that the reflectivity $r_3$ as well as the scattering behavior changes and therefore changes the coupling factor $k_f$ in presence of the retina, resulting in amplitude modulation according to \Cref{A4equ:mod_depth}. In particular, by integrating the photodiode into the back reflector of the laser cavity, IR illumination and sensing element are perfectly aligned on axis to support the bright pupil effect. In addition, the effect of speckling is clearly visible, leading to a normal distributed pattern of bright and dark speckles.

To detect the location of the pupil and therefore track the eye for each full 2D scan the following three steps are applied to each recording.

\textbf{Image reconstruction:} The photodiode signal of the LFI sensor is sampled in equidistant time steps $t$ to capture the modulated feedback power $P_0'(t)$, while the MEMS mirror scans the laser beam over the surface of the eye. To generate an image of the eye region, the modulated feedback power $P_0'(t)$ and the mirror deflection angles $\alpha_h(t)$ and $\beta_v(t)$ are sampled in the same equidistant time steps. A series of samples $(P_0'(t), \alpha_h(t), \beta_v(t))$ are used to construct an image using the mirror deflection angles as pixel coordinates ($I_x \approx \alpha_h, I_y \approx \beta_v  $) on the HOE and the modulated feedback power $P_0'$ as intensity value ($I(x,y) = P_0'$) of the pixel. In \Cref{A4im:pupil_extraction_pipeline} b) in the background, a reconstructed image is shown. The pupil appears in the center of the image as a bright pattern, marked by the red box.

\textbf{Pupil segmentation: }
Similar to VOG-based eye-tracking sensors, segmentation of the pupil is required to determine the pupil contour and center. To separate the retinal area pixels from the non-retinal area pixels of the image, a histogram-based approach is used. In \Cref{A4im:pupil_extraction_pipeline} b) the normalized histogram of the image is shown in gray and green, containing information about both the retinal area  and the non-retinal area. This histogram is overlaid by a second normalized histogram (in blue), which includes only the first ten lines of the image and represents the non-retinal area area probability density distribution (PDF) $\mathcal{N}_b \left(\mu_b,\sigma_b\right)$, since there is no pupil in the first ten lines of the image. To extract the retinal area and thus separate the pupil from the non-retinal area , the intensity limit $I_b$ (red dashed line in \Cref{A4im:pupil_extraction_pipeline} b)) is calculated by $I_b = \mu_b + \sigma_b$ based on the non-retinal area  PDF. Using this limit, the normalized histogram of the image is divided into non-retinal area  intensity values (gray) and retinal area intensity values (green).

\Cref{A4im:pupil_extraction_pipeline} c) shows a cropped area around the bright pupil pattern for illustration. The remaining retinal area pixels are highlighted in green. 

\textbf{Pupil ellipse fitting: }
The segmented bright pupil pattern is given as a set of $p_i$ tuples, containing the pixel coordinates as well as the pixel intensity $p_i = (x_i,y_i,{P_0'}_i)$. To obtain the pupil ellipse from this set of tuples, a multivariate Gaussian distribution $\mathcal{N}_p \left(\mu_p,\Sigma_p\right)$ is fitted using least squares optimization. $\mu_p$ hereby represents the pupil center and the main components of the covariance matrix $\Sigma_p$ represent the horizontal and vertical axis of the ellipse representing the pupil contour. In \Cref{A4im:pupil_extraction_pipeline} d), the resulting ellipse contour is annotated in yellow as well as the center of the ellipse as a blue dot. A pupil ellipse is therefore given by $\mathcal{E}_i = \left(\mu_{p0},\mu_{p1}, 3 \cdot \Sigma_{p00}, 3 \cdot \Sigma_{p11} \right)$.

\subsection{Evaluation}
\label{A4sec:Evaluation} 

\begin{figure}[h]
	\centering
	\includegraphics[width=0.7\linewidth]{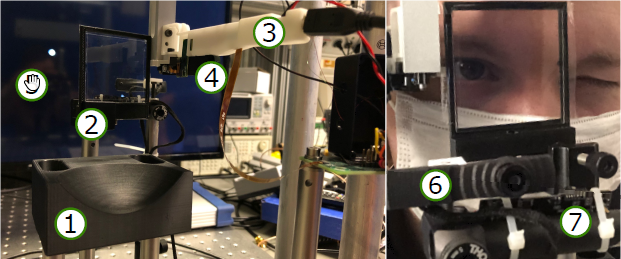}
	\caption{Laboratory setup to evaluate the proposed scanned LFI eye-tracking sensor. The left image shows the laboratory setup from the perspective of a participant and the right image shows a participant inside the setup.}
	\label{A4im:Labsetup}
\end{figure}
\Cref{A4im:Labsetup} shows the laboratory setup used to evaluate the scanned LFI eye-tracking sensor. The LFI sensor component itself is based on an research prototype adapted from an optical communication application where IR VCSELs with monitoring photodiodes in the back DBR are common. A detailed description of the sensor component is given by Grabherr et. al. \cite{grabherr2009integrated}.

The  glasses frame temple \raisebox{.5pt}{\textcircled{\raisebox{-.9pt} {3}}} with the integrated laser module and the MEMS micro mirror module \raisebox{.5pt}{\textcircled{\raisebox{-.9pt} {4}}} is based on a modified BML500P \cite{BML2021}, an optical microsystem developed for AR glasses. The MEMS mirrors are used to scan the IR laser across the surface of the HOE, which is integrated into a flat glasses lens \raisebox{.5pt}{\textcircled{\raisebox{-.9pt} {2}}}. The high transparency of the HOE allows a participant to sit in front of the laboratory setup and look through the glasses lens towards a display \raisebox{.5pt}{\textcircled{\raisebox{-.9pt} {5}}} on which stimuli markers are displayed. The participant's head is fixed in front of the HOE and the glasses frame temple by a chin rest \raisebox{.5pt}{\textcircled{\raisebox{-.9pt} {1}}} to minimize head movements that could lead to erroneous measurements. In addition a Pupil Core V1 \cite{Kassner:2014:POS:2638728.2641695} is added to the setup. The world camera \raisebox{.5pt}{\textcircled{\raisebox{-.9pt} {6}}} monitors the stimuli markers on the display and an eye camera \raisebox{.5pt}{\textcircled{\raisebox{-.9pt} {7}}} observes the participant's eye from a bottom-up perspective through the HOE. The mirror signals $\alpha_h(t)$ and $\beta_v(t)$ as well as the interference signal $P_0'(t)$ are captured in the setup by an oscilloscope.  

The Lab setup complies according to IEC 60825-1 \cite{IEC2014} regularization to a class 1 laser system and therefore does not pose any risks to the eye. The optical power of the IR laser beam surface was limited to an optical power of 142 $\mu$\,W on the eye's surface, whereas the IEC 60825-1 allows a maximum optical power of 778 $\mu$\,W for an 8-hour continuous emission to the retina. 

The low required optical power is favorable to minimizes power consumption of our scanned LFI eye-tracking sensor. Using off-the-shelf components, the power consumption of our system is estimated roughly at 30\,mW. The main components contributing to the overall systems power consumption are the transimpedance amplifier (TIA) (THS4567 ~ 10\,mW), which is used to amplify the interference signal $P_0'(t)$ measured by the integrated photodiode, and the analog digital converter (ADC) (MAX19191 with ~ 15.3 \,mW). The gain of the TIA was set to 940 during the experiments. With further integration, additional power reduction is expected. The estimated power consumption is comparable to reported power consumption of other scanned IR laser eye-tracking sensors. In example, \cite{7863402} reported a power consumption of 15\,mW for their system. A major advantage of our approach is that we reuse the existing MEMS micro mirror of the RGB projection similar to \cite{EyeWay2021}, and therefore, did not require an additional scanner which would increase the power consumption.

\subsubsection{Gaze accuracy} 
\begin{figure}[h]
	\centering
	\includegraphics[width=0.8\linewidth]{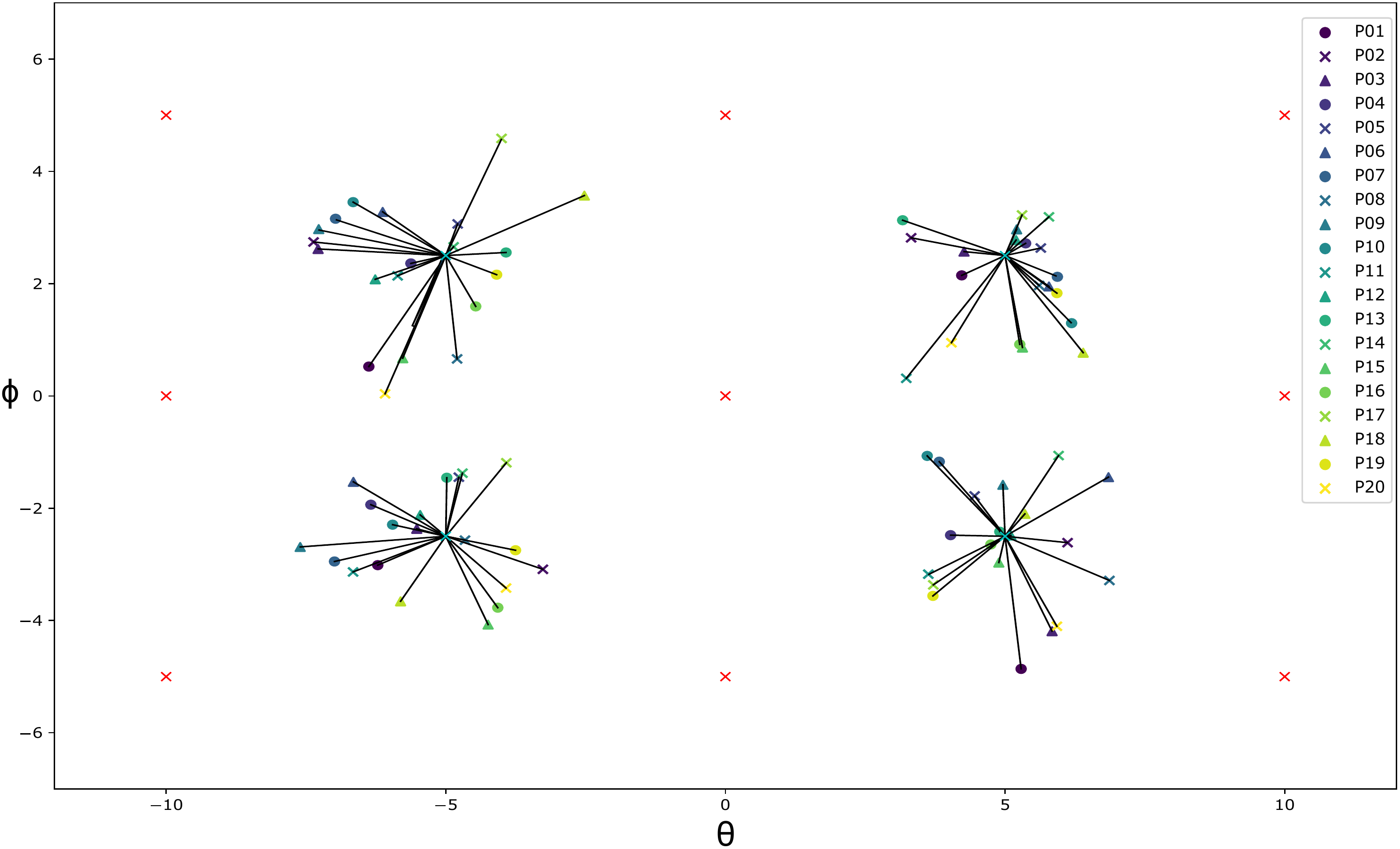}
	\caption{Results of the gaze accuracy experiment. Participants were asked to fixate the calibration markers (red crosses) and the test markers (cyan crosses). The calculated mean gaze position per test marker and participant is added as a colored marker. In addition, an arrow shows the correlation between the calculated gaze position and the test marker.}
	\label{A4im:Gaze_acc}
\end{figure}
To evaluate the performance of the scanned LFI eye-tracking sensor and prove the robustness of our approach, we conducted a study with 20 participants (4 female, 16 male, mean age 34 SD(10.83)). The participant's eye colors ranged from dark brown to blue-gray. Half of the participants required vision correction ranging from +1\,dpt to -2.75\,dpt. Except for participant P14 who wore contact lenses, participants did not wear vision correction during the study. None of the participants were of Asian ethnicity, so the robustness of the effect of reduced bright pupil response in Asian populations as reported by \cite{10.1145/507072.507099} was not tested. All participants gave their written consent after being informed about the nature of the study.

\begin{table}[h]
	\small	
	\centering
	
	\renewcommand{\arraystretch}{1.1}
	\caption{Accuracy and precision of our approach and the Pupil Core eye tracker over all participants}
	\begin{tabular}{|l|r|r|r|r|}
		\hline
		& \multicolumn{2}{c}{Scanned LFI}   &   \multicolumn{2}{c|}{Pupil Core V1}  \\
		&     \textbf{Precision $^\circ$} &   \textbf{Accuracy $^\circ$} &   \textbf{Precision $^\circ$} &   \textbf{Accuracy $^\circ$} \\
		\hline
		P1    &      1.991 &      2.591 &      0.127 &      1.802       \\
		P2    &      0.438 &      1.976 &      0.172 &      1.935       \\
		P3    &      0.718 &     1.239 &      0.644 &      1.597       \\
		P4    &     0.587 &      0.982 &     0.778 &      1.585       \\
		P5    &      1.001 &      1.122 &      0.746 &      0.988      \\
		P6    &      1.661 &      1.792 &      0.925 &      1.542       \\
		P7    &      0.512 &     1.408 &     0.830 &      1.820       \\
		P8    &      0.743 &      1.623 &      0.974 &      1.990      \\
		P9   &      1.039&      2.877 &     0.523 &      1.339      \\
		P10   &      0.905 &      2.408 &      0.411 &      1.616     \\
		P11   &      0.455 &      1.888 &      0.830 &      0.820       \\
		P12   &      1.011 &      1.082 &      0.775 &      0.875      \\
		P13   &      0.743 &      1.597 &      0.158 &      0.966      \\
		P14   &      0.960 &      1.211 &     0.892 &      1.026       \\
		P15   &      0.569 &      1.077 &     0.058 &      0.847     \\
		P16   &      1.733 &      1.812 &      0.106 &      1.126     \\
		P17   &      1.106&      1.655 &      0.477 &      1.442    \\
		P18   &      0.685 &      1.951 &      0.459 &     1.371      \\
		P19   &      1.138&      1.750 &     0.291 &      1.181       \\
		P20   &      0.914 &      1.446 &      0.791 &     2.427       \\
		\hline
		\textbf{Mean} &    0.945 &    1.674 &    0.548 &   1.415   \\
		\textbf{Std} &    0.4162 &    0.5052 &    0.3014 &   0.4305   \\
		\hline
	\end{tabular}
	\label{A4tab:data_overview}
\end{table}
During the study, participants sat approximately 0.6 m away from a 36" display and positioned their head on the chin rest. To set the calibration and test marker coordinates, participants were first asked to look straight through the HOE towards the display. Then, the center marker describing the resting position of the eye at $\theta = 0$ and $\phi = 0$ was adjusted to align with straight gaze.  After setting the calibration and test marker coordinates, participants were asked to follow and fixate on the stimuli markers on the monitor. In a sequence 9 reference markers (red crosses in \Cref{A4im:Gaze_acc}) and 4 test markers (cyan crosses in \Cref{A4im:Gaze_acc}) are presented for approximately 5 seconds each with three repetitions resulting in a total of 39 stimuli markers presented per participant. During the experiment, scanned LFI data and images from the Pupil Core eye camera were recorded for each marker location. For each point, the first and last second of recorded data were discarded to ensure that the participant had time to fixate on the next stimuli marker. In addition, scanned LFI sensor images were discarded if no pupil was detected due to blinking. In the Pupil Core data, detected pupil positions with a confidence < 0.8 are discarded in order to eliminate errors due to blinking as well. The pupil core camera was placed ~ 8\,cm away from the eye, which is rather large. To compensate the larger distance, the camera focus was tuned to receive sharp images at that distance. To compensate accuracy losses due to the increased distance we reduced the camera angle w.r.t. eye compared to a head worn configuration. 

After data acquisition and cleaning of the raw data, the standard 9-point polynomial regression algorithm is used to map the data from pupil position space to gaze angle space. The regression algorithm was trained for each user individually and for both the scanned LFI sensor and the Pupil Core VOG sensor separately. \Cref{A4im:Gaze_acc} shows the mapped gaze points for each participant and the 4 test points for the LFI eye-tracking sensor. 

To evaluate the scanned LFI eye-tracking sensor based on the captured data, we use the accuracy as evaluation metric, which is defined as the average angular offset between estimated fixation location and the corresponding marker position. In addition, we evaluate the precision, which is defined according to \cite{Kassner:2014:POS:2638728.2641695} as the root mean squared (RMS) error between successive samples. \Cref{A4tab:data_overview} summarizes precision and accuracy results of the study for the scanned LFI eye-tracking sensor and the Pupil Core.

Our scanned LFI eye-tracking sensor achieves a mean gaze accuracy of 1.674$^\circ$, which is comparable to the accuracy reported by other scanned laser eye-tracking approaches e.g. the 1.72$^\circ$ reported by \cite{10.1145/3379155.3391330}. The accuracy of the Pupil Core is 0.232$^\circ$ lower compared to our approach. In our experiments, we did not achieve the stated precision and accuracy of the Pupil Core, which is to some extent due to our laboratory setup as the scanned IR pattern appears as a varying IR illumination, which distorts the dark pupil tracking of the Pupil Core. The results of the study show that the scanned LFI eye-tracking sensor is capable to track the bright pupil with a reasonable accuracy. 

\subsubsection{Ambient light robustness}
\begin{figure}[h]
	\centering
	\includegraphics[width=0.9\linewidth]{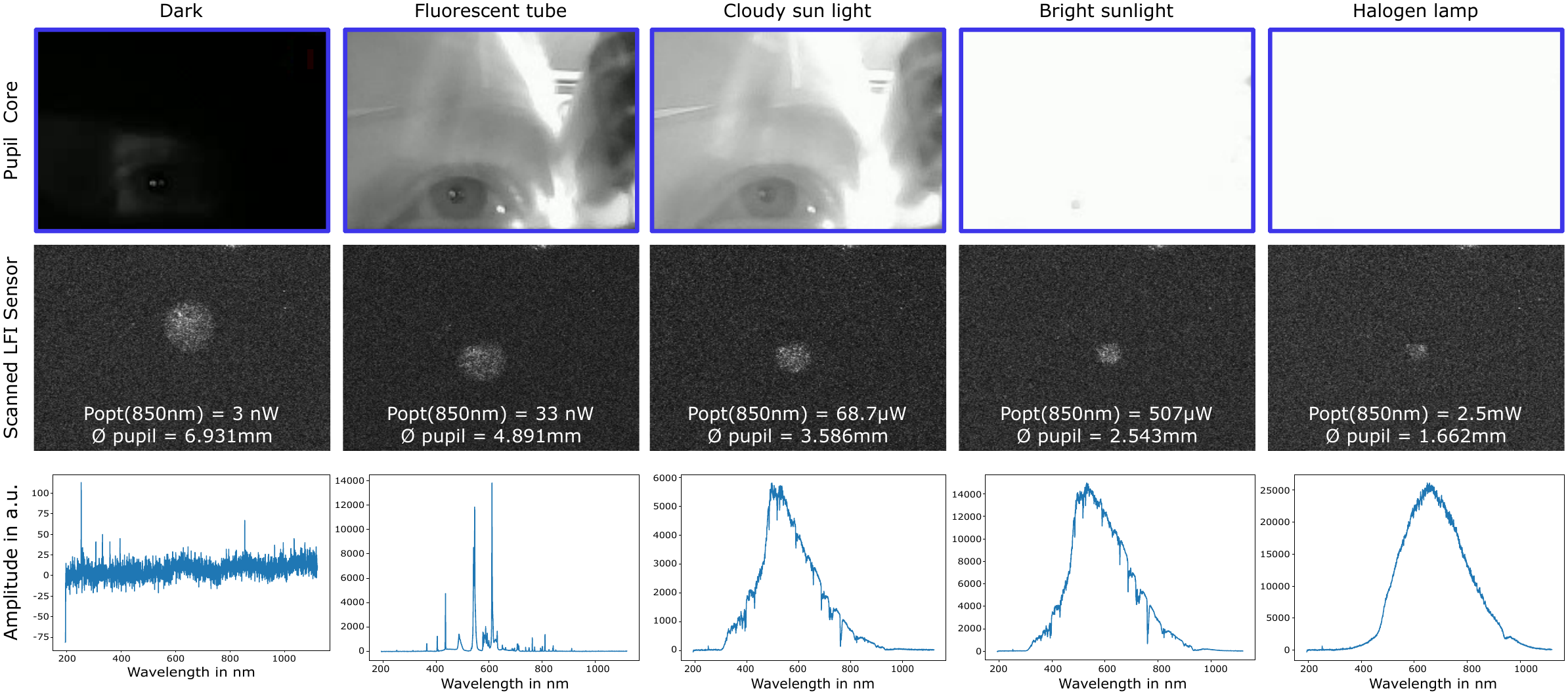}
	\caption{Comparison of the Pupil Core V1 eye-tracking sensor and our approach with respect to ambient light robustness. The first row shows images captured by the Pupil Core. The second row shows images captured by our approach with annotated optical power of the light source at 850\,nm and the estimated pupil diameter from the pupil contour. The last rows shows spectrograms of the different light sources.  }
	\label{A4im:ambient light}
\end{figure}

A further requirement to eye-tracking sensors for consumer AR glasses is a robust operation under variation of ambient light. To evaluate the ambient light robustness, our scanned LFI eye-tracking sensor as well as the Pupil Core VOG sensor are exposed to different illumination sources, while a participant was looking straight through the HOE such that the HOE and thus the parallel laser rays were perpendicular to the eye. \Cref{A4im:ambient light} summarizes the results of this study.
The first row shows images taken with the Pupil Core V1 eye-tracking sensor using the pupil capture software (V1.17.71) with default settings while the second row shows images captured with the scanned LFI eye-tracking sensor. The last row shows a spectra of each illumination source captured by an OceanOptics4000 optical spectrometer. In addition, we measured the optical power at the wavelength of 850\,nm on the eye's surface as both the the Pupil Core eye-tracking sensor and our scanned LFI eye-tracking sensor operates at 850\,nm. The results are annotated in the second row of the image.  

The first lighting situation we investigated was a completely dark laboratory with no external light sources. Under this condition, both sensors track the pupil as expected. The second lighting situation we investigated is office lighting. Under this very controlled lighting condition, both sensors also worked perfectly. Also under cloudy sunlight this condition, both sensors show stable operation. In bright sunlight ($P_{opt}$(850\,nm) = 507\,$\mu$W) the dark pupil appears only as a tiny dark spot in the camera image, which is no longer robustly detected. While the VOG camera sensor saturates, the scanned LFI eye-tracking sensor still is capable to robustly detect the bright pupil. As already a improved version of the VOG system (Pupil Core V2) is available, which we did not used for the experiment, the image quality for the bright sun light condition might improve.

 As final lightning condition, we used a halogen lamp, which is a broadband thermal radiator with characteristically high intensity in the IR wavelength region. With a measured optical power of 2.5\,mW at 850\,nm the eye region was exposed by a five times higher intensity compared to bright sun light. Even under this condition, the scanned LFI sensor is capable to detect the bright pupil reliably, leading to an outstanding dynamic range of the scanned LFI eye-tracking sensor. The observed high robustness to ambient light is in line with earlier work by \cite{Meyer2020e}.

\subsection{Discussion}
To assess the quality of our scanned LFI eye-tracking approach with respect to the state of the art of scanned IR laser eye tracking approaches and discuss the results and potential limitations, we compare our approach with other scanned IR laser eye tracking approaches in \Cref{A4tab:comparisson}.
\begin{table}[h!]
	\small	
	\centering
	
	\renewcommand{\arraystretch}{1.1}
	\caption{Comparison between different scanned IR eye tracking approaches and our approach}
	\begin{tabular}{|l|p{2.2cm}|p{2.7cm}|p{2.5cm}|p{2.7cm}|}
		\hline
		&   				\cite{7863402}  	& \cite{9149591} 	&   \cite{10.1145/3379155.3391330} &   \textbf{Ours}\\
		\hline
		Tracking method 	&   Corneal reflection 	&Dark Pupil tracking on rasterized 2D image	&Corneal reflection \& Stereo image & Bright Pupil tracking on rasterized 2D image       \\
		IR Scanner    		&      2D MEMS mirror 	&      2x 1D MEMS mirrors	&     2D MEMS mirror &       2x 1D MEMS mirrors       \\
		Accuracy    		&      >1$^\circ$ 		&     1.31$^\circ$ 		&      1.72$^\circ$  &      1.67$^\circ$      \\
		Precision    		&     - &     0.01$^\circ$ 	&     0.0091$^\circ$		&     0.945$^\circ$        \\   
		Diag. FOV    		&      35.35$^\circ$ 	&      44.72$^\circ$  		&     16.97$^\circ$  &      22.36$^\circ$      \\
		Power    			&      15\,mW 			&      11mW 		&     - &     30mW       \\
		Sample rate    		&      3300\,Hz 		&     60\,Hz		&     4000\,Hz &      60\,Hz      \\
		\hline
	\end{tabular}
	\label{A4tab:comparisson}
\end{table}

The works of Sarkar et. al. and Holmqvist et. al. differ from the work of Meyer et. al. and our approach mainly with regard to the chosen tracking method. They track corneal reflections with a rather high sampling rate while the work of Meyer et. al. and our work rely on a rasterized 2D image and tracking of either a dark or a bright pupil. All approaches are in the same range of absolute gaze accuracy and power consumption. Furthermore, they are evaluated on a comparable diagonal FOV. The main improvements in our work compared to the state of the art is the \textit{robustness of pupil detection}, which is extremely important for consumer AR glasses. By using the presented LFI measurement method, the sensor is almost immune to ambient light. Due to the signal characteristics of the bright pupil and the proposed algorithm, our approach overcomes several limitations of VOG eye-tracking systems, as it is robust against eyelashes that interfere with the pupil, mascara that causes false pupil detection and eyelids that partially occlude the pupil. In addition, the sensor works independently of eye color and iris structure.



\subsubsection{Sensor integration}
\begin{figure}[h]
	\centering
	\includegraphics[width=0.7\linewidth]{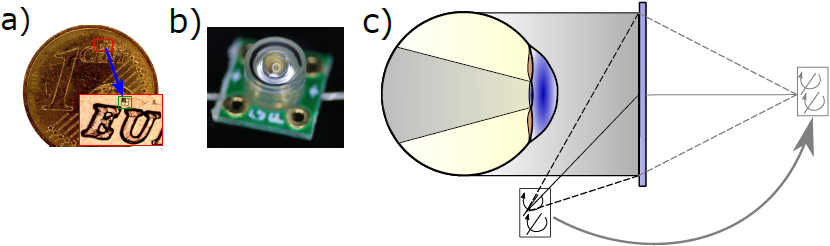}
	\caption{Sensor integration of the scanned LFI eye-tracking sensor. a) shows a microscope image of the 160\,$\mu$m x 180\,$\mu$m LFI sensing element on a coin for scale (blue arrow). b) Encapsulated optical module of the research prototype composing of the LFI sensor as well as the beam shaping lens. The lens diameter is roughly 2\,mm. c) virtual rotation of the MEMS scanner to the center of the FOV by the HOE to solve the off-axis integration issue of VOG sensors}
	\label{A4im:Sensor_integration}
\end{figure}
In addition to robust pupil detection, our approach can be fully integrated into AR glasses with retinal projection, as the optical path of the IR laser uses the same optical path as visible light. Moreover, the VCSEL as the optical transmitter and the photodiode as the optical receiver of the LFI sensor element are highly integrated in a single chip, as shown in \Cref{A4im:Sensor_integration} a). In combination with the beam shaping optics a diameter of the optical module of below 2\,mm is possible (\Cref{A4im:Sensor_integration} b)), allowing thus direct integration into the RGB laser module. Compared to other scanning laser approaches, such as e.g. shown by \cite{7863402}, our setup does not require any components to be mounted outside the spectacle temple or even in the spectacle frame. The use of an HOE allows us to virtually rotate our MEMS scanner to the front of the glasses lens as shown in \Cref{A4im:Sensor_integration} c). Images taken with the scanned LFI eye-tracking sensor therefore appear as if taken from the perspective of a camera viewing the eye centrally from the outside through the lens. Compared to VOG systems, this effect is possible without any camera arms interfere with the users FOV. In addition, this perspective allows covering the whole eye region and it is possible to track the pupil over a large FOV.
\subsubsection{Power consumption}
\cite{7863402} estimated the power consumption of VOG camera sensors at 150\,mW, while our sensor consumes only about 30\,mW, which is a significant improvement and allows real-time operation in lightweight consumer AR glasses. In addition, our proposed pupil detection algorithm requires less computationally intensive image processing steps to extract the pupil contour compared e.g. to the Pupil Core algorithm \cite{Kassner:2014:POS:2638728.2641695}. As the output of our pupil detection algorithm is an ellipse contour $\mathcal{E}$, the power consumption to derive an absolute gaze vector e.g. by using a geometrical 3D model approach as proposed by \cite{10.1145/2168556.2168585} is comparable to VOG eye-tracking systems.
\subsubsection{Glasses slippage}
A major problem that causes eye-tracking sensor accuracy to degrade is the effect of glasses slippage \cite{niehorster2020impact}. This issue also affects our sensor performance as we are working on image data. The impact of slippage to our sensor might however be less significantly affecting our results as the camera axis in our approach is close to the optical axis of the eye. Compared to the work of \cite{9149591} we only capture the bright pupil and do not gather further any information from the eye region. Thus slippage compensation by tracking landmarks like the eye corners as introduced by \cite{6595992} is not feasible. A possible solution to achieve slippage robustness for our approach is to adopt the approach of \cite{10.1145/3314111.3319835} to derive slippage robust features from a geometric 3D eye model, which we will consider and evaluate as part of our future research.
\subsubsection{Update rate, latency and motion blur}
Due to the tight coupling of the optical path of the RGB projection and the IR path the update rate is limited to 60\,Hz, which is compared to \cite{7863402}  and \cite{10.1145/3379155.3391330} rather low. A faster scanning MEMS mirror would improve the update rate to some extend. However, since the diameter of the laser beam determines the minimum required mirror diameter, mirror miniaturization is limited, resulting in a maximum technically feasible scan frequency of 120\,Hz. Compared to VOG systems latency is rather low as in our approach we capture images pixel by pixel, and therefore, the foreground background segmentation can be performed in parallel to image capturing, leading to a latency of ~ 0.0166\,s to calculate the pupil ellipse $\mathcal{E}$. A camera sensor captures all pixels in parallel while in contrast our system captures images sequential. Thus a fast saccadic movements of the pupil during image acquisition can lead to elliptical distortions of the captured ellipse $\mathcal{E}$.
 \subsubsection{Gaze angle dependency of bright pupil effect}
The bright pupil effect appears only if both the illumination axis and the sensor axis are close to each other. The scanned LFI system moves both the light source and the detector in parallel, leading to a perfect alignment of both axes.  However, a disadvantage of our system is the collimated nature of the laser beam compared to a diverging IR light source. If the laser beam is not approximately perpendicular to the retina due to the Lambertian scattering less light is back injected into the laser cavity leading to a reduced bright pupil response. This effect is independent from external illumination. A possible solution is to use a parabolic mirror function for the IR HOE which follows the curvature of the eye.
\subsection{Conclusion}
In this work, we present a novel scanned LFI eye-tracking sensor approach, which is able to track the pupil with high robustness. Compared to VOG sensors and other scanned laser approaches, we highlight the outstanding robustness to ambient light and the high integrateability of our sensor approach into retinal projection AR glasses. We introduced the sensing technology, derived the physical foundations to describe the signal occurrence and propose a pupil extraction algorithm, which is optimized for the bright pupil signal characteristics measured by our sensor approach. To validate the accuracy of our scanned laser eye-tracking sensor we build a prototype using a modified retinal projection AR glasses setup based on the BML500P, a retinal projection system.

Our eye-tracking sensor achieves a mean accuracy of 1.674$^\circ$, which is comparable to scanning laser eye-tracking approaches e.g. by \cite{10.1145/3379155.3391330}. We further solve typical problems of VOG eye-tracking sensors, e.g. the highly off axis integration of camera sensors by using an IR HOE to virtually place the laser scanner in front of the participants eye.

With the advancements especially in ambient light robustness and by the nearly invisible integration of the eye-tracking sensor we pave the way for eye-tracking sensors to become standard sensors for upcoming AR glasses, which will enable new application areas of eye-tracking e.g. long-term gaze monitoring for early detection of mental disorders.
\subsection*{Acknowledgements}
Enkelejda Kasneci is a member of the Machine Learning Cluster of Excellence, EXC number 2064/1 - Project number 390727645.

\newpage

\section{A holographic single-pixel stereo camera eye-tracking sensor for calibration-free eye-tracking in retinal projection AR glasses }
\label{APP:A5}
\subsection{Abstract}
Eye-tracking is a key technology for future retinal projection based AR glasses as it enables techniques such as foveated rendering or gaze-driven exit pupil steering, which both increases the system's overall performance. 
However, two of the major challenges video oculography systems face are robust gaze estimation in the presence of glasses slippage, paired with the necessity of frequent sensor calibration.
To overcome these challenges, we propose a novel, calibration-free eye-tracking sensor for AR glasses based on a highly transparent holographic optical element (HOE) and a laser scanner. 
We fabricate a segmented HOE generating two stereo images of the eye-region. 
A single-pixel detector in combination with our stereo reconstruction algorithm is used to precisely calculate the gaze position.\\
In our laboratory setup we demonstrate a calibration-free accuracy of 1.35$^\circ$ achieved by our eye-tracking sensor; highlighting the sensor's suitability for consumer AR glasses.

\subsection{Introduction}
\label{A5sec:Introduction}
Eye-tracking is a key technology to improve current near-eye displays such as holographic retinal projection displays \cite{Maimone:2017:HND:3072959.3073624}.
It can enable foveated rendering to increase the system's perceived resolution while minimizing the system's power consumption, or increase the overall field of view (FOV) by using exit pupil steering \cite{Jang:2017:RAR:3130800.3130889, Kaplanyan:2019:DNR:3355089.3356557,Kim:2019:FAD:3306346.3322987}. 
Thus, robust eye-tracking sensors are required for future near-eye displays such as augmented reality (AR) glasses. 

Current state of the art video-oculography (VOG) meets the requirements for gaze accuracy but is limited in sensor integration and robustness against slippage of glasses. 
In addition, VOG systems often require at least a single marker calibration to achieve high gaze accuracy; thus limiting the usability of VOG sensors in everyday AR glasses.

In order to achieve calibration-free eye-tracking \cite{10.1145/1344471.1344506} have introduced a stereo camera approach for 3D reconstruction of the pupil using a closed-form stereo reconstruction algorithm. 
However, this approach implements multiple camera sensors to cover a sufficiently large FOV; resulting in the integration of additional components with a high power consumption.

To address drawbacks of current VOG stereo approaches for AR glasses, we introduce a single-pixel holographic stereo camera sensor approach, which is capable of capturing a stereo perspective view of the eye-region. 
We embedded a holographic optical element (HOE) in the eyeglass lens, which we used to redirect a scanned laser beam to the eye-region; followed by the capturing of backscattered light using a single-pixel detector. 
The HOE is designed to achieve a stereo perspective vision by performing two holographic functions; transforming a single scanned beam into two separate beams which propagate towards the eye. 
HOEs based on photopolymer technology provide excellent transparency and low noise which, combined with their high integrability, makes them particularly interesting for future AR glasses.
In addition to that, the presented stereo camera VOG sensor consumes only a fraction of the power of current VOG stereo systems.

Our main contributions are i) we solved the sensor integration problem of headword stereo camera VOG systems by using an HOE and a single-pixel detector in a scanned laser system to generate stereo images and ii) we fabricated a suitable HOE and demonstrated the calibration-free 3D reconstruction of the eye's optical axis in our experimental setup by applying the closed-form stereo reconstruction algorithm proposed by \cite{10.1007/BF01440844}. We further evaluated the gaze accuracy and precision of the overall system.

Our main contribution compared to previous work by \cite{Meyer2020e},\cite{9149591} is the proposed and fabricated segmented HOE to create a stereo perspective of the eye to enable stereo reconstruction algorithms e.g. by  \cite{10.1007/BF01440844} to reconstruct the optical axis of the eye without calibration.

In the next section, we present related work focusing on stereo camera eye-tracking for mobile applications. 
Sec. \ref{A5sec:Holographic_Stereocamera} describes our single-pixel stereoscopic holographic camera sensor system design as well as the reconstruction algorithm to derive the 3D gaze vector from a pair of images. 
Afterwards, in Sec. \ref{A5sec:Evaluation} and Sec. \ref{A5sec:Discussion} our approach is evaluated in a laboratory setup to derive gaze accuracy and precision. 
Finally, in Sec. \ref{A5sec:Conclusion}, we conclude our work and discuss limitations and further research directions.

\subsection{Related Work}
\label{A5sec:Related_Work}
%
One of the first works introducing a stereo camera approach for calibration-free eye-tracking was published by \cite{10.1145/1344471.1344506}. The authors used a pair of infrared (IR) cameras to capture a set of images to create a stereo perspective of the left eye. The pair of cameras were then integrated into the side of the glasses' frame and an image of the eye was relayed onto the camera sensors via a semi-transparent mirror (hence resulting in a rather bulky design). By applying a closed-form stereo reconstruction algorithm  \cite{10.1007/BF01440844}, a 3D perspective of the pupil plane was reconstructed by extracting pairs of pupil ellipses. The system achieved an accuracy of 2.2$^\circ$ over a FOV of 30$^\circ$.

\cite{10.1145/2578153.2578163} combined a stereo camera approach with additional glint features and reported a gaze accuracy of 1.6$^\circ$. To solve the camera integration issue they integrated the camera pair below the glasses by adding a mechanical arm, leading thus to a highly off-axis camera integration. A similar approach is proposed by \cite{10.1145/3161166}, which combined corneal images captured by an RGB camera as well as pupil ellipse features extracted from images captured by a second IR camera in a mobile stereo camera setup. The cameras were mounted on a mechanical arm to integrate them into a head-worn setup, leading to similar highly off-axis integration. By fusion of corneal images and IR images, the authors achieved a mean accuracy of 2.19$^\circ$ over a FOV of 30$^\circ$. Additionally, this approach showed robustness to calibration drift.

Currently, the Tobii Pro glasses \cite{TobiiPro2021} are to the best of our knowledge the only commercial mobile eye tracker, working with a pair of cameras. They have embedded the camera sensors and IR illumination directly into the glass lenses inside the user's FOV. After a single point calibration, a gaze accuracy of 0.6$^\circ$ is reported combining stereo reconstruction methods as well as corneal reflection methods. The system shows to be robust to slippage \cite{niehorster2020impact}.

To summarize the state of the art, all current approaches use IR camera sensors that are either directly integrated into the lens of the eyeglasses or mechanically integrated by camera brackets, which results in an impairment of the user's vision. Only the work of \cite{10.1145/1344471.1344506} tend to reduce the impairment by using a semi-transparent mirror. However, their approach still requires two power consuming camera sensors and additional imaging optics per eye.

In summary, to address challenges related to camera integration and reduce the power consumption of the overall system, we present the combination of an HOE with a single-pixel detector. With this, we demonstrate an eye-tracking system creating a stereo holographic image with two virtual cameras. 

\subsection{Holographic Single-Pixel Stereo Camera}
\label{A5sec:Holographic_Stereocamera}
\begin{figure}[h]
	\centering
	\includegraphics[width=\linewidth]{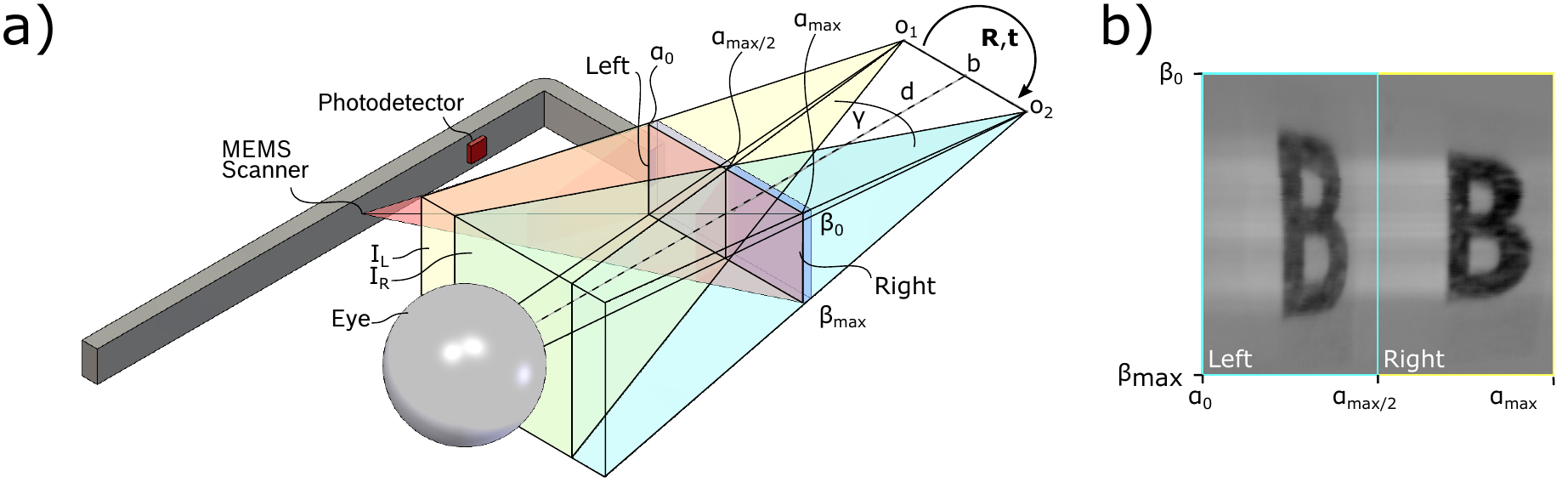}
	\caption{a) Schematic integration of the proposed eye-tracking sensor into a glasses frame. The MEMS laser scanner and the single-pixel detector are integrated into the frame temple while the HOE with its two sub-HOEs (Left and Right) with the corresponding wave fronts (yellow and blue) are integrated into one glasses lens. b) Image of a capital \textit{B} letter printed on paper. The paper is placed in the overlapping image plane ($I_L, I_R$), captured by the holographic single-pixel stereo camera system}
	\label{A5im:stereo_camera}
\end{figure}

The holographic single-pixel stereo camera eye-tracking sensor, as shown in \Cref{A5im:stereo_camera} a), consists of three main components: a 2D laser scanner with a MEMS (micro-electromechanical system) micro mirror as scanning unit, an HOE embedded into the glasses' lens and a single-pixel detector. 
The MEMS scanner with its scan angles $\alpha$ and $\beta$ deflects the laser beam towards the glasses' lens to scan along the surface of the embedded HOE. A review on MEMS scanner technology is given by \cite{6714402}.
The laser component is integrated into an RGB laser module as described in our previous work \cite{9149591}. 
The HOE is deflecting the laser beam towards the user's eye, forming the total image plane $I_i$ at the eye's surface.

The global holographic function of the HOE is defined by considering the whole scanning area as a diverging point source, originating from the scanning point of the MEMS scanner.
The HOE splits up the incoming beam into two diverging point sources, which propagate towards the user's eye under different angles.  
The origin points of both wave fronts generated by the HOE lay in the space behind the HOE's surface, stretching out two diverging cones towards the eye's surface as outlined in Fig. \ref{A5im:stereo_camera} a).
\FloatBarrier
Each sub-HOE can be regarded as a virtual laser scanner, which in combination with the single-pixel detector can be described as a virtual camera sensor with its own coordinate space $c_i$, a camera matrix $\textbf{K}_i$ and an optical center of the camera $o_i$ as the origin of $c_i$, with $i \in \{1,2\}$ for each virtual camera sensor. The coordinate spaces of both virtual cameras are linked to each other by the rotation matrix $\textbf{R}$ in conjunction with the translation vector $\textbf{t}$.

To reconstruct an image from the scanned laser beam and the HOE, the backscattered light from the eye's surface is captured by the single-pixel detector consisting of a photosensitive diode, a transimpedance amplifier and an analog-to-digital converter (ADC), which samples the photodiode current in equidistant time steps. 
Therefore, the sample rate of the ADC  determines the pixel clock rate of the outlined system, which consequently determines the resolution of the virtual camera sensors. The position of an individual pixel on the sensor plane of the virtual camera relative to the image plane $I_i$ is given by the deflection angles $\alpha$ (horizontal) and $\beta$ (vertical) of the MEMS scanner, whereas the intensity of an individual pixel is given by the scattered light measured by the single-pixel detector at each scanning position.

By splitting the HOE into two non-overlapping sub-HOEs with overlapping image planes, the same scanning point in the global image plane results in two diffracted light signals, originating from the two virtual laser scanners with different propagation angles. Thus, one full scan from $\alpha_0 - \alpha_{max}$ and $\beta_0 - \beta_{max}$ of the MEMS scanner generates two image frames of the same object as shown in \Cref{A5im:stereo_camera} b), which can be reconstructed into two images from the object under two different perspectives. The frame rate of the proposed stereo system therefore depends on the number of full scans the laser scanner performs per second.
\subsubsection{HOE fabrication}
The HOE is fabricated by means of holographic wave front printing. 
Details on the recording setup employed as an extended immersion-based holographic wave front printer, are outlined in \cite{Wilm2021}.
The HOE is made up of individual sub-holograms, so called Hogels, which are aligned in an array-based structure.
Each Hogel is recorded by sequentially relaying two coherent recording wave fronts onto a photopolymer-based holographic film.
The resulting interference pattern leads to a photopolymerization-based modulation of the local refractive index in the volume of the holographic film, which results in a manifestation of the recording wave front's characteristics in the form of a 3D diffraction grating.

The HOE is recorded in development grade Bayfol\textsuperscript{\textregistered} HX TP* photopolymer \cite{bruder_chemistry_2017} by Covestro, with a photopolymer thickness of $\SI{16}{\micro\metre}$ and a protective polyamide layer of $\SI{60}{\micro\metre}$. Individual Hogels are recorded via two monochromatic wave fronts, modulated by means of two phase-only spatial light modulators. Both wave fronts originate from a common single-mode laser source with $\lambda = \SI{639}{\nano\metre}$. 
For demonstration purposes and ease of experimentation a wavelength in the visible spectrum has been chosen; however, future HOE-based systems are planned to operate at non-visible IR wavelengths as outlined in \cite{10.1117/1.OE.60.8.085101}.
Each Hogel performs an individual optical transformation, which contributes to the HOE's global holographic function. The HOE realizes a combiner functionality, whereby each of the two sub-HOEs performs a point-source-to-point-source transformation from a large off-axis to a close to on-axis configuration. Both sub-HOEs have the same off-axis recording point source placed at the MEMS scanner position. Under reconstruction of the HOE each sub-HOE deflects a diverging wave front propagating with an angle of $\pm$ 7.1\textdegree$ $ relative to the sub-HOE's respective surface normal.
\subsubsection{Stereo calibration}
\label{A5subsec:Stereocalibration}
To reconstruct the pupil-normal-vector in space, the parameters $\textbf{K}_1, \textbf{K}_2, \textbf{R} $, and $\textbf{t}$ of the proposed single-pixel stereo holographic eye-tracking camera sensor must be known. These parameters can be determined by a camera calibration procedure using a chess pattern. In a first step, images of a checkerboard pattern are captured with both virtual cameras, varying the orientation and position of the checkerboard. Afterwards, the method of \cite{zhang1999flexible} is applied to determine the camera matrix $\textbf{K}_i$ of each virtual camera, as well as the relative position $\textbf{t}$ and orientation $\textbf{R}$ of both cameras to each other. The stereo calibration process is required only once after the HOE is embedded in the eyeglass lens and the laser projector is integrated into the eyeglass temple.
\begin{figure}[h!]
	\centering
	\includegraphics[width=0.7\linewidth]{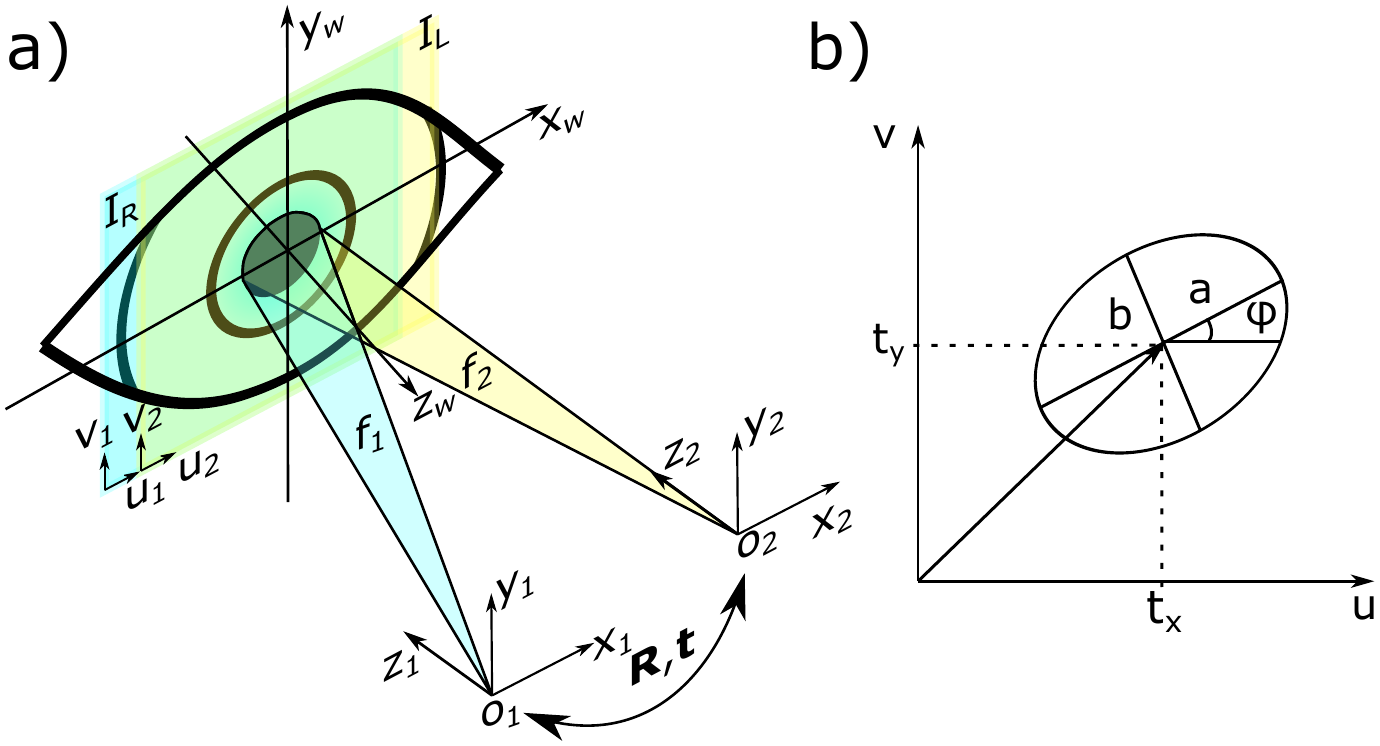}
	\caption{a) Correspondence between the world coordinate systems $c_w$ and the camera coordinate systems $c_i$. the origin of each camera coordinate system $o_i$ is the origin of one of the two cones ($f_1$, $f_2$), which are constructed from the pupil ellipse in the 2D image space. b) Parameterized description of the pupil ellipse $\mathcal{E}$ in the image plane}
	\label{A5im:coordinate_space}
\end{figure}
\FloatBarrier

\subsubsection{Stereo correspondence}
The correspondence of the pupil in the stereo system can be described by its appearance as ellipse $\mathcal{E}_i$ in each camera frame $I_i$. With the given pupil ellipse $\mathcal{E}_i$ and the known optical center of the camera, a cone, e.g. $f_1$, can be constructed to re-project the pupil ellipse from a 2D image plane to 3D, as shown in \Cref{A5im:coordinate_space} a). From a single cone, the 3D position and orientation of the pupil cannot be derived due to the fact that many valid pupil projections exist, which yield the same cone \cite{Swirski2013}. To solve this problem, we make use of the correspondence of the pupil and construct a second cone $f_2$. As both cones are constructed from the same pupil in the 3D space, the pupil position $\textbf{t}_1$ and orientation $\textbf{R}_1$ with respect to $c_1$ can be derived from the intersect of both cones. Aside from $\textbf{t}_1$ and $\textbf{R}_1$ the pupil size $a$ and $b$ is derived. To reconstruct the pupil ellipse, the closed form solution as proposed by \cite{10.1007/BF01440844} is used.

Each virtual-holographic single-pixel stereo camera operates in its own coordinate space $c_i$ with its camera center $o_i$ and an image plane $I_i$ in which the normalized pupil ellipse $\mathcal{\hat{E}}_i$ is defined. Since the contour of the pupil in the image plane is assumed to be an ellipse or circle and both cameras observe the same pupil, the intersection of the two cones $f_1$ and $f_2$ again results in a conic intersection. Therefore, the projected ellipse are defined by two conics $\textbf{A}$ given by
\begin{equation}
\label{A5equ:Projected_ellipse}
	x^T \textbf{A}_1 x = 0 \qquad x^T \textbf{A}_2 x = 0 .
\end{equation}

 An ellipse $\mathcal{E}_i$ is described in the image plane by its principal axis $a$ and $b$, its position in the image given by $t_x$ and $t_y$ as well as the orientation $\phi$ with respect to the u-axis, as shown in \Cref{A5im:coordinate_space} b). With this information the conic matrices $A_1$ and $A_2$ can be derived by applying an affine transformation $\textbf{S}$ 
 \begin{equation}
 	\textbf{S}_i = \begin{pmatrix}
 	\cos(\phi_i) & -\sin(\phi_i) & -t_{x_i}\cos(\phi_i)+t_{y_i}\sin(\phi_i)\\
 	\sin(\phi_i) & \cos(\phi_i) & -t_{x_i}\sin(\phi_i)-t_{y_i}\cos(\phi_i)\\
 	0 & 0 & 1
 	\end{pmatrix}
 \end{equation}
 
 to the ellipse matrix $\textbf{H}_i$ for virtual camera $i$ in normal form in the pupil plane 
 \begin{equation}
 	\textbf{H}_i = \begin{pmatrix}
 	\frac{1}{a^2_i} & 0 & 0\\
 	0 & \frac{1}{b^2_i} & 0 \\
 	0 & 0 & -1\\
 	\end{pmatrix}
 \end{equation} which results in $ 	\textbf{A}_i = \textbf{S}^T_i \textbf{H}_i \textbf{S}_i.$
 As $\textbf{H}_i$ is defined in the pupil plane, we can without loss of generality assume a focal distance $f=1$ and therefore rescale and translate the ellipse parameters in the image plane $I_i$ by
 \begin{equation}
 \label{A5equ:rescaling_ellipse}
 	\hat{t}_{x_i} = \frac{1}{f_{x_i}} (t_{x_i} - c_ {x_i}), \quad
 	\hat{t}_{y_i} = \frac{1}{f_{y_i}} (t_{y_i} - c_{y_i}) , \quad
 	\hat{a}_i = \frac{a_i}{f_{x_i}}, \quad
 	\hat{b}_i = \frac{b_i}{f_{y_i}}
 \end{equation}
 where $f_{x_i}$ and $f_{y_i}$ are the the focal distances and $c_{x_i}$ and $c_{y_i}$ are the camera centers from $\textbf{K}_i$. After normalization of the ellipse $\mathcal{E}_i$ towards the pupil plane, the relation between the camera coordinate system $c_i$ and the world coordinate system $c_w$ is given by $x_i  = \textbf{R}_i x_w + \textbf{t}_i.$ For points in the pupil plane $I_i$, this can be rewritten as $x_i = \textbf{G}_i \textbf{u}_w$ where $\textbf{u}_w = (x_w, y_w, 1)$ are homogeneous coordinates in the pupil plane and $\textbf{G}_i$ is a $3\times 3$ matrix consisting of the first two rows of $\textbf{R}_i$ and the last row contains the corresponding translation vector $\textbf{t}_i$
\begin{equation}
\textbf{G}_i = \left(\textbf{r}_{i1}  \textbf{r}_{i2}  \textbf{t}_i \right)\quad i=1,2.
\end{equation} 

Considering the pinhole camera model $u_i$ = $\frac{x_i}{z_i}$ and $v_i$ = $\frac{y_i}{z_i}$, the pupil plane is linked to the world coordinate system $c_w$:
\begin{equation}
\label{A5equ:World_Camera_Correspondance}
	z_i \textbf{u}_i = \textbf{G}_i \textbf{u}_w \qquad i=1,2.
\end{equation}

With this correspondence between $c_w$ and $c_i$, a pupil ellipse $\mathcal{E}$ in the pupil plane is defined according to \cite{10.1007/BF01440844} by
\begin{equation}
\label{A5equ:Cone_in_world}
	\textbf{u}^T_w \textbf{H} u_w = 0
\end{equation}

and its projection with respect to \Cref{A5equ:Projected_ellipse} by 
\begin{equation}
\label{A5equ:World_Local_Correspondance}
\textbf{u}^T_i \textbf{A}_i \textbf{u}_i = 0 \qquad i=1,2.
\end{equation}

Inserting \Cref{A5equ:World_Camera_Correspondance} into \Cref{A5equ:World_Local_Correspondance} yields
\begin{equation}
\label{A5equ:Full_correspondance}
	\textbf{u}^T_w  \textbf{G}^T_i \textbf{A}_i \textbf{G}_i \textbf{u}_w = 0 \qquad i=1,2.
\end{equation}

As both equations \Cref{A5equ:Cone_in_world} and \Cref{A5equ:Full_correspondance} describe the same cone, $\textbf{H}$ can be written in a generalized form 

\begin{equation}
\label{A5equ:Basic_correspondance}
	\textbf{G}^T_i \textbf{A}_i \textbf{G}_i = k_i \textbf{H}
\end{equation}

with $k_i$ describing an unknown scaling factor of the cone \cite{10.1007/BF01440844}.
\vspace{-.6cm}
\subsection{Stereo reconstruction}
\label{A5subsec:Stereo_reconstruction}
For a set of two virtual cameras, \Cref{A5equ:Basic_correspondance} provides 12 constraints as it contains two real valued symmetric $3\times 3$ matrices with six parameters each. As we have only 10 unknown parameters, three from $\textbf{R}_1$ and $\textbf{t}_1$, as well as the scalars $k_1$, $k_2$ ,$a$ and $b$, the system is overdetermined and we can solve it e.g. for $\textbf{R}_1$ and $\textbf{t}_1$ independently as shown by \cite{10.1007/BF01440844}. 

In a first step, \Cref{A5equ:Basic_correspondance} is reduced to 
\begin{equation}
\label{A5equ:Reduced_correspondance}
	\left(\textbf{R}^T_1 \textbf{A}_1 \textbf{R}_1\right)^{2\times 2} = k_1 \textbf{H}^{2\times 2}, 
	\left(\textbf{R}^T_2 \textbf{A}_2 \textbf{R}_2\right)^{2\times 2} = k_2 \textbf{H}^{2\times 2}
\end{equation}
where $2\times 2$ denotes the upper left submatrix of the corresponding $3\times 3$ matrix. By substituting the known stereo correspondence $\textbf{R}_2 = \textbf{R}\textbf{R}_1$ and $\textbf{t}_2 = \textbf{R} \textbf{t}_1 + \textbf{t}$ obtained from stereo calibration as described in \Cref{A5subsec:Stereocalibration} and elimination of $\textbf{H}^{2\times 2}$, \Cref{A5equ:Reduced_correspondance} can be rewritten to 
\begin{equation}
\label{A5equ:resolved_system}
	\left[\textbf{R}_1 \left(\textbf{A}_1 - k \textbf{R}^T\textbf{A}_2 \textbf{R}\right)\textbf{R}_1\right]^{2\times 2} = \mathcal{O}^{2\times 2} \quad with \quad k = \frac{k_1}{k_2}.
\end{equation}
To achieve a $(2\times 2)$ zero matrix $\mathcal{O}$, as stated by \Cref{A5equ:resolved_system}, the determinant of the corresponding $(3\times 3)$ matrix must be equal to zero. Therefore, \Cref{A5equ:resolved_system} yields 
\begin{equation}
\label{A5equ:solve_k}
	det\left(\textbf{A}_1 - k \textbf{R}^T \textbf{A}_2 \textbf{R}\right) = 0
\end{equation}
which shows that $k$ is an eigenvalue of the matrix $\textbf{R}^T \textbf{A}_2 \textbf{A}_1$. By solving for $k$ and denoting the left side of \Cref{A5equ:solve_k} by $det\left(\textbf{C}\right)$ \Cref{A5equ:resolved_system} can be rewritten as follows:
\begin{equation}
\label{A5equ:reduced_system_w_k}
	\left(\textbf{R}_1 \textbf{C} \textbf{R}^T\right)^{2\times 2} = \mathcal{O}^{2\times 2}
\end{equation}
As we already used $det\left( \textbf{C} \right) = 0$, \Cref{A5equ:reduced_system_w_k} only provides two independent solutions. With this constraint, $\textbf{R}_1$ can be found. In a first step the two non-zero eigenvalues $\lambda_1$ and $\lambda_2$ and the corresponding eigenvectors $\textbf{s}_1$ and $\textbf{s}_2$ of $\textbf{C}$ are calculated. Afterwards, the third column $\textbf{r}_{31}$ of $\textbf{R}_1$ is given by
\begin{equation}
\label{A5equ:Solution_r_31}
	\textbf{r}_{31} = \pm norm\left( \sqrt{|\lambda_1|}\textbf{s}_1 \pm \sqrt{|\lambda_2|}\textbf{s}_2 \right).
\end{equation}
\Cref{A5equ:Solution_r_31} yields four different possible solutions of $\textbf{r}_{31}$, which is due to the fact that the two intersecting cones $f_1$ and $f_2$ have two ellipses in common, as \cite{10.1007/BF01440844} shows. Considering the geometric conditions, that both virtual cameras are positioned on the same side of the ellipse and the gaze vector is directed away from the pupil plane towards the virtual camera, only one possible solution remains \cite{10.1145/1344471.1344506}. To obtain the correct $\textbf{r}_{31}$, we must ensure that both z-components of $\textbf{r}_{31}$ and $\textbf{r}_{32} = \textbf{R}\textbf{r}_{31}$ are positive. Using this criterion, in a second step the remaining vector $\textbf{r}_{31}$ is selected and the other columns of $\textbf{R}_1$, $\textbf{r}_{21}$ and $\textbf{r}_{11}$, are given by the corresponding eigenvectors of $\textbf{A}_1 - \textbf{r}_{31}\textbf{r}_{31}^T \textbf{A}_1$, which finally leads to a solution for $\textbf{R}_1$. Afterwards, the remaining parameters can be resolved in the following order
\begin{equation}
	\textbf{t}_1 = \begin{pmatrix}
	\textbf{r}^T_{11} \textbf{A}^T_1\\
	\textbf{r}^T_{12} \textbf{A}^T_1\\
	\textbf{r}^T_{21} \textbf{A}^T_2 \textbf{R}\\
	\end{pmatrix} 
	\begin{pmatrix}
	0\\
	0\\
	-\textbf{r}^T_{21} \textbf{A}^T_2 \textbf{t}\\
	\end{pmatrix}
,\quad
	k_1 = -\textbf{t}^T_1 \textbf{A}_1 \textbf{t}_1
,\quad
	a^2 = \frac{k_1}{\textbf{r}^T_{11}\textbf{A}_1 \textbf{r}_{11}}
,\quad
b^2 = \frac{k_1}{\textbf{r}^T_{12}\textbf{A}_1 \textbf{r}_{12}}
\end{equation}
to obtain the length of the principle axis $a$ and $b$ of the pupil ellipse $\mathcal{E}$ as well as the center of the pupil with respect to the camera coordinate system $c_1$. From  $\textbf{t}_1$ and $\textbf{R}_1$ the pupil-normal-vector can be derived to calculate the gaze angles.

\begin{figure}[h]
	\centering
	\includegraphics[width=\linewidth]{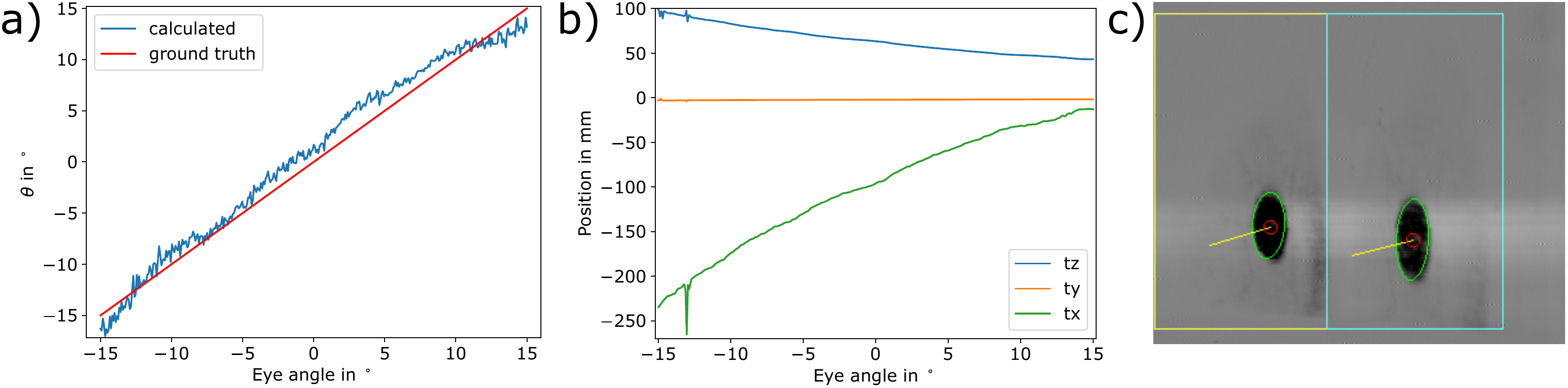}
	\caption{a) Calculated gaze angle $\theta$ (blue) based on stereo reconstruction and ground truth from rotation stage (red), b) Position of pupil center w.r.t. virtual camera 1 $\textbf{t}_1$, c) A captured stereo image with annotated ellipse contour, center, and reconstructed gaze vector   }
	\label{A5im:results}
\end{figure}
\FloatBarrier
\subsection{Evaluation}
\label{A5sec:Evaluation}
\begin{figure}[h]
	\centering
	\includegraphics[width=0.8\linewidth]{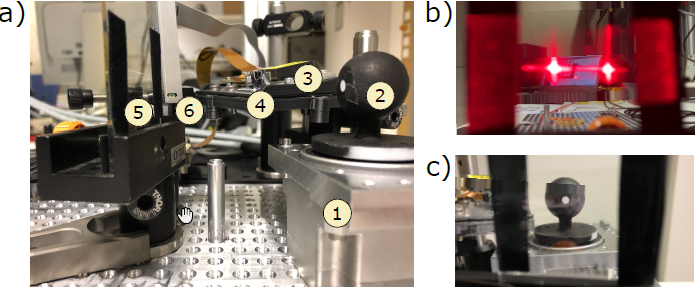}
	\caption{a) Laboratory setup consisting of the fabricated HOE (5), the MEMS laser scanner (3,4), a single-pixel detector (6) and an artificial eye (2), mounted on a precision, motorized rotation stage (1), b) Origin of the two virtual cameras captured by a camera placed in the pupil plane, c) Image of the artificial eye through the glasses' lens demonstrating the excellent transparency of the HOE.}
		\label{A5im:lab_setup}
\end{figure}
\FloatBarrier

To evaluate the accuracy and precision of our single-pixel holographic stereo camera eye-tracking sensor and to demonstrate the working principle, we implemented a laboratory setup as outlined in \Cref{A5im:lab_setup}. 

An artificial eye model (2) is attached to a precision motorized rotation stage (1) \footnote{Rxhq 50-12T0.3 Jenny Science }, which allows the artificial eye to be rotated with a resolution of $0.0089^\circ$ and a precision of $\pm 0.00083 ^\circ$. Furthermore a laser scanner\footnote{BML050 \cite{BML0502017}} consisting of a laser module (3) and a  MEMS micro mirror (4) is used to scan a 2D field across the fabricated HOE (5), which generates the stereo perspective. To capture the reflected light for image reconstruction, a photodiode (6) is used.

The HOE is fabricated to operate in accordance with the red laser wavelength ($\lambda = \SI{639}{\nano\metre}$) of the laser scanner. The two  optical center points of the two virtual cameras $o_1$ and $o_2$ are shown in \Cref{A5im:lab_setup} b) captured from the pupil plane.  Finally, \Cref{A5im:lab_setup} shows an image of the artificial eye model taken from the outside of the lens to highlight the high transparency of the HOE. The artificial eye model uses a white pupil to emphasize the  specular reflection of red light at the pupil, causing the photodetector to perceive a dark pupil, similar to the dark pupil shown by \cite{9149591} on a real eye using an IR laser scanner. 

 Tab. \ref{A5tab:lab_params} summarizes the geometrical and electrical parameters of the prototype system used in the laboratory to evaluate our approach.

To determine the camera parameters ($\textbf{K}_1$, $\textbf{K}_2$, $\textbf{R}$, and $\textbf{t}$) of the system, the stereo system is first calibrated as described in \Cref{A5subsec:Stereocalibration} using a 5$\times$3 checkerboard with a square size of 3\,mm. Afterwards, the artificial eye is rotated in 0.5$^\circ$ increments from -15$^\circ$ to 15$^\circ$ to cover a total FOV of 30 $^\circ$. For each position, 100 image pairs are acquired with the single-pixel-detector as described in \Cref{A5sec:Holographic_Stereocamera}. Then, the two sub-images per image are isolated and the contour of the pupil ellipse is extracted for each sub-image. Based on the extracted contours, the ellipse $\mathcal{E}_i$ is fitted using a least squares method and the ellipse parameters ($a_i$, $b_i$, \textbf{$t_{xi}$}, \textbf{$t_{yi}$} and $\phi_i$) are extracted. After rescaling the ellipse parameters by applying \Cref{A5equ:rescaling_ellipse} and computing \textbf{$A_1$} and \textbf{$A_2$}, the reconstruction algorithm, described in \Cref{A5subsec:Stereo_reconstruction}, is used to determine the gaze vector as well as the position of the ellipse with respect to $c_1$. \Cref{A5im:results} summarizes the results of the experiment.
\begin{table}[t]
	\caption{Summary of design parameters of the stereo setup in the laboratory setup according to \Cref{A5im:stereo_camera}.}
	\label{A5tab:lab_params}
	
	\centering
	\resizebox{\linewidth}{!}{%
		\begin{tabular}{p{1.5cm}p{1.5cm}p{2cm}p{3.2cm}p{2.8cm}p{2cm}p{1.5cm}}
			\hline
			Baseline b & Distance d  & Camera angle $\gamma$	& Camera resolution (w$\times$h)   & Size of HOE (w$\times$h)  & Pixel clock  & Frame rate  \\ 
			
			20\,mm & 80\,mm&14.2$^\circ$  & 110\,px $\times$ 240\,px   &20\,mm  $\times$ 10\,mm  &22\,MHz  &60\,Hz  \\\hline
		\end{tabular}
	}
	
\end{table}

\Cref{A5im:results} a) shows the gaze angle resulting from the reconstructed gaze vector for the artificial eye. The mean gaze accuracy is 1.35$^\circ$ with a mean precision of 0.02$^\circ$. The gaze accuracy is mainly limited by the pupil segmentation and the ellipse fitting accuracy. The accuracy can be improved by using more advanced ellipse fitting algorithms e.g. DeepVOG by \cite{YIU2019108307}. \Cref{A5im:results} b) shows the pupil center $\textbf{t}_1$ relative to $\textbf{c}_1$. While the y-axis is constant, the x-axis varies as the rotation of $\theta$ around $y_w$ leads to a shift of the pupil center along $x_w$. Furthermore, the distance $d$ between camera center $c_1$ and the pupil plane is close to the camera distance of 80\,mm described in Tab. \ref{A5tab:lab_params}.

\subsection{Discussion}
\label{A5sec:Discussion}
Overall, the experiment shows that stereo reconstruction leads to a comparably high gaze accuracy, even with the low resolution of the two virtual cameras. In terms of sensor integration, our approach has several advantages over VOG stereo systems, as the use of an HOE allows to capture image information from multiple angles.

\cite{9149591} reported a power consumption of 11\,mW of a similar setup, which is already significant lower compared to VOG systems. Therefore, our approach reduces power consumption significantly further compared to a stereo VOG setup, since we generate two virtual cameras from a single laser scanner and only have to sacrifice image resolution.

A limitation of our approach is the high dependence on robust detection of the pupil ellipse $\mathcal{E}$, since the reconstruction fails if the cones $f_1$ and $f_2$ do not intersect. \cite{Meyer2020e} addressed this general issue by replacing the IR laser and the single-pixel-detector by an laser feedback interferometry (LFI) sensor to increase the robustness of pupil detection, an approach that could be applied also in this working context.

Finally the stereo reconstruction algorithm  only estimates the normal vector of the pupil and thus the optical axis of the eye omitting the offset angle $\kappa$ between optical and visual axis \cite{1634506}. To determine the user dependent $\kappa$ at least a one time single marker calibration is required. Otherwise $\kappa$ needs to derived from average population.
\subsection{Conclusion}
\label{A5sec:Conclusion}
In this paper, we presented a novel approach to mobile stereo camera eye-tracking for AR glasses, which combines a MEMS laser scanner with a segmented HOE. We fabricated the HOE and demonstrated its functionality for the proposed use-case. We applied a stereo reconstruction algorithm to the stereo images captured by our holographic single-pixel virtual stereo camera, which achieved an accuracy of 1.35$^\circ$.

To increase the robustness of the stereo reconstruction as well as the FOV covered by the virtual cameras, we plan to increase the number of virtual cameras and optimize the orientation of the cameras w.r.t. to the eye. In addition, we plan to integrate our system into a head-worn demonstrator based on the BML500P \footnote{\cite{BML2021}} to further investigate precision and accuracy also taking human error into account.

\subsection*{Acknowledgements}
Enkelejda Kasneci is a member of the Machine Learning Cluster of Excellence, EXC number 2064/1 - Project number 390727645.
\newpage

\chapter{Static LFI HCI for AR Glassess}
\label{APP:Static_LFI_HCI}
This chapter includes the publications \cite{Meyer2021, meyer11788compact, CNN_Meyer_2021, 10.1145/3530884}:\blfootnote{Publications are included with minor template modifications. Original versions are available via the digital object identifier at the corresponding publishers. Publications 1,3 and 4 are \copyright 2021 ACM  and \copyright 2022 ACM respectively, and included with relevant permission. Publication 2 is \copyright 2021  Society of Photo‑Optical Instrumentation Engineers (SPIE) and reprinted, with permission, from Johannes Meyer, Thomas Schlebusch, Hans Spruit, Jochen Hellmig, Enkelejda Kasneci, "A compact low-power gaze gesture sensor based on laser feedback interferometry for smart glasses," Proc. SPIE 11788, Digital Optical Technologies 2021, 117880D (20 June 2021); https://doi.org/10.1117/12.2593772. }
\begin{enumerate}[label=\arabic*.]
	\item \textbf{Johannes Meyer}, Thomas Schlebusch, Hans Spruit, Jochen Hellmig, Enkelejda Kasneci. "A Novel Gaze Gesture Sensor for Smart Glasses Based on Laser Self-Mixing". In Extended Abstracts of the 2021 CHI Conference on Human Factors in Computing Systems  (2021), \url{https://doi.org/10.1145/3411763.3451621}  
	\item \textbf{Johannes Meyer}, Thomas Schlebusch, Hans Spruit, Jochen Hellmig, Enkelejda Kasneci. "A compact low-power gaze gesture sensor based on laser feedback interferometry for smart glasses". Proc. SPIE 11788, Digital Optical Technologies 2021  (2021), \url{https://doi.org/10.1117/12.2593772}
	\item \textbf{Johannes Meyer}, Adrian Frank, Thomas Schlebusch, Enkeljeda Kasneci. "A CNN-based Human Activity Recognition System Combining a Laser Feedback Interferometry Eye Movement Sensor and an IMU for Context-aware Smart Glasses". Proc. ACM Interact. Mob. Wearable Ubiquitous Technol. 5 (2021), \url{https://doi.org/10.1145/3494998} 
	\item \textbf{Johannes Meyer}, Adrian Frank, Thomas Schlebusch,  Enkelejda Kasneci. "U-HAR: A Convolutional Approach to Human Activity Recognition Combining Head and Eye Movements for Context-Aware Smart Glasses". Proc. ACM Hum.-Comput. Interact. 6 (2022), \url{https://doi.org/10.1145/3530884}  
\end{enumerate}
\section{A Novel Gaze Gesture Sensor for Smart Glasses Based on Laser Self-Mixing}
\label{APP:B1}
\subsection{Abstract}
The integration of gaze gesture sensors in next-generation smart glasses will improve usability and enable new interaction concepts. However, consumer smart glasses place additional requirements to gaze gesture sensors, such as a low power consumption, high integration capability and robustness to ambient illumination. We propose a novel gaze gesture sensor based on laser feedback interferometry (LFI), which is capable to measure the rotational velocity of the eye as well as the sensor's distance towards the eye. This sensor delivers a unique and novel set of features with an outstanding sample rate allowing to not only predict a gaze gesture but also to anticipate it. To take full advantage of the unique sensor features and the high sampling rate, we propose a novel gaze symbol classification algorithm based on single sample. At a mean F1-score of 93.44 \%, our algorithms shows exceptional classification performance.

\subsection{Introduction}







In recent years, various smart glasses have been released into the market. As a successor of virtual reality (VR) glasses, they follow a more natural design while integrating similar sensing technology into the glasses. In general, they consist of a light projection engine and a set of sensors to capture user inputs and monitor the user's state. 

To enable human computer interaction (HCI) with smart glasses, a variety of interaction concepts are possible. These concepts can be classified into touch based interactions e.g. via a track pad on the glasse's temple introduced by the Google Glass \cite{Google2013}, an external on-body device e.g. a controller like as in the Magic Leap One \cite{Magicleap2015} or via touchless interactions such as spoken commands via voice recognition as used by the Echo Frames \cite{amazon2019}. Also, hand gestures captured by a camera sensor have been shown by the HoloLens \cite{8368051}.

An additional touchless interaction concept is making use of the user's gaze by tracking their eye movements by means of eye-tracking sensors. Gaze-based interaction allows a fast and natural input, leading to an intuitive and unobtrusive way of interaction with smart glasses while maintaining social acceptance and user's privacy \cite{8368051}.

To the best of our knowledge, no commercially available smart glasses solution so far utilizes gaze gestures as input modality, mainly due to the challenges arising from integration and due to limited available power and space constraints \cite{8368051}. Furthermore, a high robustness of the sensor against variable external illumination is required to allow for operation in uncontrolled outdoor environments e.g. in bright sun light \cite{Fuhl2016}. 

To overcome these limitations and enable gaze gesture based interaction for the next generation of smart glasses, we present a novel low power gaze gesture sensing approach based on laser feedback interferometry (LFI). This multimodal sensor is capable of measuring distance towards the eye as well as eye rotational velocity with a high sample rate of up to 1\,kHz. The sensor is based on a small vertical cavity surface emitting laser (VCSEL) in the near infrared (IR) spectrum, which allows for a space constrained integration into the glasse's frames. Additionally, the LFI sensing principle is due to the coherent sensing scheme only sensitive to its own radiation, allowing a robust operation in presence of ambient radiation, as shown in \cite{Meyer2020e}.

In the next section, we give an overview of the state of the art regarding gaze gesture sensor concepts for AR glasses and discuss the limitations of existing sensor concepts. In \Cref{B1sec:Gaze gesture sensor}, we introduce the sensing principle and  the sensor concept. In addition, we provide an overview of the measurement features produced by the LFI sensor on the human eye. Based on these features, we introduce  an optimized and robust gaze symbol classification algorithm in \Cref{B1sec:gaze_gesture_algorithm}. Afterwards, we evaluate the proposed sensor concept in \Cref{B1sec:Evaluation} using a laboratory setup. In the last section, we conclude our findings and discuss further steps.





\subsection{Related Work}
\label{B1sec:Related_work}
Several eye-tracking sensor concepts have been investigated in the past. The successors of these different concepts, which are widely used in research and commercial applications, are video-oculography (VOG) and electro-oculography (EOG) \cite{majaranta2014eye}. In addition, novel low power eye-tracking sensor approaches for AR glasses based on microscanners and infrared (IR) lasers have emerged in recent years \cite{7181058, 9149591}.

 

Drews et al. used a VOG eye-tracking sensor with a 60\,Hz sample rate to capture the absolute gaze position and extract gaze gestures from the input data stream \cite{Drews2007}. They described a gaze gesture as a sequence of atomic eye movements. Atomic eye movements, also refereed to as strokes of the eye, are single unidirectional eye movements, e.g. left or upwards and can be interpreted as gaze symbols. A set of gaze symbols forms a gaze gesture protocol. Drews et al. used gaze symbols to control the user interface by linking a sequence of gaze symbols to a gaze gesture.The main advantage of their approach over dwell based interaction approaches, e.g. by \cite{Bednarik_Gowases_Tukiainen_2009}, is that an absolute calibration of the eye-tracking sensor is not required because only relative eye positions are tracked.



The main disadvantage of VOG sensor based approaches is the high power consumption required by the sensors and the image processing, as well as the limited sample rate of the sensors \cite{8368051}. Furthermore, they are sensitive to ambient light which disturbs the captured images leading to a low detection rate and therefore limiting the interaction capabilities in the wild \cite{Fuhl2016}.


To overcome this limitations Bulling et al. integrated EOG sensors into a wearable glasses demonstrator to measure relative movements of the eye and extract gaze gestures for mobile human computer interactions \cite{10.5555/1735835.1735842}. They placed two electrode pairs for horizontal and vertical eye movements around the eye and used microcontroller to sample the four EOG channels and extract the relative eye movements. In addition, they used a light sensor and an accelerometer to compensate external artefacts in the sensor data. They reported a sample rate of 250\,Hz with an overall power consumption of 769\,mW.


They use a gaze gesture protocol with 16 symbols to encode the gaze gestures. The overall correct classification rate for different gaze gestures of varying complexity was given as 87\,\%.

Compared to VOG sensor concepts, EOG sensors benefit from computational light weight signal processing. In addition, they are more robust against ambient illumination. The major drawback limiting the use of EOG sensor for gaze-based interaction in recent smart glasses is the use of electrodes on the skin \cite{majaranta2014eye}. 

Our gaze gesture sensor concept based on LFI combines the benefits of non-intrusive integration of VOG sensors and the calibration free measurement of relative eye movements by EOG sensors. Furthermore, our sensor principle is robust against external illumination and is therefore capable of operating in the wild.

\subsection{Gaze gesture sensor based on laser feedback interferometry}
\label{B1sec:Gaze gesture sensor}
\begin{figure}[h]
	 \centering
	\includegraphics[width=0.8\linewidth]{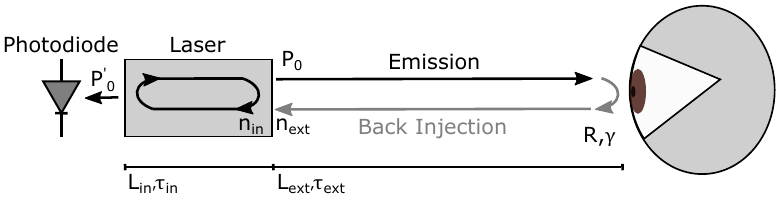}
	\caption{Coupled cavity model of a laser feedback interferometry sensor. The laser emits light which is scattered by the eye and back injected into the laser cavity. The photodiode monitors the laser power, which varies with changes in the feedback path.}
	\label{B1im:coupled_cavity_model}
\end{figure}
\FloatBarrier
Laser feedback interferometry or self-mixing interferometry (SMI) is a widely known interferometry measurement method. It is used in the industry as well as in laboratory environments to measure displacement and velocity of solid targets, as well as fluids and distance. Due to the high distance and velocity resolution, it is also widely used in vibrometry applications \cite{Giuliani2002}. 

\subsubsection{Sensing principle}
\label{B1subsec:Sensing principle}
\Cref{B1im:coupled_cavity_model} shows the coupled cavity model to introduce the basic sensing principle of LFI sensors. A laser with an optical output power $P_0$ emits a coherent laser beam towards the surface of the eye. The laser beam hits the eye under an angle of incidence $\gamma$, is attenuated by volume scattering effects and absorption described by a reflectivity $R$ and is back injected into the laser. $\tau_{ext}$ denotes the time the laser beam requires to travel over the distance $L_{ext}$ towards the eye. $\tau_{ext}$ is dependent on the speed of light $c_0$ and the external refraction index $n_{ext}$ of the external medium \cite{Taimre:15}.

The back injected light interferes with the local oscillating field, which is often referred to as self-mixing interference, resulting in a modulation of the optical power
\begin{equation}
\label{B1equ:modulated_power}
	P_0' = P_0 \left( 1 + m \cdot \cos \left(\phi_{fb}\right)\right).
\end{equation} 
The varying feedback power $P_0'$ is dependent on the optical power $P_0$, the modulation depth $m$ and a varying phase $\phi_{fb}$ of the backscattered light field. A small fraction of the varying feedback power $P_0'$ is measured by a photodiode, which is integrated into the back mirror of the laser cavity \cite{Taimre:15}. 

A more in-depth description of the coupled cavity model is given by the rate equations introduced by \cite{1070479}. A solution of the rate equation is the excess phase equation 
\begin{equation}
\label{B1equ:excess_phase_equation}
	\phi_{fb} - \phi_s + C \sin\left(\phi_{fb} + \arctan \left(\alpha\right)\right) = 0.
\end{equation}
The feedback phase is expressed as a function of the signal phase $\phi_s$, Acket's feedback parameter $C$ and Henry's line width enhancement factor $\alpha$. Considering operation of the LFI sensor in the weak feedback regime ($C<1$) and a constant $\alpha$, \Cref{B1equ:excess_phase_equation} has only a single solution and $\phi_{fb}$ is, therefore, only dependent on $\phi_s$, leading to 
\begin{equation}
\label{B1equ:phase_stimuli}
	\phi_s = \frac{4 \pi n_{ext} L_{ext}}{\lambda}
\end{equation}
with $\lambda$ describing the wavelength of the laser. Considering $n_{ext} \approx 1 $ is constant due to the operation of the sensor in free space, only changes in $\lambda$ and $L_{ext}$ lead to a varying phase $\phi_s$ and , consequently, to a varying phase $\phi_{fb}$. This results, with respect to \Cref{B1equ:modulated_power}, in a modulation of the optical power which is measured by the photodiode. Changes in the wavelength $\lambda$ occur by a modulation of the laser drive current. That leads to a periodic heating and cooling of the resonator and, thus, to a periodic change of the cavity length. The  variation of the cavity length leads to a periodic modulation of the wavelength, which allows for continuous measurement of the distance according to \Cref{B1equ:distance_equation}.

To distinguish between both effects, we compute the partial derivative of \Cref{B1equ:phase_stimuli} with respect to time, which leads to
\begin{equation}
\label{B1equ:distance_equation}
	f_0 = \left. \frac{2 L_{ext}}{\lambda^2} \frac{d\lambda}{dI} \frac{dI}{dt}\right|_{L_{ext} = const.}
\end{equation}
and
\begin{equation}
\label{B1equ:doppler_equation}
	f_d = \left. \frac{2 v_{ext} cos\left(\gamma \right)}{\lambda}\right|_{\lambda = const.}.
\end{equation}

Considering a known $\frac{d\lambda}{dI}$, which is a static process parameter of the laser, and a controlled current modulation $\frac{dI}{dt}$, the distance to the eye can be calculated by extracting the so called beat frequency $f_0$ by applying a fast Fourier transform (FFT) to the measured varying optical power and rearranging \Cref{B1equ:distance_equation} with respect to $L_{ext}$.

Movements of the eye ($\frac{d L_{ext}}{dt} = v_{ext}$) lead to a shift of the beat frequency $f_0$ by the so called Doppler frequency $f_d$. With a known angle of incidence $\gamma$ and a measured Doppler frequency, \Cref{B1equ:doppler_equation} can be rearranged with respect to $v_{ext}$ to obtain the surface velocity of the eye. 

In order to separate $f_0$ and $f_d$ and, thus, simultaneously measure the distance and velocity of the eye, a triangular modulation similar to frequency modulated continuous wave (FMCW) radar is applied to the drive current of the laser \cite{10.1117/12.775131}. By separating the up- and down ramp signals into two segments and applying an FFT on each segment, an $f_{up}$ and an $f_{down}$ frequency is captured. $f_0$ and $f_d$ are obtained from these measurements by
\begin{equation}
\label{B1equ:distance_triangular}
	f_0 = \frac{f_{up} + f_{down}}{2}
\end{equation}   
and 
\begin{equation}
\label{B1equ:doppler_triangular}
	f_d = \frac{f_{up} - f_{down}}{2}
\end{equation}
respectively. 


\begin{figure}[h]
	\centering
	\includegraphics[width=0.4\linewidth]{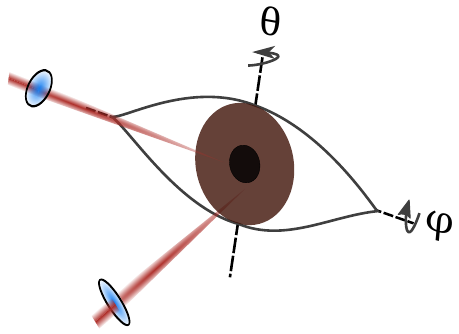}
	\caption{Positioning of the LFI sensors with respect to the rotational axes of the human eye.}
	\label{B1im:positioning_of_sensors}
\end{figure}
\FloatBarrier
The LFI sensor measures surface velocity as a dot product between the laser normal and the surface velocity vector of the moving eye at the intersection point between laser and eyeball. If the eye's rotational axis and the laser beam are aligned, a simplified description by the angle of incidence $\gamma$ as described in \Cref{B1equ:doppler_equation} is possible. Therefore, two LFI sensors are required to measure both, the horizontal movement around the $\theta$-axis and vertical movement around the $\phi$-axis. \Cref{B1im:positioning_of_sensors} shows the positioning and the laser beam directions of two LFI sensors to comply with these requirements.

From an integration point of view, the sensor for vertical movements can be integrated into the AR glasse's frame below the spectacle and the sensor for horizontal movements can be integrated into the AR glasses frame temple. The size of a single LFI sensor is mainly determined by the diameter of the lens in front of the laser chip which is roughly 4\,mm. This results in a size of the final sensor that is comparable to the twin-eye sensor shown in \cite{Taimre:15}.

\subsubsection{Sensor features on the eye} 
\label{B1subsec:Sensor features on the eye}
\begin{figure}[h]
	\centering
	\includegraphics[width=0.6\linewidth]{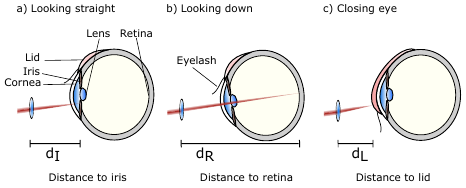}
	\caption{Different movement related position changes of the eye and the lid resulting in a change in distance measurement due to the geometry and scaffold of the eye.}
	\label{B1im:distance_features_eye}
\end{figure}
\FloatBarrier
Based on \Cref{B1equ:distance_equation} and \Cref{B1equ:doppler_equation} and the positioning of the LFI sensors shown in \Cref{B1im:positioning_of_sensors}, four features are measured by the sensors. For each rotational axis of the eye, the surface velocity and the distance are measured resulting in $v_{\theta}$ and $v_{\phi}$ as velocity related features and $d_{\theta}$ and $d_{\phi}$ as position related features.

\Cref{B1im:distance_features_eye} shows a sectional view of the eye and a fixed LFI sensor position for three different positions of the eye and the lid. In \Cref{B1im:distance_features_eye} a), the gaze is directed straight ahead. The laser beam of the LFI sensor for vertical rotations penetrates the cornea and backscattering occurs at the iris. For this setup, the LFI sensor measures the distance between sensor and iris $d_I$.

In \Cref{B1im:distance_features_eye} b), the eye is slightly rotated along the vertical direction downwards and the laser beam penetrates the cornea and the lens and is backscattered from the retina. The measured distance in this case is the distance between sensor and retina $d_R$.

In \Cref{B1im:distance_features_eye} c), the eye is directed straight forward but the lid is closed by a blink. In this case, the distance $d_L$ between sensor and lid is measured. By subtracting $d_I$ and $d_L$, the thickness of the lid can be calculated, which is around 4\,mm, and by subtracting $d_R$ and $d_L$ the diameter of the eye is approximately calculated, which is around 24\,mm \cite{gross2008human}. 
\subsection{Gaze gesture algorithm} 
\label{B1sec:gaze_gesture_algorithm}
Similar to related works by \cite{Drews2007}, \cite{10.5555/1735835.1735842} and \cite{10.1007/978-3-030-20521-8_27}, the gaze gesture algorithm is based on a gaze gesture protocol which encodes atomic movements of the eye into gaze gestures denoted by symbols. 

We define a gaze symbol set consists of four basic atomic eye movements which are annotated by small letters. For the horizontal axis, movements to the left \textit{l} and to the right \textit{r}  around $\theta$ are possible. In addition, eye movements in the vertical axis around $\phi$ are denoted by \textit{u} and \textit{d} for up and down movements of the eye, respectively. In addition to movements of the eye, blinks (\textit{b}) are considered an additional atomic movement. In order to cover all types of eye movements, fixations of the eye, as well as slow movements, are described by an additional symbol \textit{n} for non-movements. 

The entire set of symbols $S$ thus consists of six unique symbols $S=\{l,r,u,d,b,n\}$.
\subsubsection{Feature extraction} 
\label{B1subsec:Feature extraction}
The novel sensor concept described in \Cref{B1sec:Gaze gesture sensor} allows for the extraction of a unique set of features from the human eye with a high sampling rate. A major advantage over the state of the art is the use of the distance measurement between the glasse's frame and the eye as an additional feature, as discussed in \Cref{B1subsec:Sensor features on the eye}.

\begin{figure}[h]
	\centering
	\includegraphics[width=0.3\linewidth]{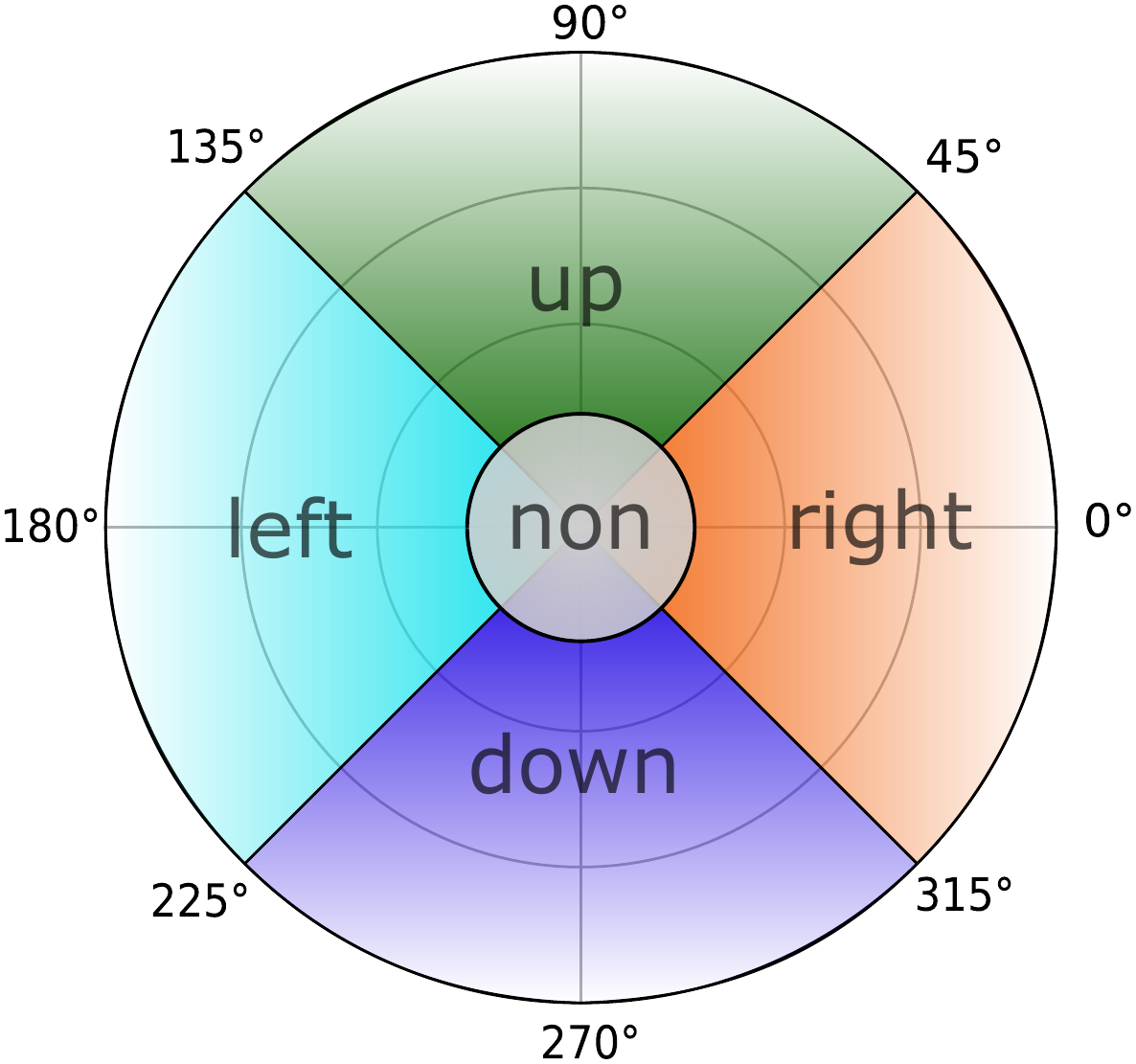}
	\caption{Gaze gesture velocity feature space shown in a polar coordinate system.}
	\label{B1im:polar_feature_space}
\end{figure}

To obtain unique features from the measured velocities, $v_\theta$ and $v_\phi$ are treated as velocity vector components. This velocity vector can be represented in polar coordinate space by an angle 
\begin{equation}
\label{B1equ:angle_feature}
	\epsilon = \arctan(\frac{v_\theta}{v_\phi})
\end{equation}
and a vector length 
\begin{equation}
\label{B1equ:velocity_feature}
	v = \sqrt{v_\theta^2 + v_\phi^2}
\end{equation}
which lead us to an two dimensional feature vector $F$ with $\epsilon$ and $v$ as gaze symbol features. \Cref{B1im:polar_feature_space} shows the feature space, which is spanned by $\epsilon$ and $v$, in polar coordinate space. 
\FloatBarrier
The grey area in the centre covers the area of fixations and slow eye movements e.g. drift or tremor. The size is determined by a certain absolute velocity threshold $v_n$ for $v$. The sensitivity and robustness of the gaze symbol algorithm can be controlled by varying $v_n$. The main advantage of this representation is the wide range of allowed angles of 90$^\circ$ per direction, leading to a robust gaze symbol detection and an easy execution of gaze gestures for the user.

To also be able to measure blinks $d_\theta$ is added as a third feature to $F$. The feature vector describes a point in cylindrical coordinate denoted by $F=\{\epsilon, v, d_\theta\}$.
\subsubsection{Gaze symbol classification} 

\label{B1subsec:Gaze symbol classification}
The first step of the classification process is to map a feature vector $F$ measured by the LFI sensors to a gaze symbol from the gaze symbol set $S$. For this purpose, a decision tree, which describes the feature space shown in \Cref{B1im:polar_feature_space} and the additional distance information $d_\theta$, is used. The main advantage of this single sample classification approach over other classification approaches, e.g. by \cite{10.5555/1735835.1735842}, is the invariance with respect to time. This allows for a robust classification, which is insensitive to sensor drift, as is the case with EOG sensors \cite{majaranta2014eye}. Furthermore, this classification approach does not require the detection of a movement sequence in the input sensor signal stream to extract significant movements which belong to a gaze symbol as proposed by \cite{10.1007/978-3-030-20521-8_27}. 

\subsection{Evaluation}
\label{B1sec:Evaluation}
\begin{figure}[ht]
	\centering
	\includegraphics[width=0.43\linewidth]{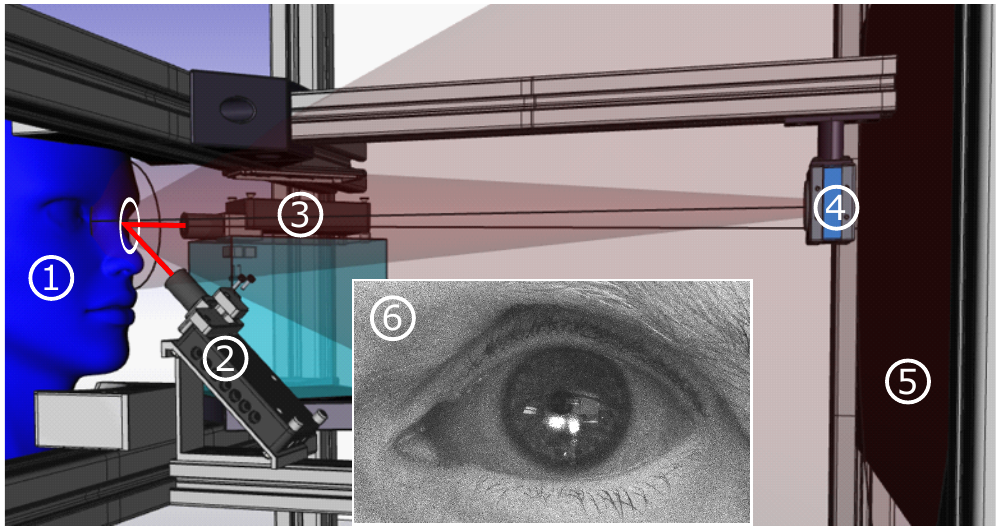}
	\caption{CAD sketch of the laboratory setup to validate the proposed LFI gaze gesture sensor approach.}
	\label{B1im:Lab_setup}
\end{figure}
\FloatBarrier
To validate the proposed features of the human eye measured by the LFI sensors and evaluate the  gaze gesture recognition algorithm, a laboratory setup is used, as shown in \Cref{B1im:Lab_setup}. 

The subject \circled{1} is placed in front of a monitor \circled{5} which is used to create stimuli for eye movements. Its head is fixated by a head- and chin rest to suppress unintended head movements. The LFI sensor at position \circled{2} is responsible for measurements of $v_\phi$ and $d_\phi$ while the LFI sensor at position \circled{3} is responsible for measurements of $v_\theta$ and $d_\theta$. An IR camera \circled{4} captures a video of the subject during the experiments. A sub Figure \circled{6} shows a frame captured by the camera showing the focused laser spots of the two IR lasers on the iris of a subject as white spots.

\begin{table}[ht]
	\caption{Properties of the LFI sensors of the laboratory setup.}
	\label{B1tab:eye_tracking_params}
		 \centering
	\begin{tabular}{ccccc}
		
		LFI Sensor& $d \lambda / dI$ &$\lambda$	&$\gamma$  &$P_0$  \\ \hline
		
		\circled{2} &	 0.406\,nm/mA &848\,nm &~45$^\circ$  &390\,$\mu W$ \\
		
		\circled{3} &	 0.396\,nm/mA &856\,nm  &~45$^\circ$  &390\,$\mu W$   \\
	\end{tabular}
	
\end{table}
\FloatBarrier

With the low optical power of the IR VCSELs, the mechanical setup of chin and head rest and the positioning of the lasers in the laboratory setup, a class 1 laser system according to IEC 60825-1 (class 1 optical power limit 780\,$\mu$W) is achieved and, therefore, the experiments do not pose any medical hazard to the subject's eye. 

The power consumption of the proposed gaze gesture sensor is roughly 140\,mW, which is estimated based on an STM32G473 microcontroller including all required peripherals to capture and process the data from the LFI sensors. With optimized logic blocks and an subsampling scheme enabled by an higher integration using a custom application specific integrated circuit (ASIC) a further power reductions to 30\,mW is expected.   

A comparable power consumption of 150\,mW is reported by \cite{7181058} for state of the art VOG eye tracking sensors excluding image processing. Compared to \cite{10.5555/1735835.1735842} a much lower power consumption can be achieved.

\subsubsection{Feature validation}
\label{B1subsec:Feature validation}
\begin{figure}[ht]
	\centering
	\includegraphics[width=0.7\linewidth]{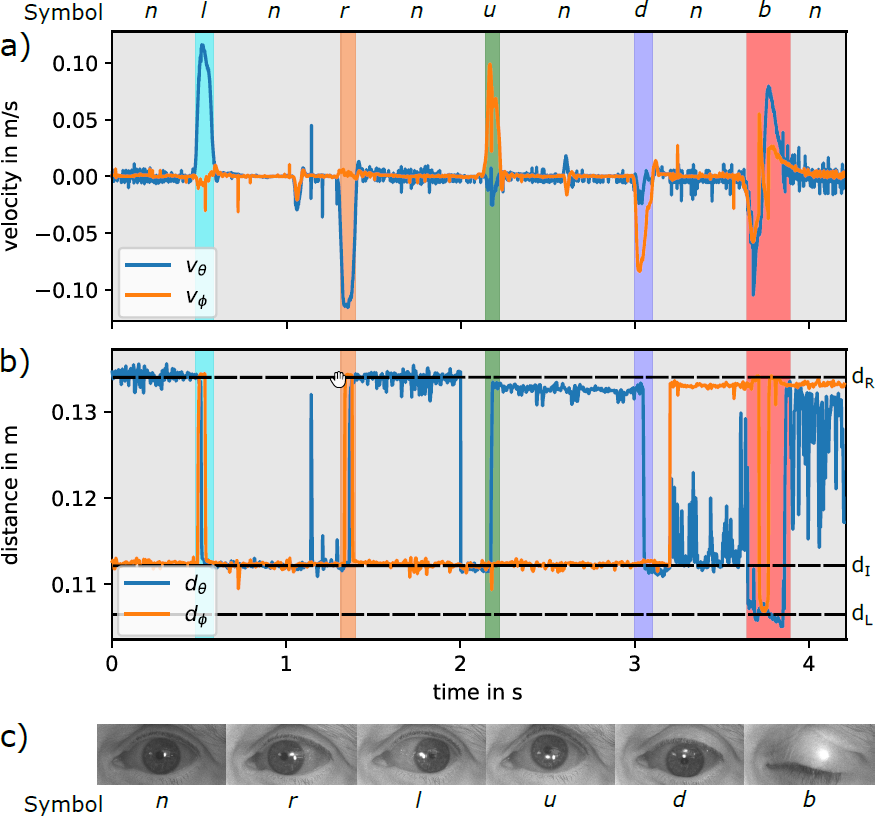}
	\caption{Data captured with the laboratory setup to evaluate the proposed features of the eye. a) shows the measured velocity, b) shows the measured distance and c) shows eye movements captured with the IR camera and the corresponding gaze symbol. The background colour refers to the corresponding atomic movement of the eye.}
	\label{B1im:evaluation_features}
\end{figure}

To validate the proposed features of the human eye from \Cref{B1subsec:Sensor features on the eye}, the laboratory setup is used. \Cref{B1im:evaluation_features} shows the measured data of the two LFI sensors from a subject performing the gaze symbol sequence $[l,r,u,d,b]$ .  
\FloatBarrier
The LFI sensors are modulated to achieve an update rate of 1\.kHz for distance and velocity measurement. \Cref{B1im:evaluation_features} a) shows the measured velocity and \Cref{B1im:evaluation_features} b) shows the measured distance. The movement threshold $v_n$ is set to 0.02\,m/s to distinguish between atomic eye movements and non-relevant movements of the eye. In the distance plot we added three distance lines corresponding to the distances described by \Cref{B1im:distance_features_eye} a) - c). The measured distance between $d_R$ (retina) and $d_I$ (iris) is 23,36\,mm and the measured distance between $d_I$ (iris) and $d_L$ (lid) is 5,01\,mm, which both correspond to the known anatomy of the human eye.




\subsubsection{Gaze gesture symbol classification}
To validate the proposed gaze gesture symbol classifier described in \Cref{B1subsec:Gaze symbol classification}, a set of gestures is recorded using the laboratory setup as described in \Cref{B1sec:Evaluation}. Two male subjects with blue and brown eyes were instructed to perform 18 eye movement gestures and 9 blinks each. To capture natural trajectories of the eye, during the execution of the gaze gestures no visual stimuli were used to guide the gaze of the subjects. 

In a first step, the ground truth is manually annotated utilizing the IR camera images to the measured data with the corresponding gaze symbols, similar to related work by \cite{Santini2016}. To distinguish between movements and non-movements, the velocity threshold $v_n = 0.02\,ms$ is applied. Afterwards, we extract the features from the measured data according to \Cref{B1subsec:Feature extraction} and use the proposed gaze symbol classification algorithm to classify atomic eye movements. This allows us to evaluate the  classification approach on the sample-level. In addition, we treat the multiclass classification problem for evaluation purpose as a binary one-vs-all classification problem. This evaluation approach is commonly used in the literature \cite{startsev20191d, hoppe2016endtoend} and allows us to compute the F1-score as evaluation metric. \Cref{B1tab:gaze_gestures} shows the evaluation results.

\begin{table}[ht]
	\caption{Properties of the LFI sensors of the laboratory setup.}
	\label{B1tab:gaze_gestures}
	\centering
	\begin{tabular}{cccccc}
		
		Gaze symbol& Gestures  & Symbols & Precision  & Recall  & F1-Score  \\ \hline
		\textit{l} &	 36 & 3783 &0.906  &0.995 & 0.949\\
		\textit{r} &	 36 & 3505  &0.916  & 0.997 & 0.955  \\
		\textit{u} &	 36 & 3229 & 0.820 & 0.976 & 0.891 \\
		\textit{d} &	 36 & 3624  &0.873  & 0.945 & 0.908   \\
		\textit{b} &	 18 & 3063  &0.971  &0.967 & 0.969  \\
	\end{tabular}
\end{table}

\subsection{Conclusion}
We present a novel low power gaze gesture sensor concept based on IR VCSELs and the LFI effect, which allows for seamless integration of gaze gesture sensors into next generation smart glasses. This sensor enables new gaze based interaction concepts like true hands-free interaction. We also introduced a computational lightweight gaze symbol classification algorithm with a sufficient classification accuracy and precision. 

Based on the promising results obtained with the laboratory setup, our next goal is to integrate the sensors into our head-mounted demonstrator. This will allow us to evaluate the gaze gesture sensor during everyday activities to investigate the robustness against glasses slippage.

Furthermore we want to develop a gaze gesture algorithm which recognizes a user interaction with the glasses by detecting a sequence of gaze symbols and investigate the robustness against unwanted interactions during everyday activities.   
	
\newpage

\section{A compact low-power gaze gesture sensor based on laser feedback interferometry for smart glasses (invited)}
\label{APP:B2}
\subsection{Abstract}
 The integration of gaze gesture sensors in next-generation smart
glasses will improve usability and enable new interaction concepts.
However, consumer smart glasses place additional requirements
to gaze gesture sensors, such as a low power consumption, high
integration capability and robustness to ambient illumination. We
propose a novel gaze gesture sensor based on laser feedback interferometry
(LFI), which is capable to measure the rotational velocity
of the eye as well as the sensor’s distance towards the eye. This sensor
delivers a unique and novel set of features with an outstanding
sample rate allowing to not only predict a gaze gesture but also to
anticipate it. To take full advantage of the unique sensor features
and the high sampling rate, we propose additionally a novel gaze
gesture classification algorithm based on single sample. At a mean
F1-score of 93.44 \%, our algorithms shows exceptional classification
performance at a negative latency between gaze gesture input and
command execution.

\subsection{Introduction}

Consumer smart glasses evolve their full potential, if the user has intuitive means to interact with and control the device. In the past a variety of interaction concepts are brought into the market, which can be classified into touch based interactions e.g. via a track pad on the glasse's temple introduced by the Google Glass \cite{Google2013}, an external on-body device e.g. a controller like as in the Magic Leap One \cite{Magicleap2015} or via touchless interactions such as spoken commands via voice recognition as used by the Echo Frames \cite{amazon2019}. Also, the use of hand and finger gestures captured by a camera sensor have been shown by the HoloLens \cite{8368051}.

An additional touchless interaction concept is making use of the user's gaze by tracking their eye movements by means of eye-tracking sensors. Gaze-based interaction allows a fast and natural input, leading to an intuitive and unobtrusive way of interaction with smart glasses while maintaining social acceptance and user's privacy \cite{8368051}. 

Gaze-based interaction allow further to adaptive control and activate the image projection system based on the users interaction, which allows to extend the battery life of lightweight augmented reality (AR) glasses to all-day operation.

To cover both aspects, user interaction and system control, low-power always-on gaze gesture sensors are well suited.

To the best of our knowledge, no commercially available smart glasses solution so far utilizes gaze gestures as input modality, mainly due to the challenges arising from integration and due to limited available power and space constraints \cite{8368051}. Furthermore, a exceptional robustness of the sensor against variable ambient illumination is required to allow for operation in uncontrolled outdoor environments e.g. in bright sun light \cite{Fuhl2016}. 

To fulfill these requirements and enable gaze gesture based interaction for the next generation of smart glasses, we present a novel low power gaze gesture sensor based on laser feedback interferometry (LFI) and extend our previous work \cite{Meyer2021} by an gaze gesture detection algorithm. This multimodal sensor is capable of measuring distance towards the eye as well as eye rotational velocity with an outstanding sample rate of up to 1\,kHz. The sensing component consits of a tiny infra red (IR) vertical cavity surface emitting laser (VCSEL) with an integrated photodiode. Due to its coherent sensing scheme, the sensor is only sensitive to its own emitted radiation, allowing a robust operation in presence of ambient radiation, as shown by Meyer et.al. \cite{Meyer2020e}. Another advantage of this sensing technology is the omission of an imaging process for gaze sensing. This enables unobtrusive, socially acceptable sensor integration and reduces privacy concerns as no images of the eyes are captured.

In the next Section, we give an overview of the state of the art regarding gaze gesture sensors as well as gaze interaction concepts for AR glasses and discuss the limitations of existing concepts. In \Cref{B2sec:Gaze gesture sensor}, we introduce the sensing principle and  the sensor concept of the proposed LFI sensor. In addition, we provide an overview of the measurement features captured by the LFI sensor on the human eye. Based on these features, we introduce  an optimized and robust gaze gesture recognition algorithm in \Cref{B2sec:gaze_gesture_algorithm}. Afterwards, we evaluate the proposed sensor concept in \Cref{B2sec:Evaluation} using a laboratory setup. In the last section, we conclude our findings and discuss further steps.

\subsection{Related Work}
\label{B2sec:Related_work}
Several eye-tracking sensor concepts have been investigated in the past. The successors of these different concepts, which are widely used in research and commercial applications, are video-oculography (VOG) and electro-oculography (EOG) \cite{majaranta2014eye}. In addition, novel low power eye-tracking sensor approaches for AR glasses based on microscanners and infrared (IR) lasers have emerged in recent years \cite{7181058, 9149591}.

Most gaze interaction concepts relie on VOG. VOG concepts for smart glasses rely on video cameras, which are mainly integrated into the glasses frame. They record the movements of the eye and track the position of the pupil and extract the gaze vector by using computer vision algorithms. 

Bednarik et.al. introduce a gaze interaction method by interpreting the gaze vector as mouse courser to interact with UI elements \cite{Bednarik_Gowases_Tukiainen_2009}. By fixating and dwelling on an UI element, the UI element is activated and an interaction is performed. The dwell time is required to differentiate between explicit interactions and random eye movements and thus solve the so-called Midas touch problem. The main disadvantage of this approach is that, due to the required dwell time, only a few interactions can be performed in a defined period of time. Furthermore, an absolute calibration of the eye tracker is necessary to obtain the absolute gaze position.
 
Drews et al. used a VOG eye-tracking sensor to capture the absolute gaze position and extract gaze gestures from the input data stream \cite{Drews2007}. They described a gaze gesture as a sequence of atomic eye movements. Atomic eye movements, also refereed to as strokes of the eye, are single unidirectional eye movements, e.g. left or upwards and can be interpreted as gaze symbols. A set of gaze symbols forms a gaze gesture protocol. Drews et al. used gaze symbols to control the user interface by linking a sequence of gaze symbols to a gaze gesture. In addition, they used a timeout of 1000\,ms to distinguish between gesture inputs and natural eye movements. The main advantage of their approach over dwell based interaction approaches is that an absolute calibration of the eye-tracking sensor is not required because only relative eye positions are tracked.

In recent years gaze based pursuit interactions are investigated \cite{Bace2020}. Pursuit interaction exploit pursuit eye movements by displaying moving objects, which are pursued by the human gaze. \cite{Bace2020} combined the gaze vector and optical flow of a series of images to obtain pursuit informations for targets shown on a display. Similar to the gaze gesture approach, they only require relative eye movements, but in order to obtain the velocities for detection of pursuit eye movement , the optical flow needs to be calculated from a series of incoming images, resulting in high computational complexity.



The main disadvantage of VOG sensor based approaches is the high power consumption required by the sensors and the image processing, as well as the limited sample rate of the sensors \cite{8368051}. Furthermore, they are sensitive to ambient light which disturbs the captured images leading to a low detection rate and therefore limiting the interaction capabilities in the wild \cite{Fuhl2016}.

Our LFI-based gaze gesture sensor concept combines the advantages of non-intrusive integration of VOG systems with the advantage of directly capturing the relative velocity of the eye with an outstanding sampling rate and a low power consumption. Due to the fact that the LFI sensor captures relative eye movements, it is ideally suited for a gaze gesture-based interaction concept similar to Drews et.al.. Furthermore, our sensor principle is robust against external illumination and is therefore capable of operating in the wild.

\subsection{Gaze gesture sensor based on laser feedback interferometry}

\begin{figure}[h]
	\centering
	\includegraphics[width=0.8\linewidth]{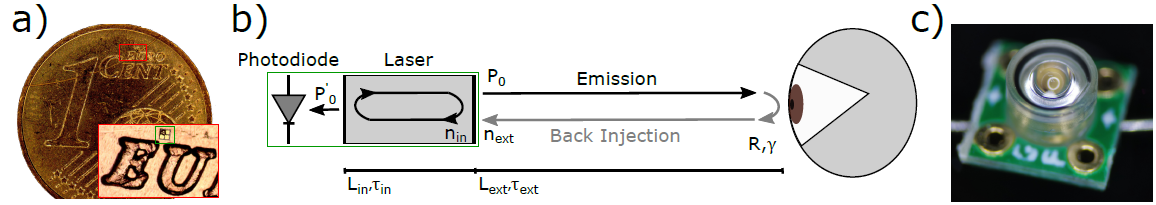}
	\caption{a) 160 $\mu$m x 180 $\mu$m VCSEL with integrated photodiode on a cent coin b) Coupled cavity model of a laser feedback interferometry sensor. The laser emits light which is scattered by the eye and back injected into the laser cavity. The photodiode monitors the laser power, which varies with changes in the feedback path. c) Encapsulated optical module including the LFI sensor and the beam shaping optics.}
	\label{B2im:coupled_cavity_model}
\end{figure}
\FloatBarrier
\label{B2sec:Gaze gesture sensor}
Laser feedback interferometry or self-mixing interferometry (SMI) is a widely known interferometry measurement method. It is used in the industry as well as in laboratory environments to measure displacement and velocity of solid targets, as well as fluids and distance. Due to the high distance and velocity resolution, it is also widely used in vibrometry applications \cite{Giuliani2002}. 

\subsubsection{Sensing principle}
\label{B2subsec:Sensing principle}
\Cref{B2im:coupled_cavity_model} b) shows the coupled cavity model to introduce the basic sensing principle of LFI sensors. A laser with an optical output power $P_0$ emits a coherent laser beam towards the surface of the eye. The laser beam hits the eye under an angle of incidence $\gamma$, is attenuated by volume scattering effects and absorption described by a reflectivity $R$ and is back injected into the laser. $\tau_{ext}$ denotes the time the laser beam requires to travel over the distance $L_{ext}$ towards the eye. $\tau_{ext}$ is dependent on the speed of light $c_0$ and the external refraction index $n_{ext}$ of the external medium \cite{Taimre:15}.

The back injected light interferes with the local oscillating field, which is often referred to as self-mixing interference, resulting in a modulation of the optical power
\begin{equation}
\label{B2equ:modulated_power}
	P_0' = P_0 \left( 1 + m \cdot \cos \left(\phi_{fb}\right)\right).
\end{equation} 
The varying feedback power $P_0'$ is dependent on the optical power $P_0$, the modulation depth $m$ and a varying phase $\phi_{fb}$ of the backscattered light field. A small fraction of the varying feedback power $P_0'$ is measured by a photodiode, which is integrated into the distributed Bragg reflector (DBR) of the laser cavity itself \cite{Taimre:15}. 

To understand the link of the phase $\phi_{fb}$ to our observation goal (distance and velocity), we consider the rate equations introduced by \cite{1070479}. A solution of the rate equation is the excess phase equation 
\begin{equation}
\label{B2equ:excess_phase_equation}
	\phi_{fb} - \phi_s + C \sin\left(\phi_{fb} + \arctan \left(\alpha\right)\right) = 0.
\end{equation}
The feedback phase is expressed as a function of the signal phase $\phi_s$, Acket's feedback parameter $C$ and Henry's line width enhancement factor $\alpha$. Considering operation of the LFI sensor in the weak feedback regime ($C<1$) and a constant line width enhancement factor $\alpha$, \Cref{B2equ:excess_phase_equation} lead to a single solution \cite{Taimre:15} and $\phi_{fb}$ is, therefore, only dependent on $\phi_s$, leading to 
\begin{equation}
\label{B2equ:phase_stimuli}
	\phi_s = \frac{4 \pi n_{ext} L_{ext}}{\lambda}
\end{equation}
with $\lambda$ describing the wavelength of the laser. Considering $n_{ext} \approx 1 $ is constant due to the operation of the sensor in free space, only changes in $\lambda$ and $L_{ext}$ lead to a varying phase $\phi_s$ and , consequently, to a varying phase $\phi_{fb}$. This results, with respect to \Cref{B2equ:modulated_power}, in a modulation of the optical power which is measured by the photodiode. Changes in the wavelength $\lambda$ occur by a modulation of the laser drive current. That leads to a periodic heating and cooling of the resonator and, thus, to a periodic change of the cavity length. The  variation of the cavity length leads to a periodic modulation of the wavelength, which allows for continuous measurement of the distance according to \Cref{B2equ:distance_equation}.

To distinguish between both effects, we compute the partial derivative of \Cref{B2equ:phase_stimuli} with respect to time, which leads to
\begin{equation}
\label{B2equ:distance_equation}
	f_0 = \left. \frac{2 L_{ext}}{\lambda^2} \frac{d\lambda}{dI} \frac{dI}{dt}\right|_{L_{ext} = const.}
\end{equation}
and
\begin{equation}
\label{B2equ:doppler_equation}
	f_d = \left. \frac{2 v_{ext} cos\left(\gamma \right)}{\lambda}\right|_{\lambda = const.}.
\end{equation}

Considering a known $\frac{d\lambda}{dI}$, which is a static process parameter of the laser, and a controlled current modulation slope $\frac{dI}{dt}$, the distance to the eye can be calculated by extracting the so called beat frequency $f_0$ by applying a fast Fourier transform (FFT) to the measured varying optical power and rearranging \Cref{B2equ:distance_equation} with respect to $L_{ext}$.

Movements of the eye ($\frac{d L_{ext}}{dt} = v_{ext}$) lead to a shift of the beat frequency $f_0$ by the so called Doppler frequency $f_d$. With a known angle of incidence $\gamma$ and a measured Doppler frequency, \Cref{B2equ:doppler_equation} can be rearranged with respect to $v_{ext}$ to obtain the surface velocity of the eye. 

In order to separate $f_0$ and $f_d$ and, thus, simultaneously measure the distance and velocity of the eye, a triangular modulation similar to frequency modulated continuous wave (FMCW) radar is applied to the drive current of the laser \cite{10.1117/12.775131}. By separating the up- and down ramp signals into two segments and applying an FFT on each segment, an $f_{up}$ and an $f_{down}$ frequency is captured. $f_0$ and $f_d$ are obtained from these measurements by
\begin{equation}
\label{B2equ:distance_triangular}
	f_0 = \frac{f_{up} + f_{down}}{2}
\end{equation}   
and 
\begin{equation}
\label{B2equ:doppler_triangular}
	f_d = \frac{f_{up} - f_{down}}{2}
\end{equation}
respectively. Recalling \Cref{B2equ:doppler_triangular}, the triangle modulation allows to extract the direction of the velocity as well. The modulation frequency of the triangle signal therefore corresponds to the sample frequency of the LFI sensor. Under this consideration, the upper sample frequency of the LFI sensor is determined by thermal time constants of the laser cavity, which is for spacial confined semiconductor VCSELs in the range of 100\,kHz \cite{Michalzik2013}.

\subsubsection{Sensor concept}
\label{B2subsec:Sensor concept} 
\begin{figure}[b]
	\centering
	\includegraphics[width=0.5\linewidth]{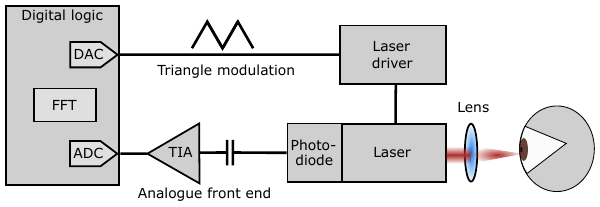}
	\caption{System diagram of the whole LFI sensor including the sensing element, the modulation circuit and the data processing element \cite{Taimre:15}.}
	\label{B2im:system_diagram}
\end{figure}  
\Cref{B2im:system_diagram} shows a system diagram of the whole LFI sensor. The output of the photodiode is AC coupled to an transimpedance amplifier (TIA) to reject the DC component of the varying optical output power and only measure the  cosine part of \Cref{B2equ:modulated_power}.

Inside the digital logic the output of the TIA is sampled by an analogue digital converter (ADC). The sampled data is synchronized with the up- and down ramp of the triangle signal synthesized by an internal digital to analogue converter (DAC) and fed to the laser driver. Afterwards, an FFT is applied to the two signal segments to obtain $f_{up}$ and $f_{down}$ and use \Cref{B2equ:distance_triangular} and \Cref{B2equ:doppler_triangular} to calculate $f_0$ and $f_d$, respectively.

\begin{figure}[h]
	\centering
	\includegraphics[width=0.4\linewidth]{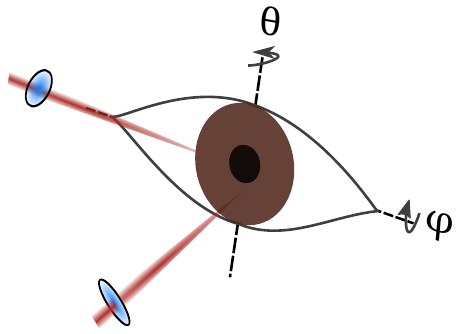}
	\caption{Positioning of the LFI sensors with respect to the rotational axes of the human eye.}
	\label{B2im:positioning_of_sensors}
\end{figure}

The LFI sensor measures surface velocity as a dot product between the laser normal and the surface velocity vector of the moving eye at the intersection point between laser and eyeball. If the eye's rotational axis and the laser beam are aligned, a simplified description by the angle of incidence $\gamma$ as described in \Cref{B2equ:doppler_equation} is possible. Therefore, two LFI sensors are required to measure both, the horizontal movement around the $\theta$-axis and vertical movement around the $\phi$-axis. \Cref{B2im:positioning_of_sensors} shows the positioning and the laser beam directions of two LFI sensors to comply with these requirements. 
\FloatBarrier
From an integration point of view, the sensor for vertical movements can be integrated into the AR glasse's frame below the spectacle and the sensor for horizontal movements can be integrated into the AR glasses frame temple. The size of a single LFI sensor is mainly determined by the diameter of the lens in front of the laser chip, shown in \Cref{B2im:coupled_cavity_model} c) is in the range of 1\,mm - 2\,mm.
\begin{figure}[h]
	\centering
	\includegraphics[width=0.5\linewidth]{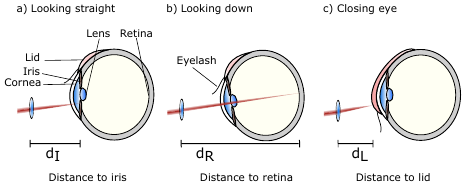}
	\caption{Different movement related position changes of the eye and the lid resulting in a change in distance measurement due to the geometry and scaffold of the eye.}
	\label{B2im:distance_features_eye}
\end{figure}
\FloatBarrier
\subsubsection{Sensor features on the eye} 
\label{B2subsec:Sensor features on the eye}

Based on \Cref{B2equ:distance_equation} and \Cref{B2equ:doppler_equation} and the positioning of the LFI sensors shown in \Cref{B2im:positioning_of_sensors}, four features are measured by the sensors. For each rotational axis of the eye, the surface velocity and the distance are measured resulting in $v_{\theta}$ and $v_{\phi}$ as velocity related features and $d_{\theta}$ and $d_{\phi}$ as position related features.

\Cref{B2im:distance_features_eye} shows a sectional view of the eye and a fixed LFI sensor position for three different positions of the eye and the lid. In \Cref{B2im:distance_features_eye} a), the gaze is directed straight ahead. The laser beam of the LFI sensor for vertical rotations penetrates the cornea and backscattering occurs at the iris. For this setup, the LFI sensor measures the distance between sensor and iris $d_I$.

In \Cref{B2im:distance_features_eye} b), the eye is slightly rotated along the vertical direction downwards and the laser beam penetrates the cornea and the lens and is backscattered from the retina. The measured distance in this case is the distance between sensor and retina $d_R$.

In \Cref{B2im:distance_features_eye} c), the eye is directed straight forward but the lid is closed by a blink. In this case, the distance $d_L$ between sensor and lid is measured. By subtracting $d_I$ and $d_L$, the thickness of the lid can be calculated, which is around 4\,mm, and by subtracting $d_R$ and $d_L$ the diameter of the eye is approximately calculated, which is around 24\,mm \cite{gross2008human}. 
\subsubsection{Gaze gesture algorithm} 
\label{B2sec:gaze_gesture_algorithm}
Similar to related works by Drews et. al.\cite{Drews2007}   Bulling et. al. \cite{10.5555/1735835.1735842} and Findlin et.al. \cite{10.1007/978-3-030-20521-8_27}, the gaze gesture algorithm is based on a gaze gesture protocol which encodes atomic movements of the eye into gaze gestures denoted by symbols. 

\Cref{B2im:gesture_alphabet} a) shows a graphical description of our gaze gesture protocol.  We define four basic atomic eye movements which are annotated by small letters. For the horizontal axis, movements to the left \textit{l} and to the right \textit{r}  around $\theta$ are possible. In addition, eye movements in the vertical axis around $\phi$ are denoted by \textit{u} and \textit{d} for up and down movements of the eye, respectively. In addition to movements of the eye, blinks (\textit{b}) are considered an additional atomic movement. In order to cover all types of eye movements, fixations of the eye, as well as slow movements, are described by an additional symbol \textit{n} for non-movements. The entire set of symbols $S$ thus consists of six unique symbols $S=\{l,r,u,d,b,n\}$.
\subsubsection{Feature extraction} 
\label{B2subsec:Feature extraction}
The novel sensor concept described in \Cref{B2subsec:Sensor concept} allows for the extraction of a unique set of features from the human eye with a high sampling rate. A major advantage over the state of the art is the use of the distance measurement between the glasse's frame and the eye as an additional feature, as discussed in \Cref{B2subsec:Sensor features on the eye}.
\begin{figure}[h]
	\centering
	\includegraphics[width=0.5\linewidth]{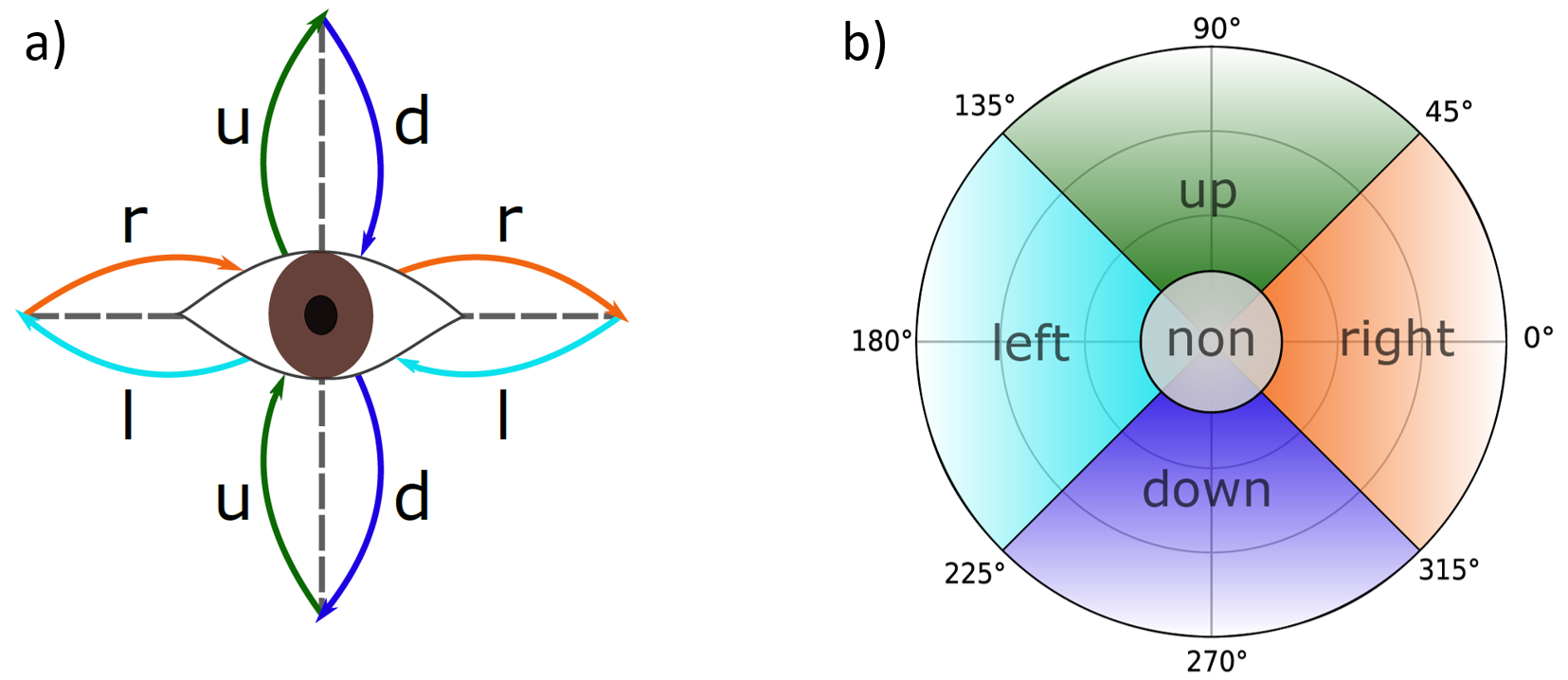}
	\caption{ a) Graphical description of the proposed gaze gesture alphabet. The small letters denote atomic unidirectional movements of the eye. b) Gaze gesture velocity feature space shown in a polar coordinate system. }
	\label{B2im:gesture_alphabet}
	\label{B2im:polar_feature_space}
\end{figure}
\FloatBarrier
To obtain unique features from the measured velocities, $v_\theta$ and $v_\phi$ are treated as velocity vector components. This velocity vector can be represented in polar coordinate space by an angle 
\begin{equation}
\label{B2equ:angle_feature}
	\epsilon = \arctan(\frac{v_\theta}{v_\phi})
\end{equation}
and a vector length 
\begin{equation}
\label{B2equ:velocity_feature}
	v = \sqrt{v_\theta^2 + v_\phi^2}
\end{equation}
which lead us to two dimensional feature vector $F$ with $\epsilon$ and $v$ as gaze symbol features. \Cref{B2im:polar_feature_space} b) shows the feature space, which is spanned by $\epsilon$ and $v$, in polar coordinate space. 

The grey area in the centre covers the area of fixations and slow eye movements e.g. drift or tremor. The size of this area is independent on the angle $\epsilon$ and is, therefore, only determined by a certain absolute velocity threshold $v_n$ for $v$. The sensitivity and robustness of the gaze gesture algorithm can be controlled by varying $v_n$. The other areas in the feature space are each associated with exactly one symbol of the symbol set $S$. 

The main advantage of this representation is the wide range of allowed angles of 90$^\circ$ per direction, leading to a robust gaze symbol detection and an easy execution of gaze gestures for the user. In addition, the use of speed as a feature for extracting gaze symbols allows atomic eye movements to be performed individually without exceeding a certain angle of eye's rotation.

Taking \Cref{B2equ:doppler_equation} into account, it appears that the LFI sensors are capable of measuring any kind of velocity at the eye. This includes blinking, which would be misclassified using the approach previously described as an opening or closing movement of the eye due to the up and down movement introduced by the moving eyelid. To avoid this misclassification, the distance measurement $d_\theta$ is evaluated for differentiation, as already described in \Cref{B2subsec:Sensor features on the eye}.  By adding the distance measurement $d_\theta$ as a third feature to $F$, the feature vector describes a point in cylindrical coordinate denoted by $F=\{\epsilon, v, d_\theta\}$.
\subsubsection{Gaze symbol classification} 
\label{B2subsec:Gaze symbol classification}
The first step of the classification process is to map a feature vector $F$ measured by the LFI sensors to a gaze symbol from the gaze symbol set $S$. For this purpose, a decision tree, which describes the feature space shown in \Cref{B2im:polar_feature_space} b) and the additional distance information $d_\theta$, is used. The main advantage of this single sample classification approach over other classification approaches, e.g. by Bulling et. al. \cite{10.5555/1735835.1735842}, is the invariance with respect to time. This allows for a robust classification, which is insensitive to sensor drift, as is the case with EOG sensors \cite{majaranta2014eye}. Furthermore, this classification approach does not require the detection of a movement sequence in the input sensor signal stream to extract significant movements which belong to a gaze symbol as proposed by Findling et. al. \cite{10.1007/978-3-030-20521-8_27}.
\subsubsection{Gaze symbol preprocessing} 
\label{B2subsec:Gaze symbol preprocessing}
\begin{figure}[h]
	\centering
	\includegraphics[width=0.4\linewidth]{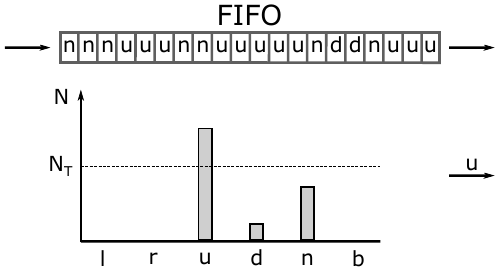}
	\caption{Gaze gesture symbols accumulated in a FIFO memory.}
	\label{B2im:FIFO_Speicher}
\end{figure}

One drawback of the single sample classification approach is that even slight movements of the eye in the wrong direction lead to a different gaze symbol. To overcome this limitation and increase robustness of the gaze gesture recognition, the gaze symbol stream is preprocessed. \Cref{B2im:FIFO_Speicher} shows the preprocessing of the gaze symbol stream.
\FloatBarrier
A first in first out (FIFO) buffer with length $N$ is filled with incoming gaze  symbols. In addition, for every symbol in $S$, a corresponding bin is created. These bins contain the number of associated symbols from the FIFO. If the number of symbols in a bin exceeds a certain threshold $N_T$, the corresponding gaze gesture symbol is set as an active symbol until $N_T$ is undershot. The FIFO is initialized at beginning of the algorithm with \textit{n} as default symbol to always output an active symbol.  
\subsubsection{Gaze gesture recognition} 
\label{B2subsec:Gaze gesture recognition}
\begin{figure}[h]
	 \centering
	\includegraphics[width=0.25\linewidth]{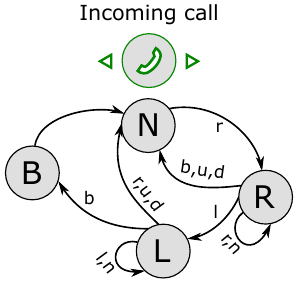}
	\caption{Description of a gaze gesture using a FSM to accept a incoming call displayed in the AR glasses user interface.}
	\label{B2im:Incoming_call}
\end{figure}
\FloatBarrier
Similar to Drews. et. al. \cite{Drews2007}, we define a gaze gesture as a sequence of atomic movements of the eye described by the corresponding gaze gesture symbols. To describe a gaze gesture, a finite state machine (FSM) is used. This allows for a flexible design of custom gaze gestures. To manipulate the state of the FSM, the preprocessed gaze symbol stream is used.

\Cref{B2im:Incoming_call} shows an FSM with a gaze symbol sequence $[r,l,b]$ as gaze gesture to accept a incoming call. When the state $R$ is reached, a timer is started with a given time $t_0$ to reset the FSM to its initial state $N$ if a timeout is reported before reaching the final state $B$. This increases the robustness and reduces the sensitivity to unintended interactions. Furthermore, a higher robustness of gaze gesture recognition can be achieved by increasing the length of the gaze symbol sequence of a gaze gesture. 

The transition to the final state $B$ is issued as soon as the blink gaze symbol \textit{b} is detected in the preprocessed gaze symbol stream, which occurs as soon as enough \textit{b} symbols have entered the FIFO and $N_T$ is exceeded. This allows for the execution of the gaze gesture command before the blink is even finished, resulting in a negative latency between finishing the gesture and execution of the command.

Due to the fact that no sliding window or time and power consuming transformation of time series data, as proposed by Bulling et. al. \cite{5444879}, is used to recognize the gaze gesture, a low power gaze gesture recognition for smart glasses is achieved.

\subsection{Evaluation}
\label{B2sec:Evaluation}
\begin{figure}[h]
		\centering
	\includegraphics[width=0.5\linewidth]{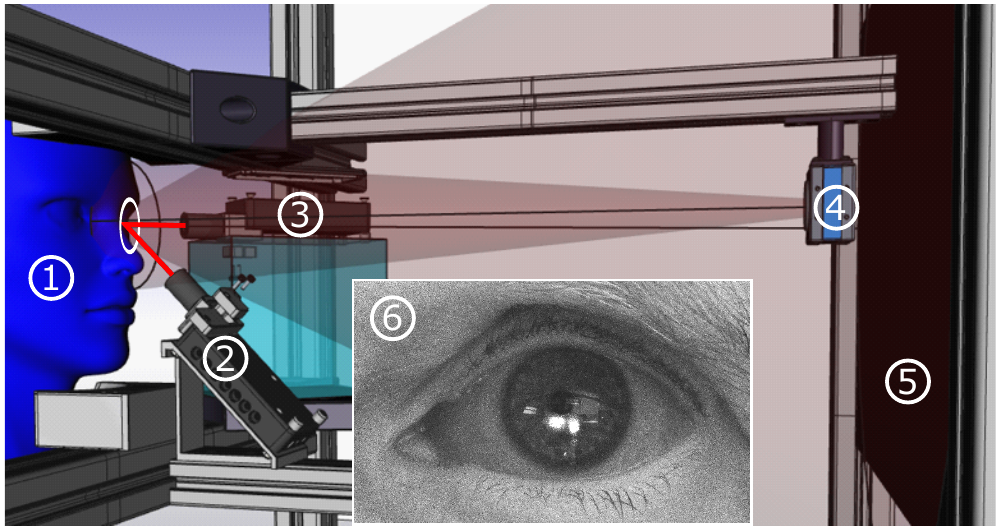}
	\caption{CAD sketch of the laboratory setup to validate the proposed LFI gaze gesture sensor approach. The grey scale image shows the eye of a subject with the two laser spots on the iris captured by the reference camera.}
	\label{B2im:Lab_setup}
\end{figure}
\FloatBarrier
To validate the proposed features of the human eye measured by the LFI sensors and evaluate the  gaze gesture recognition algorithm, a laboratory setup is used, as shown in \Cref{B2im:Lab_setup}. 

The subject \circled{1} is placed in front of a monitor \circled{5} which is used to create stimuli for eye movements. Its head is fixated by a head- and chin rest to suppress head movements, which would lead to a faulty velocity measurement. The LFI sensor at position \circled{2} is responsible for measurements of $v_\phi$ and $d_\phi$ while the LFI sensor at position \circled{3} is responsible for measurements of $v_\theta$ and $d_\theta$. An IR camera \circled{4} captures a video of the subject during the experiments. A sub Figure \circled{6} shows a frame captured by the camera showing the focused laser spots of the two IR lasers on the iris of a subject as white spots. The lasers are not aligned to hit the eye at the same position, as already described in \Cref{B2subsec:Sensor concept}.

\begin{table}[h!]
	\caption{Properties of the LFI sensors of the laboratory setup.}
	\label{B2tab:eye_tracking_params}
		\centering
	\begin{tabular}{ccccc}
		
		LFI Sensor& $d \lambda / dI$ &$\lambda$	&$\gamma$  &$P_0$  \\ \hline
		
		\circled{2} &	 0.406\,nm/mA &848\,nm &~45$^\circ$  &390\,$\mu W$ \\
		
		\circled{3} &	 0.396\,nm/mA &856\,nm  &~45$^\circ$  &390\,$\mu W$   \\
	\end{tabular}
	
\end{table}

Considering the sensor system diagram shown in \Cref{B2im:system_diagram}, the laser together with the laser driver and the lens are integrated into the black housing shown in \Cref{B2im:Lab_setup}. The triangular modulation signals are synthesized by an external waveform generator and the photodiode signal of the sensors is sampled by an oscilloscope together with the synthesized triangular modulation signal to extract $f_{up}$ and $f_{down}$, which are required to calculate $f_d$ and $f_0$ by \Cref{B2equ:distance_triangular} and \Cref{B2equ:doppler_triangular}, respectively. Afterwards the distance- and the velocity are calculated by \Cref{B2equ:distance_equation} and \Cref{B2equ:doppler_equation} with the known laser parameters shown in \Cref{B2tab:eye_tracking_params}. 

With the low optical power of the IR VCSELs, the mechanical setup of chin and head rest and the positioning of the lasers in the laboratory setup, a class 1 laser system according to IEC 60825-1 is achieved and, therefore, the experiments do not pose any medical hazard to the subject's eye.

The power consumption of the proposed gaze gesture sensor is roughly 140\,mW using of the shelf components. With optimized logic blocks and a subsampling scheme enabled by a higher integration using a custom application specific integrated circuit (ASIC), a further power reductions to 30\,mW is expected. A comparable power consumption of 150\,mW is reported by Sarkar et.al. \cite{7181058} for state of the art VOG eye tracking sensors, however excluding image processing. Compared to the EOG approach of Bulling et al. \cite{10.5555/1735835.1735842}, a much lower power consumption can be achieved.

\subsubsection{Feature validation}
\label{B2subsec:Feature validation}
\begin{figure}[h]
		\centering
	\includegraphics[width=0.6\linewidth]{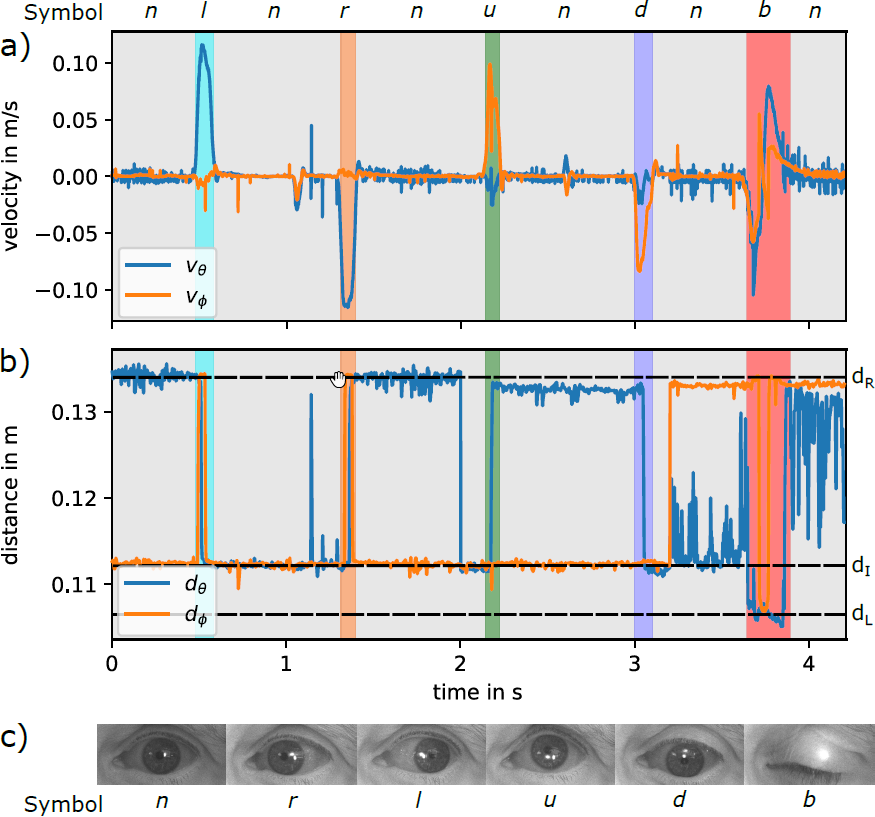}
	\caption{Data captured with the laboratory setup to evaluate the proposed features of the eye. a) shows the measured velocity, b) shows the measured distance and c) shows eye movements captured with the IR camera and the corresponding gaze symbol. The background colour refers to the corresponding atomic movement of the eye.}
	\label{B2im:evaluation_features}
\end{figure}

To validate the proposed features of the human eye from \Cref{B2subsec:Sensor features on the eye}, the laboratory setup is used. \Cref{B2im:evaluation_features} shows the measured data of the two LFI sensors from a subject performing the gaze symbol sequence $[l,r,u,d,b]$ .  
\FloatBarrier
The LFI sensors are modulated to achieve an update rate of 1\.kHz for distance and velocity measurement. \Cref{B2im:evaluation_features} a) shows the measured velocity and \Cref{B2im:evaluation_features} b) shows the measured distance. The movement threshold $v_n$ is set to 0.02\,m/s to distinguish between atomic eye movements and non-relevant movements of the eye. In the distance plot we added three distance lines corresponding to the distances described by \Cref{B2im:distance_features_eye} a) - c). The measured distance between $d_R$ (retina) and $d_I$ (iris) is 23,36\,mm and the measured distance between $d_I$ (iris) and $d_L$ (lid) is 5,01\,mm, which both correspond to the known anatomy of the human eye. The distance measurement shows that the iris of the eye scatters IR light and the cornea and the lens of the eye let IR light pass through without significant reflection and backscattering only occurs on the retina. 

In \Cref{B2im:evaluation_features} c), the images captured by the IR reference camera show the laser spots on the subject's eye for different atomic eye movements described by the corresponding gaze symbol. Similar, the background colours of the plots in a) and b) refer to the corresponding atomic eye movements.  



\subsubsection{Gaze gesture symbol classification}
To validate the proposed gaze gesture symbol classifier described in \Cref{B2subsec:Gaze symbol classification}, a set of gestures is recorded using the laboratory setup as described in \Cref{B2sec:Evaluation}. Two male subjects with blue and brown eyes were instructed to perform 18 eye movement gestures and 9 blinks each. To capture natural trajectories of the eye, during the execution of the gaze gestures no visual stimuli were used to guide the gaze of the subjects. 

In a first step, the ground truth is manually annotated utilizing the IR camera images to the measured data with the corresponding gaze symbols, similar to related work by Santini et. al. \cite{Santini2016}. To distinguish between movements and non-movements, the velocity threshold $v_n = 0.02\,ms$ is applied. Considering Emsley’s reduced eye model (reye = 11.11mm) this equals to a velocity threshold of 103.14 $^\circ$/s, close to the suggested velocity threshold used in the velocity threshold Identification algorithm (I-VT). Afterwards, we extract the features from the measured data according to \Cref{B2subsec:Feature extraction} and use the proposed gaze symbol classification algorithm to classify atomic eye movements. This allows us to evaluate the  classification approach on the sample-level. In addition, we treat the multiclass classification problem for evaluation purpose as a binary one-vs-all classification problem. This evaluation approach is commonly used in the literature e.g. Startsev et.al. \cite{startsev20191d} or, Hoppe et. al. \cite{hoppe2016endtoend} and allows us to compute the F1-score as evaluation metric. \Cref{B2tab:gaze_gestures} shows the evaluation results.

%
%
%
%
%
%

\begin{table}[h]
	\caption{Results of the gaze gesture symbol classification algorithm.}
	\label{B2tab:gaze_gestures}
	\centering
	\begin{tabular}{cccccc}
			
		Gaze symbol& Gestures  & Samples & Mean duration  & F1-Score  \\ \hline
		
		\textit{l} & 36 & 3783 & 105.08\,ms & 0.949\\
		
		\textit{r} & 36 & 3505 & 97.36\,ms& 0.955  \\
		
		\textit{u} & 36 & 3229 & 89.69\,ms & 0.891 \\
		
		\textit{d} & 36 & 3624 & 100.67\,ms & 0.908   \\
		
		\textit{b} & 18 & 3063 & 170.17\,ms & 0.969  \\
	\end{tabular}

\end{table}

The overall high F1-scores, shows high precision and recall capability of the proposed algorithm and underlines the robustness of the features acquired with our novel LFI gaze gesture sensor.
\begin{figure}[h]
	 \centering
	\includegraphics[width=0.5\linewidth]{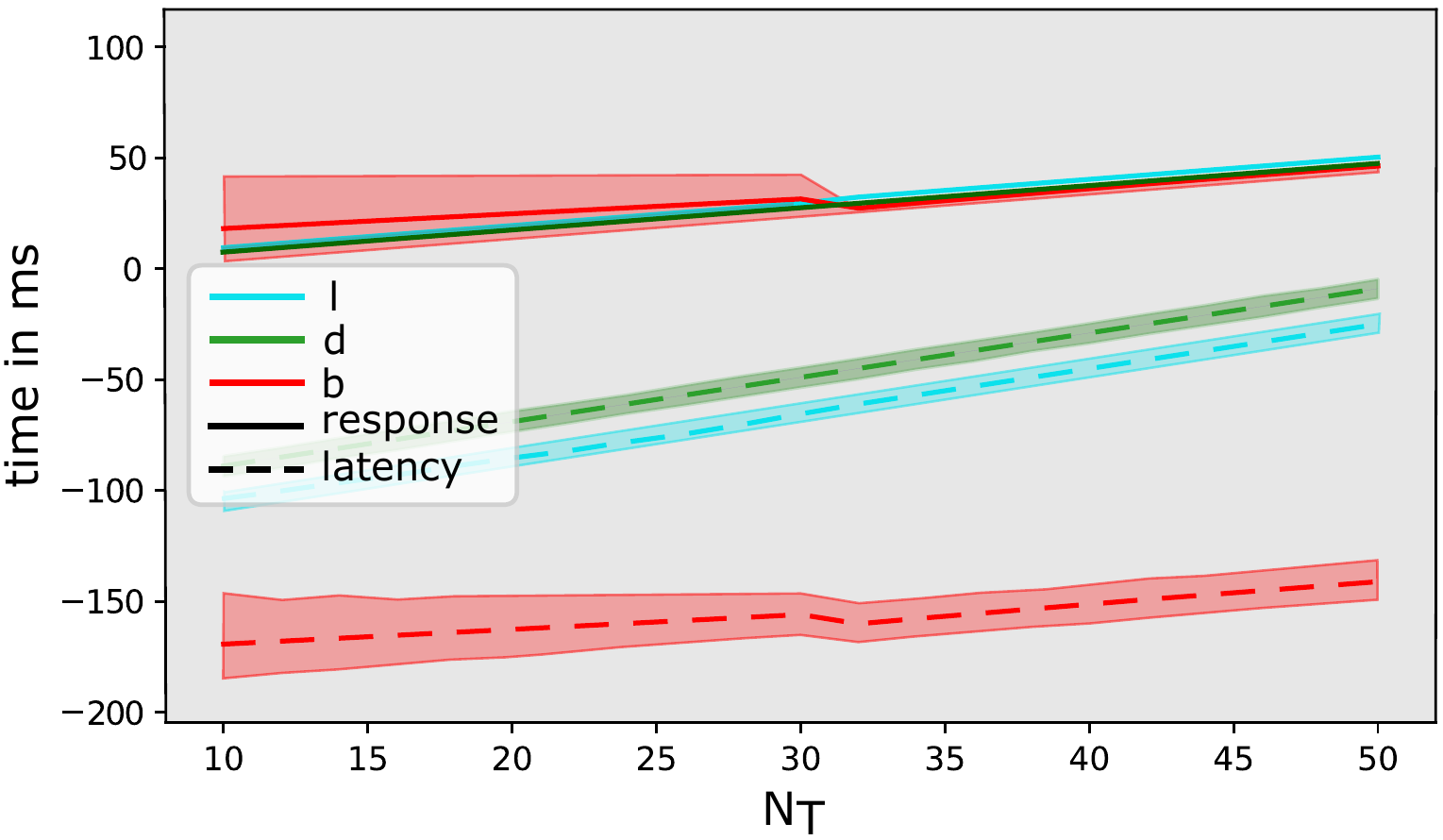}
	\caption{Response time $t_{res}$ (solid line) and latency $t_{lat}$ (dashed line) for the left-, up- and blink atomic eye movements evaluated for different thresholds $N_T$. The light shaded regions around the lines represent the standard deviation of all atomic eye movements captured during the experimental validation.}
	\label{B2im:response_and_latency}
\end{figure}
\FloatBarrier
\subsubsection{Gaze symbol preprocessing validation}
 To increase the robustness  of the recognition algorithm, the gaze symbols are preprocessed according to the proposed method in \Cref{B2subsec:Gaze symbol classification}. Based on the results shown in \Cref{B2tab:gaze_gestures}, a FIFO size $N$ of 100 elements is used, which represents an 100\,ms window. This preprocessing smooths the incoming gaze symbol stream to increase the robustness, but also introduces a response and latency time to the gaze gesture symbol output, which is used to control the FSM. We define the response time $t_{res}$ as time between the start of a gesture ($v$ exceeds $v_n$) and the first indication of a gaze gesture symbol in the output symbol stream ($N_T$ is exceeded). In addition the latency, $t_{lat}$ is defined as the time between the end of a gesture ($v$ falls below $v_n$) and the first indication of a gaze gesture symbol in the output symbol stream ($N_T$ is exceeded). \Cref{B2im:response_and_latency} shows the latency and response time for different thresholds $N_T$. A part of the gestures described in \Cref{B2tab:gaze_gestures} is used for the evaluation. The solid and dashed lines represent the mean response time and latency, respectively, while the light shaded regions around the lines represent the standard deviation. 

\Cref{B2im:response_and_latency} shows another significant advantage of the proposed gaze gesture algorithm over state of the art approaches. Beside the low response time, we achieve a negative latency, meaning that the proposed algorithm is able to recognize a single gaze gesture before the user has finished its execution. This allows for a seamless interaction with the smart glasses UI because the integrated control unit reacts before the user finishes the gesture input.  
\subsubsection{Gaze gesture recognition validation} 
\begin{figure}[t]
	\centering
	\includegraphics[width=0.5\linewidth]{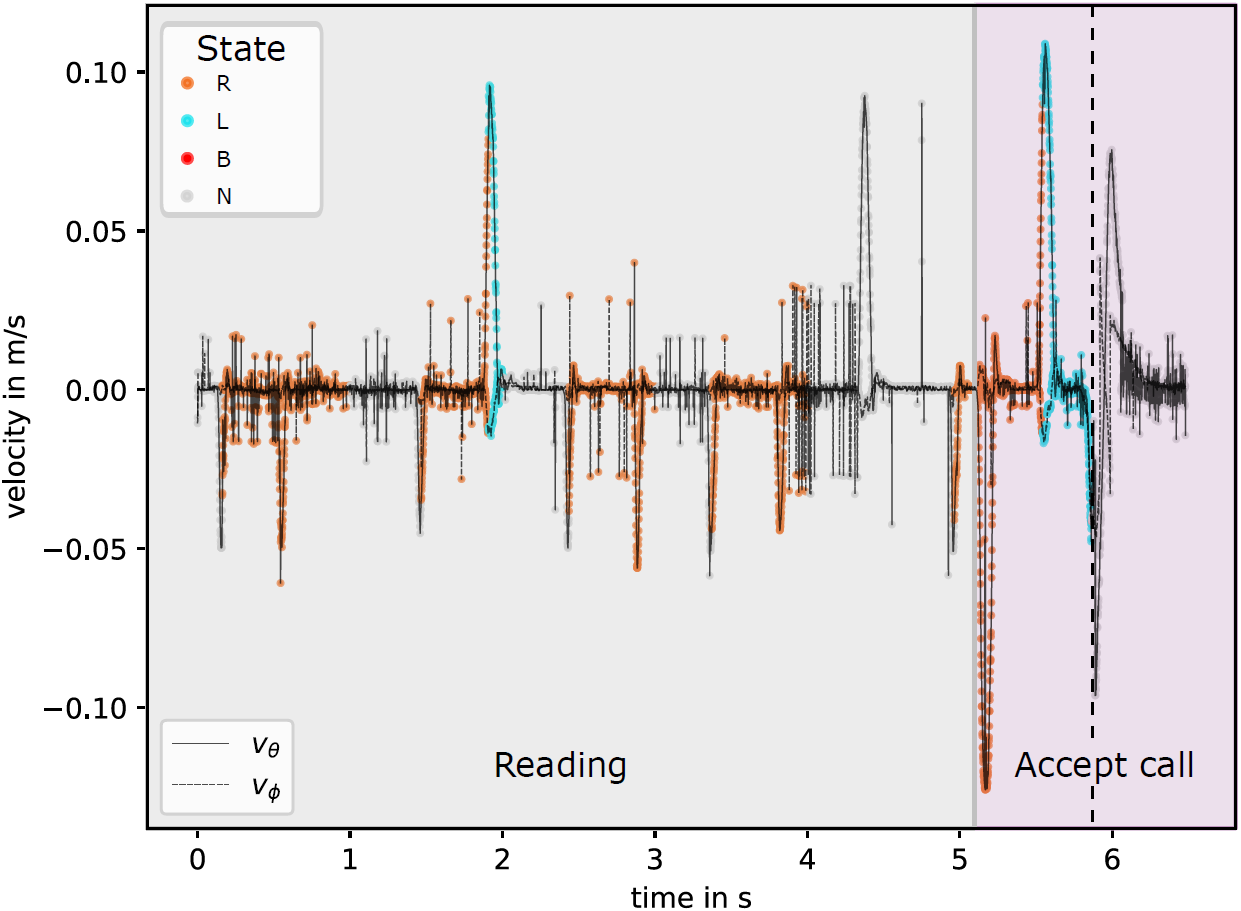}
	\caption{Application of the proposed gaze gesture recognition approach to an example user interaction. The colourized dots on the velocity trajectory denotes the corresponding state of the FSM described in \Cref{B2im:Incoming_call}. }
	\label{B2im:gaze_gesture_recognition}
\end{figure}

 Based on the example in \Cref{B2subsec:Gaze gesture recognition}, the proposed FSM based gaze gesture recognition is validated. A subject is reading an text until a incoming call appears. When the incoming call appears, the subject performs the gaze gesture as described in \Cref{B2im:Incoming_call}. \Cref{B2im:gaze_gesture_recognition} shows the measured velocities $v_\theta$ and  $v_\phi$ as solid and dashed lines respectively. Every data point is classified, preprocessed and fed into the FSM. The corresponding state of the FSM is annotated to the velocity trajectories. The black dashed vertical line indicates the time at which the gesture is successfully recognized. Similar to Drews et.al. \cite{Drews2007}, we set the timeout $t_0$ to one second.  

This example shows the robustness of the detection and preprocessing algorithm. Furthermore, the negative latency between gesture recognition and the completion of the gesture is illustrated. The blink gesture and, thus, the execution of the user interaction already starts while the eyelid is closing, which is described by the negative speed of $v_\theta$ and $v_\phi$. 
Especially during reading activities, the eye primarily makes small movements to the right when scanning over a line of text and larger movements to the left when jumping back to the beginning of a new line. In this experiment, the FSM reaches the state $L$ at the 2 seconds mark on the time axis. If a blink occurs at this position, a false gesture input is executed.

To address this issue and make this approach more robust against unintended activations, an incoming call can be classified as an event-based interaction and the FSM for user interaction detection is armed only when the event occurs. Furthermore, the time-out $t_0$ can be decreased to reset the FSM earlier or a more complex gesture consisting of a longer sequence of atomic eye movements can be used.

\subsection{Conclusion}
We present a novel low power gaze gesture sensor concept based on IR VCSELs and the LFI effect, which allows for seamless integration of gaze gesture sensors into next generation smart glasses. This sensor enables new gaze based interaction concepts like true hands-free interaction. In addition, the high sampling rate of the sensor, combined with our gaze gesture recognition algorithm, results in negative latency and ,thus ,ensures a fast response time and a high user experience.

Based on the promising results obtained with the laboratory setup, our next goal is to integrate the sensors into our head-mounted demonstrator. This will allow us to evaluate the gaze gesture sensor during everyday activities to minimize unwanted activations and investigate the robustness against glasses slippage.  

In addition the proposed LFI sensor opens options for further applications like fatigue detection based on blink speed and duration, recognition of user's emotions, classification ofthe user's mental health as well as early detection of eye diseases.

\newpage

\section{A CNN-based Human Activity Recognition System Combining a Laser Feedback Interferometry Eye Movement Sensor and an IMU for Context-aware Smart Glasses}
\label{APP:B3}
\subsection{Abstract}
Smart glasses are considered the next breakthrough in wearables. As the successor of smart watches and smart ear wear, they promise to extend reality by immersive embedding of content in the user's field of view. While advancements in display technology seems to fulfill this promises, interaction concepts are derived from established wearable concepts like touch interaction or voice interaction, preventing full immersion as they require the user to frequently interact with the glasses. To minimize interactions, we propose to add context-awareness to smart glasses through human activity recognition (HAR) by combining head- and eye movement features to recognize a wide range of activities.
To measure eye movements in unobtrusive way, we propose laser feedback interferometry (LFI) sensors. These tiny low power sensors are highly robust to ambient light. We combine LFI sensors and an IMU to collect eye and head movement features from 15 participants performing 7 cognitive and physical activities, leading to a unique data set. To recognize activities we propose a 1D-CNN  model and apply transfer learning to personalize the classification, leading to an outstanding macro-F1 score of 88.15\,\% which outperforms state of the art methods. Finally, we discuss the applicability of the proposed system in a smart glasses setup.

\subsection{Introduction}
\label{C1sec:Introduction}
Followed by smart watches and smart ear wear smart glasses are considered as the next breakthrough in the smart wearable domain because they are able to enhance human perception and embedding user interfaces seamlessly into the user's field of view (FOV). In certain situations, e.g. when driving, this highly integrated display causes distraction to the user as information e.g. notifications appear in the FOV. State of the art smart glasses try to adapt established user interface control concepts from the smartphone e.g. by adding tactile input modalities like capacitive sliders, buttons or touch sensors to the glasses frame \cite{GoogleGlass}, \cite{Espon2021}  or voice based control \cite{amazon2019}. Furthermore, human gaze is investigated as a natural input modality for controlling user interfaces with smart glasses \cite{Meyer2021},\cite{Drews2007}.

All of these concepts require active interaction with the glasses. Furthermore, for voice based interaction the user's privacy is not covered, what might reduce the social acceptance of smart glasses \cite{8368051}. To address this drawback, we propose a pervasive computing approach to control and adapt the user interface based on the user's activities. This leads to a reduction of required user interaction and enables unconscious non-intrusive control of the user interface of smart glasses. 

A major advantage for HAR with smart glasses is that the glasses provide access to the human eyes and thus the human gaze to derive features for activity recognition. Combined with head positions or head movements, the result is a unique set of features that represents a wide range of human perception. This allows recognizing a rich set of human activities ranging from mainly cognitive activities like reading e.g. shown by Islam et. al. \cite{islam2020self} to activities of physical nature like walking \cite{BullingTutorial}.

Besides of these advantages, smart glasses also introduce challenges arising from the sensor integration due to space constraints as well as the limited available power for a continuous operation in a HAR task. While head movements can be easily captured by accelerometers and gyroscope sensors, which are available as highly integrated microelectromechanical systems (MEMS) with low power consumption and small sensor size, challenges arise for eye movement sensors. 

Eye movement sensors, also referred to as eye tracking sensors, are dominated by video oculography (VOG) sensors and electro oculography (EOG) sensors \cite{majaranta2014eye}. VOG sensors, e.g. the Pupil Invisible \cite{10.1145/3130971} rely on video cameras, which are integrated into the glasses frame or the frame temple. They capture video frames from the eye and use computer vision algorithms to extract the pupil contour and estimate the user's gaze. The main disadvantages of VOG sensors are the size of the camera sensors and the required optics as well as the required angle under which the sensor needs to be positioned to capture images of the eye. In addition, the computer vision algorithms add significantly to the overall power budget for VOG systems. Camera sensors itself are furthermore sensitive to varying external illumination by limited dynamic range which occurs during every day activities, reducing the pupil detection accuracy \cite{Fuhl2016}. EOG sensors measure the electrical potential of the electrical dipole between the retina and the cornea with electrodes placed on the user's skin to estimate the gaze vector. While EOG sensors are insensitive to external illumination and consume less power compared to VOG sensors, they cannot be integrated unobtrusively and contactless into glasses as they require attaching electrodes to the skin \cite{majaranta2014eye}.

To overcome these limitations and to be able to reliably measure eye movements to enable HAR for smart glasses, we introduce laser feedback interferometry (LFI) sensors. The LFI sensors have three main advantages compared to VOG and EOG sensors.

i) \textbf{Sensor integration} The sensor is based on a small vertical cavity surface emitting laser (VCSEL)  in the near infrared (IR) spectrum with a cavity-integrated photodiode, which allows for a space constrained integration into the glasses frame. Due to the invisible IR laser beam, the sensor is unobtrusive and contactless for the user.

ii) \textbf{Sensed features} The multimodal sensor is capable of measuring distance towards the eye as well as eye rotational velocity with an outstanding sample rate of up to 1\,kHz. From the distance information, characteristic blink patterns can be extracted, while the rotational velocity is directly related to eye activity without any further signal processing.

iii) \textbf{Ambient light robustness} The VCSEL with cavity-integrated photodiode enables to apply the LFI sensing principle, a self-coherent sensing scheme. This leads to a high sensitivity to its own radiation while suppressing most other radiation entering the laser cavity. This allows a robust operation in the presence of ambient radiation, as shown by Meyer et. al. \cite{Meyer2020e}.

To summarize the advantages of the LFI sensor compared to camera-based VOG systems, we like to highlight the low power consumption of the sensors and the ability to measure the rotational velocity without any further signal processing. In addition, the coherent sensing scheme of the LFI sensors allows for high robustness to ambient light, enabling robust operation in a wide range of lighting conditions from darkness to bright sunlight, as shown by Meyer et. al. \cite{Meyer2020e}. 

In this work, we propose a novel way to capture eye movements for HAR utilizing LFI sensors to enable a pervasive control of the user interface of upcoming smart glasses. We show that the combination of eye and head movements lead to a high activity recognition rate for a rich activity set which consists of seven different activity classes from physical as well as cognitive domains. To prove this, we conduct experiments and collect simultaneously LFI sensor data as well as data from an inertial measurement unit (IMU) consisting of an accelerometer as well as a gyroscope sensor from 15 participants performing seven different activities. We further design a one dimesional convolutional neural network (CNN) based classifier to recognize the activities. We investigate the effect of different sensing modalities as well as the effect of transfer learning on the classification accuracy and discuss the proposed sensors as well as the proposed classifier with respect to sensor integration and power consumption.

We highlight the contribution of this work as follows. First, we integrate a novel LFI based sensor modality to smart glasses to capture eye movements and combine them with an IMU to obtain a unique HAR data set. To the best of our knowledge, this is the first work of HAR systems, which combines LFI sensor modalities and IMU sensor modalities for smart glasses. Secondly, we implemented the activity recognition system into a smart glasses demonstrator. Experiments with 15 participants were conducted to evaluate the performance of the system. Using state of the art leave one participant out cross validation (LOPOCV) we obtain a macro F1-score of 88.15\,\%, which is outstanding for the chosen diverse activity set.

In the next section, we discuss the related work in body-worn HAR, eye-tracking based HAR and related work that combines head and eye tracking to perform HAR in a head-worn setup. In Section \ref{C1sec:Laser Feedback interferometry} we provide background knowledge of the LFI sensing principle as well as the features the LFI sensor is capable to capture from the human eye. Afterwards, in Section \ref{C1sec:Data collection} we introduce our data recording setup and describe the experimental design to generate our data set. In addition, we give an overview of the data processing and the obtained data set. In Section \ref{C1sec:Evaluation} we introduce our 1D-CNN model to classify the human activities and compare the classification performance with a baseline model. Furthermore, we show the effect of transfer learning as well as the influence of both sensor modalities on the classification accuracy for different activities. We conclude this work with a discussion on the applicability and the main challenges of the proposed system.

\subsection{Human activity recognition}
\label{sec:Related work}
HAR is a well established field of research which deals with concepts of inferring the current goals and behavior of one or multiple human individuals based on a series of observations \cite{DeepLearningModelsForHAR}. The main motivation for building and implementing a precise HAR system is its wide range of applicability. Domains such as active and assisted living (AAL), healthcare monitoring or surveillance can derive a clear benefit from employing HAR algorithms. What is more, within the last decade HAR found its way into consumer electronics with products such as Microsoft Kinect \cite{KinectWindows, KinectAzure} or Nintendo Switch \cite{NintendoSwitch} and also tapped into the consumer sports sector with smart footwear providing feedback during running or golfing \cite{Runtopia, Sensoria, Salted}. 

At present, there are two main approaches of how human activity data is collected: video-based systems and sensor-based systems \cite{DeepLearningForHAROverview}. In video-based systems cameras are installed to gain insights on human behavior from images and videos, typically employed for surveillance or recognizing activities of a group of people \cite{videoHARconv}. On the other hand, sensor-based systems rely on observations stemming from sensors either attached to an individual's body \cite{HARFromBodyWornAccData} or from ambient sensing devices such as RFID-tags \cite{HARRFID} installed within a participant's environment. 

In this work, we present a sensor-based HAR system and therefore limit our literature review to sensor-based HAR approaches.

\subsubsection{HAR Based on Body Worn Sensors}
\label{subsec:HAR based on body worn sensors}
Modern smart wearables from smart phones to smart ear wear are equipped with MEMS IMU sensors capable of measuring body motion in a ubiquitous and unobtrusive fashion. This intensified research has led to a multitude of publications in the area of sensor-based HAR. One of the most renown works was published by Kwapisz et al. \cite{HARAndroidAcc}. They collected an activity data set with the six physical activities \textit{walking}, \textit{jogging}, \textit{ascending stairs}, \textit{descending stairs}, \textit{sitting} and \textit{standing} from 29 participants. They sampled the smart phone's accelerometer with 20\,Hz, which the participants wore in their front trouser pocket. The raw data stream was split into windows of 10\,s duration with each window containing 200 samples. For each window they derived the statistical features \textit{mean}, \textit{standard deviation} and \textit{time between peaks}  and used them for training a decision tree classifier. They achieve an overall accuracy of 91\,\%. This high accuracies only drops for \textit{climbing stairs} as it frequently was misclassified as \textit{walk} \cite{HARAndroidAcc}. This effect is also known as inter-class similarity where the sensor features of two activities are not distinguishable by the classifier \cite{BullingTutorial}.

To increase classification performance, Wahl et. al. \cite{10.4108/eai.28-9-2015.2261470} integrated an IMU sensor, a light sensor as well as an heart rate sensor into a glasses frame and recorded data of nine participants during a full day. They derive 25 statistical features to classify nine activities where two mainly cognitive activities where present in their activity set. They reported a overall accuracy of 77\,\% utilizing a LOPOCV validation scheme.

In a more recent work by Hayashi et al. \cite{HARAudioAccApartment}  audio data was added as further sensor modality and captured together with smart phone's accelerometer data at a sample rate of 16\,kHz and 200\,Hz, respectively.

The participants were asked to pursue everyday occupation in a dedicated apartment. Data was recorded for 19 different participants and labelled to match 22 activity classes such as \textit{cooking}, \textit{eating}, \textit{reading}, \textit{sleeping}, etc. leading to a diverse activity set, with 12 physical activities, 4 cognitive activities and 6 mixed activities. Similarly to Kwapisz et al. \cite{HARAndroidAcc}, Hayashi et al. \cite{HARAudioAccApartment} made use of statistical features (\textit{mean}, \textit{variance}, \textit{entropy}, \textit{correlation}) which were extracted from non-overlapping sliding windows with a duration of one second each. These features were then used to train a deep neural network for activity classification. For training and predicting on data originating from the same participant, F1-scores of 80\,\% are reported. However, classification performance drops as soon as the classifier is reviewed using a LOPOCV scheme as different participants perform the same activity in different ways, leading to variation in statistical characteristics. This effect is also called intra-class heterogeneity \cite{BullingTutorial}.

To counteract intra-class heterogeneity, Hayashi et al. \cite{HARAudioAccApartment} followed the idea of transfer learning and retraining a subset of the deep neural network's parameters with a few samples from the left-out participant. This adaptation method reduces the accuracy degradation in the LOPOCV validation scheme and the F1-Score of 80\,\% is obtained.
\subsubsection{HAR Based on Eye Tracking}
\label{subsec:HAR based on eye movement data}
One of the first works utilizing eye-tracking for HAR was published by Bulling et al. in 2008 \cite{BullingReadingTransit}. They classify whether a participant was \textit{reading}, solely based on the participant's eye movements recorded with an EOG sensor. Within the collected data set of eight participants they achieved an overall recognition rate of 80.2\,\% by using a hidden Markov model as classifier.

In a follow-up study, they extended the activity set by office activities like \textit{reading}, \textit{copying}, \textit{handwriting}, \textit{watching a video} and \textit{web browsing}. For classification, they extracted statistical features from the raw EOG readings of eight participants such as \textit{mean}, \textit{variance} or \textit{maximum peak}. Applying a support vector machine (SVM), lead to a precision of 76 \%  across multiple participants \cite{5444879}. 

In 2013,  Bulling et al. \cite{Bulling2013HighLevelCues} stepped away from a discrete activity set and abstracted distinct activities with high level cues which group a set of activities. They choose \textit{spatial}, \textit{physical}, \textit{social} and \textit{cognitive} cues as high-level abstraction. Eye movements of four participants were recorded using an EOG sensor. Detected saccades were encoded into a string representation based on the saccade's direction. For each context and participant, a separate binary classifier (string kernel SVM) was trained. The reported results show a high F1 score of 91.21\,\% for the recognition of social interactions but less accurate performance when predicting a physical activity context (F1 score of 74.78\,\%).

Steil et al. \cite{BullingLDA} approached the challenge of HAR in an unsupervised way. 10 participants wore a VOG head-mounted eye-tracking sensor \cite{Kassner:2014:POS:2638728.2641695} throughout a full day of their ordinary life and were not restricted in terms of the activities to be performed. Saccadic direction, fixation duration and blink rate were encoded into strings and altogether form a bag-of-words representing an individual's visual behavior. This bag-of-words was passed to a Latent-Dirichlet-Allocation (LDA) topic model, which tried to find reoccurring topics, i.e. activities, within the visual data. With the help of an exemplary ground truth annotation of activity classes, detected topics could be assigned to an activity class. Steil et al. \cite{BullingLDA} reported the highest average F1-score for the activity of reading (74.75 \%) but mentioned that good recognition performance is highly dependent on the individual participant, the activity duration and the number of topics to be discovered by the LDA model. Nevertheless, unsupervised approaches have the advantage of being able to deal with an arbitrary set of activities.

Braunagel et al. \cite{DAR_Scanpath} investigated eye tracking based HAR in the context of conditionally autonomous driving. They recored the eye movements of 84 participants using a VOG sensor while they performed secondary tasks (\textit{reading}, \textit{video}, \textit{idle}) in a driving simulator. For classification, Braunagel et al. \cite{DAR_Scanpath} used visual scanpaths, i.e. they mapped the recorded gaze data to a string of symbols, with each symbol representing a certain region within the gaze space. The advantage of visual scanpaths compared to statistical feature extraction is that scanpaths preserve the temporal order of the gaze signal. The maximum reported average F1-score of 84\,\% supports this argument.

A very recent publication from Lan et al. \cite{GazeGraph} stepped away from using hand-crafted statistical features for HAR and leveraged the automatic feature learning capabilities of deep learning models. An individual's visual behavior was encoded into a spatial-temporal graph. Each node in the graph represents a gaze vector and has connecting edges to temporally adjacent gaze vectors. Edge weights were computed based on differences in angular orientation and euclidean distance of the connected nodes. The resulting weight matrix was passed to a CNN which learns meaningful features such that recognition performance is maximized. The gaze vectors of eight participants where captured by an VOG sensor from Pupil Labs while the participants perform six mainly cognitive activities (\textit{Browsing}, \textit{playing online games}, \textit{reading}, \textit{searching answers in a list}, \textit{watching video}, \textit{typing}). The reported F1-scores were most promising, 96\,\% on average over data of all participants, when considering gaze data within windows of 30 seconds in duration.
\subsubsection{HAR Combining Eye-Tracking and Body Worn Sensors}

Bulling et. al. \cite{10.1145/2134203.2134205} were the first to combine information from eye and head movements to classify \textit{reading} activities in an everyday environment, using an EOG sensor to extract eye movement features and a body-worn IMU sensor to extract body movements. Using a combined evaluation of body and eye movement features, they achieved 87.8\,\% performance with an SVM across 8 participants, while performance degraded by 24\,\% when head movement features were discarded.

Ishimaru et al. \cite{BullingGoogleGlasses} advances the initial study of Bulling et. al. \cite{10.1145/2134203.2134205} by a larger set of in total 5 activities from cognitive as well as physical domain. For data acquisition, eight participants wore the commercially available Google Glass and performed the activities \textit{talking}, \textit{watching}, \textit{solving} a mathematical problem, \textit{reading} and \textit{sawing}. Head motion was recorded using Google Glass' built-in IMU. Actual eye movement recording is not possible with Google Glass which is why Ishimaru et al. \cite{BullingGoogleGlasses} resorted to the data provided by the glasses' proximity sensor for blink detection. The recorded accelerometer data was converted into a motion feature by using the averaged variance across all axes. The mean blink frequency as well as the center of distribution of all recorded blink frequencies were used as eye motion features. Ishimaru et al. \cite{BullingGoogleGlasses} evaluated the performance for both sensor modalities and their combination by training a person-dependent decision tree. Results indicate that the combination of head motion and blink features yields and average F1-score of 82\,\% whereas restricting the features space to head or eye features resulted in F1-scores of 63  \% and 67  \% respectively.

In a later study Ishimaru et. al. replaced the Google Glass with commercial EOG glasses \cite{ActRecogWithCommercialEOGGLasses} leading to EOG signals as well as head acceleration signals as input modalities for their k-nearest neighbor (KNN) classifier. They collected an activity data set of four activities (\textit{typing}, \textit{eating}, \textit{reading} and \textit{talking}) from two participants and reported and overall accuracy of 70\,\%.

The analysis of related work shows that the area of HAR with body worn sensors and the area of HAR based on eye tracking are well established fields with a lot of previous work. Furthermore, it is shown that HAR with body worn sensors most likely emphazise the features of physical activities and HAR with eye tracking more likely emphazise the features of cognitive activities. Therefore, high recognition accuracies are present in both subfields. The combination of both sensor modalities in a head worn setup is only investigated by a few works of Ishimaru et al. where he was also mainly focusing on cognitive tasks. To capture eye movements he used an EOG sensor in his latest study, which has disadvantages in terms of product integration for future smart glasses.

In our work, we replace the EOG sensor with a promising LFI sensor and furthermore use state of the art method to improve the HAR accuracy in a balanced activity data set and show the classification robustness over a larger set of participants by applying the state of the art LOPOCV method.

\subsection{Laser Feedback interferometry}
\label{C1sec:Laser Feedback interferometry}
\begin{figure}[h]
	\centering
	\includegraphics[width=0.8\linewidth]{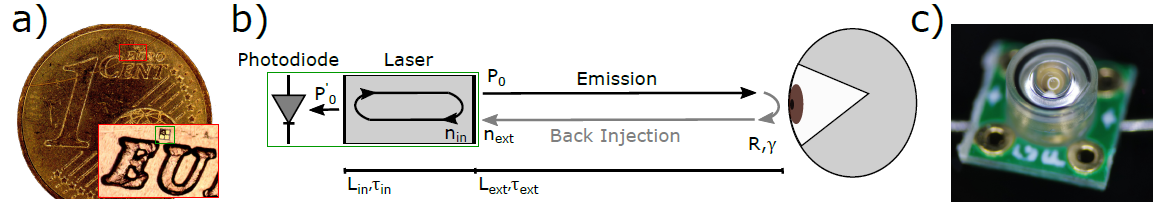}
	\caption{a) 160 $\mu$m x 180 $\mu$m VCSEL with integrated photodiode as sensing element on a cent coin b) Coupled cavity model of a laser feedback interferometry sensor. The laser emits light which is scattered by the eye and injected back into the laser cavity. The photodiode monitors the laser power, which varies with changes in the feedback path. c) Encapsulated optical module including the LFI sensor and the beam shaping optics used in the experiments \cite{Meyer2021}.}
	\label{C1im:coupled_cavity_model}
\end{figure}
\FloatBarrier
\label{C1sec:Gaze gesture sensor}
Laser feedback interferometry is a widely known interferometry measurement method \cite{Taimre:15}. It is used in the industry as well as in laboratory environments to measure displacement and velocity of solid targets, as well as fluids and distance. Due to the high distance and velocity resolution, it is also widely used in vibrometry applications \cite{Giuliani2002}. Recent works by Meyer et. al. \cite{meyer11788compact},\cite{Meyer2021} apply LFI sensors in a smart glasses setting to recognize gaze gestures and show the basic functionality of this sensing modality in a near eye setup. 

\subsubsection{Sensing Principle}
\label{C1subsec:Sensing principle}
Figure \ref{C1im:coupled_cavity_model} b) shows the coupled cavity model to introduce the basic sensing principle of LFI sensors. A laser with an unattenuated optical baseline output power $P_0$ emits a coherent laser beam towards the surface of the eye. The laser beam hits the eye under an angle of incidence $\gamma$, is attenuated by volume scattering effects and absorption described by a reflectivity $R$ and is injected back into the laser. $\tau_{ext}$ denotes the time the laser beam requires to travel over the distance $L_{ext}$ towards the eye. $\tau_{ext}$ is dependent on the speed of light $c_0$ and the external refraction index $n_{ext}$ of the external medium \cite{Taimre:15}.

The back injected light interferes with the local oscillating field, which is often referred to as self-mixing interference, resulting in a modulation of the optical power
\begin{equation}
\label{C1equ:modulated_power}
P_0' = P_0 \left( 1 + m \cdot \cos \left(\phi_{fb}\right)\right).
\end{equation} 
The modulated feedback power $P_0'$ is dependent on the optical power $P_0$, the modulation depth $m$ and a varying phase $\phi_{fb}$ of the backscattered light field. A small fraction of the modulated feedback power $P_0'$ is measured by a photodiode, which is integrated into the distributed Bragg reflector (DBR) of the laser cavity itself  \cite{Taimre:15}. 

To understand the link of the phase $\phi_{fb}$ to our observation goal (distance and velocity), we consider the rate equations introduced by Lang and Kobayashi \cite{1070479}. A solution of the rate equation is the excess phase equation 
\begin{equation}
\label{C1equ:excess_phase_equation}
\phi_{fb} - \phi_s + C \sin\left(\phi_{fb} + \arctan \left(\alpha\right)\right) = 0.
\end{equation}
The feedback phase is expressed as a function of the signal phase $\phi_s$, Acket's feedback parameter $C$, describing the coupling strength between target (eye) and laser cavity, and Henry's line width enhancement factor $\alpha$. As we operate the LFI sensor in the weak feedback regime ($C<1$) and a constant line width enhancement factor $\alpha$, Equation \ref{C1equ:excess_phase_equation} leads to a single solution \cite{Taimre:15} and $\phi_{fb}$ is, therefore, only dependent on $\phi_s$, leading to 
\begin{equation}
\label{C1equ:phase_stimuli}
\phi_s = \frac{4 \pi n_{ext} L_{ext}}{\lambda}
\end{equation}
with $\lambda$ describing the wavelength of the laser. Considering $n_{ext} \approx 1 $ is constant due to the operation of the sensor in free space, only changes in $\lambda$ and $L_{ext}$ lead to a varying phase $\phi_s$ and consequently, to a varying phase $\phi_{fb}$. This results, with respect to Equation \ref{C1equ:modulated_power}, in a modulation of the optical power which is measured by the photodiode. Changes in the wavelength $\lambda$ occur by a modulation of the laser drive current. That leads to a periodic heating and cooling of the resonator and, thus, to a periodic change of the cavity length. The variation of the cavity length leads to a periodic modulation of the wavelength, which allows for continuous measurement of the distance according to Equation \ref{C1equ:distance_equation}.

To distinguish between both effects, we compute the partial derivative of Equation \ref{C1equ:phase_stimuli} with respect to time, which leads to
\begin{equation}
\label{C1equ:distance_equation}
f_0 = \left. \frac{2 L_{ext}}{\lambda^2} \frac{d\lambda}{dI} \frac{dI}{dt}\right|_{L_{ext} = const.}
\end{equation}
and
\begin{equation}
\label{C1equ:doppler_equation}
f_d = \left. \frac{2 v_{ext} cos\left(\gamma \right)}{\lambda}\right|_{\lambda = const.}.
\end{equation}

Considering a known $\frac{d\lambda}{dI}$, which is a static process parameter of the laser, and a controlled current modulation slope $\frac{dI}{dt}$, the distance to the eye can be calculated by extracting the so called beat frequency $f_0$ by applying a Fast Fourier Transform (FFT) to the measured varying optical power and rearranging Equation \ref{C1equ:distance_equation} with respect to $L_{ext}$.

Movements of the eye ($\frac{d L_{ext}}{dt} = v_{ext}$) lead to a shift of the beat frequency $f_0$ by the so called Doppler frequency $f_d$. With a known angle of incidence $\gamma$ and a measured Doppler frequency, Equation \ref{C1equ:doppler_equation} can be rearranged with respect to $v_{ext}$ to obtain the surface velocity of the eye. 

In order to separate $f_0$ and $f_d$ and, thus, simultaneously measure the distance and velocity of the eye, a triangular modulation similar to frequency modulated continuous wave (FMCW) radar is applied to the drive current of the laser \cite{10.1117/12.775131}. By separating the up- and down ramp signals into two segments and applying an FFT on each segment, an $f_{up}$ and an $f_{down}$ frequency is captured. $f_0$ and $f_d$ are obtained from these measurements by
\begin{equation}
\label{C1equ:distance_triangular}
f_0 = \frac{f_{up} + f_{down}}{2}
\end{equation}   
and 
\begin{equation}
\label{C1equ:doppler_triangular}
f_d = \frac{f_{up} - f_{down}}{2}
\end{equation}
respectively. Recalling Equation \ref{C1equ:doppler_triangular}, the triangle modulation allows extracting the direction of the velocity as well. The modulation frequency of the triangle signal therefore limits the update rate of the LFI sensor. 

The main advantage of the LFI sensor compared to Time-of-Flight (ToF) sensors is its FMCW operating method, which allows distance and velocity to be measured simultaneously, whereas ToF sensors only measure absolute distance \cite{8067701}.  A rotation of the sclera surface in a fixed distance is therefore not measurable by TOF sensors as the absolute distance does not change.

\subsubsection{Sensor Features on the Eye} 
\label{C1subsec:Sensor features on the eye}
\begin{figure}[h]
	\centering
	\includegraphics[width=0.7\linewidth]{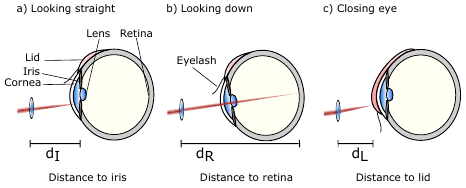}
	\caption{Different movement related position changes of the eye and the lid resulting in a change in distance measurement due to the geometry and scaffold of the eye \cite{Meyer2021}.}
	\label{C1im:distance_features_eye}
\end{figure}
\FloatBarrier
Based on Equation \ref{C1equ:distance_equation} and Equation \ref{C1equ:doppler_equation} the LFI sensor is capable to measure the surface velocity in laser beam axis $v_{eye}$ and the distance between the sensor and the eye $d_{eye}$. Both contain valuable features for HAR. Figure \ref{C1im:distance_features_eye} describes the distance features measurable on the eye.  A sectional view of the eye and a fixed LFI sensor position is shown for three different positions of the eye and the eyelid. In Figure \ref{C1im:distance_features_eye} a), the eye is directed straight ahead. The laser beam of the LFI sensor penetrates the cornea and backscattering occurs at the iris. For this arrangement of sensor and eye, the LFI distance measurement gives the distance between sensor and iris $d_I$ as feature.

In Figure \ref{C1im:distance_features_eye} b), the eye is slightly rotated downwards, the laser beam penetrates cornea and lens and the main backscattering occurs from the retina, leading to a second distance feature $d_R$ describing the distance between the sensor and the retina. In a temporal context, this feature occur if the pupil crosses the laser beam during eye activities.

Lastly in Figure \ref{C1im:distance_features_eye} c), the lid is closed by for blinking. During a blink the distance $d_L$ between sensor and lid is measured. This lead to a blink feature and in a temporal context it contains information about blink duration and blink frequency, which are used e.g. by the works of Ishimaru et al. \cite{BullingGoogleGlasses} or Steil et al. \cite{BullingLDA}.

\subsection{Recording head and eye movement data set}
\label{C1sec:Data collection}
Our data set was captured from 15 voluntary participants (age (years) = 25.9 (SD 4.5); 10 male; 5 female). Before taking part in the experiment, all participants provided written consent for using their data for research purposes. Two participants wore contact lenses during the experiment. The data set was anonymized to remove any personally identifiable information. 

\subsubsection{Apparatus}
\begin{figure}[h]
	\centering
	\includegraphics[width=0.95\linewidth]{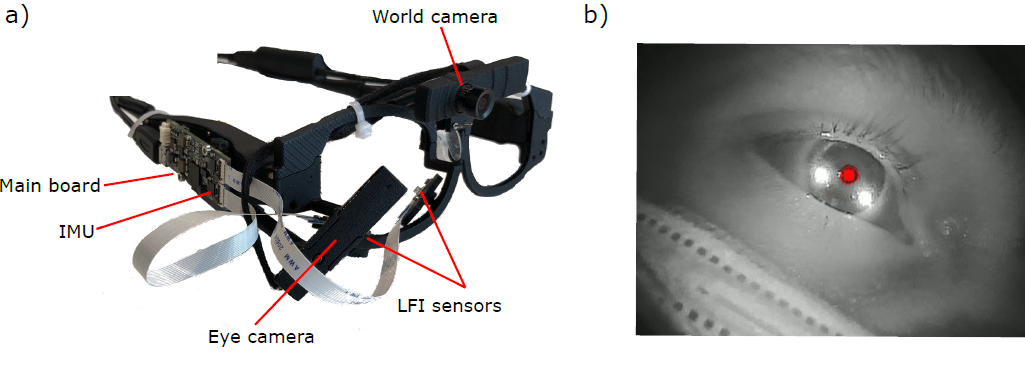}
	\caption{a) Research apparatus used to capture eye movement data by LFI sensors as well as head movement data by an IMU sensor. The colored arrows denote the coordinate space of the accelerometer. The coordinate space of the gyroscope is aligned to it. b) Image of the eye of an participant captured with the eye camera showing the detected pupil (red mark) by the pupil labs algorithm as well as the two LFI laser spots on the eye (bright spots).   }
	\label{C1im:cubeglasses}
\end{figure}

As LFI sensors are not available in commercial smart glasses the data set was captured with a custom head worn research apparatus shown in Figure \ref{C1im:cubeglasses}.
\FloatBarrier
The apparatus consists of a main board which reads out and supervises the two LFI sensors (shown in Figure \ref{C1im:coupled_cavity_model} c)) attached to the lower glasses frame. The laser beams of the LFI sensors are directed towards the eye ball to measure the eye's rotational velocity as well as the distance towards the eye with a sample rate of 1\,kHz. To ensure an eye safe operation of the lasers, the optical power of the lasers is limited to 360\,$\mu$W. With this low optical power a class 1 laser system according to IEC 60825-1 (well below class 1 optical power limit of 770\,$\mu$W) is achieved and, therefore, the experiments do not pose any medical hazard to the participant's eye. 

Furthermore, an IMU sensor (BMI270) containing a MEMS gyroscope as well as a MEMS accelerometer is attached to the main board. The IMU is used to capture head movements with a sampling rate of 860\,Hz. Both data streams are synchronized by the main board and streamed via USB to a laptop, which is used during the experiment to record the data. 

In addition to the IMU and LFI sensors we attach a Pupil Labs Core eye tracker \cite{Kassner:2014:POS:2638728.2641695}to the apparatus to capture world video frames as well as eye video frames. To record the video data the Pupil Capture (v1.17.71) software was used. The eye video camera is IR sensitive and we used it to align the laser spots of the LFI sensors to the surface of the eye as shown in Figure \ref{C1im:cubeglasses} b). This step is required to adjust the alignment of the LFI sensors to different head shapes before the actual measurement, as our rigid glasses frame of our research apparatus is adaptable to any interpupillary distance (IPD). The world video frames were used to annotate the captured head and eye movements with the corresponding activity label.

\subsubsection{Experiment Design}
Our activity set consists of seven activities (\textit{talk}, \textit{read}, \textit{video}, \textit{walk}, \textit{type}, \textit{solve} and \textit{cycle}) combining cognitive as well as physical activities. The experiment itself was split into two parts, a stationary part on the laptop and an outdoor part where the laptop was carried in a bag pack. After a short introduction into the functionality of the setup and an alignment of the LFI sensors towards the participant's eye, the experimenter initiated the recording session and starts a casual conversation with the participant. After the \textit{talk} part of the experiment the experimenter left the room and the participant followed for the remaining stationary part of the experiment the instructions on the laptop, where a website guides the participant through the \textit{solve}, \textit{read}, \textit{video} and \textit{type} parts of the experiment. During the \textit{solve} activity the participant was asked to solves a logic test. Afterwards, a text about smart glasses was presented to the participant to \textit{read} and a video about smart glasses to watch. Based on the video and the text read, the participants were then asked questions, which they had to answer by \textit{typing} their answer in text boxes on the screen.

\begin{figure}[h]
	\centering
	\includegraphics[width=0.7\linewidth]{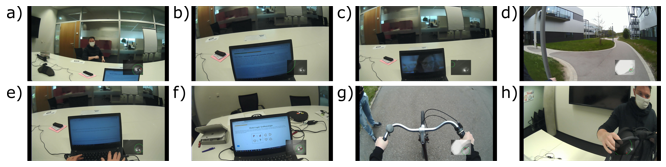}
	\caption{a) - g) show world camera frames captured during the activities \textit{talk}, \textit{read}, \textit{video}, \textit{walk}, \textit{type}, \textit{solve} and \textit{cycle}. In addition, h) shows a world camera frame captured between the indoor and the outdoor part of the experiment where the recording laptop was stowed into the backpack.}
	\label{C1im:activities}
\end{figure}

With the stationary part completed, the experimenter returned and stowed the recording laptop in a backpack to be worn by the participant and asked the participant to \textit{walk} around indoor and outdoor and afterwards to \textit{cycle} around the research campus. Figure \ref{C1im:activities} shows world camera frames of different activities captured during the experiment.      
\FloatBarrier
\subsubsection{Data Processing}
In order to classify activities from the head and eye movement data in a supervised fashion, it is required to associate each data sample to an activity label, which corresponds to an activity that the participant was performing at each sample's time of recording.

This activity labels are derived from the world video. In the first step, each participant's recording is loaded into the Pupil Player (v2.4). Next, the videos recorded with the eye and world cameras as well as the corresponding time stamps are exported and the time stamps of the LFI sensors as well as the IMU sensors are synchronized to the world video timestamps. Afterwards the exported world video is compressed and then loaded into a labeling software for annotation purpose. 

In order to assign an activity label to the head and eye movement recordings, the labels saved in the time domain of the world video need to be transcribed to the time domain of the IMU and LFI sensors. Since human activity usually extend over several seconds at a rather low frequency compared to the sensor output rate, and in order to reduce the amount of data, both the LFI and IMU data streams were down sampled to a common data rate of 120\,Hz. Due to the remaining difference in sampling frequency of head and eye movements with ($\approx$120\,Hz) and world ($\approx$30\,Hz) camera, multiple samples of the LFI sensors and the IMU correspond to a single frame in the world video. For label transcription, the start and end frame of each labeled activity block in the time domain of the world video is fetched. Next, all LFI and IMU samples that are associated to a world frame within the range of the given start and end frame receive the label of the corresponding activity block. This process results in labeled LFI and IMU features.

\subsubsection{Data Exploration}
\begin{figure}
	\centering
	\includegraphics[width=1\textwidth,height=0.95\textheight,keepaspectratio]{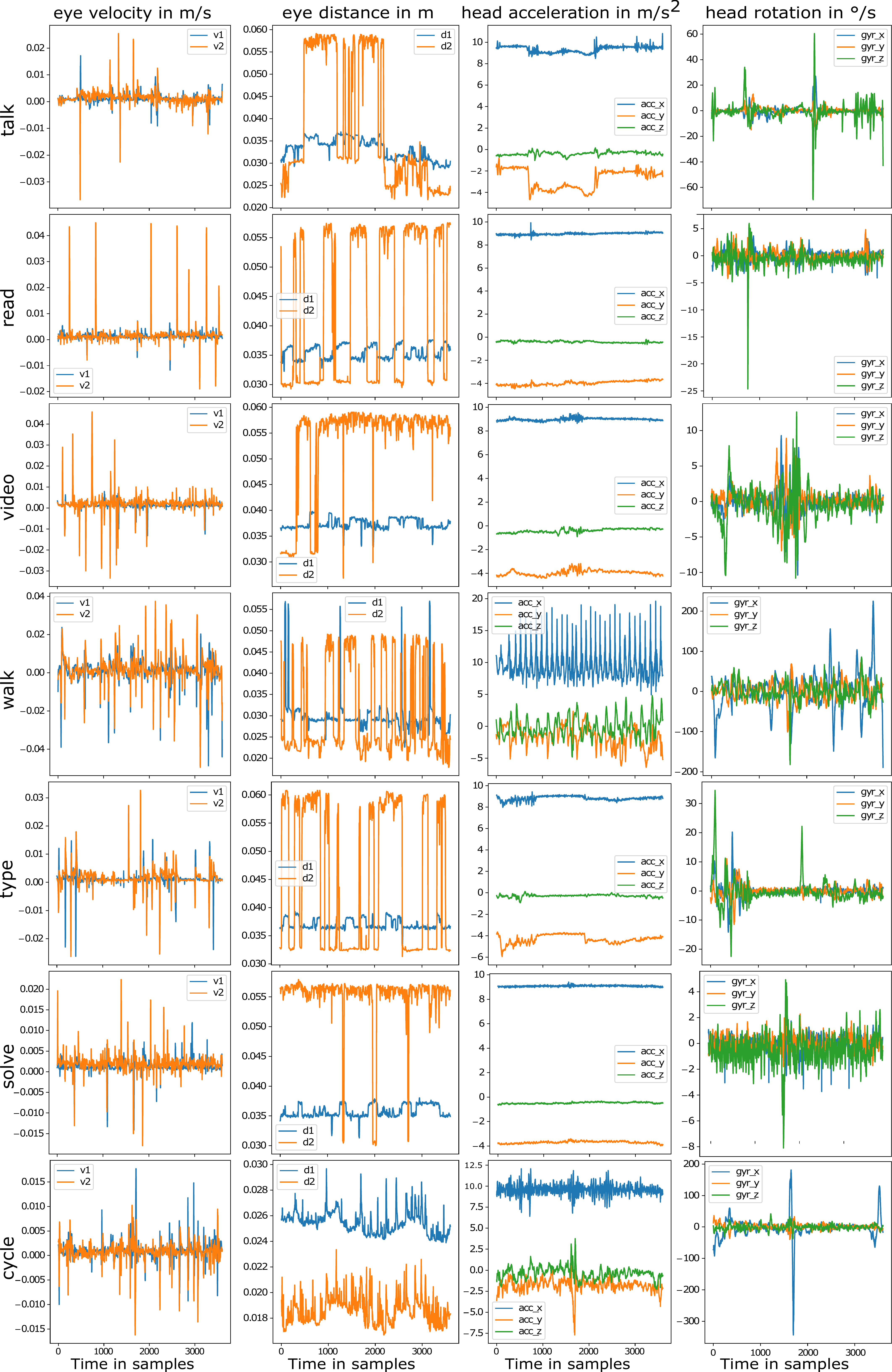}
	\caption{30 second windows of the raw head and eye movement features of participant P1 for the seven activities}
	\label{C1im:raw_features}
\end{figure}

Table \ref{C1tab:data_overview} gives an overview of the data set captured during the HAR experiment. A total of 18,5\,h head and eye movements were recorded across 15 participants and 7 activities. The mean duration per class are balanced, while the distribution of duration across participants is skewed. Participants spend the most time \textit{typing} (865\,s), while \textit{cycling}(534\,s) and \textit{reading}(506\,s) were sampled the shortest. This skew occurs because the duration of each activity was not limited during the experiment to ensure that participants behave as naturally as possible without being pressured by an expiring clock. The only exception was the activity \textit{video}, which was naturally limited by the playtime of the video. Furthermore, the experimenter paid attention that the activities \textit{walk} and \textit{cycle} do not take unnecessarily long to prevent the participants from becoming bored.
\begin{table}
	\small	
	\centering
	\caption{Duration in seconds of each individual experiment itemized by activity.}
	\begin{tabular}{|l|r|r|r|r|r|r|r|r|}
		\hline
		&   \textbf{talk} &   \textbf{read} &   \textbf{video} &   \textbf{walk} &   \textbf{type} &   \textbf{solve} &   \textbf{cycle} &   \textbf{total} \\
		\hline
		P1    &            251 &      408 &      620 &     623 &      385 &       774 &       547 &      3608 \\
		P2    &         475 &      268 &      621 &     547 &     628 &      383 &       611 &      3533 \\
		P3    &           620 &      541 &      618 &     857 &     841 &      857 &       920 &      5254 \\
		P4    &          693 &      412 &      622 &     680 &     376 &      617 &       382 &     3782 \\
		P5    &           664 &     222 &      619 &     475 &     806 &      313 &       262 &     3361 \\
		P6    &            609 &      691 &      620 &     503 &     1296 &      743 &       426 &      4888 \\
		P7    &            408 &      454 &      618 &     371 &     943 &       483 &       439 &      3716 \\
		P8    &          269 &     605 &      617 &      605 &     1735 &      869 &       295 &      4995 \\
		P9    &            772 &      760 &      618 &     760 &     589 &      730 &       605 &      4834 \\
		P10    &            745 &      536 &      616 &     632 &      1080 &       769 &       454 &      4832 \\
		P11   &            867 &     528 &      618 &      846 &     875 &      799 &       780 &      5313 \\
		P12   &            714 &      466 &      621 &     742 &     580 &      447 &       496 &      4066 \\
		P13   &            722 &      448 &      620 &     785 &      1452 &      880 &       628 &      5325 \\
		P14   &            552 &      690 &      620 &     684 &      843 &      557 &       631 &      4577 \\
		P15   &            598 &      562 &      619 &     626 &     547 &      871 &       527 &      4350 \\
		\hline
		$\varnothing$  &    597 &    506 &     619 &    649 &    865 &     673 &     534 &    4443 \\
		\hline
	\end{tabular}
	\label{C1tab:data_overview}
\end{table}
To get better understanding of the recorded data and the underlying pattern of the extracted features, Figure \ref{C1im:raw_features} shows a 30\,s time span of all input features for all activities. 
\FloatBarrier
The two left columns show the participants eye movement features eye velocity (v1, v2) as well as distance (d1, d2) towards the eye, captured by the LFI sensors. The two right columns show the participants head movement features captured by the accelerometer (accx, accy, accz) and the gyroscope (gyrx, gyry, gyrz). 

The first apparent pattern one can observe is the difference in the variance of the head-mounted sensors amplitudes. Whilst the gyroscope and accelerometer measurements depicted in Figure \ref{C1im:raw_features} show a quite steady line with few to no variation for the stationary activities, the corresponding sensor readings oscillate across all three axes for the physical activities \textit{cycle} and \textit{walk}. 

Considering the detected eye velocities, the strongest re-occurring pattern can be observed for the activity \textit{read}. Especially the velocity of the second sensor shows a well-defined pattern with sequentially high positive velocity peaks and small negative velocities in-between, which align well with the human reading pattern. Using small saccades, the eye slowly moves towards the end of a line of text before jumping back with a large saccade to the start of the next line. In addition, the distance signal shows a regular step pattern which occurs if the laser beam crosses the sclera and iris and drops into the pupil and gets back reflected from the retina of the eye, as described in Figure \ref{C1im:distance_features_eye}.

\subsection{Evaluation}
\label{C1sec:Evaluation}
Within this section, the synchronized eye and head movement data stream is used to build and train models for HAR. Prior to training, the data stream needs to be prepared to make classification feasible.

\subsubsection{Data Preparation}
Because different individuals may perform the same activity with varying intensity, the sensor readings may be offset or scaled differently across participants. This introduces undesired bias and inconsistency into the data. To avoid this, the distributions of all features within each participant's data stream are standardized to zero mean and unit variance.

Furthermore, the duration of human activity typically spans from several seconds up to numerous minutes. Thus, a single sensor sampled at sub-second resolution will most likely lack the necessary information to recognize its corresponding human activity. To overcome this issue, multiple consecutive sensor readings are combined using a sliding window of fixed duration. For each window, all containing sensor readings are extracted and considered as one sample and the corresponding label is determined by the most frequent label occurring during the window. In this way, a set of labeled windows is generated for each participant used for HAR.

Due to the varying duration of each individual activity, resampling on window scale is used to balance the data set by up sampling all under represented classes to match the most frequent class. As a result, all activity classes are represented equally often in the data set. 

With this data preparation steps, the data set has been transformed into a three-dimensional tensor $ D \in \mathbb{R}^{N\times T \times S} $, with $N$ being the total number of extracted and resampled windows across all participants' data streams, $T$ being the number of sensor readings per window and $S$ being the number of sensor modalities considered.  The sliding window size was fixed similar to previous work by Lan et al. \cite{GazeGraph} to 30 seconds with an overlap of 30\,\%. This results in a total number of $N = 4377$ sliding windows each containing $T = 3600$ sensor readings. As two triaxial head features (accelerometer and gyroscope) and four eye features (eye velocity and distance) from two sensors are considered, a total of $S = 10$ sensor modalities are used.

\subsubsection{Classification Approach}
\label{C1subsec:Classification approach}
The data set $D$ is split into a train and a test set by applying the concept of LOOCV. To be precise, $D$ is split into $D_{train} \in \mathbb{R}^{N_{train}\times T \times F}$ and $D_{test} \in \mathbb{R}^{N_{test}\times T \times F}$, with $N_{train} + N_{test} = N$, such that $D_{test}$ consists of all windows assigned to the left-out participant's data stream and $D_{train}$ contains the windows from all remaining participants. With 15 participants, this leads to 15 possible splits for $D$. Each classification model is trained and tested separately for each permutation of $D_{train}$ and $D_{test}$ and the model's overall performance is computed by averaging across the individual performances for each permutation, leading to a macro F1-score as classification performance indicator for each model. This method allows getting an understanding of how well a trained model generalizes to data stemming from an unknown participant. To ease comparison, each classifier's performance is visualized using the confusion matrices of the same three participants. These participants were chosen to represent test sets with overall low (P12), average (P6) and high (P11) macro F1-scores across the investigated classifiers.  
\subsubsection{Baseline Model}
As Ishairmaru et.al \cite{BullingGoogleGlasses} reported a high F1-score of 82\%  with a person-dependent decision tree for HAR based on handcrafted head and eye features, we chose a random forest classifier (RFC) as our base line model. As the RFC is not capable to learn features automatically, manual feature engineering is necessary. As suggested by previously mentioned works \cite{5444879, BullingReadingTransit, DAR_HeadAndEye, SteilAttentionForecasting, BullingGoogleGlasses}, a statistical analysis of the computed windows yields the desired input feature vector for the RFC. To be specific, \textit{mean}, \textit{variance} and \textit{L2-norm} are computed along the temporal axis for each sensor modality and window in $D_{train}$ and $D_{test}$. The resulting feature values are stacked into a feature vector $ v \in \mathbb{R}^{F*S}$ where $S$ is the number of considered sensor modalities and $F$ the number of statistical features to be computed (here $F = 3, S = 10$). Thus, the RFC is trained and tested on the matrices $V_{train} \in \mathbb{R}^{N_{train}\times 30}$ and $V_{test} \in \mathbb{R}^{N_{test}\times 30}$. 
\begin{figure}[h!]
	\centering
	\includegraphics[width=1\linewidth]{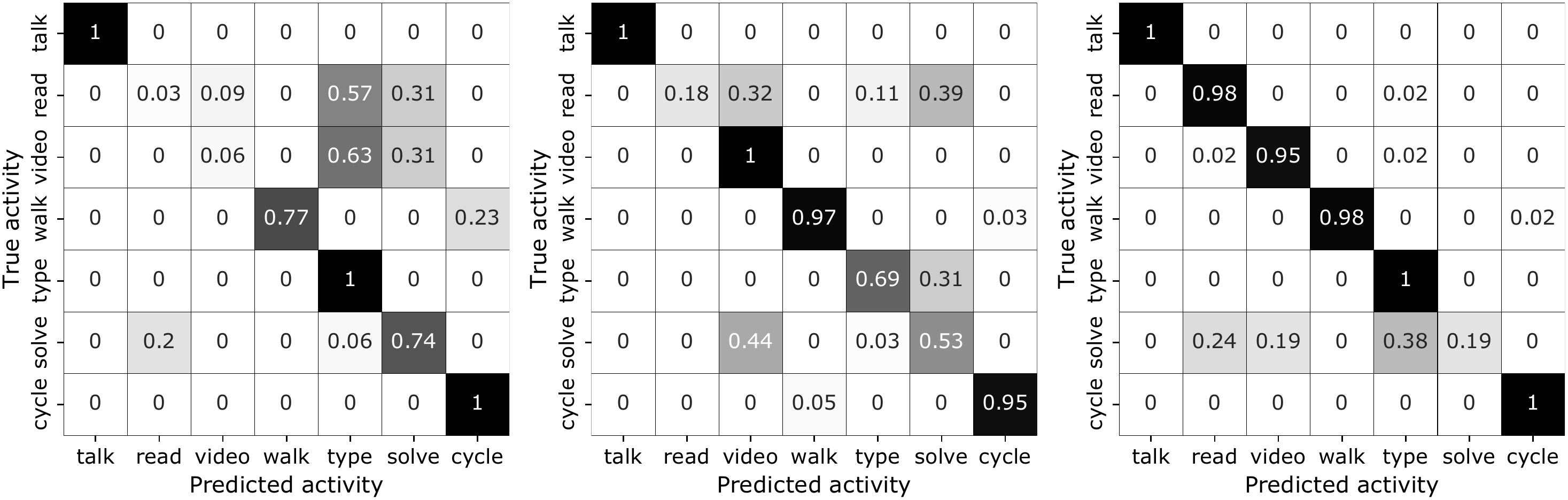}
	\caption{Normalized confusion matrix for the RFC for participants P12 (F1-score 59.77\,\%), P6 (F1-score 73.36\,\%) and P11 (F1-score 88.31\,\%).}
	\label{C1im:conf_RFC}
\end{figure}
\FloatBarrier
Figure \ref{C1im:conf_RFC} shows the average macro F1-score after LOOCV has run for each participant. The RFC achieves an average accuracy of 74.32\,\% and an average macro F1-score of 71.64\,\%. From the confusion matrices depicted in Figure \ref{C1im:conf_RFC} it can be seen that the classifier's confusion is only present within the set of stationary (\textit{solve}, \textit{read}, \textit{talk}, \textit{video} and \textit{type}) and physical (\textit{walk} and \textit{cycle}) activity classes, respectively. Furthermore, it can be observed that within the stationary activities, the activities that frequently involve head and eye movements, namely \textit{type} and \textit{talk}, lead to a higher recognition rate than eye movement intensive activities such as \textit{solve}, \textit{video} and \textit{read}. One possible reason for this observation could be that the statistical features do not take into account the time-dependent pattern of eye movements which could be helpful in distinguishing between \textit{solve} and \textit{read}, as shown in Figure \ref{C1im:raw_features}.     

\subsubsection{1D-CNN Model} 
To better represent time-dependent patterns of the head and especially the eye movement patterns and get rid of the hand crafted statistical features, we chose a CNN classification approach to improve our classification result on activities.
CNNs use a set of convolving kernels that move across an input plane, in our case an $TxF$ matrix. CNNs typicaly consists of several convolutional layers which are chained together to build up a classifier. The lower convolutional layers extract local salient patterns of the input signal, whereas the deeper layers operate on high-level, abstract patterns provided from the previous layers \cite{CNN1D_Inspiration1}. Using the concept of backpropagation, kernel weights are learned and updated such that a certain error metric is minimized. This is also referred to as automated feature learning and removes the necessity of manual feature selection. CNNs are well known for image classification. However, multiple works have managed to successfully apply CNNs for HAR tasks \cite{CNN1D_Inspiration1, CNN1D_Inspiration2, CNN1D_Inspiration3_starting_params, CNN-LSTM_Inspiration}. Here, the CNN operates on a small subset of data by gradually moving a convolving kernel of fixed size across the window. To capture temporal relationships, one-dimensional convolutions along the time axis are used. Each sensor modality is processed by a separate set of kernels. Each kernel then aggregates multiple sequential data points of that modality.
\begin{figure}[h]
	\centering
	\includegraphics[width=0.6\linewidth]{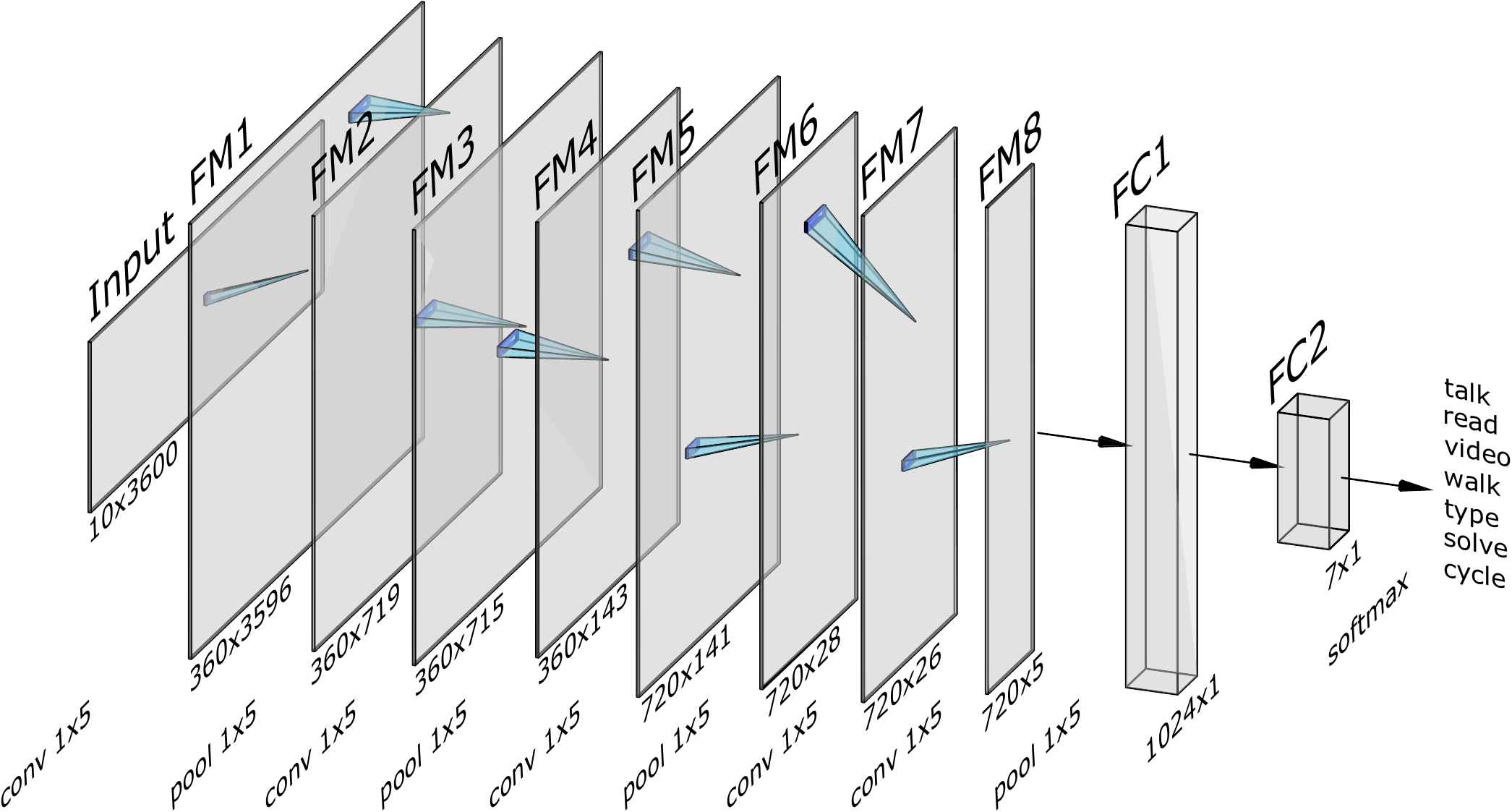}
	\caption{Architecture of the 1D-CNN model. A batch normalization layer and a leaky ReLU follow each convolutional operation, which are omitted for visualization purpose. The feature maps (FM) height corresponds to the number of convolutional layers whereas their width represents the temporal axis.}
	\label{C1im:CNN-1D_Model}
\end{figure}
\FloatBarrier

\begin{table}[h]
	\small	
	\centering
	\caption{Hyperparameters of the 1D-CNN model.}
	\begin{tabular}[]{|p{0.32\textwidth}|p{0.1\textwidth}|}
		\hline
		Parameter name                  & Value  \\ \hline
		Learning rate                & 1E-3  \\
		Epochs                       & 9      \\
		Exponential learning rate decay& 0.95   \\
		Weight decay                 & 1E-4      \\
		Window size & 30s\\
		Window overlap & 0.3\\
		\hline
	\end{tabular}
	
	\label{C1tab:CNN1D_hyperparams}
\end{table}
The 1D-CNN model architecture, shown in Figure \ref{C1im:CNN-1D_Model}, is inspired by the work of Yang et al. \cite{CNN1D_Inspiration1} and was adjusted to fit for the HAR task. The CNN1D-model is trained and evaluated using all windows in $D_{train}$ and $D_{test}$ respectively. To process a single window, the CNN1D-model utilizes four single strided convolutional layers (FM1, FM3, FM5, FM7), each followed by a maximum pooling layer (FM2, FM4, FM6, FM8), a leaky ReLU activation and a $1\times5$ max-pooling operation. The latter leads to the fact that the temporal resolution becomes coarser with increasing network depth, which allows extracting more coarse time-dependent features with increasing convolutional layers. This coarse-grained, high-dimensional time series features computed by the last pooling layer (FM8) is passed through two fully connected layers (FC1, FC2) with a dropout layer in-between to map the feature sequence to the desired amount of classes. In the final step, a softmax activation function is applied to transform the network's output into a probability distribution across class labels. The model is trained using cross entropy loss and the Adam optimizer \cite{Adam} with the hyper parameters given in Table \ref{C1tab:CNN1D_hyperparams}.

\begin{figure}[h]
	\centering
	\includegraphics[width=1\linewidth]{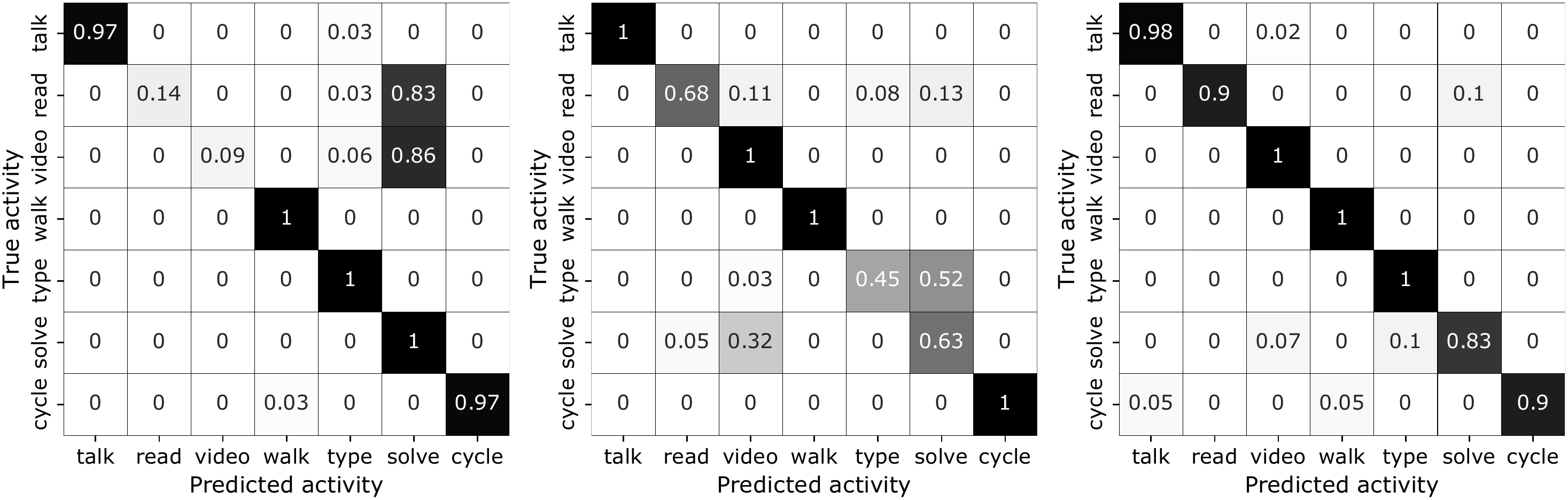}
	\caption{Normalized confusion matrix for the 1D-CNN model for participants P12 (F1-score 73.87\,\%), P6 (F1-score 82.25\,\%) and P11 (F1-score 94.55\,\%).}
	\label{C1im:conf_CNN}
\end{figure}
\FloatBarrier
Figure \ref{C1im:conf_CNN} shows the confusion matrices of the selected participants. After LOOCV, the 1D-CNN model achieves an average accuracy of 82.13\,\% and an macro F1-score of 80.98\,\%. Compared to the RFC baseline model, the overall HAR rate increases for all participants. This is mainly caused by the improved accuracy for the reading class. The 1D-CNN seems to be able to extract patterns that are specific to the \textit{read} class, which are not represented by the statistical features alone. In addition, the 1D-CNN model handles intra class similarities between the eye movement-intensive activities \textit{read}, \textit{solve} and \textit{video} better.

\subsubsection{1D-CNN with Transfer Learning} 
\begin{figure}[h]
	\centering
	\includegraphics[width=1\linewidth]{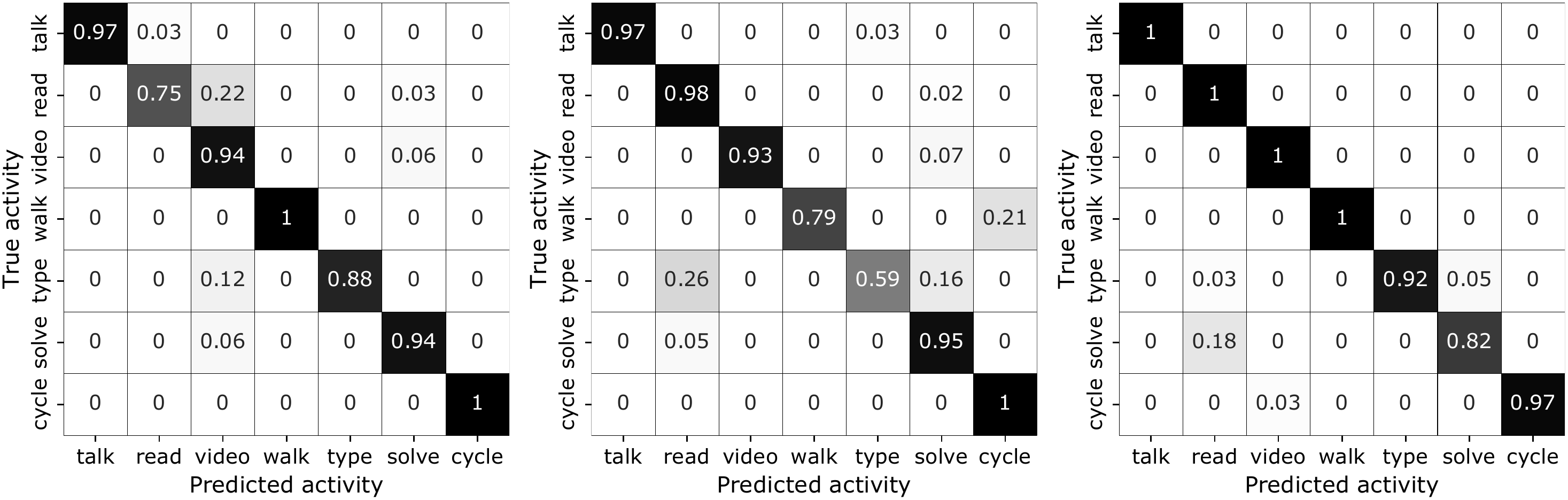}
	\caption{Normalized confusion matrix for the 1D-CNN model after three transfer samples for participants P12 (F1-score 92.41\,\%), P6 (F1-score 88.34\,\%) and P11 (F1-score 95.97\,\%).}
	\label{C1im:CNN1D_Trans_low_mean_max_conv}
\end{figure}
\FloatBarrier
Human activity is highly diverse and versatile because different participants perform the same activity in a different manner. This leads to intra class heterogeneity \cite{BullingTutorial} and a general model, which shows high recognition accuracy over all participants, is hard to build. To tackle this issue we apply transfer learning as proposed by Chikhaoui et al. \cite{HAR_CNN_Transfer_Learning} to further increase the recognition accuracy of the 1D-CNN model. The 1D-CNN model  architecture is modified as follows. For every permutation of LOOCV, the model is trained as described in Section \ref{C1subsec:Classification approach}. After training has finished, the weights in the convolutional layers and in the first linear layer (FC1) are frozen such that they will not be updated during further backpropagation. Furthermore, the dropout layer between the two linear layers (FC1, FC2) is removed. Next, only the last linear layer (FC2) is reinitialized and trained for a few epochs, using three 30 second windows per class from the left out participants data set. 

To increase generalization the three transfer learning samples are chosen randomly within an activity class. With the chosen sample size of three transfer samples per activity only a small subset of all samples per activity are used to retrain the model to minimize overfitting. 

The model still relies on the feature extractor built from all other participants, which is mainly in the convolutional layers but learns to interpret the extracted features in the context of the current test participant by personalizing their last linear layer to the participant-specific data. Figure \ref{C1im:CNN1D_Trans_low_mean_max_conv} shows the confusion matrices of the three selected participants after applying transfer learning. With this approach, the overall classification accuracy increases, leading to a macro F1-score of 88.15\,\%.

Especially for participant P12 the classification accuracy improves as the classifier now resolves the disambiguity between \textit{solve}, \ and \textit{read} in a better way. Also for participant P6 the classification accuracy increases slightly. Overall transfer learning leads to an improvement of $\approx$ 7\,\% with only three samples per class. The hyper parameters used during the transfer learning step are denoted in Table \ref{C1tab:CNN1D_trans_learn_hyperparams}. 
\begin{table}[h]
	\small	
	\centering
	\caption{Hyper parameters of the transfer learning part of the 1D-CNN model.}
	\begin{tabular}[]{|p{0.32\textwidth}|p{0.1\textwidth}|}
		\hline
		Parameter name                  & Value  \\ \hline
		Learning rate                & 1E-3  \\
		Epochs                       & 10      \\
		Exponential learning rate decay& 0.95   \\
		Weight decay                 & 1E-4      \\
		Transfer samples per class & 3 \\
		\hline
	\end{tabular}
	
	\label{C1tab:CNN1D_trans_learn_hyperparams}
\end{table}
\subsubsection{Model Performance Comparison}
In Table \ref{C1tab:comp_class} the macro F1-scores of the three models (RFC, 1D-CNN and 1D-CNN Trans) are shown for each participant.

\begin{table}[h]
	\small	
	\centering
	\caption{Macro F1-score of the three models, investigated for HAR, for each participant. The bold marked scores show the highest achieved macro F1-score over the three classifiers while the scores marked with an underline show the lowest macro F1-score.}
	\begin{tabular}{|l|r|r|r|}
		\hline
		&   \textbf{RFC} &   \textbf{1D-CNN}  &   \textbf{1D-CNN Trans}  \\
		\hline
		P1    &            \underline{71.88} &      81.85 &          \textbf{87.48}  \\
		P2    &         \underline{62.26} &      80.84&          \textbf{87.79}  \\
		P3    &          \underline{ 87.96} &      90.56 &          \textbf{95.81}  \\
		P4    &          \underline{60.05} &      69.17 &          \textbf{75.25}  \\
		P5    &           \underline{71.67} &     90.09&          \textbf{88.61} \\
		P6    &            \underline{73.97} &      81.97 &        \textbf{88.34 }  \\
		P7    &            \underline{63.55} &      79.57 &           \textbf{95.63}  \\
		P8    &          \underline{72.91}&     \textbf{84.09}&            82.78 \\
		P9    &            \underline{73.09} &      78.34 &          \textbf{83.94}\\
		P10    &            \underline{65.08} &      76.43 &          \textbf{92.72}  \\
		P11   &            \underline{84.01}&     94.49&           \textbf{95.97 }\\
		P12   &            \underline{59.36} &      69.33 &         \textbf{92.60} \\
		P13   &            \underline{74.94} &      \textbf{81.78} &         81.07 \\
		P14   &            78.47&      \underline{69.65} &           \textbf{86.22} \\
		P15   &            \underline{75.38}&      86.54 &      \textbf{88.04}  \\
		\hline
		$\varnothing$  &    \underline{71.64} &    80.98 &    \textbf{88.15}  \\
		\hline
	\end{tabular}
	
	\label{C1tab:comp_class}
\end{table}
The RFC shows in nearly all cases the lowest macro F1-score, which indicates that the extraction of statistical features from sampled windows of the sensors do not represent the patterns required to recognize activities well. With this approach, we achieve similar results as related work by Ishimaru et. al. \cite{ActRecogWithCommercialEOGGLasses}.

As we introduce a representation of the sequence and extract features automatically by applying the 1D-CNN model, the overall macro F1-score increases to a similar region as reported by Ishimaru et. al. \cite{BullingGoogleGlasses}. 

By further personalization of the classifier to the participant using transfer learning, on the 1D-CNN model a further increase in performance was observed, which outperforms related work by Ishimaru et. al. \cite{BullingGoogleGlasses, ActRecogWithCommercialEOGGLasses}. It leads to a similar performance like related work focusing either on physical activities recognized with body worn sensors e.g. \cite{HARAndroidAcc} or cognitive activities recognized with eye-tracking sensors \cite{DAR_Scanpath}.

In Table \ref{C1tab:comp_class_activites} the three classifiers are compared on an activity level.

\begin{table}[b]
	\small	
	\centering
	\caption{Averaged macro F1-score for each activity over all participants of the three models, investigated for HAR. The bold marked scores show the highest achieved average macro F1-score over the three classifiers while the scores marked with an underline show the lowest average macro F1-score.}
	\begin{tabular}{|l|r|r|r|r|r|r|r|r|}
		\hline
		&   \textbf{RFC} &   \textbf{1D-CNN} &     \textbf{1D-CNN Trans}  \\
		\hline
		talk    &            94.59 &      \underline{93.39} &         \textbf{95.06}  \\
		read    &        \underline{ 34.92} &      68.01&         \textbf{ 91.78}  \\
		video    &           \underline{70.02} &      74.88 &           \textbf{87.85 } \\
		walk    &          95.08 &      \textbf{96.12} &         \underline{95.03}  \\
		type    &           84.53 &     \textbf{85.83}&          \underline{77.17} \\
		solve    &           \underline{47.83 }&      60.14 &        \textbf{74.74 }  \\
		cycle    &           \underline{85.13} &      94.92 &       \textbf{96.59}  \\
		\hline
	\end{tabular}
	
	\label{C1tab:comp_class_activites}
\end{table}
The primary physical activities \textit{cycle} and \textit{walk} show an overall high recognition rate. In addition, \textit{talk} as an activity consisting of head and eye movements shows a high recognition rate over all classifiers. Especially the \textit{read} and \textit{solve} activity take advantage of the extraction of sequential features by the 1D-CNN model. Transfer learning improves mainly the recognition rate for the \textit{read} and \textit{video} activities, which are heavily dependent on eye movements. This indicates that these activities are dominated by an intra class heterogeneity over all participants. Another observation can be made for the activity \textit{type}. For this activity transfer learning leads to a degradation of the classification accuracy compared to the other classifiers. This shows that there is an inter class similarity (also present in Figure \ref{C1im:conf_CNN} for P6) between \textit{type} and \textit{solve} and by personalization, the decision boundaries shifts either to \textit{solve} or \textit{type} activity and therefore one increases while the other decreases.
\subsubsection{Impact of Sensor Modalities}
\label{C1subsec: Impact sensor modalities}
To investigate the impact of the sensor modalities, especially the impact of head movement features versus the impact of eye movement features, feature importance is studied. During this study all three models are retrained with only a subset of all features, more precisely the features were separated into IMU features (triaxial accelerometer and gyroscope features) and LFI features ($d_{eye}$ and $v_{eye}$) per sensor. Afterwards, we retrain and evaluate all models with only LFI features from both sensors (LFI1 \& LFI2) and only IMU features and compared the achieved macro F1-score with the macro F1-score reached if all sensor modalities where considered. In addition, the effect on the macro F1-score when only a single LFI sensor is used is investigated to answer the question of how the change in angle of incidence $\gamma$ of the laser beam affects the classification accuracy. 

\begin{figure}[h]
	\centering
	\includegraphics[width=0.6\linewidth]{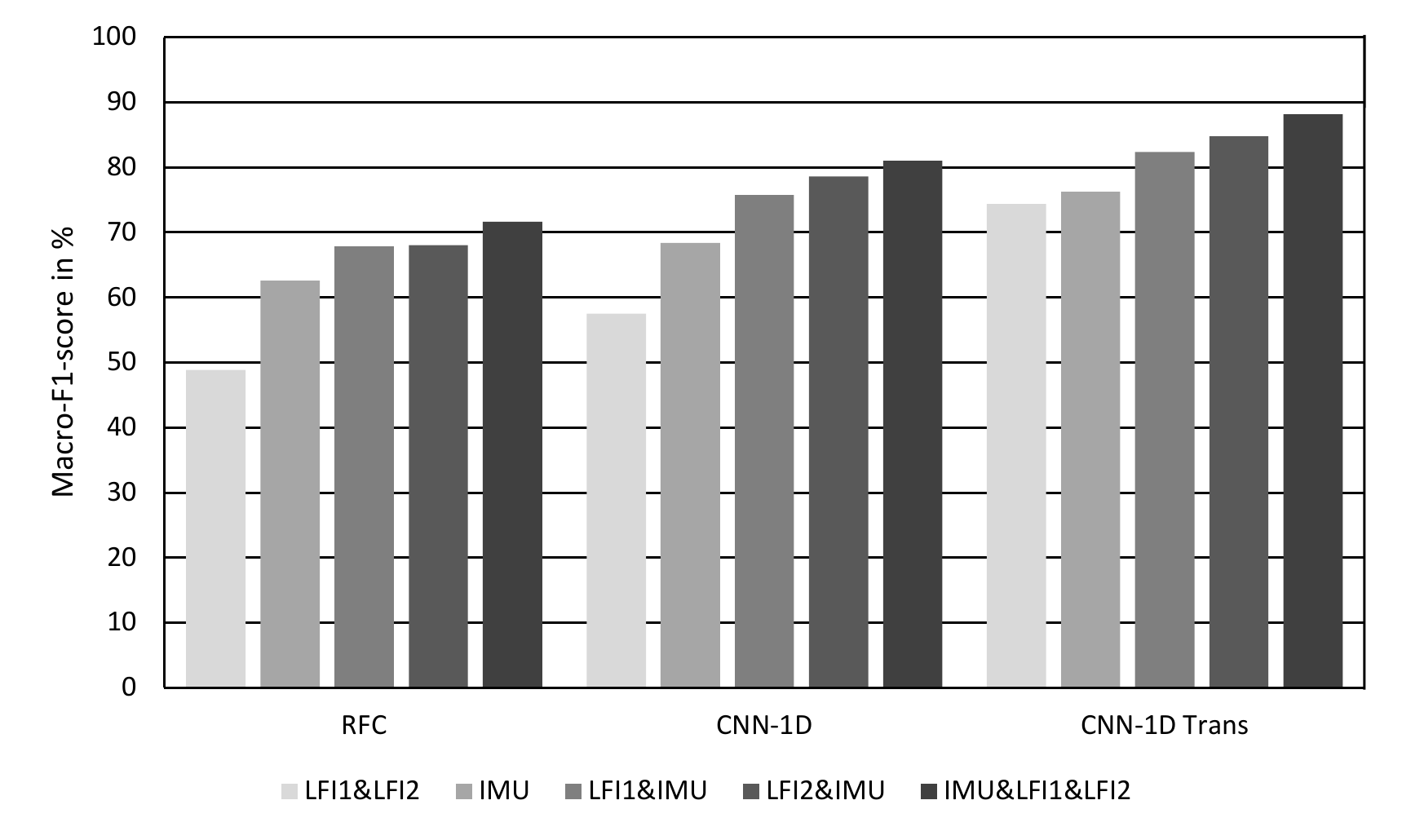}
	\caption{Impact of the sensor modalities on the macro F1-score for the three models}
	\label{C1im:Abliation study}
\end{figure}

Figure \ref{C1im:Abliation study} shows the results of the influence of the sensor modalities on the macro F1-score.
\FloatBarrier
Is it clearly visible that the combination of head and eye movement features are mandatory to reach high classification accuracy for all three models. For the 1D-CNN model and the RFC it is also shown that the impact of the IMU features is higher than the impact of the LFI features. A further interesting point to note is that the influence of the angle of incidence $\gamma$ of the laser beam shows only small contribution to the overall performance of the models but it has to be noted that the sensor position of LFI2 leads to a slightly better performance.

In Table \ref{C1tab:ablation_activities} the effect of head and eye movement features on the macro F1-score are investigated per activity. 
\begin{table}[!h]
	\small
	\centering
	\caption{Macro F1-scores per activity class averaged across all participants when solely considering eye movement features (LFI1  \& LFI2) and head movement features (IMU) as well as their combination. For each classifier and sensor modality the best and worst values are emphasized in bold and underlined respectively.}
	\renewcommand{\arraystretch}{1.1}
	\begin{tabular}{|l|c|c|c||c|c|c||c|c|c|}
		\hline
		& \multicolumn{3}{c}{RFC}   &   \multicolumn{3}{c}{1D-CNN}   &   \multicolumn{3}{c|}{1D-CNN Trans} \\
		& LFI1\& LFI2 & IMU & both & LFI1\& LFI2 & IMU  & both & LFI1\& LFI2 & IMU& both \\
		\hline
		talk  &62.39 			&92.12 				&94.59 				&55.01 				&95.02 				&93.39 &			\textbf{81.74} 		&92.26 & 95.06\\
		read  &40.50 			&\underline{23.61}	&\underline{34.92}  &58.20 				&\underline{31.74 }	&68.01 &			81.53 				&\underline{49.25} & 91.78 \\
		video &33.67 			&60.13 				&70.02 				&\underline{46.99} 	&59.26 				&74.88 &			73.42 				&82.77 & 87.85 \\
		walk  &60.60 			&\textbf{89.93} 	&\textbf{95.08} 	&\textbf{74.21 }	&\textbf{97.54} 	&\textbf{96.12} &	76.27 				&93.25 & 95.03\\
		type  &70.31 			&70.21 				&84.53				&67.96 				&81.05 				&85.53 &			62.36				 &78.63 & 77.17\\
		solve &\underline{33.54}&36.61 				&47.83 				&48.02 				&38.57 				&\underline{60.14} &\underline{60.45}	 &51.01 & \underline{74.74} \\
		cycle &\textbf{63.45} 	&79.41 				&85.13 				&71.89 				&91.84 				&94.92 &			88.16 				&\textbf{93.87} & \textbf{96.59}  \\
		\hline
	\end{tabular}
	\label{C1tab:ablation_activities}
	
\end{table}
As expected from related work, the physical activities \textit{walk} and \textit{cycle} are dominant in the IMU data, while the LFI data mainly cover the cognitive activities \textit{video} and \textit{reading}. To some extend the LFI features provide information about physical activities. This might result from glasses slippage during physical activities, which is encoded in the distance information $d_{eye}$.

Looking only at the features of the LFI sensor, the velocity feature is the most relevant feature for HAR, as it describes the eye movement pattern of the different activities. Especially for the activity \textit{reading}, which is dominated by eye movements, a clear velocity pattern can be observed, as shown in Figure \ref{C1im:raw_features}.

The distance feature mainly contains information about blinking and passing of the pupil through the laser beam. This information is only weakly correlated with activities, since an increased blink frequency per window correlates with higher cognitive load as well as drowsiness.

\subsection{Discussion}
Despite the overall high accuracy on the diverse activity set that we reach by combining LFI sensors and IMU sensor, there are still limitations, we will discuss below and point out a direction of how to improve them.\\

\subsubsection{Power Consumption}
Power consumption is an important factor since the glasses are battery powered and for continued HAR a steady stream of sensor data is mandatory. The power consumption per LFI sensor prototype is $\approx$ 70\,mW with a sample rate of 1\,kHz using of-the-shelf components. This can be reduced to about 15\,mW by using an application specific integrated circuit (ASIC) \cite{meyer11788compact}, while the power consumption of the IMU is $\approx 1.2mW$ at full operation mode \cite{BMI270}. As we down sample the data to 120\,Hz a small reduction of power consumption is achievable by adjusting the sampling rate of the sensors. In addition, the power consumption can be further reduced by using just a single LFI sensor, as the study of impact on sensor modalities in Section \ref{C1subsec: Impact sensor modalities} shows that using a single LFI sensor already a reasonable classification performance is achieved. 

However, the main driver of power consumption is the classification algorithm, which should run on a resource constrained embedded processor in the frame temple. For the RFC algorithm Kumar et. al. \cite{Bonsai_RFC} present a resource and power-efficient implementation while keeping a high classification performance. 

With respect to the 1D-CNN model, energy-efficient implementations for CNN-based models are established, e.g. in dedicated neural network accelerators or a tensor processing units \cite{garofalo2020pulp}, \cite{Bringmann_CNN_acc}. The remaining drawback of the 1D-CNN model for use on a resource constrained embedded processor is the memory footprint mainly driven by the fully connected layers (FC1, FC2) which hold 3686400 out of 3703816 model parameters. To reduce them a further network optimization could be achieved by model pruning or other optimization techniques \cite{Edge_Comp_Review}. Another method could be to rely on a pure CNN based model like proposed by Perslev et. al. \cite{perslev2021u}. \\

\subsubsection{Transfer Learning}
To exceed the state of the art with respect to classification performance we apply transfer learning to improve the 1D-CNN model. This requires the user to perform each activity of the activity set once for 90 seconds to adapt the final layer of the 1D-CNN model to the user. This may degrade the user's experience even if it is a known procedure for smart wearables e.g. the initial setup of face recognition or fingerprint recognition in a smart phone \cite{wang2020deep}. 

In addition, the last layer (FC2) of the 1D-CNN model needs to be re-trained, which needs to be done on the device. While the inference on the device is a computationally simple process, the post-training to fit the model is computationally intensive and therefore takes some time. One solution to improve model adaptation time is to perform the post-training in the cloud. \\

\subsubsection{Glasses Slippage}
Glasses slippage is a well-known issue in the eye tracking domain  \cite{niehorster2020impact} and it also affects the LFI sensors in some cases. Glasses slippage was to some extend present during our experiments, mainly during the physical activities \textit{walk} and \textit{cycle} as well as \textit{talk}. Therefore, the effect of slippage is already reflected in our reported macro F1-scores. But in some cases we observed that the eye was covered by the eye lids of a participant and the laser spots of the LFI sensors hit the eye lid. Under these circumstances no robust velocity features were captured by the LFI sensors leading to a reduction in classification performance. To ensure that the laser beam of the LFI sensor hits the eye, the laser beam can be scanned over the surface of the eye \cite{7863402}, \cite{9149591}. Another solution could be achieved by modification of the optics to split the laser beam of the sensor to several beams in a line as proposed by Riemensberger et. al. \cite{riemensberger2020massively}.

\subsection{Conclusion}
\label{C1sec:Conclusion}
In this work we introduce the LFI sensor for eye movement detection. The LFI sensor is very advantageous as it consumes very low power, is fully unobtrusive as the IR light is invisible to the human eye, and is very robust to ambient light. 

We attached the LFI sensors together with an IMU to a glasses demonstrator to capture eye and head movements and recorded a large dataset of 15 participants performing seven activities. With the proposed 1D-CNN model and by applying transfer learning, we outperform the state of the art and achieve a macro F1-score of 88.15\,\% on a diverse activity set consisting of both physical and cognitive as well as social activities. We further investigate which sensor modality leads to high classification performance on each activity level. Furthermore, we discuss our work critically with respect to power consumption, transfer learning and the impact of slippage. Based on our results, we are confident that this work will advance the path towards enabling always-on context awareness based on HAR for smart glasses.

\newpage
\section{U-HAR: A Convolutional Approach to Human Activity Recognition Combining Head and Eye Movements for Context-Aware Smart Glasses}
\label{APP:B4}
\subsection{Abstract}
Smart glasses are considered the next breakthrough in wearables. Their advancement lies in the ubiquitous embedding of content in the user's field of view, promising an immersive extension of reality. While advances in display technology seem to fulfill this promise, interaction concepts are inherited from existing wearable solutions that require the user to actively interact with the glasses, limiting the user experience. One way to improve the user experience and drive immersive augmentation is to reduce user interactions to a necessary minimum by adding context awareness to smart glasses.

To achieve context awareness, we propose an approach based on human activity recognition which incorporates features derived from the wearer's eye and head movement. Towards this goal, we combine an eye-tracker and an IMU to collect eye and head movement features from 20 participants performing seven cognitive and physical activities to derive context information. From a methodological perspective, we introduce U-HAR, a convolutional network optimized for activity recognition on a device with constrained power and memory resources. By applying few shot learning, we achieve an outstanding macro-F1 score of 86.59\.\%, allowing us to derive contextual information that will pave the wave for a more immersive user experience in future smart glasses.

\subsection{Introduction}
After the success of smartphones, smart earbuds and smartwatches, smartglasses are expected to be the next breakthrough in the domain of smart wearables  and will replace the smartphone as personal smart wearable in long term as recently announced by Apple \footnote{\url{https://www.computerworld.com/article/3642649/analyst-apples-ar-glasses-will-run-mac-chips.html}}. Their advantage over other wearables is the seamless embedding of visual content into the field of view (FOV) of a user, resulting in a true augmentation of the environment. A major drawback of smart glasses, preventing full immersive augmentation of the environment, arises from the state of the art interaction concepts experienced by the user.  The user still has to actively interact with the glasses, e.g. through spoken commands as shown in the Echo frames \cite{amazon2019}, by hand gestures used in the Hololens \cite{kress2020optical}, touch interaction concepts introduced by Google Glass \cite{GoogleGlass}  or gaze gestures \cite{Meyer2021}.

One way to improve the user experience and drive immersive augmentation is to minimize user distraction from interacting with smart glasses. A possible solution to reduce user interaction and thus drive immersive augmentation is to add context awareness to smart glasses. A possible way to achieve context awareness in a smart glass environment is to derive contextual information from the wearer's activities by using Human Activity Recognition (HAR). 

In this work, we investigate the use of HAR to derive contextual information in a smart glasses setup. For this purpose, we built up an apparatus to simultaneously collect head movement data with an inertial measurement unit (IMU) and eye movement data with a commercial video oculography (VOG) eye-tracking sensor during our experiment. For evaluation purposes,  20 participants performed seven physical or cognitive tasks (i.e. talk, solve, read, watch a video, type, walk and cycle), leading thus to an unique dataset with a total duration of 1514 minutes. In addition to this novel dataset, we propose a convolutional neural network (CNN) model to temporal fuse head- and eye-movement information to recognize human activities, which we coined \textbf{U-HAR}. By applying few shot learning, the CNN model is further personalized to achieve an macro F1-score of 86.59\,\% in a leave-one-participant-out cross validation (LOPOCV) scheme. With this outstanding accuracy our model outperforms earlier works, e.g. by \cite{ActRecogWithCommercialEOGGLasses, BullingGoogleGlasses} by a large margin.  In addition to the above methodological novelty, we investigate the relevance of the sensor features for different human activities and discuss the proposed system in terms of its applicability for smart glasses. Our contribution is three fold:

(i) \textbf{Context aware smart glasses:} We propose to employ HAR to derive contextual information to minimize user interaction with smart glasses to increase user experience, to minimize power consumption through contextual aware display control and improve privacy as contextual information can be derived without the need of a world camera sensor.

(ii) \textbf{U-HAR:} We adapt a CNN (\textbf{U-HAR}) model architecture and add few shot learning to efficiently recognize a rich set of human activities with a high accuracy from eye- and head movement features.

(iii) \textbf{Dataset:} We collected a comparable large dataset of 20 participants with commercially available VOG and IMU sensors commonly used in AR glasses to demonstrate the performance of our system and publish the dataset for further research \footnote{\url{https://atreus.informatik.uni-tuebingen.de/seafile/d/978f6631b6b34f7c9139/}}. 



In the following Section, we provide an overview of related work with a focus on HAR combining both features from head- and eye-movements. In Section \ref{C2sec:Material Methods}, we describe the apparatus and the experiment used to record the dataset. In Section \ref{C2sec:U-HAR Modell} \textbf{U-HAR} is introduced, our proposed CNN model to classify activities from the gathered dataset and describe the network architecture. To assess the performance of our network, we compare in Section \ref{C2sec:Evaluation} the proposed model with a baseline model and investigate the performance improvement that can be achieved by adapting the model with few shot learning. Furthermore, we investigate the impact of sensor modalities on HAR accuracy for the different activities of our set of activities. During the final Section, HAR for context aware smart glasses with an focus on the applicability of our model highlighting challenges w.r.t. sensor integration, power consumption and few shot adaption is discussed and a conclusion from our work is drawn.

\subsection{Related Work}
\label{C2sec:Related work}
A huge amount of smart wearables like smartphones, smart earbuds and smartwatches are equipped with an IMU sensor to measure body motion in an unobtrusive and ubiquitous fashion. This stimulates research on HAR using body motion features. One of the probably most renowned works in this context was the work of \cite{HARAndroidAcc}. In this work, the authors  sampled an IMU sensor of a smartphone with 20\,Hz while 29 participants performed 6 mainly physical activities:  \textit{jogging}, \textit{ascending stairs}, \textit{walking},  \textit{sitting}, \textit{standing} and \textit{descending stairs}. The time series data they gathered was split over the time axis into equidistant slices of 10 seconds duration. From each of the slices the authors derived the following statistical features: \textit{mean}, \textit{time between peaks} and \textit{standard deviation}. The individual features from each windowed time series slice were used to train a decision-tree-classifier. \cite{HARAndroidAcc} achieved with this method an accuracy of 91\,\% by applying 10-fold-cross-validation.

To expand the activity space compared to the physical activity dominated activity space by \cite{HARAndroidAcc}, \cite{HARAudioAccApartment} added a microphone as additional sensor modality to an IMU setup to gather audio data and derive audio features at a sampling frequency of 16\,kHz along with IMU features at a sampling frequency of 20\,Hz.  The authors, recorded IMU and audio data for 19 participants in a rather unconstrained experiment environment and labelled 22 activity classes e.g. \textit{reading}, \textit{cooking}, \textit{sleeping},  \textit{eating}, etc., forming thus a diverse activity dataset composed of 4 cognitive activities, 12 physical activities and 6 mixed activities containing physical as well as cognitive elements. Similar to \cite{HARAndroidAcc}, \cite{HARAudioAccApartment} derived statistical features like (\textit{variance}, \textit{mean}, \textit{correlation} as well as \textit{entropy}) from non-overlapping windows spanning over 1\,s of the total data stream. They achieved an F1-score of 80\,\% using a deep neural network (DNN) while train and predict on windows extracted from the same participant. However, this rather high accuracy drops significantly as soon as a LOPOCV validation scheme is applied. This effect also known as interclass heterogeneity\cite{BullingTutorial}, shows that human activities vary across participants as the same activities are performed in a slightly different ways by different participants, leading to a variation of the statistical features.

To counteract accuracy degradation due to interclass heterogeneity, \cite{HARAudioAccApartment} apply transfer learning and thus retrained a few of the final layers of the DNN model using a small amount of samples stemming from the left-out participant. With the use of transfer learning the authors achieved again an F1-score of 80\,\% even while using the LOPOCV validation method.

In contrast to the previous mentioned works of \cite{HARAndroidAcc} and \cite{HARAudioAccApartment}, \cite{GazeGraph} used an video based eye-tracking sensor (VOG) from Pupil Labs \cite{Fuhl2016} to capture eye-movement data and extract gaze vector information as features during an experiment with eight participants. The participants perform during the experiment the following six mainly cognitive activities: \textit{reading}, \textit{playing online games}, \textit{Browsing}, \textit{typing}, \textit{watching video} and \textit{searching in a list}.

 The authors reported an macro F1-scores of 96\,\% over all participants, when using sliding windows with 30\,s duration to slice the data stream.

The work of \cite{BullingGoogleGlasses} is one of the first works, which recognize human activities by combining  eye- and head-movement features. According to the authors this combination of features allows to classify physical as well as cognitive activities. During the experiment eight participants wore the Google glasses, a commercial smart glasses to acquire data while the participants perform the following activities: \textit{solving a mathematical puzzle}, \textit{watching media}, \textit{talking}, \textit{reading} and \textit{sawing}.
While head motion was recorded with the builtin accelerometer of the google glasses, the recording of eye movements was not possible as the google glasses only provide a proximity sensor to detect blink events. Thus only blink information is considered as eye-movement feature.

From the captured accelerometer data the \textit{averaged variance} over all axes was extracted as head-movement feature. As eye movement features \cite{BullingGoogleGlasses}  derived the \textit{mean} blink frequency and the \textit{center of distribution} from the detected blinks. To evaluate theire approach they trained an individual decision tree classifier per participant using features from both sensor modalities.
The combined evaluation using both sensor modalities yielded an macro F1-score of 82\,\% while \cite{BullingGoogleGlasses} reported a degradation of accuracy to 63\,\% and 67\,\% if they restrict the feature space to solely to head- or eye-movement features.

To gather a richer feature set from eye movements \cite{ActRecogWithCommercialEOGGLasses} used a commercial available glasses with integrated electro oculography (EOG) sensor instead of the Google Glasses. In addition the captured head movement data using an accelerometer. Data from both sensing modalities are combined and finally fed into a k-nearest neighbor (KNN) classifier to classify \textit{talking}, \textit{reading}, \textit{eating} and \textit{typing} gathered from two participants. They achieved an overall accuracy of 70\,\%.

Most recently \cite{CNN_Meyer_2021} propose the combination of an IMU sensors with an optical laser feedback interferometry (LFI) sensor to capture head- and eye-movement data from 15 participants, which performed the seven activities \textit{reading}, \textit{watching media}, \textit{solving}, \textit{typing}, \textit{talking}, \textit{walking} and \textit{cycling}. Instead of statistical features they extract the features automatically using a CNN model and achieve an overall  F1-score of 88.15\,\% by personalizing their network using transfer learning.

Related work indicates that significant effort is made to classify physical activities using data from an body- or headworn IMU sensor and cognitive activities from an eye-tracking sensor. Thus, high recognition accuracy is reported isolated in both sub fields using either IMU sensors for physical activities or eye tracking sensors for cognitive activities. The few works dealing with both eye- and head movement data in a headworn setup rely either on special sensors e.g. the LFI sensor from \cite{CNN_Meyer_2021} or EOG sensors which did not allow for a robust measurement of gaze signals in an everyday device, as the dry electrodes of an EOG sensor require constant good contact with the skin to obtain an optimal EOG signal \cite{ActRecogWithCommercialEOGGLasses}. Furthermore, related work restricted their studies to a small number of participants thus limiting evaluation with respect to generalization for a larger set of participants.

To overcome this limitations and provide context awareness for a broad spectra of smart glasses with integrated VOG sensors, we used a commercial VOG sensor similar to the sensor used by \cite{GazeGraph} and add a commercial available IMU sensor for further collection of head movement data from a large number of participants; resulting in a unique dataset that includes data from physical and cognitive activity domains. Inspired from related work we further propose a state of the art CNN model together with the aspect of transfer learning to improve HAR accuracy in a balanced activity dataset containing physical, cognitive as well as mixed activities, and show the robustness of classification in a large number of participants by applying the LOPOCV method

\subsection{Collection of head- and eye-movement dataset}
\label{C2sec:Material Methods}
The dataset was collected from 20 participants (10 male, 10 female) with a mean age of 27 ($\pm 4.3$). All participants gave their written consent to use their data for research purpose before taking part in the experiment. Further, participants affected by visual impairment were asked to wear contact lenses to ensure a high data quality.
\subsubsection{Apparatus}
\begin{figure}[h]
	\centering
	\includegraphics[width=0.4\linewidth]{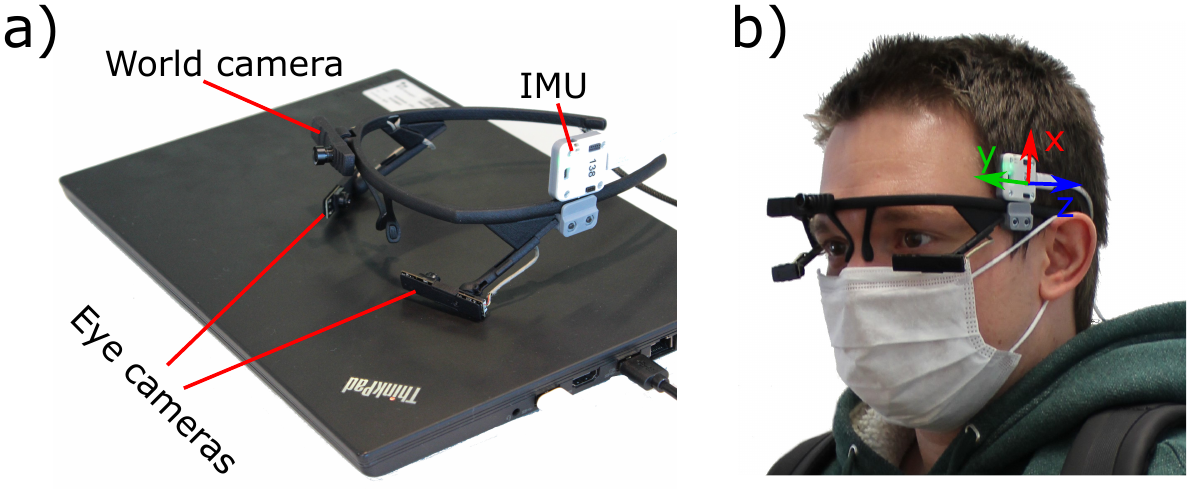}
	\caption{Research apparatus used during the experiments to collect eye- and head-movement data.}
	\label{C2im:Apparatus}
\end{figure}

To collect the data, the participants wore a research apparatus  during the experiment which is shown in Figure \ref{C2im:Apparatus} a).
Similar to \cite{GazeGraph}, we used the commercially available Pupil Labs Core \cite{Kassner:2014:POS:2638728.2641695} VOG eye-tracking system as the basis of our research apparatus, as shown in Figure \ref{C2im:Apparatus} a). The VOG sensor system includes two eye cameras pointed towards the eye, capturing images at 120\,Hz of a participant's left and right eyes. In addition, the frame holds a third camera, which points at the world scene. To collect head movement data, a custom board consisting of a microcontroller and an IMU sensor (BMX055) is mechanically attached to the frame of the Pupil Labs Core. The microcontroller controls and reads the IMU's triaxial accelerometer, the triaxle gyroscope and the triaxial magnetometer at a sampling rate of 120\,Hz. In Figure \ref{C2im:Apparatus} b), a participant wears the visualized apparatus and the axis orientation of the accelerometer and the magnetometer are drawn. The rotational axes of the gyroscope are perpendicular to the indicated acceleration axes.  Both the Pupil Labs Core and the custom IMU board are connected to a laptop via USB to simultaneously collect eye and head movement data during the experiment. 
\subsubsection{Experiment design}
The experiment itself is divided into two parts, a stationary part and a physical part. During the stationary part, participants sit in front of the recording laptop and perform the following activities in the following order \textit{talk}, \textit{read}, \textit{solve}, \textit{watching video} and \textit{type on the keyboard}, which are mainly cognitive and mixed activities. In this part of the experiment, the execution time per activity was not limited in order to elicit a natural behavior of the participants during the execution of the activities.
\begin{figure}[h]
	\centering
	\includegraphics[width=0.7\linewidth]{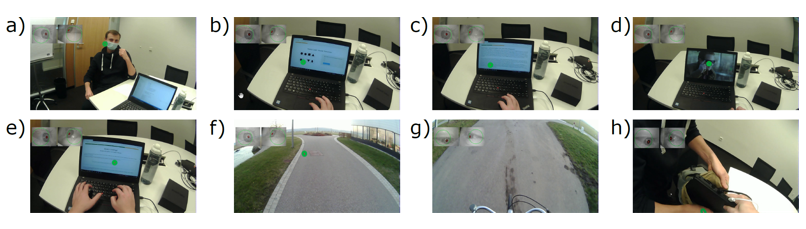}
	\caption{a) - g) frames captured by the world camera during the 7 activities \textit{talk}, \textit{solve}, \textit{read}, \textit{video}, \textit{type}, \textit{walk} and \textit{cycle} in this order. In addition, h) shows a captured frame during the switching between the stationary and the physical experiment part.}
	\label{C2im:Activities}
\end{figure}
During the physical part of the experiment, participants go for a \textit{walk} and ride a bike to \textit{cycle} outdoors. Before the start of the experiment, the Pupil Labs Core eye tracker is calibrated using the built-in single marker calibration method \cite{Calibme} to provide accurate gaze estimation in the participants' FOV . Afterwards, the recording is started and eye movements are recorded utilizing the Pupil Capture (v1.15) software. At the beginning of the experiment, the experimenter initiates a conversation to \textit{talk} with the participant. Afterwards the experimenter leaves the participant alone and the participant is guided by a website on the laptop through the stationary part of the experiment. For the \textit{solve} activity, a logic test with a set of tasks containing a mathematical quiz and extension of a series of geometric patterns was presented to the participant.
Afterwards, the participants \textit{reads} a text about recent advances in the domain of smart glasses and watches a \textit{video}. Afterwards the participant has to answer questions related to the shown text and the video, by \textit{typing} answers on the keyboard of the laptop. After finishing the stationary part,  the laptop is stashed in a backpack by the experimenter to start the physical part of the experiment which consists of the activities \textit{walk} and \textit{cycle}. 

The activities were selected based on related work e.g. by \cite{CNN_Meyer_2021} as well as to span a large variation between pure eye movement related activities (\textit{read}, \textit{video}), mixed activities (\textit{type}, \textit{solve}), social activities (\textit{talk}) and pure head movement dominated activities (\textit{walk}, \textit{cycle}) . 
\subsubsection{Data processing}
After a successful recording, the recorded world and eye videos are exported using the Pupil Player (v2.4). For each frame of the eye videos, the pupil center is extracted based on the provided 3D eye model pupil detection method \cite{Swirski2013}.

\begin{figure}[h]
	\centering
	\includegraphics[width=0.7\linewidth]{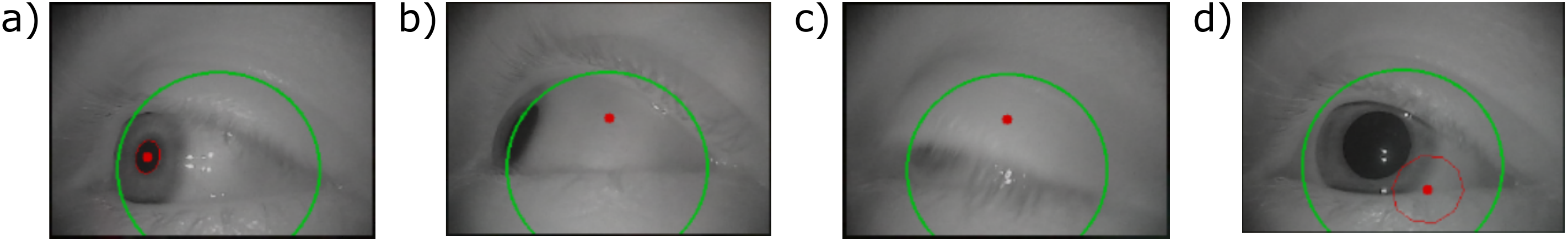}
	\caption{Four different cases of pupil detection (red dot) affect data quality. a) Well detected pupil  b) no detected pupil due to steep camera angle c) no detected pupil during blink d) false detected pupil due to eye lashes}
	\label{C2im:Pupil_errors}
\end{figure}

\begin{figure}[h]
	\centering
	\includegraphics[width=0.9\textwidth,height=0.9\textheight,keepaspectratio]{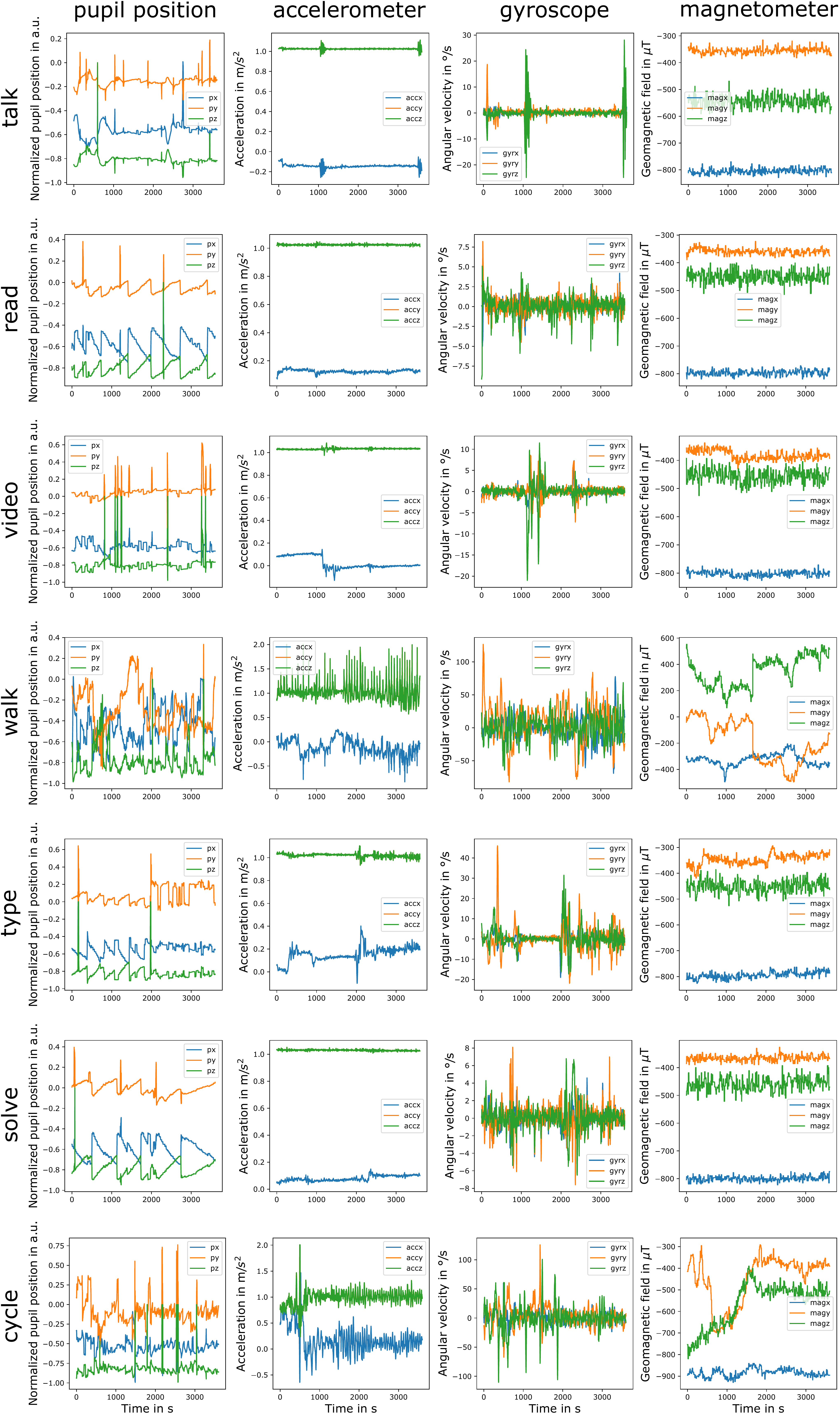}
	\caption{30 second windows of the raw pupil position and IMU features of participant P1 for the seven activities}
	\label{C2im:Raw_data}
\end{figure}

Figure \ref{C2im:Pupil_errors} shows images of the extracted pupil center for four different cases. In some cases, e.g. during a blink (Figure \ref{C2im:Pupil_errors} c)), the pupil could not be detected, also indicated by a low pupil detection confidence value of less than 0.9. In these cases, we forward filled the current pupil position with the last valid pupil position (pupil detection confidence of more than 0.9) to obtain a continuous stream of features over time. Furthermore, we applied a moving median filter with kernel size 10 on the pupil position data stream to smooth out high saccades or jumps due to falsely detected pupils e.g. shown in Figure \ref{C2im:Pupil_errors} d).
\FloatBarrier
In Figure \ref{C2im:Activities}, images captured by the world camera are shown for each activity.  The images are used to manually assign an activity label to each timestamp and thus label the record. Each world video frame is assigned with one of the seven activity labels or \textit{none} if no activity is performed by the participant e.g. during the transition between the stationary part of the experiment and the physical part. As the world camera samples at a sampling rate of $\approx$ 30\,Hz and the eye video frames are sampled at $\approx$ 120\,Hz, four eye video frames are assigned with the activity label extracted from one world video frame. The same labels are assigned to the IMU data stream, which is also sampled at $\approx$ 120\,Hz and synchronized with the pupil labs data stream. Figure \ref{C2im:Raw_data} shows 30\,s windows of the sensor features captured from participant P1 for each performed activity.

To get insights into the collected dataset, Table \ref{C2tab:data_overview} summarizes total duration, mean duration and the standard deviation per activity. A more detailed version on individual participant level is given in \Cref{C2tab:data_overview_large}.

\begin{table}[h!]
	\small	
	\centering
	
	\renewcommand{\arraystretch}{1.1}
	\caption{Total duration, mean duration and standard deviation in minutes of the individual activity over all participants.}
	\begin{tabular}{|l|r|r|r|r|r|r|r|r|r|}
		\hline
		&   \textbf{NULL} &   \textbf{talk} &   \textbf{read} &   \textbf{video} &   \textbf{walk} &   \textbf{type} &   \textbf{solve} &   \textbf{cycle} &   \textbf{total} \\
		\hline
		\textbf{total} &    98 &    137 &    165 &     209 &    237 &    284 &     266 &     119 &    1514 \\
		\textbf{mean} &    1.96 &    6.85 &    8.25 &     10.45 &    11.85 &    14.2 &     13.3 &     5.95 & -    \\
		\textbf{std} &    2.06 &    2.58 &    2.36 &     0.42 &   2.32  &  5.15    &     5.88 &     1.02 &  -   \\
		\hline
			\end{tabular}
	\label{C2tab:data_overview}
\end{table}
\FloatBarrier
In total 1514 minutes of raw data were recorded from 20 participants. The order of stationary and physical part were not fixed. Also experiments take place in different environments like in the office or at the participants home. Thus e.g. cycling was performed on different bikes and walking at different locations to reduce bias by experiment design. The large difference in mean and deviation between activities caused by the experiment design as participants were not restricted in duration per activity to ensure a naturally behavior during the experiment.
\subsection{U-HAR Network architecture}
\label{C2sec:U-HAR Modell}
\begin{figure}[h]
	\centering
	\includegraphics[width=0.8\linewidth]{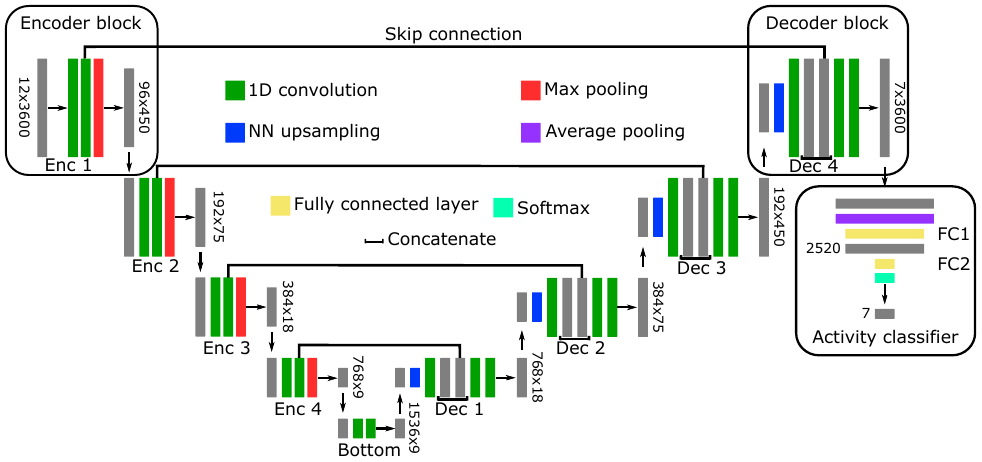}
	\caption{Network architecture of the proposed \textbf{U-HAR} model consisting of a U-Time \cite{perslev2019u} like encoder and decoder structure and a activity classifier based on two fully connected layers}
	\label{C2im:U_HAR}
\end{figure}
\FloatBarrier
To classify the collected data according to the seven activities in the activity set, we propose a CNN structure followed by a fully connected classification structure as shown in Figure \ref{C2im:U_HAR}. The basic network architecture of the so-called U-Nets was introduced by \cite{ronneberger2015u} for two-dimensional biomedical image segmentation and adapted by \cite{perslev2019u} for segmentation of one-dimensional time series data. Perslev et. al. shows a reasonable classification accuracy of sleep stages  \cite{perslev2021u}. Other works, e.g. \cite{elspas2021time}, adapt the network architecture to the domain of autonomous driving to classify lane changes. As HAR refers to the concept of inferring activities based on a time series of observations \cite{BullingTutorial}, which is similar to the applications of Perslev et. al. and Eslpas et. al., we decided to adapt the model architecture and apply it to HAR.
The idea of the encoder and decoder structure is to transform a time series input with $T$ samples from sensors providing $F$ features into a time series output with $T$ samples for $C$ classes, resulting in a single sample classification. Since human activities normally ranges from several seconds to minute scale, classification on a higher time scale above a single sample is mandatory. To this end, an activity classifier was added to the network architecture, which allows activities to be classified over a longer time period. More details about the classifier are provided in Section \ref{C2subsec:Activity_classifier}.

\subsubsection{Data preparation}
To avoid bias introduced by sensor offsets or the scaling of sensor data introduced by different participants performing activities with non-uniform intensity, all data streams of each participant are normalized to unit variance and zero mean. A sequence of normalized sensor readings are combined to create a window with a fixed number of sensor readings $T$ by applying similar to related work a sliding window approach to the input data stream. All sensor readings (pupil position, accelerometer, gyroscope and magnetometer) of one window are extracted and handled as a single sample for classification. The class label $C$ is derived from the most frequent occurrence of a label within a single window. As shown in Table \ref{C2tab:data_overview}, the duration of each individual activity varies across participants. Therefore, the dataset on window scale has to be rebalanced to match the most frequent class by upsampling all underrepresented classes $C$. After applying the data preparation steps, the dataset is transformed into a 3D tensor $ D \in \mathbb{R}^{N\times T \times F} $ with $N$ as the total number of samples and $F$ the number of features per sample. Similar to the work of \cite{GazeGraph} we choose a window length of 30\,s and an overlap of 30\,\% between windows, which leads at a sample rate of 120\,Hz to $T = 3600$. This leads to a total number $N$ of 3318 samples. The IMU provides in total 9 features from the 3D accelerometer, gyroscope and magnetometer and the VOG sensor a 3D pupil position vector, leading in total to $F=12$ features.
\subsubsection{Encoder block}
\label{C2subsec:Encoder}
The encoder consists of four equally built up convolutional blocks. Similar to the initial architecture proposed by \cite{perslev2019u}, each block contains two convolutional layers each with a five dimensional kernel and same padding to preserve the input dimensions. All convolutional layers are using batch normalization. Afterwards, the temporal resolution is decreased by applying max pooling with kernel sizes (8, 6, 4, 2) after each corresponding layer, while increasing the number of filters respectively. In this way, the temporal resolution is decreased from 8.33\,ms to 400\,ms allowing deeper layers to learn more abstract features of human activities. Through this aggressive down-sampling, the number of trainable parameters and therefore the memory footprint is reduced, desirable for operating HAR on embedded hardware inside the glasses frame temple. 
\subsubsection{Decoder block}
\label{C2subsec:Decoder}
Analogous to \cite{perslev2019u}, the decoder consists of four transposed convolution blocks. Each of these blocks performs nearest neighbor (NN) up-sampling to increase the time resolution. Each up-sampling block is followed by a convolutional layer with kernel size of (2, 4, 6, 8) and batch normalization for Dec1 - Dec4, respectively. Afterwards, the feature maps of the up-sampled tensor and the corresponding encoder tensor are concatenated along the filter axis and fed into two convolutional layers, each followed by batch normalization as well to process the feature maps in each decoder block. After Dec4, a pointwise convolution is performed to map the feature map obtained by the encoder decoder structure to a $TxC$ tensor, which assigns each sensor reading in the given window $T$ a classes $C$ confidence score.      
\subsubsection{Activity classifier}
\label{C2subsec:Activity_classifier}
The activity classifier serves as a trainable link between the single sample level class confidence score and the label level assigned on the window size $T$. 

Therefore, the activity classifier maps the class confidence scores of size $TxC$ to a classification output of size $1xC$, effectively removing the temporal axis from the intermediate representation. For this purpose, the activity classifier is added to the output of Dec4,  consisting of an average pooling layer with kernel size of $S = T/K$ and two fully connected layers FC1 and FC2. The average pooling layer calculates the average along the temporal axis leading to an output with size $SxC$ which is flattened and fed into the first fully connected layer FC1, followed by a dropout layer. FC2 maps the output of FC1 to the desired $1xC$ feature vector with a softmax activation. 

Within the proposed structure, $K$ serves as an additional hyper parameter, which can be used to adjust the count of trainable parameters of the activity classifier.  This allows the optimization of the network in terms of recognition accuracy and power consumption as well as memory requirements, making it possible to adapt the network structure, to a certain extent, to the available embedded hardware. 
\subsubsection{Few shot adaption of the activity classifier} 
\label{C2subsec:FSL}
As shown by \cite{HARAudioAccApartment}, a major challenge in human activity recognition is intraclass heterogeneity \cite{BullingTutorial} leading to a degeneration of HAR accuracy. To counteract this degeneration, \cite{HARAudioAccApartment} proposed transfer learning to personalize the HAR classifier. \cite{HAR_CNN_Transfer_Learning} introduced few shot learning (FSL), a distinct form of transfer learning, to personalize a generalized HAR model.

We adapt this method to the \textbf{U-HAR} model by fetching a small fraction of samples per class from the left-out participant after training the \textbf{U-HAR} model using the LOPOCV scheme to personalize the decision boundaries of the activity classifier. During this model adaption phase, the weights of the convolutional layers of the encoder blocks and the decoder blocks and the first fully connected layer (FC1) are frozen. Furthermore, the dropout layer between FC1 and FC2 is removed and the weights of FC2 are reinitialized. Afterwards FC2 is retrained by a few epochs using a small set of samples per class stemming from the leave out participant. By freezing the weights of the encoder and decoder the model keeps its feature extractor build from all but the left out participant and learns a personalized activity classifier bound to the individual participant.

\subsection{Evaluation}
\label{C2sec:Evaluation}
To evaluate the proposed \textbf{U-HAR} model as well as the effect of few shot learning (\textbf{U-HAR-FSL}), we split the dataset $D$ into a test and train subset. to be precise, $D$ is split into $D_{test} \in \mathbb{R}^{N_{test} \times T \times F}$ and $D_{train} \in \mathbb{R}^{N_{test}\times T \times F}$, with $N_{train} + N_{test} = N$. As we used LOPOCV $D_{test}$ contain all samples of the left out participant. With 20 participants, this leads to 20 permutations of $D_{train}$ and $D_{test}$ for which the model is trained and tested separately. To rate the model's overall accuracy the macro F1-score is calculated across all permutations. The macro F1-score gives insights how well the model generalizes across participants and how well it classifies data from a unknown subject. \Cref{C2tab:comp_class_activites} gives an high level overview of the resulting F1-Score of all classifiers per activity as well as the overall macro F1-score.

\begin{table}[]
	\small	
	\centering
	\caption{Average macro F1-score over all participants for each activity for the investigated HAR models.}
	\begin{tabular}{|l|r|r|r|r|r|r|r|r|}
		\hline
		&   \textbf{SVM} &   \textbf{U-HAR} &     \textbf{U-HAR-FSL}  \\
		\hline
		talk    &            \underline{73.03} &      92.15 &         \textbf{97.20}  \\
		read    &         \textbf{91.97} &     \underline{77.74}&         89.42  \\
		video    &           \underline{66.89} &      67.38 &           \textbf{78.60} \\
		walk    &          \underline{89.88} &      95.59 &         \textbf{96.37}  \\
		type    &           \underline{88.07} &     \textbf{88.15}&        86.89 \\
		solve    &           \underline{20.16 }&      \textbf{66.57} &        64.40  \\
		cycle    &           \underline{77.99} &      92.29&       \textbf{95.87}  \\
		\hline
		Overall & \underline{66.86} & 82.09 & \textbf{86.59} \\
		\hline
	\end{tabular}
	
	\label{C2tab:comp_class_activites}
\end{table}
For better comparison between the proposed \textbf{U-HAR} model and the state-of-the-art, we benchmark our dataset using a SVM as baseline model. Afterwards the \textbf{U-HAR} model without and with transfer learning \textbf{U-HAR-FSL} is used to classify activities from the dataset. To ease comparison between the three models, the performance of each classifier is visualized based on the confusion matrices in the upcoming sections stemming from three participants, selected to represent an overall low (P15), an average (P7) and a well (P13) performing participant.
\subsubsection{Baseline model}
\cite{5444879} reported an F1-Score of 76\%  using a SVM model to predict human activities from statistical eye movement features extracted from EOG sensor readings. We extract similar to previous works \cite{5444879, BullingReadingTransit, DAR_HeadAndEye, SteilAttentionForecasting, BullingGoogleGlasses}, the following statistical features \textit{L2-norm}, \textit{variance} and \textit{mean} from the temporal axis of each sensor reading to derive $3xF$ input feature vector from each sample. The SVM is trained on all but the the leave out participant and tested on the left out participant and classification is approached in a one-vs-all setting. As hyper parameters a RBF kernel function is used and we tune the regularization parameter to 1 to maximize classification performance of the SVM. Further optimization of the hyper parameters or the use of additional features derived from the time series data might lead to further improvements of the overall performance.
\begin{figure}[ht]
	\centering
	\includegraphics[width=0.85\linewidth]{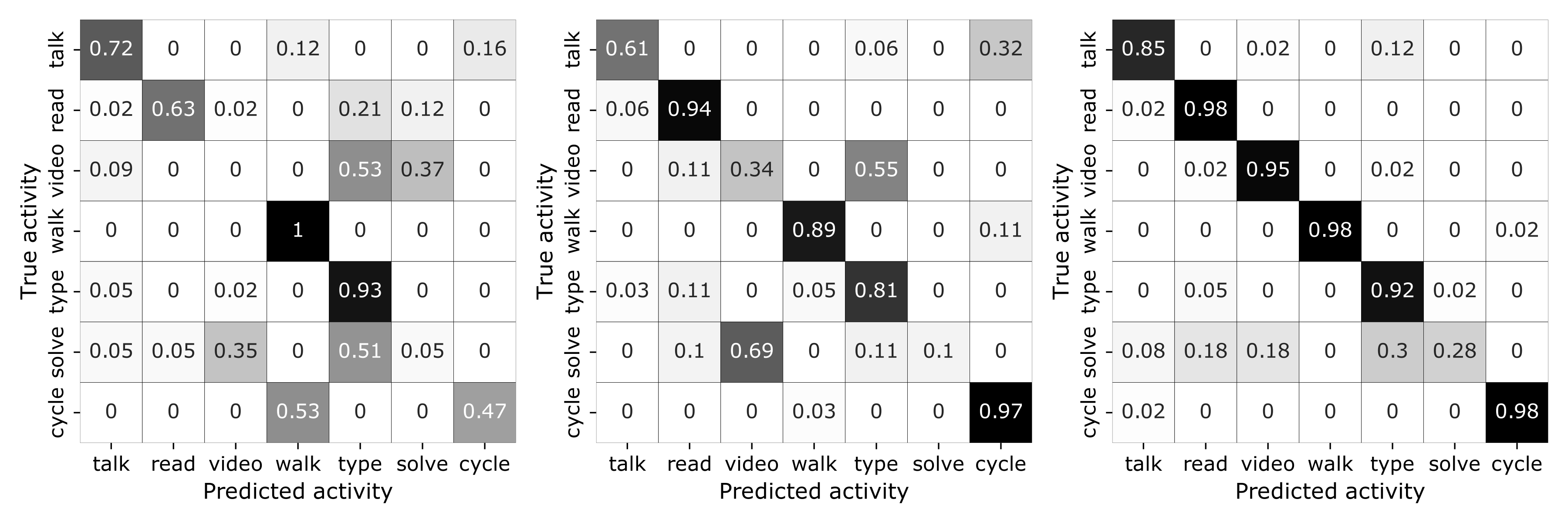}
	\caption{Confusion matrices of P15 (macro F1-score 50.81\,\%), P7 (macro F1-score 59.53\,\%) and P13 (macro F1-score 82.13\,\%) obtained by the SVM classifier based on statistical features.}
	\label{C2im:SVM_Confmatrices}
\end{figure}
\FloatBarrier
Figure \ref{C2im:SVM_Confmatrices} shows three confusion matrices obtained from the selected participants to provide an overview of the classifier's performance with respect to the different activities. The macro F1-score of the SVM classifier is 66.86\,\% with an accuracy of 69.63\,\%, which is roughly 10\,\% below the reported performance from \cite{5444879}. One possible reason could be that the activity set consists of cognitive as well as physical activities, and therefore, the activities are harder to recognize. Another reason could be that statistical features did not serve as an effective representation of the underlying patterns. Of particular note is that miss classification exists only in the subset of cognitive and physical activities for P15. More specifically, the activity \textit{walk} is never miss classified as \textit{read} while confusion exists for \textit{walk} and \textit{cycle}. The confusion matrices of participant P7 as well  as the successfully performing participant P13 show confusion between \textit{solve}, \textit{read} and \textit{type},exhibiting that these classes share a common set of statistical features. This is a well-known problem in HAR and \cite{BullingTutorial} denote it as interclass similarity.
\subsubsection{U-HAR model} 
To get rid of statistical features and improve on the representation of time-dependent patterns, the dataset $D$ is evaluated in the following using the proposed \textbf{U-HAR} model, introduced in Section \ref{C2sec:U-HAR Modell}.
\begin{figure}[ht]
	\centering
	\includegraphics[width=0.85\linewidth]{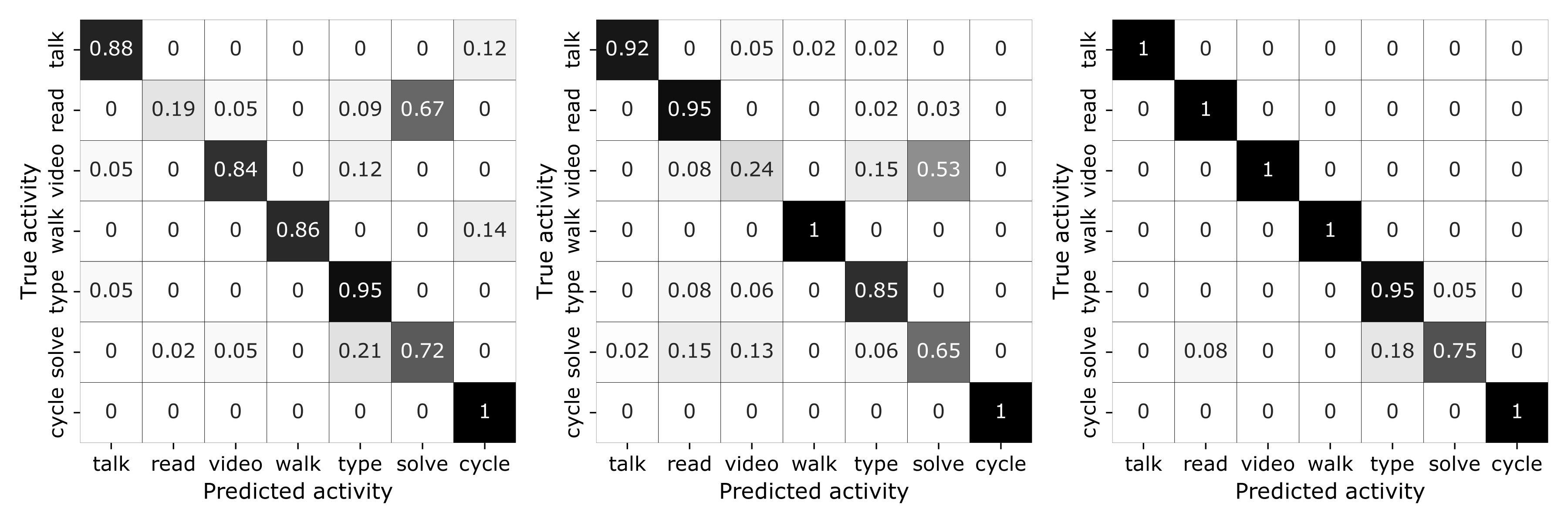}
	\caption{Confusion matrices of P15 (macro F1-score 75.52\,\%), P7 (macro F1-score 75.52\,\%) and P13 (macro F1-score 95.59\,\%) obtained by the \textbf{U-HAR} model introduced in Section \ref{C2sec:U-HAR Modell} without few shot learning.}
	\label{C2im:U_HAR_Confmatrices}
\end{figure}
\FloatBarrier
Similar to the SVM classification approach, the dataset is split into a training set consisting of all participants except the left-out participant, resulting in 20 permutations of training and test splits.
For all 20 permutations, the model is trained over 6 epochs using cross-entropy-loss and Adam optimization with a learning rate of 4E-5 and a weight decay of 1E-6. Furthermore the learning rate was reduced by 1E-6 per epoch. In comparison with the SVM baseline model, the macro F1-score improves significantly by 15.23\,\%, reaching 82.09\,\% with a accuracy of 82.84\,\%. For the average pooling layer, we choose a kernel size of 10 ($K$\,=\,360), effectively down sampling the segmented output to $360x7$, as a good trade off between network size and classification accuracy. Comparing the confusion matrices in Figure \ref{C2im:U_HAR_Confmatrices} with the corresponding confusion matrices in Figure \ref{C2im:SVM_Confmatrices}, there is a clear improvement in recognition for the activity \textit{talk} among the three participants. Furthermore, the physical activities \textit{cycle} and \textit{walk} are well recognized among all participants as well. As with the SVM model for participants P15 and P7, there still exists confusion between the cognitive activities \textit{read}, \textit{solve} and \textit{video}, which hints to interclass similarity.
	
\subsubsection{U-HAR model with few shot learning}
\begin{figure}[ht]
	\centering
	\includegraphics[width=0.85\linewidth]{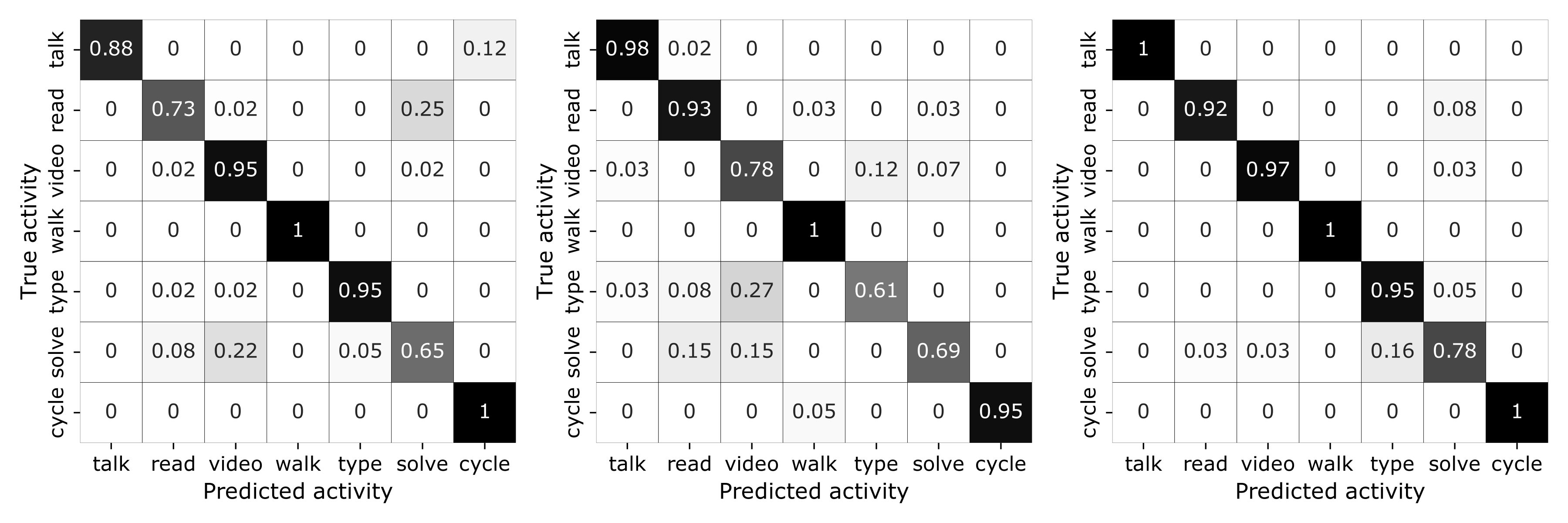}
	\caption{Confusion matrices of P15 (macro F1-score 89.87\,\%), P7 (macro F1-score 84.74\,\%) and P13 (macro F1-score 94.57\,\%) obtained by the U-HAR model after adapting the decision boundaries of the activity classifier block by applying few shot learning.}
	\label{C2im:TL_Confmatrices}
\end{figure}
\FloatBarrier
To further improve the accuracy of activity recognition, we apply few shot learning as proposed in Section \ref{C2subsec:FSL} to the activity classifier. For this purpose, we randomly select three samples (windows of 30 seconds) for each activity class of the omitted participant and retrain the last fully connected layer over 10 epochs using a learning rate of 1E-4 and an exponential learning rate decay of 0.99. Figure \ref{C2im:TL_Confmatrices} shows the three confusion matrices derived from the chosen participants after few shot learning. The macro F1-score increased to 86.59\,\% and the mean accuracy increased to 86.99\,\%. Across all three participants, the recognition of the activities \textit{read}, \textit{solve} and \textit{video} improves. As we personalize the activity classifier to some extent by adjusting the decision boundaries through few shot learning, the interclass similarity observed in  Figure \ref{C2im:SVM_Confmatrices} and Figure \ref{C2im:U_HAR_Confmatrices} resolved.  

\subsubsection{Model comparison}
To compare the performance of the classification approaches, we analyze the performance of the three models across all participants in the experiment. Table \ref{C2tab:comp_class} summarizes the model performances for each participant. In nearly all cases, the SVM classifier performs worst leading to the overall lowest macro F1-score, indicating that the used statistical features does not lead to a good representation of the underling temporal patterns to  recognize activities. The SVM model reaches similar performance as related work by \cite{ActRecogWithCommercialEOGGLasses}. The \textbf{U-HAR} model improves the performance as it automatically extract features to represent the underling temporal patterns much better and due to the convolutional layers it is capable of to also gather information from the temporal sequence of incoming data. The macro F1-score increases to be comparable with related work e.g. \cite{BullingGoogleGlasses}. The additional personalising of the activity classifier through transfer learning, as mentioned in Section \ref{C2subsec:Activity_classifier}, an additional increase in activity recognition performance is observable. The proposed \textbf{U-HAR} model combined with few shot learning outperforms the related work e.g. by \cite{BullingGoogleGlasses, ActRecogWithCommercialEOGGLasses}. It also reaches similar activity recognition performance when compared to related work that solely focus on  activities from the physical domain recognized by IMUs e.g. by \cite{HARAndroidAcc}.
\begin{table}[h]
\centering
\setlength{\tabcolsep}{1pt}
\small
\caption{Macro F1-score of the baseline model, the \textbf{U-HAR} as well as \textbf{U-HAR-FSL} model over all participants. The highest achieved macro F1-scores for each participant over all models are bold marked while the lowest F1-scores are underlined.}
\resizebox{\linewidth}{!}{%
\begin{tabular}{|l|llllllllllllllllllll|l|} 
\hline
                   & P1             & P2             & P3             & P4             & P5             & P6             & P7             & P8             & P9             & P10            & P11            & P12            & P13            & P14            & P15            & P16            & P17            & P18            & P19            & P20            & $\varnothing$   \\ 
\hline
\textbf{SVM}       & \underline{56.22}  & \underline{47.77}  & \underline{57.84}  & \underline{62.45}  & \underline{69.87}  & \underline{88.89}  & \underline{59.43}  & \underline{80.87}  & \underline{63.02}  & 72.62          & \underline{62.44}  & \underline{82.12}  & \underline{60.47}  & \underline{50.81}  & \underline{68.97}  & \underline{67.56}  & \underline{73.3}   & \underline{78.47}  & \underline{57.70}  & \underline{76.35}  & \underline{66.86}   \\ 

\textbf{U-HAR}     & 75.40          & 80.31          & 83.13          & 74.69          & 76.66          & 93.67          & 78.71          & 81.78          & 80.55          & \textbf{79.99} & 77.56          & \textbf{95.59} & 81.98          & 75.52          & 77.98          & 91.65          & \textbf{83.36} & 81.54          & \textbf{79.12} & 92:59          & 82.09           \\ 

\textbf{FSL} & \textbf{91.59} & \textbf{88.28} & \textbf{87.11} & \textbf{78.56} & \textbf{89.47} & \textbf{95.76} & \textbf{84.74} & \textbf{87.96} & \textbf{87.96} & 71.30          & \textbf{77.67} & 94.57          & \textbf{86.20} & \textbf{89.87} & \textbf{86.03} & \textbf{95.20} & 75.03          & \textbf{93.91} & 77.63          & \textbf{95.04} & \textbf{86.59}  \\
\hline
\end{tabular}
}
\label{C2tab:comp_class}
\end{table}
To analyze which activity profits from each model design we further calculate per activity and model the average macro F1-score over all participants.  In Table \ref{C2tab:comp_class_activites} the results are summarized. The physical dominated activities \textit{walk} and \textit{cycle} show together with the mixed activity \textit{talk} the an overall high accuracy over all HAR models. Especially the \textit{solve} activity improves by automatic extraction of features while taking into account the sequence data. The few shot adaption scheme improves the recognition of \textit{read} and \textit{video}, which are cognitive activities mainly described by eye movement features. The improvement through few shot adaption indicates that in particular for this classes intraclass heterogenity is dominating. Especially for reading intraclass heterogenity is well known as each individual has a slightly different scan pattern during reading mainly influenced how trained an individual is on this task. Furthermore a high interclass similarity exists for the activities \textit{video} and \textit{solve}, which leads to a high confusion between the activities and thus a low accuracy. To improve on this additional features are required. This can either be derived from the existing sensors e.g. pupil diameter variation or blinks as used by \cite{BullingGoogleGlasses} or by adding additional senor modalities like an microphone as proposed by \cite{HARAudioAccApartment}. Another potential improvement is to step away from fine grained activities to classification on a higher contextual hierarchy like proposed by \cite{Bulling2013HighLevelCues} e.g. classify activities into physical context or cognitive context to control the glasses UI based on this higher context levels. Finally the \textbf{U-HAR} model provides in the final stage of the decoder as intermediate representation  a sample wise activity classification. From this representation further information can be drawn e.g. to distinguish composed activities containing multiple sub activities. Furthermore inter class similarity could be investigated in at this stage. 
\begin{figure}[ht]
	\centering
	\includegraphics[width=0.5\linewidth]{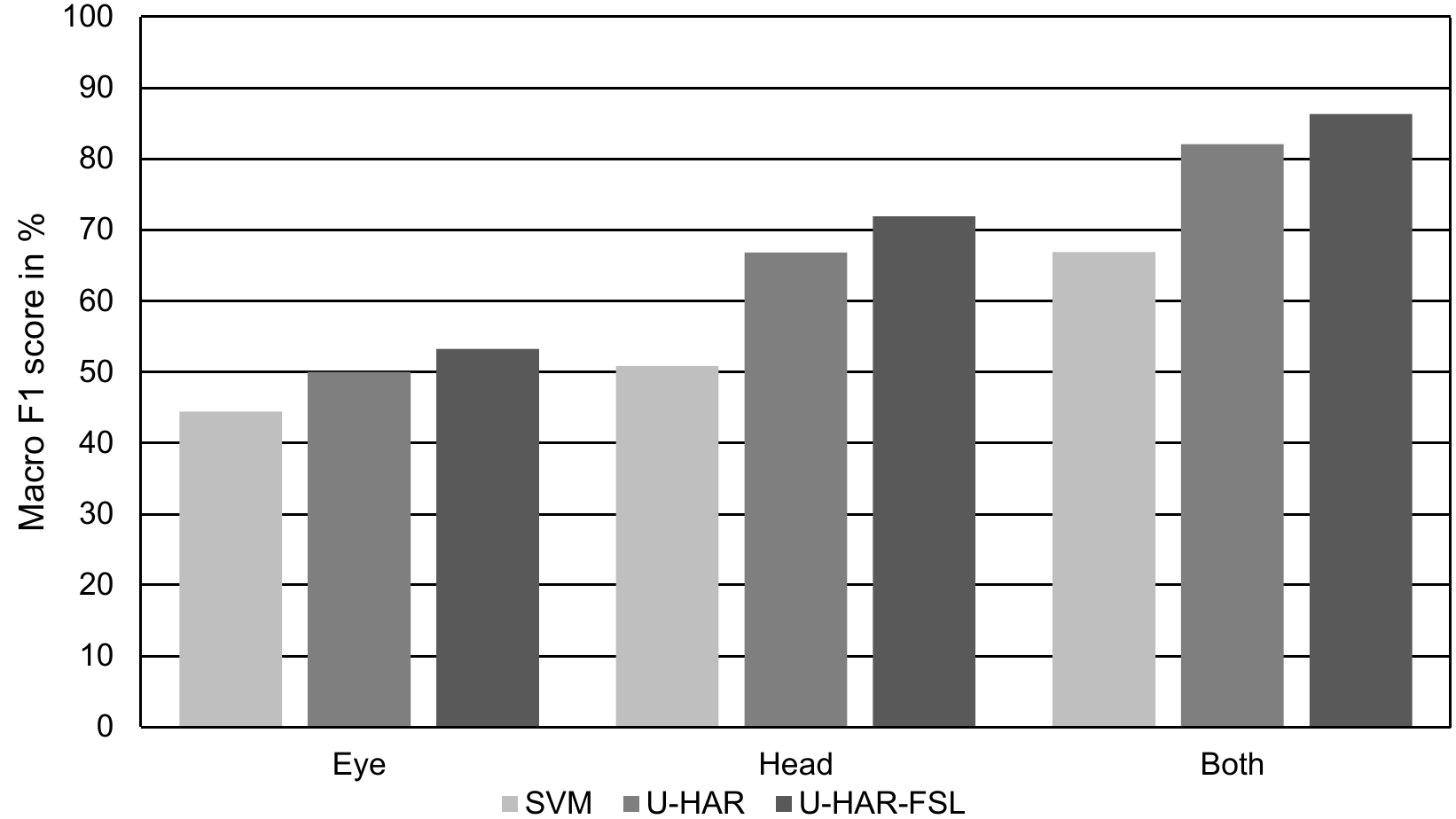}
	\caption{Effect of eye related features (pupil position) and head related IMU features on the macro F1-score.}
	\label{C2im:Abliation study}
\end{figure}
\FloatBarrier
\subsubsection{Impact of the sensing modalities}
To analyze effects of both sensing modalities,  we conducted a study on the importance of these sensor modalities. More specifically, the three models (i.e., SVM, \textbf{U-HAR} and \textbf{U-HAR-FSL}) are retrained with either the head movement related features (IMU) or the eye movement related features (VOG). Figure \ref{C2im:Abliation study} summarizes the resulting macro F1-scores after retraining the three models with only head or eye movement features. Our results in Table \ref{C2tab:ablation_activities} show that a high classification score is only obtained if both head and eye movement features are taken into account. At least for activity sets which contain physical and cognitive activities the use of both sensor modalities is mandatory.  The head movement features have a higher impact on classification accuracy compared to the eye movement features for all three classifiers. Furthermore it is shown that the high classification accuracy of cognitive activities \textit{read} and \textit{video} is mainly driven by the eye movement features while the mainly  physical activities \textit{walk} and \textit{cycle} are mainly classified by the head movement features.
\begin{table}[ht]
	\small
	\centering
	\caption{Averaged resulting macro F1-scores per activity when solely considering head movement features or eye movement features and their combination. For each classifier  and sensor modality, both the lowest and highest F1-scores are highlighted.}
	\renewcommand{\arraystretch}{1.1}
	\begin{tabular}{|l|c|c|c||c|c|c||c|c|c|}
		\hline
		& \multicolumn{3}{c}{SVM}   &   \multicolumn{3}{c}{U-HAR}   &   \multicolumn{3}{c|}{U-HAR-FSL} \\
		& Eye & Head & Both & Eye & Head & Both & Eye & Head & Both \\
		\hline
		talk  &24.05			&86.15 				&73.03				&30.38	&90.96 				&92.15 &			\underline{31.92} 		&97.21 & 97.20\\
		read  &\textbf{92.84}	&\underline{14.07}	&\textbf{91.79}  	&\textbf{80.59}		&\underline{28.43}	&77.74 &			\textbf{88.45} 				&\underline{27.30} & 89.42 \\
		video &73.85 			&58.49 				&66.89				&40.26				&77.25				&67.38 &			51.86 				&78.04 & 78.60 \\
		walk  &68.14 			&\textbf{88.11} 	&89.88 				&65.36				&\textbf{94.61} 	&\textbf{95.59} &	58.15 				&96.54& 96.37\\
		type  &72.14 			&51.18 				&88.07				&76.94				&71.35 				&88.15 &			68.78				 &84.43 & 86.89\\
		solve &\underline{8.45}	&14.46				&\underline{20.16 }	&52.59 				&35.29 			&\underline{66.57} &42.55	 &38.39 & \underline{64.40} \\
		cycle &11.71 	&75.67 		&77.99 			&\underline{25.22}			&91.90 				&92.29 	&			45.93 				&\textbf{95.87} & \textbf{95.37}  \\
		\hline
	\end{tabular}
	\label{C2tab:ablation_activities}
\end{table}
\subsection{Discussion and limitations}
As shown in Section \ref{C2sec:Evaluation}, the combination of eye and head motion features leads to higher HAR accuracy. Therefore, the proposed approach of combining eye and head movements and fusing them through the \textbf{U-HAR} network enables HAR-based context awareness for smart glasses. In order to apply the proposed system in a real application, several aspects needs to be considered, which are discussed below.

\subsubsection{Sensor integration}
While integrating IMU sensors into smart glasses is straightforward as they are already integrated into every smart wearable, integrating eye-tracking sensors seems to be more challenging. The power consumption of VOG eye tracking sensors is magnitudes higher compared to IMU sensors, and the size and the position requirements of the video cameras in order to obtain robust pupil detection over the whole FOV of an participant are challenging, as shown in Figure \ref{C2im:Pupil_errors}. To overcome the limitations of power consumption and sensor integration, \cite{7181058, Mindlink2021} and \cite{9149591, Meyer2020e} proposed new eye tracking sensor approaches to replace the camera sensor and facilitate integration by using scanned infrared lasers to track a person's eye.

\subsubsection{Power consumption}
The power consumption of smart glasses is dominated by the power consumption of the display, similar to smartphones. Therefore, a smart glass system directly benefits from a reduction of the display's duty cycle. The contextual control of the display proposed in this paper is a possible solution to this issue, as it allows the display to be activated contextually. This implies an always-on HAR with continuous classification of activities by the \textbf{U-HAR} model. Besides power memory is a further constrained resource on an embedded processor on the glasses. However, as the \textbf{U-HAR} model mainly rely on convolutional layers, energy efficient deployment of the model e.g. by using tensor processing units (TPUs) or dedicated DNN accelerators is possible \cite{garofalo2020pulp}, \cite{Bringmann_CNN_acc}. The remaining constrain is the small available memory on embedded devices. The memory footprint of the \textbf{U-HAR} model is mainly determined by the parameters of FC1 and FC2. With an average pooling kernel size of 10 ($K$\,=\,360), this two layers hold 6.373.087 parameters, which are 98.61\% of the total network's parameters. A straightforward way to decrease this rather large number of parameters is the adoption of the hyper parameter $K$. In fact ($K$\,=\,50), results in a kernel size of 72 of the average pooling layer, reduces the parameters by a factor of 24.6 while keeping a sufficient macro F1- score of 80.53\,\% . Further optimization of the model architecture e.g. by improving the 1D CNN layers e.g. by adaption of bottleneck layers as proposed by \cite{sandler2018mobilenetv2} or by model pruning or other model optimization techniques are still possible \cite{Edge_Comp_Review} to reduce model size and thus computational complexity. Furthermore the IMU features can be fused into a 3D orientation vector of the head to further reduce network parameters.

\subsubsection{Few shot learning}
As suggested by related work to improve classification performance and handle intraclass heterogeneity we used few shot learning to enhance our \textbf{U-HAR} model. This require to obtain labels from the user to adapt FC2 and thus personalize the activity classifier. The label collection may degrade user's experience although it is a known operation from other wearables e.g. the initial configuration of face-recognition in a smartphone \cite{wang2020deep}. 

\subsubsection{Privacy}
The proposed \textbf{U-HAR} system relies on privacy preserving sensor features as no direct link between the features and the individual exists. If inference of contextual information occurs on the edge,  as proposed in the previous sections, no raw potential private data  needs to be transmitted.  Furthermore the proposed system did not require a world camera sensor to derive contextual information, which adds on the social acceptance of such a glasses system.    

\subsection{Conclusion}
In this work, we present an approach which enables context-awareness through human activity recognition for smart glasses. We built a research apparatus and conducted experiments with 20 participants to record and published a unique activity dataset. Our dataset contains a distinctive combination of seven activities stemming from physical and cognitive domains. We recorded eye movement features using a VOG sensor and head movement features using an IMU sensor for all activities leading to a total recording duration of 1514 minutes. We further adapted the U-Time CNN model architecture proposed by \cite{perslev2019u}, applied it to human activity recognition and proposed the \textbf{U-HAR} model structure. We advanced the network architecture to enable few shot adaptation to personalize the activity classifier to each individual and counteract accuracy losses due to intraclass heterogeneity. With the few shot adaptation approach, we achieved a macro F1-score of 86.59\,\% using LOPOCV. This result significantly outperforms related work. As part of our future work, we will focus on deployment of the classifier in a smart glasses prototype. In addition, we like to investigate how to gather labels from users in a every-day setting to apply the proposed few shot adaption to increase classification performance and thus minimize missclassification to optimize usability. Finally we like to investigate how a user can extend the classifier with its personal activities in a on device setting. 

\subsection*{Acknowledgements}
Enkelejda Kasneci is a member of the Machine Learning Cluster of Excellence, EXC number 2064/1 - Project number 390727645.

%

\subsection{Supplement Material}
\FloatBarrier
\newpage
\begin{table}[h!]
	\small	
	\centering
	
	\renewcommand{\arraystretch}{1.1}
	\caption{Duration in minutes of each individual experiment itemized by activity.}
	\begin{tabular}{|l|r|r|r|r|r|r|r|r|r|}
		\hline
		&   \textbf{NULL} &   \textbf{talk} &   \textbf{read} &   \textbf{video} &   \textbf{walk} &   \textbf{type} &   \textbf{solve} &   \textbf{cycle} &   \textbf{total} \\
		\hline
		P1    &      1 &      4 &      6 &      11 &     16 &      9 &       6 &       7 &      59 \\
		P2    &      6 &      8 &      7 &      11 &     14 &     18 &      12 &       5 &      79 \\
		P3    &      6 &     12 &      9 &      10 &     13 &     28 &      17 &       7 &     103 \\
		P4    &     11 &      6 &     13 &      11 &     10 &     13 &      35 &       5 &     104 \\
		P5    &      4 &      9 &      8 &      10 &     11 &     17 &      12 &       5 &      78 \\
		P6    &      4 &      5 &      7 &      10 &     14 &     13 &       7 &       6 &      67 \\
		P7    &      8 &     14 &     11 &      10 &      8 &     22 &      18 &       6 &      96 \\
		P8    &      5 &      6 &      7 &      10 &     10 &      9 &       7 &       7 &      60 \\
		P9   &      6 &      5 &     10 &      11 &      8 &     14 &      13 &       7 &      74 \\
		P10   &      4 &      7 &      5 &      10 &     11 &     14 &      12 &       5 &      68 \\
		P11   &      2 &      7 &      6 &      10 &     12 &      7 &      11 &       6 &      61 \\
		P12   &      3 &      9 &      5 &      10 &     14 &      6 &      14 &       6 &      68 \\
		P13   &      4 &      6 &      9 &      10 &     14 &     13 &      11 &       5 &      72 \\
		P14   &      6 &      4 &     12 &      11 &     10 &     15 &      12 &       5 &      75 \\
		P15   &      5 &      5 &     11 &      10 &     14 &     12 &      16 &       6 &      78 \\
		P16   &      4 &      4 &      8 &      10 &     13 &     13 &      12 &       6 &      71 \\
		P17   &      5 &      6 &      7 &      10 &     13 &     22 &      14 &       9 &      87 \\
		P18   &      6 &      8 &      8 &      10 &     10 &     11 &      13 &       5 &      72 \\
		P19   &      5 &      5 &     11 &      11 &      8 &     15 &      16 &       5 &      75 \\
		P20   &      4 &      5 &      5 &      10 &     14 &     13 &       9 &       6 &      67 \\
		\hline
		\textbf{total} &    98 &    137 &    165 &     209 &    237 &    284 &     266 &     119 &    1514 \\
		\hline
	\end{tabular}
	\label{C2tab:data_overview_large}
\end{table}

\newpage

\chapter{Static Laser Feedback Interferometry Eye Tracking}
\label{APP:Static_LFI_Eye_Tracking}
This chapter contains the manuscript:
\begin{enumerate}[label=\Roman*.]
	\item \textbf{Johannes Meyer}, Stefan Gehring,  Enkelejda Kasneci. "Static Laser Feedback Interferometry based Gaze Estimation for Wearable Glasses". Submitted to IEEE Transactions on Systems, Man, and Cybernetics: Systems (2022)  
\end{enumerate}
\section{Static Laser Feedback Interferometry based Gaze Estimation for Wearable Glasses}
\label{APP:C1}
\subsection{Abstract}
Fast and robust gaze estimation is a key technology for wearable glasses, as it enables novel methods of user interaction as well as display enhancement applications, such as foveated rendering. State-of-the-art video-based systems lack a high update rate, integrateability, slippage robustness and low power consumption. To overcome these limitations, we propose a model-based fusion algorithm to estimate gaze from multiple static laser feedback interferometry (LFI) sensors, which are capable of measuring distance towards the eye and the eye's rotational velocity. The proposed system is ambient light robust and robust to glasses slippage. During evaluation a gaze accuracy of 1.79$^\circ$ at an outstanding update rate of 1\,kHz is achieved while the sensors consume only a fraction of the power compared to state-of-the-art video-based system.

\subsection{Introduction}
\label{D1Introduction}
Eye tracking is a key sensing technology for wearable glasses required to enable applications ranging from gaze contingent interaction \cite{7464272, Bace2020, Meyer2021} to display enhancement methods like foveated rendering \cite{Kaplanyan:2019:DNR:3355089.3356557,Kim:2019:FAD:3306346.3322987, 9005240} or exit pupil steering\cite{Ratnam:19, Xiong:21}. State of the art eye tracking systems rely on video oculography (VOG), which is a camera-based system to infer the users gaze from a sequence of images.  However, due to power constraints imposed by the wearability of these glasses, the update rate of VOG systems is limited by the high power consumption of the camera sensors and the image processing algorithms required to estimate user's gaze from captured images \cite{fuhl2020tiny}. Especially see-through AR glasses further require an eye tracking system, which operates robustly in the presence of ambient light \cite{Fuhl2016}. This further limits the applicability of VOG systems. Finally, VOG systems are prone to glasses slippage \cite{10.1145/3314111.3319835}, possibly leading to a degradation of gaze estimation accuracy \cite{niehorster2020impact,holmqvist2022eye}.

To overcome these limitations of VOG systems, we introduce a static laser feedback interferometry (LFI) sensor approach. The LFI sensor itself consists of a tiny (160\,$\mu$m $\times $ 180\,$\mu$ m) vertical cavity surface emitting laser (VCSEL) with  infrared (IR) light at 850\,nm (invisible to the user) and a photo detector integrated into the laser cavity \cite{grabherr2009integrated}. By applying a frequency modulated continuous wave (FMCW) modulation scheme \cite{NORGIA201731}, the sensor is capable to measure the distance $d$ towards the eye as well as the eye's rotational speed in beam axis $v_T$ simultaneously with an outstanding update rate of 1\,kHz, while consuming only a fraction of the power of VOG systems. Due to the coherent sensing scheme, the sensor in addition is robust against ambient light, as shown in \cite{Meyer2020e}. Our contribution in this work is two-fold:

(i) We introduce the \textbf{static LFI sensor} modality for gaze estimation, characterize a static LFI sensor with respect to distance and velocity resolution in a near-eye setting, propose an geometric eye model required for sensor fusion, and build up a simulation tool to generate measurements of multiple static LFI sensors.

(ii) We propose a slippage robust, calibration-free \textbf{gaze estimation algorithm}, fusing multiple static LFI sensors in order to reconstruct the gaze vector based on a geometric eye model.

The remainder of this work is organized as follows: \Cref{D1Related_Work} gives an overview over model-based eye tracking approaches of VOG systems as well as IR laser-based eye tracking approaches with focus on LFI sensors. Afterwards in \Cref{D1sec:LFI}, the LFI sensing principle is introduced together with a geometric eye model in order to link LFI sensor measurements to an eye pose. In addition, a LFI simulation tool is introduced to generate LFI sensor measurements in a multi LFI sensor setting. In \Cref{D1Gaze Reconstruction Algorithm}, the gaze reconstruction algorithm is introduced to fuse multiple LFI sensors for gaze estimation. In \Cref{D1Evaluation}, we characterize distance and velocity noise of the LFI sensor and evaluate the gaze estimation accuracy of the proposed system in presence of glasses slippage. In the final sections, system limitations are discussed, and a final conclusion is drawn.

\subsection{Related Work}
\label{D1Related_Work}
Related work is divided into two parts. During the first part, geometric model-based gaze estimation algorithms used in VOG systems are introduced to highlight the differences between existing geometric model-based gaze estimation algorithms and our novel approach. In the second part, related work using the LFI sensor in a near-eye setting is presented.
\subsection{Geometric model-based Eye Tracking}
Geometric model based eye tracking approaches used in VOG systems can be divided into corneal reflection eye models and glint-free eye models \cite{8003267}.
\begin{figure}
	\centering
	\includegraphics[width=0.7\linewidth]{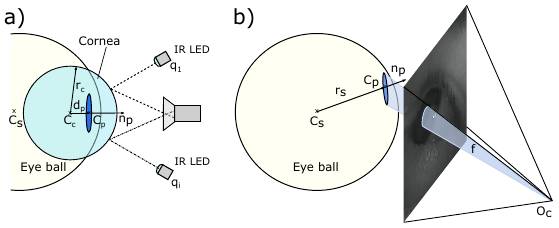}
	\caption{a) Corneal reflection eye model based on \cite{Hennessey2006}.  b) Glint-free eye model based on \cite{Swirski2013}.}
	\label{D1fig:geometric_models}
\end{figure}
Corneal reflection models, as shown in \Cref{D1fig:geometric_models}a), estimate the gaze vector, described as normal vector of the pupil $n_p$, by extracting the center of the pupil $P_c$ from a camera image and estimate the center of the cornea $C_\mathrm{c}$. To estimate the cornea center, IR LEDs ($q_1$ - $q_n$) are added to the system. The diverging IR light of the IR LEDs leads to specular reflections on the surface of the cornea, so-called glints. The positions of the glints on the cornea are extracted from the camera images in order to estimate the cornea center $C_\mathrm{c}$ \cite{1634506}. According to Hennessey et al. \cite{Hennessey2006} at least two glints are required to estimate the cornea center under the assumption that the system geometry is fixed and known in advance. For reconstruction of the cornea center $C_\mathrm{c}$ furthermore the radius of the cornea $r_c$, the distance between pupil and corneal center $d_p$ and the refraction index of the aqueous humor need to be known or derived from average population \cite{Hennessey2006}.  

In contrast, glint-free models rely solely on a camera sensor as shown in \Cref{D1fig:geometric_models}b) \cite{Swirski2013}. To estimate the gaze vector $n_p$, the center of the sclera $S_\mathrm{c}$ and the center of the pupil disk $P_\mathrm{c}$ in space need to be estimated from camera images. Assuming a pinhole camera model and a projected ellipse representing the pupil in the image plane, the pupil disk with its center $P_\mathrm{c}$ is derived from a cone projection $f$ in space. The cone $f$ is constructed from the pupil ellipse representation in the image plane and the camera focal point $o_c$. As the pupil diameter varies with the portion of ambient light, and the diameter of the pupil ellipse representation depends on the distance between camera sensor and the eye, several singularities exist, which prevent the estimation of the pupil disk and thus the gaze vector from a single image. By introducing geometric constraints like a fixed pupil diameter, possible pupil disk candidates fitting to the projected cone can be derived. By adding further constraints, the correct pupil disk can be estimated from a sequence of frames \cite{Swirski2013}.

Under the assumption that the eye ball diameter  $r_s$ is fixed and known in advance, the center of the eye ball $S_\mathrm{c}$ is derived from a sequence of pupil disk observations, assuming a stationary relation between camera position and eye ball center $S_\mathrm{c}$. This assumption holds for a short time period, while for longer time periods, glasses tend to slip on the nose and thus the model parameter $S_\mathrm{c}$ needs to be updated continuously  \cite{10.1145/3204493.3204525, niehorster2020impact}. 

In summary, VOG eye tracking approaches rely on a reduced anatomical model of the human eye with minimum physical dimensions and optical quantities in order to link the sensor features to an eye pose. In addition, fixed model parameters derived from the average population are used as additional system constraints. Hennessey et. al. \cite{Hennessey2006} assume a known system geometry and a rigid glasses frame with mounted IR LEDs and the camera sensor. Swirski et. al. \cite{Swirski2013} further assume a stationary fixed relation between eye coordinate space and glasses coordinate space. In addition, both model-based approaches estimate the optical axis of the eye as normal vector  $n_p$ of the pupil.

\subsection{Static LFI eye tracking}
Static LFI sensors are well known from literature and are used in a wide range of applications, like vibrometry or velocimetry \cite{Giuliani2002, NORGIA201731}. The first work which proposes a static LFI sensor to measure eye movements was released by Capelli et al. \cite{5954803}. The authors used a 1310\,nm IR laser with an optical power of 4\,mW to measure velocity of a rotating disk to simulate eye rotational velocities. Measurements on a real human eye were not performed, as the laser source did not fulfill eye safety requirements. The authors apply a triangular current modulation scheme to the drive current of the laser to be able to measure rotational velocities from 0\,$^\circ$/s to 500\,$^\circ$/s at a sample rate of 60\,Hz. They reported a velocity measurement error of 10\,\%. 

Meyer et al. \cite{Meyer2020e} used an unmodulated LFI sensor together with a 2D micro scanner to capture reflectivity images from the eye and to measure the bright pupil response. They reported an outstanding ambient light robustness of LFI sensors due to the coherent sensing scheme of the sensor.

Meyer et al. \cite{meyer11788compact, Meyer2021, CNN_Meyer_2021} were also the first to create an eye safety compliant static LFI sensor system and reported measurements on a real human eye. They used two LFI sensors at a wavelength of 850\,nm with an optical power below 500\,$\mu$W to measure distance towards the eye as well as relative eye velocities with an update rate of 1\,kHz. The authors restricted their velocity measures to relative eye velocities, because the static LFI sensor was only capable to measure speed in direction of the laser beam axis $v_T$ and thus rotational movements with perpendicular components could not be measured with their setup.

This work builds up on previous works by Meyer et. al. \cite{Meyer2020e, meyer11788compact, Meyer2021, CNN_Meyer_2021}  to apply static LFI eye sensors in near-eye setting to develop a high-speed gaze estimation method based on LFI sensor technology.
%

%
%
%
%
%
%
%
%
%
%
%

\subsection{Laser Feedback Interferometry Eye Tracking}
\label{D1Material_and_Methods}
\label{D1sec:LFI}

\begin{figure}
	\centering
	\includegraphics[width=0.7\linewidth]{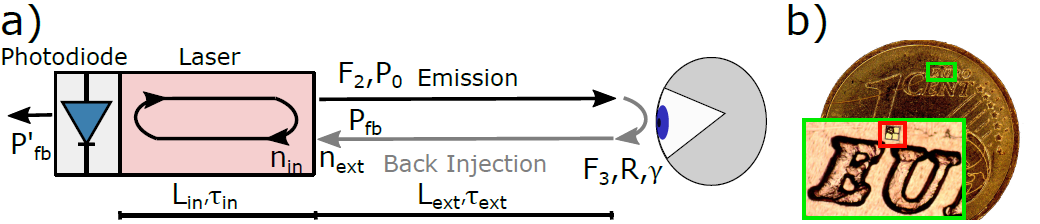}
	\caption{Left: Sensing principle of an LFI sensor described by the coupled cavity model \cite{Taimre:15}. Right: LFI sensing element on a coin \cite{meyer11788compact, Meyer2021}.}
	\label{D1fig:3MirrorModell}
\end{figure}
The basic sensing principle of an LFI sensor can be described by the coupled cavity model as shown in \Cref{D1fig:3MirrorModell}. A single mode VCSEL emits coherent light with optical power $P_0$ through the laser facet $F_2$. The light propagates along the optical axis of the laser until it hits a target (the eye) at distance $L_\mathrm{ext}$. Dependent on the target's reflectivity $R$, a portion of the emitted light is  back-scattered and injected back into the laser cavity after the round trip time $\tau_{ext}$. The round trip time is dependent on the external refractive index $n_\mathrm{ext}$ and the speed of light $c_0$.

The back injected light interferes with the local oscillating field of the laser, leading to a perturbation of the lasers optical power through constructive or destructive interference resulting in a modulation of the optical power $P_{fb}$
\begin{equation}
\label{D1equ:mod_power}
P_{fb} = P_0 \left(1- m\cos\left(\phi_{ext}\right)\right).
\end{equation}
The modulation of optical power depends on the modulation index $m$, which attributes an amplitude modulation, and the phase modulation $\phi_{ext}$ through the cosine term. A small fraction of the modulated power $P'_{fb}$ is measured by a photo detector integrated into the laser. 

According to Taimre et al. \cite{Taimre:15}, the external phase $\phi_{ext}$ is linked via the signal phase $\phi_0$ of the unperturbed laser with the desired observations (distance and velocity) by

\begin{equation}
\label{D1equ:Excess_phase_equation}
\phi_{ext} - \phi_0 + C \sin \left(\phi_{ext} + \arctan\left(\alpha\right)\right) = 0
\end{equation}

where $C$ describes \textit{Acket's feedback parameter} and $\alpha$ \textit{Henry's linewidth enhancement factor} \cite{1071522}. Considering operation of the LFI sensor in the weak feedback regime ($C < 1$) and $\alpha$ to be constant \Cref{D1equ:Excess_phase_equation} has only a single solution and $\phi_{ext}$ is directly linked to 
\begin{equation}
\label{D1equ:signal_phase}
\phi_0 = \frac{4 \pi n_{ext} L_{ext}}{\lambda},
\end{equation} 
with laser wavelength $\lambda$ \cite{Taimre:15}. Assuming operation in free space ($n_{ext} \approx 1$), frequency modulation of optical power through a modulation of phase $\phi_0$ requires either a change in distance $L_\mathrm{ext}$ or a change of the laser's wavelength $\lambda$ w.r.t. time. To separate both effects, the partial derivatives  $\frac{\partial \lambda}{\partial t}$ and $\frac{\partial L_{ext}}{\partial t}$ of \Cref{D1equ:signal_phase} are calculated leading to
\begin{equation}
f_0 = \frac{2 L_{ext}}{\lambda} \frac{d \lambda}{d t}
\end{equation}
and
\begin{equation}
f_d = \frac{2 v_{ext} \cos \left(\gamma \right)}{\lambda},
\end{equation}
with $f_0$ as distance corresponding beat frequency, $f_d$ as velocity corresponding Doppler frequency, $v_{ext}$ as eye velocity and $\gamma$ as incident angle between eye and laser beam \cite{NORGIA201731}.

A modulation of the laser's wavelength is achieved through modulation of the laser's drive current while a modulation of the $L_{ext}$ is achieved through a moving target. To measure target distance as well as the target velocity simultaneously, a triangular modulation pattern of the laser drive current is applied similar to FMCW LIDAR systems \cite{8067701}. Therefore the update rate of the LFI sensor directly corresponds to the triangle modulation rate, which can reach up to 100\,kHz for spatial confined semiconductor VCSELs \cite{NORGIA201731}. 
\subsubsection{Geometric Eye Model}
\label{D1subsec:Geometric_eye_model}
Similar to the discussed model based eye tracking approaches in \Cref{D1Related_Work}, we derive a geometric model to link the LFI sensor measurements, in particular the distance measures, to the pose of the eye and thus the gaze vector. To derive our eye model, a linear scan across the surface of a healthy human eye is made while the eye is fixating on an object. 
\begin{figure}
	\centering
	\includegraphics[width=0.8\linewidth]{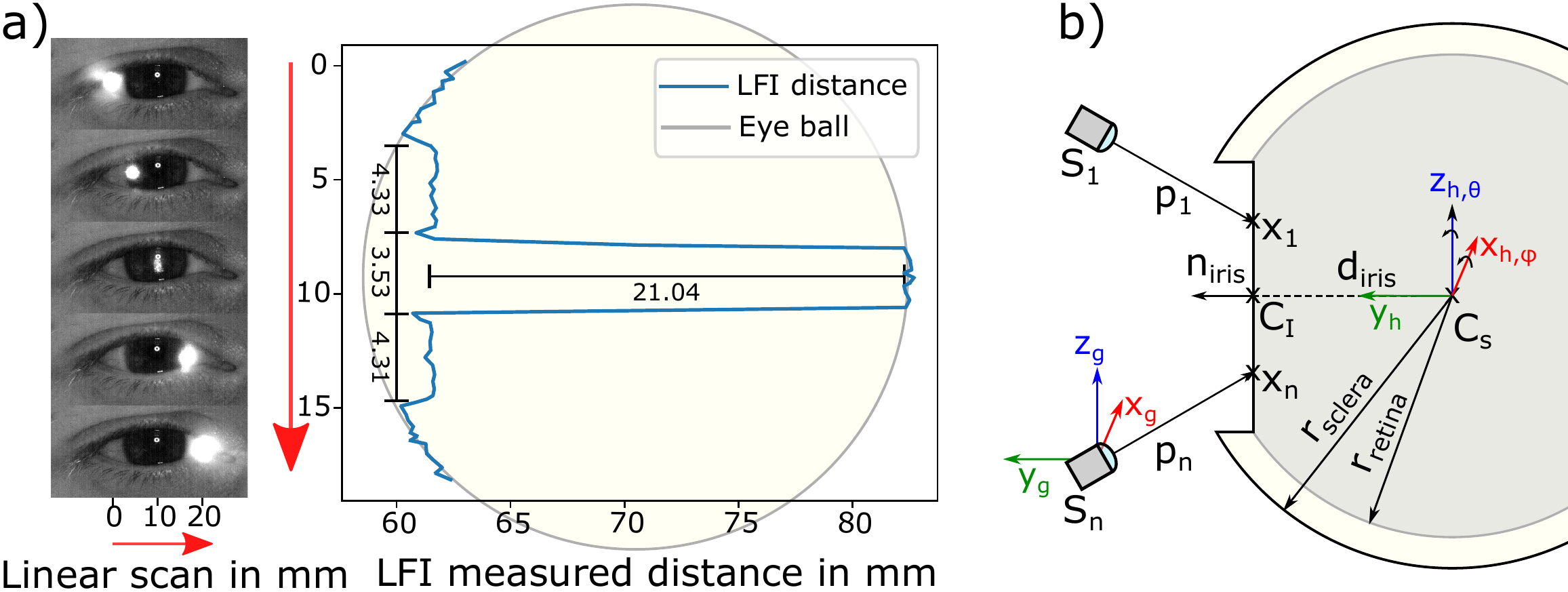}
	\caption{Left: Images from reference camera. Center: Distance measurements of the LFI sensor for different positions on the eye's surface. Right: Derived geometrical system model.}
	\label{D1fig:eye_model}
\end{figure}
\Cref{D1fig:eye_model} a) shows images taken by an IR reference camera while the IR laser beam\footnote{Class 1 laser system according to IEC 60825-1} (white spot) was scanned over the participants eye. The scan direction is indicated by the red arrow. For each scan position, the measured distance between laser and eye surface is shown. The distance measurement follows the curvature of the sclera until it reaches the limbus. At this point, the beam penetrates the cornea and hits the iris. If the beam hits the pupil, light is scattered back from the retina and thus the distance towards the retina is measured. The measured total diameter of the iris is 12.17\,mm, which fits well to average population  \cite{6645910}. From this observation we derived a geometric model as shown in \Cref{D1fig:eye_model} b). The iris is modeled as plane inside the eyeball with a distance $d_{iris}$ between the center of rotation $C_\mathrm{S}$ and the iris center $C_\mathrm{I}$. The eye ball is modeled as two spheres, one sphere with radius $r_{sclera}$ to link measured distances from the outer eye to the model, and a second sphere with radius $r_{retina}$ to link measured distances from the inner eye to the model. In addition, the LFI sensors are modeled as point sources with a pose $S_n$ and a linear laser beam, hitting the eye at an intersection point $x_{n}$. The gaze vector is described as normal vector $n_{iris}$ perpendicular to the iris plane. Similar to the geometric eye model proposed by Hennessy et al. \cite{Hennessey2006}, $d_{iris}$, $r_{sclera}$ and $r_{retina}$ are considered as fixed parameters of the model, which are derived from average population.
\subsubsection{Multi LFI Sensor Simulation tool}
\label{D1Subsec:Multi_LFI_sensor_simulation_tool}
\begin{figure}[h]
	\centering
	\includegraphics[width=0.8\linewidth]{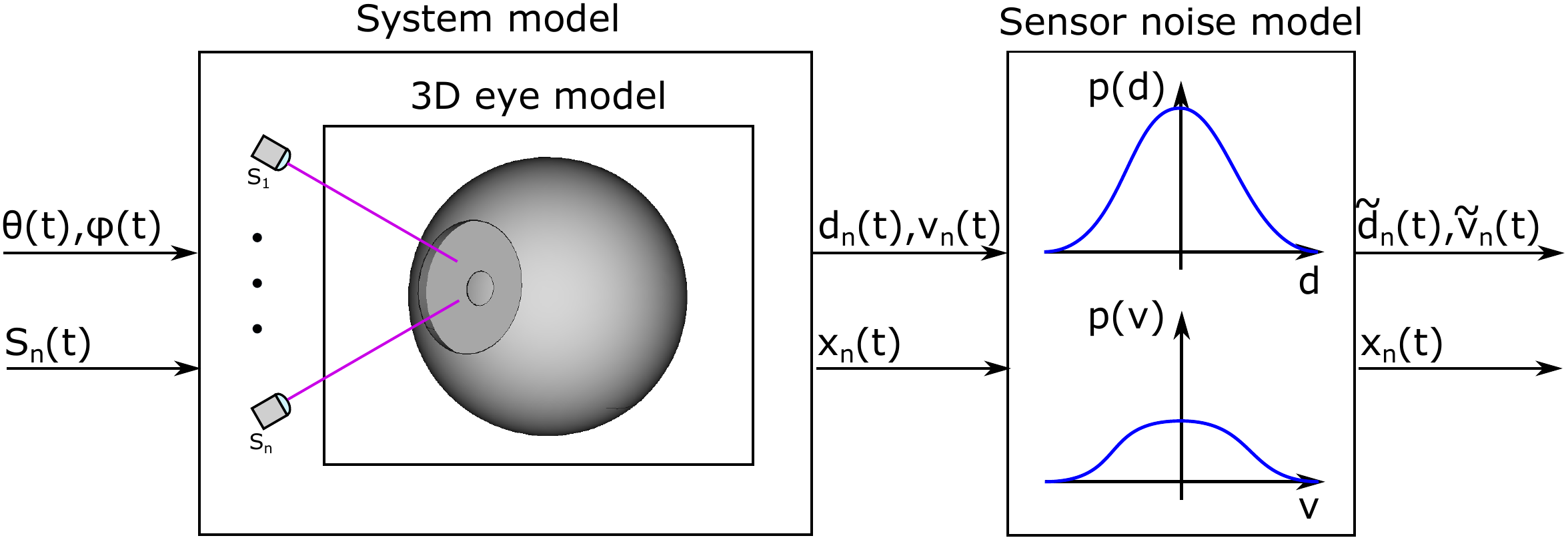}
	\caption{Block diagram of the LFI simulation tool used to generate LFI measurement data without human error.}
	\label{D1fig:simulation_model}
\end{figure}
\FloatBarrier
In order to evaluate the proposed LFI gaze estimation system, we set up a simulation tool to generate sensor data from arbitrary sensor poses $S_n$ and gaze trajectories denoted by the tuple ($\theta$,$\phi$). Compared to a laboratory setup, the simulation allows to evaluate the gaze estimation accuracy while excluding human error e.g. through noise. Thus a simulated environment allows for systematic investigation of the different algorithm steps on gaze accuracy.

 Input of the simulation is a trajectory of eye positions described by a tuple of gaze angles $(\theta(t)$,$\phi(t))$ and the sensor pose $S_n(t)$ of $n$ sensors. The sensor pose $S_n$ of the n-th sensor consists of a sensor position $s_n$ and a direction $p_n$ describing the propagation direction of the laser beam in the glasses' coordinate space. The sensor pose can vary over time to investigate the effect of glasses slippage on gaze estimation accuracy of the proposed system.

The input parameters are used to setup our system simulation inside a CAD program. The sensors are modeled as point sources in space with a laser beam propagating towards the 3D eye model. The parameters of the eye model e.g. $r_{sclera}$, $r_{retina}$ or $d_{iris}$ are derived from a 3D scan of an human eye \cite{10.1093/nar/gkn613}. To fit the 3D eye model to the proposed geometric eye model derived in \Cref{D1subsec:Geometric_eye_model}, the cornea and the lens are removed. 

During the simulation, the eye is rotated by $\theta(t)$ and $\phi(t)$ while the sensor pose is updated for each time step $t$ sequentially according to $S_n(t)$. At each time step, the beam of each LFI sensor is propagated from the laser position $s_n$ along the laser beam direction $p_n$ to calculate the intersection point $x_n(t)$ with the 3D eye model. The corresponding distance measure $d_n(t)$ is calculated as the Euclidean distance between sensor position and intersection with the eye. In addition, based on the intersection point $x_n(t)$, the region of intersection (none, sclera, iris, retina), which is hit by the laser beam, is extracted. The barely occurring region \textit{none} summarizes distance measurements not stemming from the eyeball e.g. distance measurements from the lashes or the eye lid during a blink. To calculate the velocity $v_n(t)$ measured by the LFI sensor $S_n$, the rotation axis $\omega(t)$ is calculated from the difference between two gaze trajectory points. Afterwards, the velocity is calculated, e.g., for the sclera by
\begin{equation}
v_n(t) = S_n(t) \cdot \left(\omega(t) \times x_n(t) \right).
\end{equation}

However, the system model does not take into account sensor noise e.g. due to speckling \cite{kliese2012spectral} leading to ideal LFI measurements, leading to ideal senor measures. Therefore we add a sensor noise model to the simulation tool to incorporate sensor noise described by $\mathcal{N}(0,\,\sigma_d^{2})$ and $\mathcal{N}(0,\,\sigma_v^{2})$. The parameters $\sigma_d^{2}$ and $\sigma_v^{2}$ of the sensor noise model are derived from sensor characterization measurements. The final output of the simulation is a set of LFI sensor measurement for each sensor and each time stamp, which is used as input for the gaze reconstruction algorithm. 

\subsection{Gaze Reconstruction Algorithm}
\label{D1Gaze Reconstruction Algorithm}
The general structure of the gaze estimator consists of three individual stages: 
\begin{enumerate}[label=\Alph*]
	\item \emph{Region of Intersection Classification}: For each laser, determine the region of intersection with the eye (none, sclera, iris, retina).
	\item \emph{LFI-Sensor Position and Pose Estimation}: From measurements of all lasers with known class-labels, estimate the sensor position and orientation of the lasers.
	\item \emph{Gaze Angle Estimation}: For known sensor positions and with all labeled measurements, estimate the gaze.
\end{enumerate}
In the following sections, each of the three stages is detailed. 

%


\subsubsection{Region of Intersection Classification}

For the classification, one crucial question is which signal(s) may be used for \emph{robust} classification.
From the measurements shown in \Cref{D1fig:eye_model} a) it is readily seen, that by means of the distance signal, one could easily differentiate between the intersection point being on the sclera or on the retina. On the other hand, classification solely based on distance would require some threshold value, which might have to be adapted in case of slippage.



\begin{figure}
	\centering
	\includegraphics[width=0.5\linewidth]{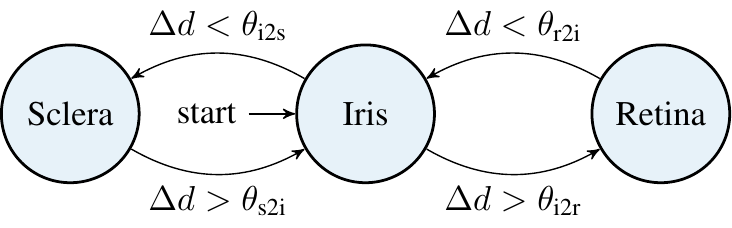}
	\caption{Deterministic automaton on difference of distance signal for naive classification.}
	\label{D1fig:deterministicAutomaton}
\end{figure}

Taking the time difference $\Delta d_n$ of the distance signal could reveal useful information about the laser position. Because the outer shape of the eye contains discontinuities, these will affect $\Delta d_n$ significantly. Therefore, a possible naive classifier could be the deterministic automaton shown in \Cref{D1fig:deterministicAutomaton}, operating solely on $\Delta d_n$.

The drawbacks of such automaton are that absolute distance information is neglected completely, and its susceptibility to noise in the distance signal. 
Also, the transition from sclera to iris will cause a much weaker signal in $\Delta d_n$ compared to a transition from iris to retina. 

\begin{figure}[h]
	\centering
	\includegraphics[width=0.5\linewidth]{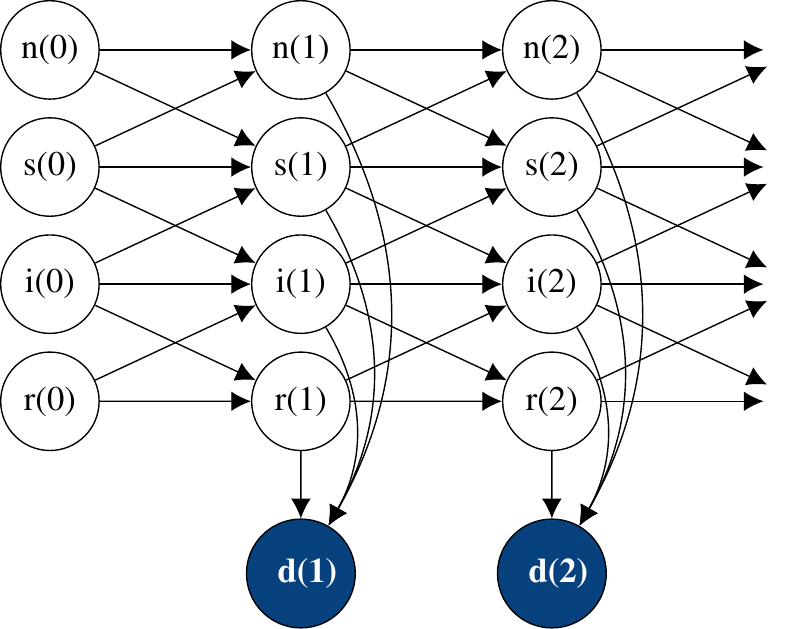}
	\caption{Hidden Markov Model for classification, depicted as lattice diagram.}
	\label{D1fig:HMM}
\end{figure}
\FloatBarrier
We therefore propose to use both signals $d_n$ and $\Delta d_n$ for classification in a probabilistic framework.
To do so, we set up a Hidden Markov Model (HMM) (see, e.g. \cite{Bishop2006}), which is depicted in \Cref{D1fig:HMM}.
Therein, the hidden (latent) states $z(k) = \begin{bmatrix} n(k) & s(k) & i(k) & r(k) \end{bmatrix}^T$ represent the intersection region (none, sclera, iris, retina) at time $k$ with a 1-of-K representation, i.e., $z_i(k) \in \{0,1\}$ and $\sum_i z_i(k)$ = 1.
We assume a multinomial distribution for $z$ with $\bE{}{\zk{i}} = \pi_i$ and $p(z) = \Pi_i \pi_i^{z_i}$.

For the  \emph{emission probabilities}, we assume normal conditional distributions 
\begin{equation}
\condProb{d}{z} = \prod_{i=1}^4 \npdf{d}{\mu_i}{\sigma_i}^{z_i}
\label{D1eq:condDistr}
\end{equation}
of the measured distance $d$ given a latent state $z$.
Therein, $\mu_i, \sigma_i$ are treated as unknown parameters, which are to be inferred online. 
This resembles the fact that the absolute position of the laser sensors are unknown a-priori, and might even change over time due to slippage.
With \Cref{D1eq:condDistr}, the joint distribution of latent and observed state at time instant $k$ follows as Gaussian mixture distribution
\begin{equation}
p(d) = \sum_z p(z) \condProb{d}{z} = \sum_{i=1}^4 \pi_i \npdf{d}{\mu_i}{\sigma_i}.
\end{equation}

In contrast to the emission probabilities with unknown parameters, the \emph{transition probabilities} ${p(z_i(k) | z_i(k-1), \Delta d(k-1))}$  can be specified well before-hand, if based on the measured difference of distance.
This is mainly due to the fact that eye-size varies only weakly among healthy adults \cite{Bekerman2014}.
Hence, distance signal jumps during traversing of the laser signal from one eye part to another may be treated as quasi-constant quantity, akin to the automaton in \Cref{D1fig:deterministicAutomaton}. 
The strongest assumption in this regard is the spot size of the laser, which is not infinitesimally small.
Hence, for slow eye movements or short sampling times, the laser might hit two neighboring eye-parts simultaneously, which might hinder transition detection.

\begin{figure}[h]
	\centering
	\includegraphics[width=0.7\linewidth]{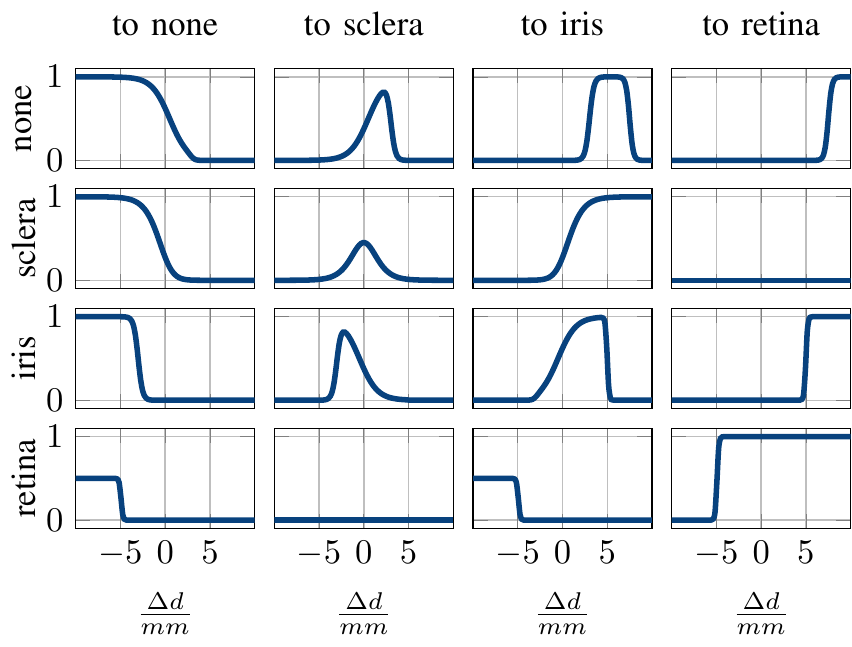}
	\caption{Modeled transition probabilities.}
	\label{D1fig:transitionProbabilities}
\end{figure}

The conditional probability of a latent variable given its previous value 
\begin{equation}
\condProb{z(k)}{z(k-1)} = \prod_{i=1}^4 \left( \sum_{j=1}^4 a_{ij} z_j(k-1) \right)^{z_i(k)}
\end{equation}
are assumed to be multinomially distributed with transition probabilities $A = \left[ a_{ij}(\Delta d)\right]$ shown in \Cref{D1fig:transitionProbabilities}. 
As can be seen, starting from the laser hitting the \textit{none} region, zero or negative distance change favors the classifier to remain in the \textit{none} region, whereas a significant positive distance change is likely due to a transition to sclera, iris or retina region dependent on the magnitude of the distance change. 
A direct jump from sclera to retina is given zero probability due to high sampling rate and limited eye movement.
Same reasoning holds for the other transition probabilities.

\FloatBarrier
For the HMM, (approximate) inference has to be carried out online for sequential data, i.e. measured distances $d(k)$ and difference of distances $\Delta d(k)$.
We make use of a rather standard maximum likelihood-approach, namely Expectation Maximization (EM).
A number of EM-like algorithms have been developed for the case of HMMs, e.g., the \emph{forward-backward algorithm}, and the \emph{Viterbi algorithm} (see \cite{Bishop2006} and the references therein). 
However, since the transition probabilities $A(k)$ are assumed to be known, and multinomial and normal distributions are used, the implementation is significantly simplified, which is derived next.

If the complete-data $\{D,Z\} = \{d(k),z(k)\}, k=1,\ldots, T, $ was known, inference of the unknown parameters $\theta = \{\pi_i, \mu_i, \sigma_i \}$ could simply be carried out by maximizing the complete-data log-likelihood function
\begin{multline}
\ln \condProb{D,Z}{\theta} =  \ln \prod_k  \condProb{\zk{}}{\zkm{}, A(k)} \cdot \condProb{d(k)}{\zk{},\theta} =  \\
\sum_k \ln \condProb{\zk{}}{\zkm{}, A(k)} + \ln \condProb{d(k)}{\zk{}, \theta}
\label{D1eq:completeDataLogLike}
\end{multline}

However, since the latent variables $Z$ are not known a-priori, an iterative EM scheme is used. 
During the \emph{expectation-step}, the posterior $\condProb{Z}{D,\theta'}$
over latent variables are computed for previous parameter set $\theta'$ and measurements $D$.
From \Cref{D1eq:completeDataLogLike}, one may readily derive the relation
\begin{multline} 
\condProb{Z}{D,\theta'} \propto \prod_{k=1}^T \condProb{\zk{}}{\zkm{}} \condProb{d(k)}{\zk{}, \theta'}  \\
= \prod_{k=1}^T \prod_{i=1}^4 \left(  \pi_i \npdf{d(k)}{\mu_i'}{\sigma_i'}  \sum_{j} a_{ij}(k) z_j(k-1) \right)^{\zkm{i}}.
\end{multline}
Furthermore, by taking the expected value, the \emph{responsibility} 
\begin{multline}
\gfun{\zk{i}} := \bE{}{\zk{i}} = 
\frac{\left(\sum_{j} a_{ij} \zkm{j}\right) \pi_i \npdf{d(k)}{\mu_i'}{\sigma_i'} }{\sum_{h} \left(\sum_{j} a_{ij} \zkm{j} \right) \pi_h \npdf{d(k)}{\mu_h'}{\sigma_h'}}
\label{D1eq:gamma}
\end{multline}
is obtained.
Note that for each time instance $k$ \Cref{D1eq:gamma} depends on the current data $d(k)$ and due to the transitions on the latent variable of the previous time step $\zkm{}$.

During the \emph{maximization-step}, the latent variables are fixed in order to recompute the parameters $\theta$.
This is achieved by maximizing the log-likelihood function $\ln \condProb{D}{\theta}$, which is obtained by marginalizing over \Cref{D1eq:completeDataLogLike} for $Z$, yielding the expectation w.r.t $Z$:
\begin{equation}
\bE{Z}{\ln \condProb{D,Z}{\theta}}
= \sum_Z \condProb{Z}{D,\theta} \cdot \ln \condProb{D,Z}{\theta}.
\end{equation}
In order to make the computation tractable, $\theta'$ is used within posterior $\condProb{Z}{X,\theta'}$, yielding
\begin{equation}
\Qfun{\theta}{\theta'} := \sum_Z \condProb{Z}{D,\theta'} \cdot \ln \condProb{D,Z}{\theta}.
\label{D1eq:Qfun1}
\end{equation}
Thus, during iteration of the EM-scheme, 
\begin{equation}
\lim_{\theta' \rightarrow \theta} \Qfun{\theta}{\theta'} = \bE{Z}{\ln \condProb{D,Z}{\theta}}.
\end{equation}

Substituting the complete-data log-likelihood \Cref{D1eq:completeDataLogLike} into \Cref{D1eq:Qfun1} and marginalizing over $Z$, 
\begin{multline}
\Qfun{\theta}{\theta'} = \sum_{k=1}^T \sum_{Z} \condProb{\zk{}}{d(k),\theta'} \ln \condProb{d(k)}{\zk{},\theta}  \\
+ \sum_{k=1}^T \sum_{Z} \condProb{\zk{},\zkm{}}{D} \ln \condProb{\zk{}}{\zkm{},D} 
\label{D1eq:Qfun2}
\end{multline}
is obtained, which is to be maximized w.r.t $\theta$.
Apparently, the second summand does not depend on $\theta$, hence only the first term has to be considered.
Taking the derivative of \Cref{D1eq:Qfun2} w.r.t $\theta$ and solving for optimal parameter values yields known results for Gaussian mixture distributions \cite{Bishop2006}:
\begin{subequations}
	\begin{align}
	\pi_i &= \frac{N_i}{T} = \frac{1}{T} \sum_{k=1}^T \gamma(\zk{i}), \\
	\mu_i &= \frac{1}{N_i} \sum_{k=1}^T \gfun{\zk{i}} d(k), \\
	\sigma_i &= \frac{1}{N_i} \sum_{k=1}^T \gfun{\zk{i}} \left(d(k) - \mu_i \right)^2.
	\end{align}
\end{subequations}
These update equations for the M-step assume availability of the full data set $d(k), k=1,\ldots,T$. They can however easily be rewritten as sequential update-equations for the online case:
\begin{subequations}
	\begin{align}
	N_i &= \underbrace{\sum_{k=1}^{T-1} \gfun{\zk{i}}}_{N_i'} + \gfun{z_i(T)} \\
	\pi_i &=  \frac{N_i}{T}, \\
	N_i \mu_i &=  \underbrace{\sum_{k=1}^{T-1} \gfun{\zk{i}} d(k)}_{N_i'\mu_i'} + \gfun{z_i(T)} d(T), \\
	N_i \sigma_i &= \underbrace{\sum_{k=1}^{T-1} \gfun{\zk{i}} \left(d(k) - \mu_i \right)^2}_{N_i'\sigma_i'} + \gfun{z_i(T)} \left(d(T) - \mu_i \right)^2.
	\end{align}
	\label{D1eq:statistics}
\end{subequations}
Hence, only the statistics $N_i', N_i'\mu_i, N_i'\sigma_i'$ have to be stored, rather than the entire set of measurements growing over time.

The EM-procedure for an offline-available batch of data usually requires iteration between the E- and M-steps until a certain convergence criterion is fulfilled, e.g., the log-likelihood function or the parameters themselves.

Since we carry out the inference in an online fashion, it is justified to perform only one iteration with each new datum. 
If $\theta$ is quasi-constant over time and sample rate is high, convergence of the parameter estimation may be achieved over time.

Furthermore, the statistics \Cref{D1eq:statistics} are suitable in case of the parameters $\pi_i, \mu_i, \sigma_i$ being constant over time, since inherently, mean value computations are carried out in \Cref{D1eq:statistics}.
For the present application, however, parameters might vary over time in case of slippage of the wearable device. 
In this case, a certain \emph{forgetting property} (see, e.g., \cite{Johnstone1982}, \cite{Shaferman2021}), such that the most recent measurements have a more dominant impact on the estimated parameters compared to outdated measurements.

A constant exponential forgetting factor however would require a persistence of excitation.
For the considered application, this requires that each laser to collect measurements from sclera, iris and retina region sufficiently often in order not to completely forget any learned parameters.

We therefore modify the update equations in such a way that the old statistics are weighted by a forgetting factor $\lambda_i$ depending on the current responsibility $\gfun{\zk{i}}$ i.e., number of supporting samples for cluster $i$ and a desired number $N^*$ of past supporting samples, yielding:
\begin{subequations}
	\begin{align}
	\lambda_i &= \min\left\{ \frac{N^* - \gfun{\zk{i}}}{N_i'} , 1\right\} \\
	N_i &= \lambda_i N_i' + \gfun{z_i(T)} \\
	\pi_i &=  \frac{\lambda_i (T-1) \pi_i' + \gfun{z_i(T)} }{\lambda_i (T-1) + 1}, \\
	\mu_i &= \frac{\lambda_i N_i' \mu_i' + \gfun{z_i(T)} d(T)}{\lambda_i N_i' + \gfun{z_i(T)}}  \\
	\sigma_i &= \frac{\lambda_i N_i' \sigma_i' + \gfun{z_i(T)}\left(d(T) - \mu_i \right)^2}{\lambda_i N_i' + \gfun{z_i(T)}} 
	\end{align}
	\label{D1eq:statisticsWForgetting}
\end{subequations}

Loosely speaking, these modified update equations collect $N^*$ data points for each cluster and forget about previous ones. 
Hence, forgetting is explicitly designed for each cluster depending on the actual information, rather than having a constant forgetting independent of the cluster support.

Updating \Cref{D1eq:statisticsWForgetting} will infer the model parameters over time. 
In contrast to the emission probabilities, the transition probabilities $a_{ij}$ are assumed to be given. 
This reasonable assumption leads to a low number of parameters to be inferred.
One shortcoming of this approach however is the different treatment of sorting of clusters: 
The transition probabilities assume a sorting $\mu_1 < \mu_2 < \mu_3 < \mu_4$, such that the labeling $l_i$ $z \rightarrow \begin{bmatrix}\text{none}, \text{sclera}, \text{iris}, \text{retina} \end{bmatrix}$ becomes possible. 
This sorting however does not hold a priori for the inferred parameters of \Cref{D1eq:statisticsWForgetting}.
Depending on the initialization of $z(0)$, the means of emission distributions may not be sorted.
A solution to this problem is to rearrange the clusters during each time step, such that their means have an ascending order.

\begin{figure}[h]
	\centering
	\includegraphics[width=0.7\linewidth]{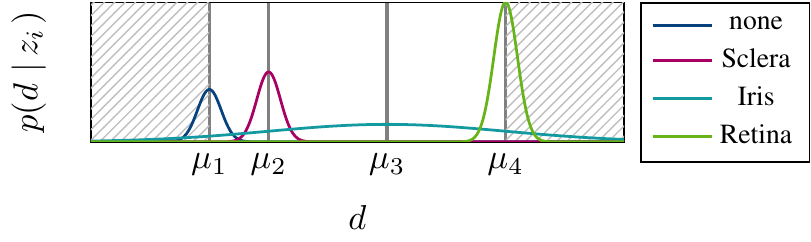}
	\caption{Example normal distributions for conditional emission probabilities.}
	\label{D1fig:interpolExtrapol}
\end{figure}
\FloatBarrier
Another shortcoming of this classifier is its limitation to the interpolation region. 
\Cref{D1fig:interpolExtrapol} depicts an exemplary mixture of normal distributions defining the emission probabilities. 
For this known distribution, one can easily predict the most probable class for a measurement $d(k)$. 
However in the extrapolation region, shown as shaded area in \Cref{D1fig:interpolExtrapol},  this might lead to miss-classification. 
Consider in this example $d\ll \mu_1$ or $d \gg \mu_4$. 
Then, one would have highest probability for $d$ belonging to the iris, but it is a \textit{none} or retina point.
As an easy solution, clamping the measured distances to $d \in [\mu_1, \mu_4]$ when evaluating the conditional emission probability will lead to the desired behavior.

The classification is summarized as follows:

\begin{algorithm}[H]
	Init: $N^*$, $z_0$
	\begin{algorithmic}
		\FOR{k=1, \ldots, T} \STATE 
		\begin{itemize}
			\item get measurements $d(k), \Delta d(k)$\\
			\item update transition probability matrix $A(\Delta d(k))$ \\
			\item update statistics \Cref{D1eq:statisticsWForgetting} \\
			\item sort clusters depending on $\mu_i$ in ascending order\\		
			\item classify measurement $d(k)$ by means of \Cref{D1eq:gamma} with $d$ limited to the interpolation region.
		\end{itemize}
		\ENDFOR		
	\end{algorithmic}
	\caption{Classification Algorithm.}
	\label{D1alg:classification}
\end{algorithm}

As a result, one obtains labels for each measurement, and in addition a vague model of the eye.

This information is exploited in following steps, namely estimation of the sensor positions and actual gaze angle estimation.

\subsubsection{LFI-Sensor Position and Pose Estimation}

In order to estimate the gaze angle, it is first necessary to determine absolute position and orientation of the laser sensors mounted to the glasses.
In the sequel, we denote the absolute sensor position in glasses-fixed coordinates as $\sg \in \bR^3$, a unitary orientation vector as $p$, and denote $\xg$ as intersection point of laser beam with the eye surface in glasses-fixed coordinates.
These intersection points
\begin{equation}
x_{\mathrm{g},n} = s_{\mathrm{g},n} + d_n p_n
\end{equation}
can be easily computed from the known sensor position in the glasses frame and the (scalar) distance measurement $d_i$.

The task is now to find the absolute position of the sensors $s_{\mathrm{g},n}, n=1,\ldots,N$ in head-fixed coordinates, with $N \geq 3$, and the origin of head-fixed coordinates being the eye ball center point $C_\mathrm{S}$. 
The transformation between both coordinate systems may be denoted as
\begin{equation}
\xh = R(\vartheta, \varphi, \varrho)\xg + t,
\label{D1eq:affineTrafo}
\end{equation}
with rotation matrix $R \in \bR^{3 \times 3}$ and translation $t \in \bR^3$.
For the rotation matrix, we use the common decomposition in roll, pitch and yaw
\begin{equation}
R(\vartheta, \varphi, \varrho) = R_z(\varrho) R_y(\varphi) R_x(\vartheta).
\end{equation}

The idea is to solve the task by means of trilateration given those measurements were classified to be either a sclera or retina point. 
For these measurements, we have 
\begin{equation}
\| x_{\mathrm{h},n} \| = r = 
\begin{cases}
\rsc, \quad l_i = s, \\
\rrt, \quad l_i = r.
\end{cases}
\label{D1eq:casesNormConstraint}
\end{equation}
due to the ball shape of sclera and retina.
Substituting \Cref{D1eq:affineTrafo} into \Cref{D1eq:casesNormConstraint} yields
\begin{equation}
x_{\mathrm{g},n}^T R^T R x_{\mathrm{g},n} + 2 x_{\mathrm{g},n}^T R^T t + t^T t - r^2 = 0,
\end{equation}
which can be further simplied to 
\begin{equation}
x_{\mathrm{g},n}^T x_{\mathrm{g},n} + 2 x_{\mathrm{g},n}^T R^T t + t^T t - r^2 = 0,
\label{D1eq:constraintsTrafoEstimation}
\end{equation}
since $R$ is unitary (i.e., $R^T R = I$), with the unity matrix $I$.

Collecting a batch of measurements of all lasers over time, in which the glasses position is constant, \Cref{D1eq:constraintsTrafoEstimation} constitutes a non-linear (possibly over-determined) system of equations in the six unknowns $\vartheta, \varphi, \varrho, t_x, t_y, t_z$.

Although non-linear in nature, one has good chances to solve this root-finding problem, e.g., by means of a Levenberg-Marquardt algorithm \cite{More1978}.

\subsubsection{Gaze Angle Estimation}

\begin{figure*}[h]
	\centering
	\includegraphics[width=\linewidth]{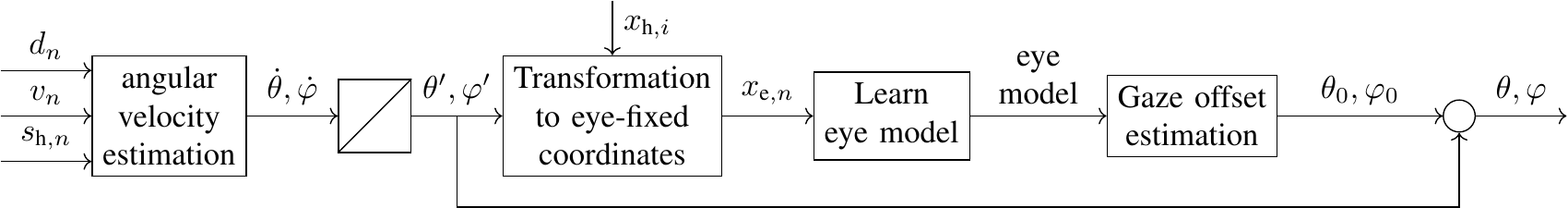}	
	\caption{Gaze estimation principle.}
	\label{D1fig:blockDiagramGazeEstimation}
\end{figure*}

Having carried out classification of measurements and estimation of absolute sensor position, the final step is now the actual gaze estimation.
The main idea is to estimate and integrate the angular velocity of the eye.
Due to the unknown starting value of the integration, the initial estimation $\theta', \varphi'$ will be corrupted by an offset.
In order to compensate for this systematic error, a surrogate model of the eye in eye fixed coordinates is learned over time, by which the quasi-constant offset can be determined.
This principle is depicted in \Cref{D1fig:blockDiagramGazeEstimation}.

\FloatBarrier
In order to estimate the angular velocity of the eye, we first denote the intersection points of the laser sensors with the eye surface in spherical coordinates
\begin{equation}
x_n(r,\theta, \varphi) = 
r \begin{bmatrix} \cos \theta \cos \varphi \\ \sin \theta \cos \varphi  \\ \sin\varphi\end{bmatrix}.
\end{equation}
In addition to a distance proportional signal, each laser returns a velocity signal, which is the projection of the surface velocity at intersection point $x_i$ in direction $p_i$ of the laser beam.
Thus, 
\begin{multline}
v_n = -p_n^T\left( \frac{\partial x_n}{\partial r}\dot{r} + \frac{\partial x_n}{\partial \theta}\dot{\theta} + \frac{\partial x_n}{\partial \varphi} \dot{\varphi} \right) 
= -p_n^T\left( \frac{ x_n}{\| x_n \|} \dot{r} + \frac{\partial x_n}{\partial \theta} + \frac{\partial x_n}{\partial \varphi} \right)
\label{D1eq:velocitySignal}
\end{multline}
holds.
Obviously, for a laser beam hitting the sphere-like sclera or retina, $\dot{r} = 0$
follows in case of no slippage, i.e., the measured surface velocity is only due to eye rotation.
For a laser hitting the iris however, the change of $r_i = \| x_i \|$ has to be taken into consideration for the velocity signal.
In the following, we will regard the change of radius by means of a first order approximation
\begin{equation}
\dot{r} \approx \| x_n(k) - x_n(k-1) \| / \Ts.
\end{equation}
Defining $\theta$ and $\varphi$ as rotations about the x- and z-axes, i.e.
$e_\theta = e_x = \begin{bmatrix} 1 & 0 & 0\end{bmatrix}^T$
$e_\varphi = e_z = \begin{bmatrix} 0 & 0 & 1\end{bmatrix}^T$
\Cref{D1eq:velocitySignal} constitutes for all lasers $n=1,\ldots,N$ a linear system of equations
\begin{equation}
\begin{bmatrix}
-p_n^T \left( e_\theta \times x_n \right) &
-p_n^T \left( e_\varphi \times x_n \right)
\end{bmatrix}
\begin{bmatrix}
\dot{\theta} \\ \dot{\varphi}
\end{bmatrix}
=
v_n - \frac{x_n}{\| x_n \|} \dot{r}_n,
\label{D1eq:linSysEqAngVelEst}
\end{equation}
which can be solved during each time step by means of a (pseudo-) inverse. At this point, we omit a lengthy discussion on existence and uniqueness of the solution of \Cref{D1eq:linSysEqAngVelEst}.

Integrating the estimations of angular velocities yields
$\theta' = \int_0^T \dot{\theta} \dt$,  $\varphi' = \int_0^T \dot{\varphi} \dt$,
with yet unknown starting values $\theta_0, \varphi_0$.
In order to compensate for these offsets, all eye-surface points $\xhi$ measured in head-fixed coordinates are transformed into eye-fixed coordinates by means of 
\begin{equation}
\xe = R_z(\varphi') R_x(\theta') \xh.
\end{equation}
Note that we have neglected rotation around the $\yh$-axis (rolling) of the eye.

\begin{figure}[h]
	\centering
	\includegraphics[width=0.3\linewidth]{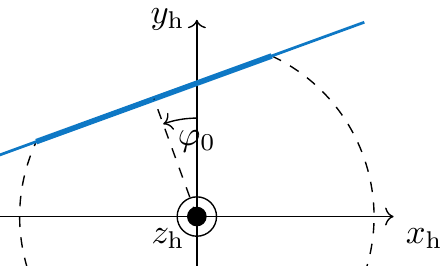}	
	\caption{Simple iris model.}
	\label{D1fig:surrogateModel}
\end{figure}

The $\xe$ are then used to parameterize a surrogate model for the eye shape in order to determine the quasi constant offset $\theta_0, \varphi_0$ with regard to a neutral viewing direction ($\theta=\varphi=0$).
In principle, several surrogate models are applicable to achieve this task. 
2D-tensor splines for example could be used to represent the entire outer shape of the eye. 
We are, however, interested in a minimal representation of the eye, to reconstruct the initial gaze direction.
Since information about the viewing direction is mainly given by means of iris-intersection points, it is sufficient to consider only these points.
Then, a simple plane representation 
\begin{equation}
\niris^T x = \diris
\label{D1eq:irisModel}
\end{equation}
of the iris as depicted in \Cref{D1fig:surrogateModel} is a suitable choice.
There are four parameters in total, which might be reduced to three if $\diris$ is preset to a reasonable value.
In any case, since the problem is parametric linear, it can be solved by means of a 
recursive least squares (RLS) with exponential forgetting \cite{Johnstone1982}.

\FloatBarrier
From the estimated parameters of \Cref{D1eq:irisModel}, the desired offset parameters for the gaze are then given by
\begin{subequations}
	\begin{align}
	\theta_0  &= \arcsin{\frac{\nirisz}{\| n \|}}, \\
	\varphi_0 &= \arctan{\frac{-\nirisx}{\phantom{-}\nirisy}},
	\end{align}
\end{subequations}
which in a last step gives the final gaze estimates
\begin{subequations}
	\begin{align}	
	\theta &= \theta' + \theta_0, \\
	\varphi &= \varphi' + \varphi_0.
	\end{align}
\end{subequations}

\subsection{Evaluation}
\label{D1Evaluation}
In this section, the proposed static LFI gaze estimation approach is evaluated including the LFI sensor measurement noise characterization, the region of intersection classification accuracy as well as the gaze estimation accuracy in presence of glasses slippage.

\subsubsection{LFI Sensor Characterization}
\label{D1Eval:LFI_sensor_characterization}
For accurate simulation of multi LFI sensor measurements generated by the simulation tool introduced in \Cref{D1Subsec:Multi_LFI_sensor_simulation_tool}, the sensor distance and velocity measurement noise need to be characterized. For this purpose, a similar lab setup as proposed by Capelli et al. \cite{5954803} is used. A disk with a diameter of 24\,mm is rotated with a precision rotation stage\footnote{Jenny Science ROTAX Rxhq 50-12}, while the LFI sensor is mounted on a precision linear stage\footnote{Jenny Science ELAX Ex 50F20} so that the laser beam hits the disk with an incident angle $\gamma$ of 45\,$^\circ$. The linear stage simulates distance variations of 20\,-\,30\,mm, which is reasonable for a near-eye setting, while the rotating disk simulates rotational eye movements with velocities up to 500\,$^\circ$/s.
\begin{figure}[h]
	\centering
	\includegraphics[width=0.8\linewidth]{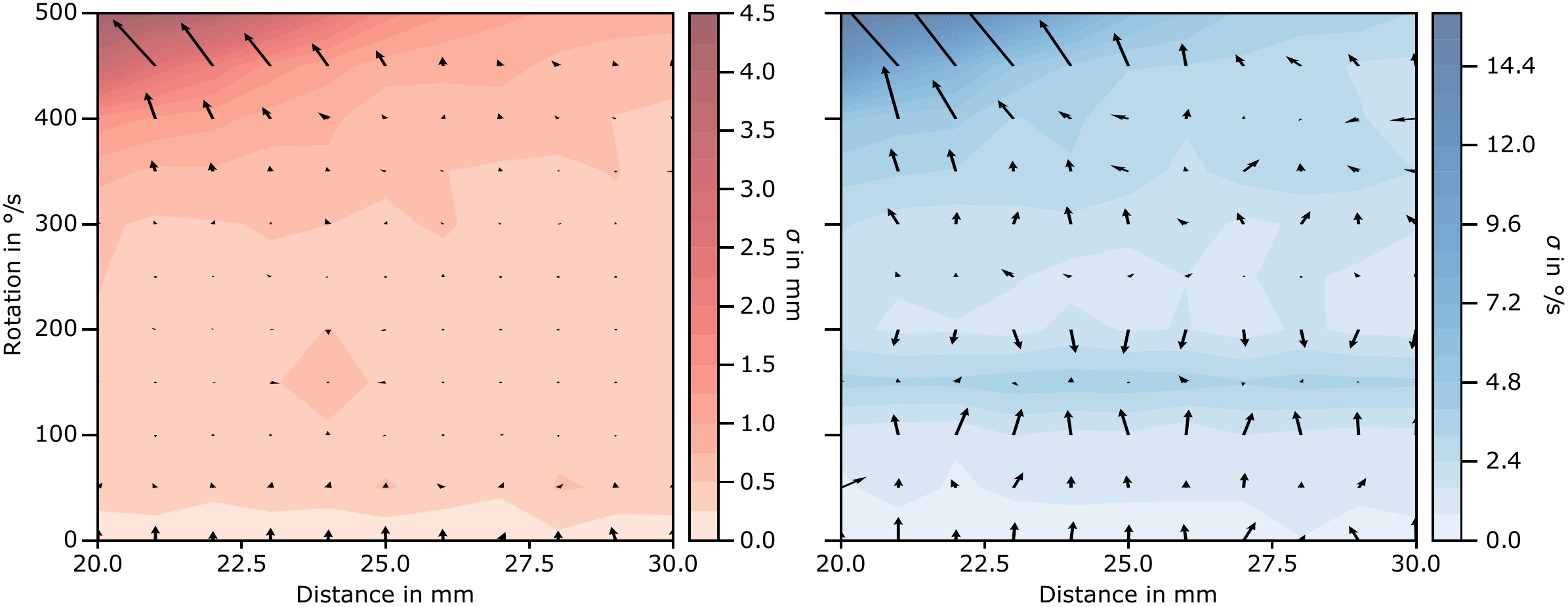}	
	\caption{Sensor noise characterization results over the whole range of eye rotational velocities and distances. Left: Variance of distance measurements, Right: Variance of velocity measurements.}
	\label{D1fig:sensor_charac}
\end{figure}
\FloatBarrier
\Cref{D1fig:sensor_charac} shows the resulting sensor noise for distance (red) as well as velocity (blue) measurements over the whole parameter space. The LFI sensor is capable of resolving the whole range of eye velocities up to rare occurring saccades with amplitudes up to 500\,$^\circ$/s. Especially for small distances and high velocities above 300\,$^\circ$/s the sensor noise increases as indicated by the gradient vectors in \Cref{D1fig:sensor_charac} due to limitations of the current electronics to resolve the signal. Compared to the work of Capelli et al. \cite{5954803}, who reported a maximum velocity error of 10\,\%, we achieved one of 2.8\,\% while measuring distance and velocity simultaneously. To exclude errors stemming from electronic limitations, we limit the velocity range to 300\,$^\circ$/s from which  $\sigma^2_\mathrm{d}$ and $\sigma^2_\mathrm{v}$ are derived. The resulting sensor noise characteristics used for further simulation are 66.85\,$\mu$m and 2.95$^\circ$/s for $\sigma^2_\mathrm{d}$ and $\sigma^2_\mathrm{v}$, respectively. 

\subsubsection{Region of Intersection Classification}
The region of intersection classification is the first stage of the proposed gaze estimation algorithm, introduced in \Cref{D1Gaze Reconstruction Algorithm}. To evaluate the performance of the classifier, we simultaneously sample a Pupil Labs V2 Core eye camera at 200\,Hz and an LFI sensor at 1\,kHz. Both sensors are pointing towards the eye of an participant\footnote{Class 1 laser system according to IEC 60825-1}. From the camera images, ground truth labels denoting the region of intersection hit by the IR laser were derived. 
\begin{figure}[h]
	\centering
	\includegraphics[width=0.6\linewidth]{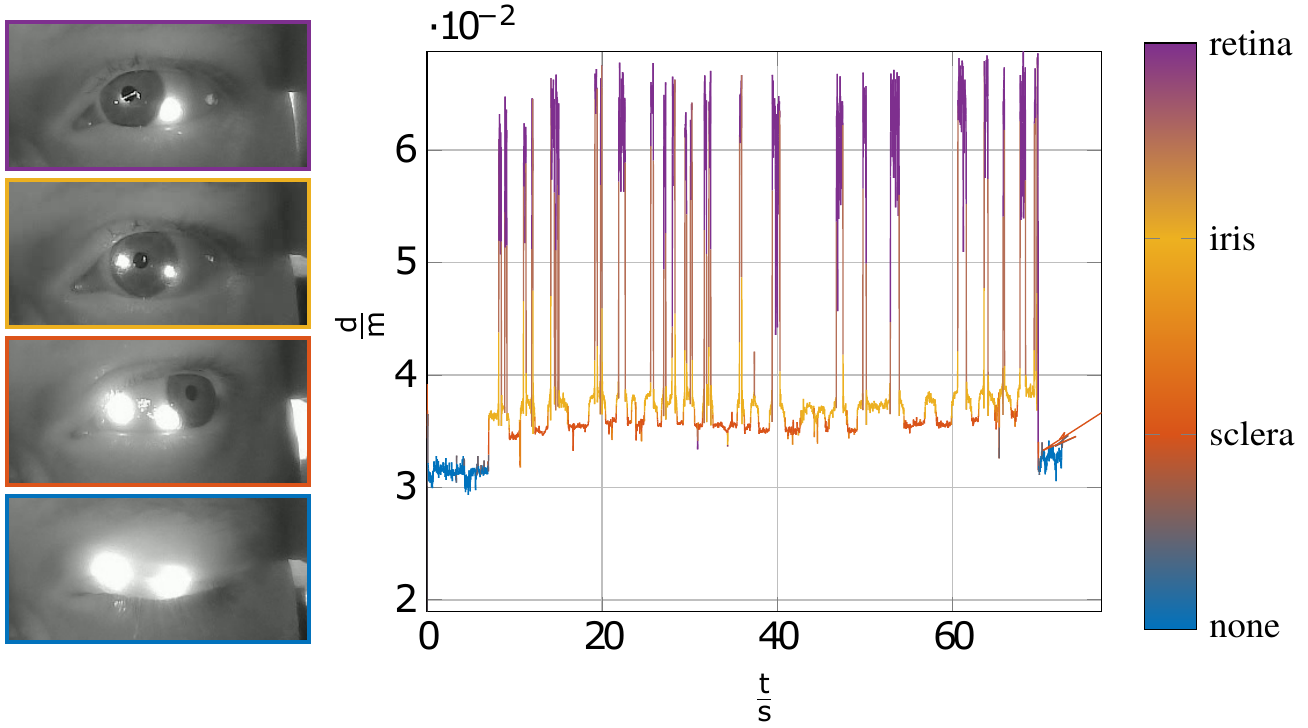}	
	\caption{Left: Reference camera images taken to derive ground truth labels of the left LFI sensor. Right: Measured distance of the left LFI sensor with corresponding label derived by the proposed classifier highlighted in color.}
	\label{D1fig:meas_Classifier}
\end{figure}
\FloatBarrier
The measured distances $d$ of the LFI sensor are fed into the proposed HMM classifier to derive classification labels. As the sample rates of LFI sensor and camera sensor differ, we up-sample the labels derived from the camera sensor from 200\,Hz to 1\,kHz by using nearest neighbor (NN) up-sampling.
\Cref{D1fig:meas_Classifier} shows the distances measured by the LFI sensor and the corresponding labels estimated by the classifier highlighted in color. Additionally, images of the VOG eye camera are shown, where the intersection of the LFI sensor with the eye is marked by a bright spot in the IR camera image. Classification was made for the left IR spot in the camera frames. The proposed HMM classifier reaches an macro F1-score of 94.42\,\%.
Classification mismatches occurs mainly between sclera and iris due to the overlapping probability distributions of the two classes. In addition, the ground truth labels are derived from the 200\,Hz camera frames and up-sampled using NN up-sampling, leading to an  inaccuracy of ground truth labels and thus affect the classification accuracy.

\subsubsection{LFI-Sensor Position and Gaze Estimation}
To evaluate the sensor position and gaze estimation accuracy of the proposed system, we used the multi LFI sensor simulation introduced in \Cref{D1Subsec:Multi_LFI_sensor_simulation_tool} to generate distance as well as velocity measurement trajectories for a setup of 6 LFI sensors. In order to take into account sensor noise, we initialized the sensor noise model with the parameters obtained by the sensor characterization as described in \Cref{D1Eval:LFI_sensor_characterization}. To evaluate the robustness of the proposed gaze estimation algorithm, we used a human eye movement trajectory captured by a Pupil Labs V2 Core eye tracker sampled at 200\,Hz. We then up-sampled it similar to the ground truth labels in the previous section to 1\,kHz by NN up-sampling before feeding the trajectory into the system simulation.
\begin{figure}[h]
	\centering
	\includegraphics[width=\linewidth]{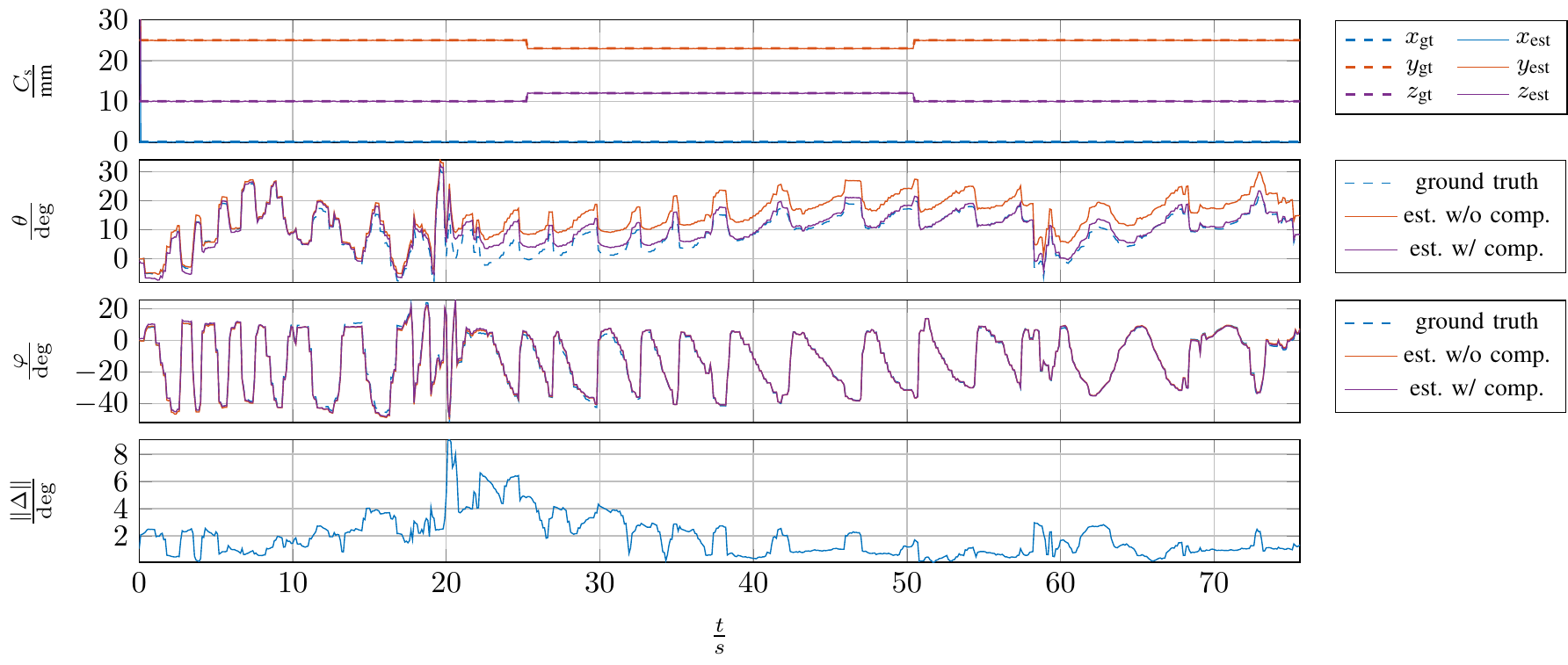} 
	\caption{ Top: Original and estimated eye ball center $C_\mathrm{S}$. At $t=25\,s$ the glasses slip 2\,mm towards the eye and 2\,mm up the nose. Center: Original gaze angle $\theta$ and $\varphi$ (dashed blue), estimated gaze angle $\theta'$ and $\varphi'$ (orange) and estimated gaze angle  $\theta$ and $\varphi$ with offset compensation (purple) Bottom: Absolute gaze angle estimation error in $^\circ$.}
	\label{D1fig:Gaze_estimation}
\end{figure}
\FloatBarrier
In addition, we let the sensors slip at $t=25\,s$ $\Delta C_{\mathrm{S}y}=-2\,mm$ towards the eye and $\Delta C_{\mathrm{S}z}=2\,mm$ along the nose to investigate robustness against slippage. At t=52\,s the glasses slip back to their initial position on the nose. We choose a step like slippage trajectory, as worst case assumption for glasses slippage.  \Cref{D1fig:Gaze_estimation} a) shows the original sclera center $C_\mathrm{S}$ (dashed) and the corresponding estimation from our algorithm. As the estimation relies on trilateration, our approach is capable of immediately re-estimating the eyeball center in presence of slippage. The root mean squared error (RMSE) of the eye ball center estimation is 76.39\,$\mu$m and the mean absolute error (MAE) is 123.11\,$\mu$m, which both are close to the distance noise limit of an individual sensor. Therefore, the eye model can be reconstructed precisely with a minimum error, which is mandatory as inaccuracies in the eye model reconstruction directly impact the gaze estimation accuracy. 

\Cref{D1fig:Gaze_estimation} b) and c) show the gaze estimation for both angles $\theta$ and $\varphi$. The original human eye movement trajectory (dashed blue) starts with random eye movements with different velocities during the first 20 seconds, followed by a reading part between $t=30\,s$ and $t=60\,s$ and finally slow eye movements and two fixations at the end of the trajectory. With this eye movement trajectory profile, a variety of eye movements from fixations to saccades are covered. 

The reconstruction of the gaze angles by integration of the measured velocity (orange) introduces especially for $\theta$ a large offset error. The RMSE and MAE error of the uncompensated gaze estimation is 5.28\,$^\circ$ and 6.14$^\circ$ respectively. Therefore, the compensation of the offset is mandatory. For this purpose, we introduced the surrogate model as given by \Cref{D1eq:irisModel} to continuously compensate gaze estimation offsets arising from integration of measured rotational velocity. With this additional step, the gaze estimation error declines and the estimated trajectory follows the original trajectory (cyan dashed) closely. The RMSE and MAE error of the compensated gaze estimation is 1.79\,$^\circ$ and 2.32\,$^\circ$, respectively. 

The absolute error shown in \Cref{D1fig:Gaze_estimation} d) rises during fast saccadic eye movements and starts declining over time if slower and controlled eye movements dominate the trajectory and the offset compensation starts to correct the velocity integration error.

\subsection{Discussion and Limitations}
The proposed LFI eye tracking approach is capable of tracking the eye robust in presence of glasses slippage. Furthermore, the sensor fusion and gaze estimation approach is able to estimate the optical axis of the eye without additional calibration, which fosters immersion of AR glasses and thus improves user experience. 

Compared to a slippage robust VOG eye tracking system like the Tobii Pro Glasses 2 with stereo camera sensors and additional IR LEDs, our LFI eye tracking approach achieves comparable gaze accuracy \cite{niehorster2020impact} with a lower power consumption. Furthermore, our system is not limited in the FOV, as we do not solely rely on tracking the pupil, but on the measurement of rotational speed on the entire eyeball.

The main limitation of the proposed LFI eye tracking system is that the surface of the eye must be hit by the laser beams to measure the rotational velocity of the eye. Therefore, it is possible that incorrect velocity measurements stemming from the \textit{none} region (lashes or lids) could lead to a gaze estimation error if the laser sensor e.g., during blinking or squinting the eye. Therefore, measurements classified as \textit{none} need to be handled separately during gaze estimation.

Finally a calibration routine is required to calibrate once the individual sensor poses $S_n$ within the glasses coordinate system, as the proposed algorithm assumes a fixed known sensor pose within the glasses coordinate system.

\subsection{Conclusion}
\label{Conclusion}
In this work, we present a novel eye tracking sensor system based on static LFI sensors. In addition to the novel sensing modality, an eye tracking algorithm is proposed, which is capable to fuse distance and velocity measurements of multiple static LFI sensors by means of an optimized geometric eye model. We further focus on a lightweight implementation of the different stages of the algorithm to be able to integrate the whole system into lightweight AR glasses. In addition, we derived a geometric eye model and characterized the sensor noise to build up a multi LFI sensor simulation tool to generate measurement data for a multi LFI sensor setting while excluding human error. With the generated data we evaluated the proposed LFI system and achieved a gaze accuracy of 1.79\,$^\circ$ at an outstanding sampling rate of 1\,kHz.

Based on the promising results of this work, we will investigate the achievable gaze estimation accuracy including human error by building up a head worn demonstrator in future work. In addition, we will investigate the combination of the proposed multi LFI system with a low frame rate camera sensor as reference  to combine the high accuracy of VOG systems with the outstanding update rate of the proposed LFI sensor system.

\newpage
\backmatter
\cleardoublepage
\phantomsection
\addcontentsline{toc}{chapter}{Bibliography}
\bibliographystyle{apalike}
\bibliography{bibliography}
\addcontentsline{toc}{chapter}{Bibliography}


\end{document}